%% file: thesis.tex
\documentclass[12pt,a4paper,onecolumn,twoside,openright,final]{book}
\usepackage[includehead,top=20mm, bottom=20mm, bindingoffset=7mm, marginparwidth=2.3cm, marginparsep=2mm]{geometry}

\newcommand{\imagepath}{../imgs} 

\usepackage{preamble}    
\usepackage[thesis]{optional}

\begin{document}

    \onehalfspacing

    \frontmatter

    \pagestyle{plain}
    \include{pads/title}

    \cleardoublepage
    \phantomsection
    \include{pads/copyright}

    \cleardoublepage
    \phantomsection
    \include{pads/examiners}

    \cleardoublepage
    \phantomsection
    \addcontentsline{toc}{chapter}{Preface}
    \include{pads/preface}

    \cleardoublepage
    \phantomsection
    \addcontentsline{toc}{chapter}{Dedication}
    \include{pads/dedication}

    \cleardoublepage    
    \phantomsection
    \addcontentsline{toc}{chapter}{Acknowledgements}
    \include{pads/acknowledgements}        

    \cleardoublepage
    \phantomsection
    \addcontentsline{toc}{chapter}{Abstract}
    \include{pads/abstract}


    \cleardoublepage
    \phantomsection
    \tableofcontents

    \cleardoublepage
    \phantomsection
    \addcontentsline{toc}{chapter}{List of Figures}
    \listoffigures

    \cleardoublepage
    \phantomsection
    \addcontentsline{toc}{chapter}{List of Tables}
    \listoftables

    \cleardoublepage
    \phantomsection
    \addcontentsline{toc}{chapter}{Nomenclature}    
    \printnomenclature  


    \mainmatter
    \pagestyle{fancy}
    \input{pads/nomenclature}
    \include{chapters/1_intro}

    \include{chapters/2_related_work}

    \pagestyle{plain}
    \include{pads/partI}
    \pagestyle{fancy}
    \include{chapters/3_offline_detection}
    \include{chapters/4_online_detection}
    \pagestyle{plain}
    \include{pads/partII}
    \pagestyle{fancy}
    \include{chapters/5_tramnet}

    \include{chapters/6_tpnet}
    \pagestyle{plain}
    \include{pads/partIII}
    \pagestyle{fancy}
    \include{chapters/7_rcn}
    \include{chapters/8_conclusion}


    \backmatter
    \pagestyle{fancy}
    \renewcommand{\bibname}{References}
    \cleardoublepage
    \phantomsection
    \addcontentsline{toc}{chapter}{\bibname}
    \bibliographystyle{ieeetr}
    \bibliography{ref}


\end{document}

%% file: pads/title.tex
\begin{titlepage}

	\begin{minipage}{0.9\textwidth}
	\end{minipage}
	
	\begin{center}
		\Large
		\textbf{Online Spatiotemporal Action Detection and Prediction via Causal Representations} 
	\end{center}	
		
	\vspace{0.5cm}

	\begin{center}		
		\Large
		{Gurkirt Singh}
	\end{center}
	
	\vspace{0.05cm}
	
	\begin{center}		
            A Thesis Submitted in Partial Fulfilment\\
            of the Requirements for the Degree of \\
            Doctor of Philosophy    \\
            in \\
	Computer Science and Mathematics
	\end{center}

	\vspace{0.3in}

	\begin{center}
	Visual Artificial Intelligence Lab\\
	School of Engineering, Computing and Mathematics\\
	Oxford Brookes University
	\end{center}
	
	\vspace{1.0in}

	\begin{center}
	Supervised by :\\
	Prof. Fabio Cuzzolin and Prof. Nigel Crook \\
	\end{center}

	\vspace{0.1in}
	
	\begin{center}		
		\small
		September 2019
	\end{center}
	
\end{titlepage}

%% file: pads/copyright.tex
\newpage
\begin{minipage}{0.9\textwidth}
\vspace{5cm}

\begin{center}
\emph{The author holds the \textbf{\emph{copyright}} of this thesis. 
Any person(s) intending to use a part or whole of the materials in the thesis in a proposed publication must seek copyright release from the author.}
\end{center}

\end{minipage}

%% file: pads/examiners.tex
\newpage

\begin{minipage}{0.9\textwidth}
\vspace{5cm}
\end{minipage}

\begin{center}
    \large
    \textbf{Thesis/Assessment Committee}
\end{center}


\vspace{0.3in}
\begin{center}
Professor Andrea Vedaldi (External Examiner)\\
Visual Geometry Group (VGG)\\
University of Oxford\\
\end{center}

\vspace{0.3in}
\begin{center}
Dr. Fridolin Wild (Internal Examiner)\\
School of Engineering, Computing and Mathematics\\
Oxford Brookes University\\
\end{center}

%% file: pads/preface.tex
\newpage
\begin{center}	
{\large
\textbf{Preface}
}
\end{center}
\vspace{1cm}

This dissertation is submitted for the degree of Doctor of Philosophy at The Oxford Brookes University of United Kingdom.
The research presented herein was undertaken under the kind supervision of 
Prof. Fabio Cuzzolin and Prof. Nigel Crook between September 2015 and August 2019.
To the best of my knowledge, this work is original, excluding those previous works for which acknowledgement and reference have been made.
Neither this, nor any substantially similar dissertation has been or is being submitted for any other degree, diploma or qualification at any other university. 
Part of this work are published in the following publications:

\begin{itemize}    
    \item \textbf{Gurkirt Singh} and Fabio Cuzzolin, 
 ``Recurrent Convolutions for Causal 3D CNNs'', 
 in Proceedings of International Conference on Computer Vision Workshop (ICCVW) on Large Scale Holistic Video Understanding, 2019.
 
 \item \textbf{Gurkirt Singh}, Suman Saha and Fabio Cuzzolin, 
 ``TraMNet - Transition Matrix Network for Efficient Action Tube Proposals'', 
 in Proceedings of Asian Conference on Computer Vision (ACCV), 2018.

 \item \textbf{Gurkirt Singh}, Suman Saha and Fabio Cuzzolin,
 ``Predicting Action Tubes'', in Proceedings of European Conference on Computer Vision Workshop (ECCVW) on Anticipating Human Behaviour, 2018.
 
 \item \textbf{Gurkirt Singh}, Suman Saha, and Fabio Cuzzolin, 
 ``Online Real-time Multiple Spatio-temporal Action Localisation and Prediction'', 
 in proceedings of International Conference on Computer Vision (ICCV) 2017.
 
 \item Suman Saha, \textbf{Gurkirt Singh} and Fabio Cuzzolin,
 ``AMTnet: Action-Micro-Tube Regression by end-to-end Trainable Deep Architecture'', 
 in proceedings of International Conference on Computer Vision (ICCV) 2017.

 \item Suman Saha, \textbf{Gurkirt Singh}, Michael Sapienza, Philip Torr and Fabio Cuzzolin,
 Deep Learning for Detecting Multiple Space-time Action Tubes in Videos,
 proceedings of British Machine Vision Conference (BMVC) 2016.

 \item \textbf{Gurkirt Singh} and Fabio Cuzzolin,
 ``Untrimmed Video Classification for Activity Detection: Submission to ActivityNet Challenge'', 
 in Tech-report arXiv:1607.01979, presented at ActivityNet Challenge Workshop in CVPR 2016.

\end{itemize}


%% file: pads/dedication.tex
\newpage
\begin{minipage}{0.9\textwidth}


\vspace{50mm}

\begin{center}

    To my parents\\
    And\\
    my wife\\
\vspace{10mm}
    Without whom none of my success would be possible.
\end{center}





%

\end{minipage}
\vfill

%% file: pads/acknowledgements.tex
\newpage
\heading{Acknowledgements}
\noindent
First and foremost I would like to express my heartfelt thanks to 
my director of studies Prof. Fabio Cuzzolin, and co-supervisor Prof. Nigel Crook
for providing such an exceptional opportunity 
for me to pursue a doctorate in such an interesting topic as Computer Vision! 
I also thank my supervisors for their invaluable time, guidance and support.
In particular, 
Fabio's strive for excellence, openness and positive criticism
always inspired me to push myself beyond my limits
and allowed me to gradually improve my scientific, technical and academic skills
through several projects.
Moreover, 
human values like elegance and generosity that I found in Prof. Cuzzolin
have inspired me to work in the Visual Artificial Intelligence Lab,
School of Engineering, Computing and Mathematics, Oxford Brookes University.
Furthermore, I am grateful to Professor Philip H. S. Torr for allowing me to work in a close
collaboration with the world-renowned Torr Vision Group (TVG), the Department of Engineering Science, University of Oxford. 

I would also like to thank my co-author Dr. Michael Sapienza for his time and effort 
during our long technical discussions.
A special thanks to my lab-mate and co-author Suman Saha for his close 
collaboration and daily discussions which led to several exciting works and many publications.


I want to thank: Tjeerd Olde Scheper for his invaluable guidance and support as research tutor;
Jill Organ, Catherine Joyejob, Lynn Farrell who helped me in all administrative tasks;
Gerald Roy for excellent technical, software and hardware support;
students at Brookes with whom I had a chance to work, especially Stephen Akrigg, Valentina Fontana, Kurt Degiorgio, Manuele Di Maio, Shantanu Rathod.
I am grateful to all my colleagues and friends 
I met over the past few years: Mohamed Idries, Bedour Alshaigy, Mireya Munoz Balbontin, Alla Vovk, Will Guest, Jalawi Alshuduki, Cristian Roman
at the TDE department,
School of Engineering, 
Computing and Mathematics,
Oxford Brookes University.

My special thanks go to Prof. Leonid Sigal for offering me an internship at Disney Research Pittsburgh (DRP), USA. 
Many thanks to Leonid and Andreas Lehrmann for their invaluable time and guidance during those my stay at DRP lab.
I would also like to thank Prof. Greg Mori and Prof. Leonid Sigal for providing me with the opportunity to do another internship at BorealisAI, Canada.
I feel fortunate to have had the chance to work in a highly stimulating environment at DRP and BorealisAI.
Special thanks go to my DRP and BorealisAI colleagues and friends for their help and support
Andreas Lehrmann, Rajitha Navarantha, Ziad Al-Halah, Hongsuck Seo, Judith Butepage, He Jiawei, Suhail Mohamad, Thibaut Durand, Mengyao Zhai, Lei Chen, Gabrial, Lilli meng, Huyen Mori.

Due acknowledgement goes to my MSc supervisors Prof. Radu Horaud and Dr. Georgios Evangelidis
for their time, advice and support during my MSc dissertation, 
and for providing me the opportunity to work as a research intern 
at INRIA, Grenoble.
My acknowledgement would not be complete without mentioning a few names of my colleagues 
I worked with for the past few years at Siemens, Bangalore, India.
I am thankful to my Siemens colleagues for teaching me so many things:
Parmeet Bhatia, Prabhu Teja, Yogesh, Todd Wegner.
My heartfelt thanks go to Dr. Amit Kale for his guidance throughout my time at Siemens.

I would also like to thank my friends, who have greatly supported me for many years, Rajjan Singh Thakur, Vijendra Kumar, Abhishek Kumar, Rajesh Ranjan, Prince Sharma, 
Mohamed Idries, Noemi Dreksler, Mireya Munoz Balbontin, Bedour Alshaigy, Rishab Mehta, Rajvee Mehta.

Last but not least,
I am thankful for my \emph{parents} and my \emph{family} in India for their unending patience, encouragement and generosity.
Especially, my parent for fighting for my life during the initial ten years of my life and those countless hospital visits; 
my sister \emph{Mandeep Kaur} for homeschooling me during those hard times.
Without their perseverance, I would not be writing this today.
My heartfelt thanks go to my wife \emph{Meenakshi} for her unbounded patience and pushing me to be my best self. 
She has been a rock in my life during hard times of late night paper submissions. 
She enjoyed my success as her own, she was always there. I can not thank her enough for being so supportive during this journey.

%% file: pads/abstract.tex
\newpage
\cmtx{Research focus, methods used, results/findings, main conclusions and recommendations.}
\heading{Abstract}
\cmtx{Research focus}
In this thesis, we focus on video action understanding problems from an online and real-time processing point of view. 
We start with the conversion of the traditional offline spatiotemporal action detection pipeline into an online spatiotemporal action tube detection system. 
An action tube is a set of bounding connected over time, which bounds an action instance in space and time.
Next, we explore the future prediction capabilities of such detection methods by extending the an existing action tube into the future by regression.
Later, we seek to establish that online/causal representations can achieve similar performance to that of offline three dimensional (3D) convolutional neural networks (CNNs) on various tasks, 
including action recognition, temporal action segmentation and early prediction.

To this end, we propose various action tube detection approaches from either single or multiple frames. We start by introducing supervised action proposals for frame-level action detection and solving two energy optimisation formulations to detect the spatial and temporal boundaries of action tubes.
Further, we propose an incremental tube construction algorithm to handle the online action detection problem. There, the real-time capabilities are made possible by introducing real-time frame-level action detection and real-time optical flow in the action detection pipeline for efficiency.
Next, we extend our frame-level approach to multiple frames with the help of a novel proposal to for predicting flexible action 'micro-tubes' from a pair of frames.
We extend the micro-tube prediction network in order to regress the future of each micro-tube, which is then fed to our proposed future action tube prediction framework.
We convert 3D CNNs to causal 3D CNNs by replacing every 3D convolution with recurrent convolution, 
and by making use of sophisticated initialisation to handle the problems of recurrent modules.

We show that our action tube detectors perform better than previous state-of-the-art methods, while exhibiting online and real-time capabilities. 
We evaluate each action tube detector and predictor on publicly available benchmarks to show the comparison with other state-of-the-art approaches.
We also show that our flexible micro-tube proposals not only improve action detection performance but can also handle sparse annotations.
Finally, we demonstrate the causal capabilities of our causal 3D CNN.

%% file: pads/nomenclature.tex
\Nomenclature[A]{1D: }{One Dimensional}
\Nomenclature[A]{2D: }{Two Dimensional (spatially for image)}
\Nomenclature[A]{3D: }{Three Dimensional (spatiotemporal for video)}
\Nomenclature[A]{BoVW: }{Bag-of-Visual-Words.}
\Nomenclature[A]{DPM: }{Deformable Part-based Model.}
\Nomenclature[A]{CNN: }{Convolutional Neural Network.}
\Nomenclature[A]{R-CNN: }{Region-based CNN.}
\Nomenclature[A]{RNN: }{Recurrent Neural Network.}
\Nomenclature[A]{LSTM: }{Long Short-Term Memory recurrent neural network.}
\Nomenclature[A]{FPN: }{Feature Pyramid Network.}
\Nomenclature[A]{CONV: }{Convolution.}
\Nomenclature[A]{IoU: }{Intersection over Union.}
\Nomenclature[A]{RCU: }{Recurrent Convolutional Unit} 
\Nomenclature[A]{T-IoU: }{Temporal Intersection over Union.}
\Nomenclature[A]{ST-IoU: }{Spatiotemporal Intersection over Union.}
\Nomenclature[A]{ST: }{Spatiotemporal.}
\Nomenclature[A]{FC (or fc) layer: }{Fully connected layer.}
\Nomenclature[A]{Cls layer: }{Classification layer.}
\Nomenclature[A]{Reg layer: }{ Regression layer.}
\Nomenclature[A]{SVM: }{ Support Vector Machine.}
\Nomenclature[A]{HMM: }{ Hidden Markov Models.}
\Nomenclature[A]{SS: }{ Selective Search.}
\Nomenclature[A]{AMTnet: }{ Action Micro Tube Regression network.}
\Nomenclature[A]{SSD: }{ Single Shot Detector.}
\Nomenclature[A]{NL-N: }{ Non-Local Network.}
\Nomenclature[A]{TraMNet: }{ Transition Matrix Network.}
\Nomenclature[A]{TPNet: }{ Tube Predictor Network.}
\Nomenclature[A]{RCN: }{ Recurrent Convolutional Network.}
\Nomenclature[A]{ROI: }{ Region of Interest.}
\Nomenclature[A]{RPN: }{ Region Proposal Network.}
\Nomenclature[A]{NMS: }{ Non-maximal Suppression.}
\Nomenclature[A]{2PDP: }{ Two Pass Dynamic Programming.}
\Nomenclature[A]{MBH: }{ Motion Boundary Histograms.}
\Nomenclature[A]{FV: }{ Fisher Vectors.}
\Nomenclature[A]{C3D: }{ Convolutional 3D Network.}
\Nomenclature[A]{I3D: }{ Inflated Convolutional 3D Network.}
\Nomenclature[A]{S3D: }{ Separated Convolutional 3D Network.}
\Nomenclature[A]{S3Dg: }{ Separated Convolutional 3D gated Network.}
\Nomenclature[A]{ReLU: }{ Rectified Linear Unit.}
\Nomenclature[A]{BN: }{ Batch Normalisation.}
\Nomenclature[A]{AUC: }{ Area Under the Curve.}
\Nomenclature[A]{mAP: }{ mean Average Precision.}
\Nomenclature[A]{RGB: }{ Red Green Blue, 3-Channel Camera output.}

%% file: chapters/1_intro.tex
\chapter{Introduction}
\label{chapter:intro}
\renewcommand{\imagepath}{/home/gurkirt/Saha_PhD_Thesis_Latex/imgs/introduction}

A video is a set of natural images ordered according to the time they were captured by a camera device. Videos can play the role of modern-day journals, personal journals, record books, 
novels, radio for sports, and letters to friends, for they can store a vast variety of information in a single format. The recording of videos is becoming increasingly simple and cheap by readily available mobile devices. There are various video libraries available these days, e.g. YouTube\footnote{www.youtube.com}, Netflix\footnote{www.netflix.com}. Recordings from millions of closed-circuit cameras (CCTV) also exist. Almost 70\% of the world internet traffic is video content. 
Because of their multi-modal nature, it is difficult to interpret videos automatically. 
This makes video content understanding a major problem in artificial intelligence (AI).

\section{Understanding video content}
Video understanding is a difficult task because it inherits all the complexity of the video capturing process and because of its multi-modal (video, audio, subtitles, commentary) nature.
Sometime these modalities could be very useful complementary signals to visual features, e.g. audio in the professional broadcasting of sports videos could be very useful. The script of a movie may help us understand its content, along with the video.
Most of the time, however, these complementary signals are either not available or confusing, e.g. music playing in a sports video.
In the purview of this thesis, we will only concern ourselves with the visual content understanding of videos. 


Video understanding entails multiple problems. Here, we itemise a few of the most visited problems by the research community.
\begin{itemize}
 \item Human action/activity understanding
 \item Video retrieval
 \item Video compression
 \item Object tracking
 \item Optical flow computation
 \item Video content description
 \item Semantic video segmentation
\end{itemize}

The above list is by no means exhaustive, but it gives us an idea of the  different tasks associated with video understanding. These problems pose multiple different types of challenges to a researcher. For instance, keeping track of moving components is necessary for object tracking, optical flow estimation and video compression. Content identification is equally necessary for video content description and action understanding. 

\section{Why human actions?}
In this thesis, we focus on human action understanding in particular for two main reasons. Firstly, most of the video content is created around humans and their activities/actions, as in the case of a security guard having to flag the unusual movements of humans. Secondly, action understanding implicitly requires to concurrently solve several of the above-listed problems. Algorithms need needs to understand the content of video if a human is interacting with some objects (which involves implicit object tracking and content recognition), but also need to understand motions (which requires optical flow computation and video compression).

A variety of terms are used to describe human behaviour understanding e.g. actions, events, and activities. We think it is useful to define these terms before moving forward.

\begin{itemize}
 \item An \emph{event} is something that happens, not necessarily because of the will of agent (human/robot/machine); examples are: a person falling, road accident, and raining etc.
 \item An \emph{action} is an event which happens because an agent (typically, a human) wanted it to happen, e.g. walking, running, throwing a ball, playing piano, working on a laptop etc.
 \item An \emph{activity} is an ensemble of atomic actions performed by one or more agents, e.g. protesting on the street, playing basketball, and javelin throwing etc.
\end{itemize}




\section{Types of problems in human action understanding}\label{intro:sec:problemtypes}

Human action understanding can be classified into various subcategories. 
Next, we will define some of the tasks involved in human action understanding and link them with some examples of the datasets used to study them.

\begin{enumerate}
 \item \emph{Action recognition} is one of the most fundamental and studied problems in human action understanding. Here, given a video clip (typically of limited duration, such as a few seconds) we need to label it  with one of the categories in an action vocabulary defined by the user or the dataset. Example datasets are UCF101\cite{ucf101}, Sports1M~\cite{karpathy2014large}, Kinetics~\cite{kay2017kinetics}. 
 \item \emph{Temporal action segmentation} consists in labelling each frame in the video with one or more labels from a list of action categories. The task is sometimes called `frame-wise action classification'. Example datasets include Charades~\cite{sigurdsson2016hollywood} and MultiThumos~\cite{yeung2015every}.
 \item \emph{Temporal action detection} is a task in which we need to identify the temporal boundaries (start and end timestamps) of each action instance, together with its label.
 Given a video, we need to detect all action instances with their category labels, start time, and end time. Example datasets include, again, Charades~\cite{activitynet2015heilbron} and THUMOS~\cite{thumos14}.
 \item \emph{Spatiotemporal action detection} indicates the case in which we need to identify the label, temporal boundaries and spatial bounding box in each frame associated with an action instance. Each action instance is called an \emph{action tube} in the literature~\cite{georgia2015tube,singh2017online}. 
 An action tube is a set of bounding boxes linked across time. Some example datasets includes UCF101-24~\cite{ucf101}, J-HMDB-21~\cite{jhmdbj2013towards}, AVA~\cite{ava2017gu}.
 \item \emph{Spatiotemporal action segmentation} is similar to action tube detection, but also includes the pixel-level segmentation of the human involved in each action instance. 
 J-HMDB-21~\cite{jhmdbj2013towards} is an example of a suitable dataset.
 \item \emph{Early action label prediction} is the problem of predicting the action label of a clip as early as possible, after observing a few frames from the initial part of the video.
 This problem can be considered as a more advanced version of the recognition problem, for
 here we recognise the action category by observing as few frames as possible. Most of the recognition and temporal segmentation datasets can be used to evaluate such a task.
 \item \emph{Causal/Online spatiotemporal action detection} is the online version of the spatiotemporal action detection problem, in which we detect action instances in the observed part of the video. Spatiotemporal reasoning is limited to the frames observed up to that point. Spatiotemporal action detection datasets are used to evaluate these tasks as a function of video observation percentage.
 \item \emph{Future action prediction} has many interpretations. Some of them include next action label prediction, i.e., what will be the action label a certain amount of time in future, when will the next action start, and so on. We are seeing a surge in methods designed to address these kinds of problems -- we will discuss them in more detail in the next chapter, in the related work section. Recognition and detection datasets can be modified to validate these tasks.
 \item \emph{Future action tube prediction} is the problem of predicting the future bounding boxes and label of an action tube by only observing the current and past video frames. Action tube detection datasets can be employed to evaluate solutions to this problem.
\end{enumerate}

In this thesis, we will touch upon most of the above problems, while mainly focusing on spatiotemporal action detection. We will start with our proposed solution to the spatiotemporal detection problem. We will then tackle the problem of solving action detection in an online and real-time fashion. We will consider multiple approaches to the online action detection problem and its extension to early label prediction and future tube prediction. Finally, towards the end of the thesis, we will present a spatiotemporal representation which is online/causal in nature.

\section{Why online action tube detection}~\label{intro:sec:whyOnlineTubeDetection}

We argue that the problem of detecting action tubes in space and time in untrimmed videos is a rather comprehensive human action understanding problem. 
It encompasses problems such as action recognition, temporal action detection, the spatial detection and tracking of each actor. 
All these tasks require a complex understanding of the video content, as well as of its context, for context is important to understand what humans are doing: for instance, basketball and tennis are sports played in specific types of courts. 

\begin{figure}[ht!]
 \centering
 \includegraphics[width=0.96\textwidth]{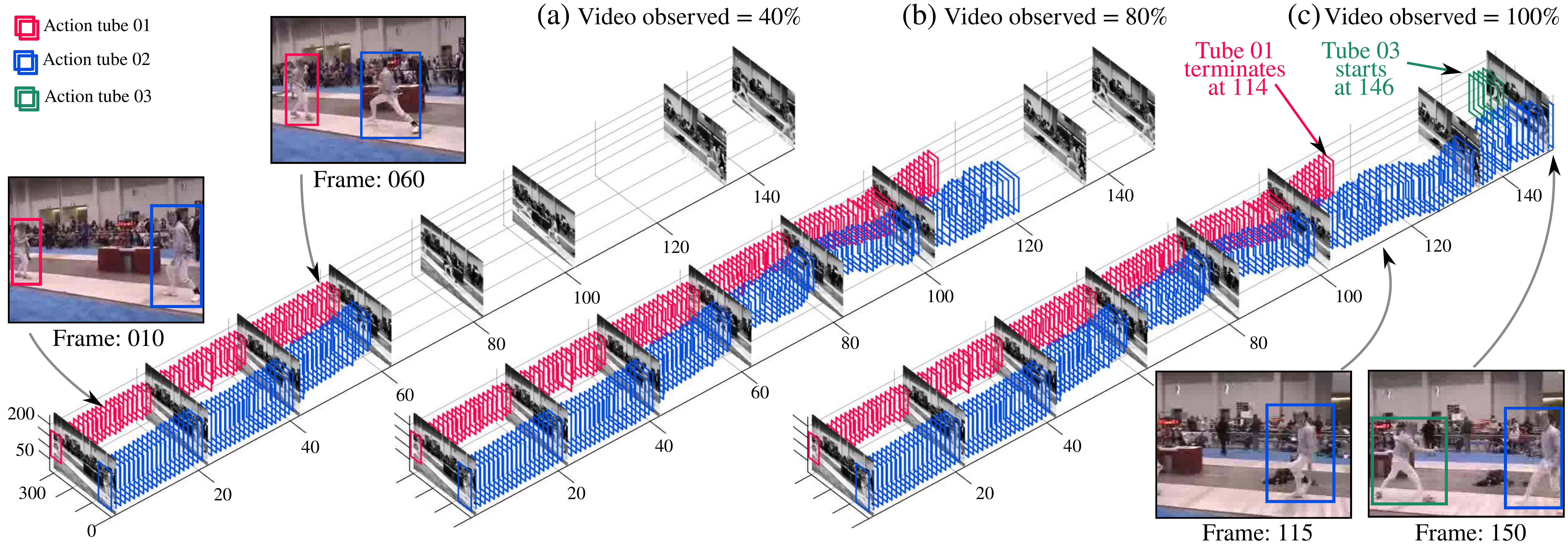}
 \caption[Visualisation of online spatiotemporal action detection problem in a test video.]{
 Online spatio-temporal action detection 
 and early action label prediction 
 in a test `fencing' video from UCF-101-24~\cite{ucf101}.
 \textbf{(a)} to \textbf{(c)}: A 3D volumetric view of the video showing detection boxes and selected frames.
 At any given time, a certain portion (\%) of the entire video is observed by the system, 
 and the detection boxes are linked up to incrementally build space-time action tubes.
 }
 \label{intro:fig:introTeaser}
\end{figure} 

Further, we consider action-tube detection in an online mode (or `causal' setting), where we need to process the video as it comes, based on its present and past observations, as shown in Figure~\ref{intro:fig:introTeaser}.
Such a style of processing is crucial for action detection system to be deployed in various real-world scenarios, including for instance action detection of other road agents for self-driving cars, human-robot interaction, surgical robotics.
Moreover, if we are able to detect actions in an online fashion then it becomes relatively easier to extend the same method to the future action prediction task --
we will show such an example in Chapter~\ref{chapter:tpnet}.

\begin{figure}[ht!]
 \centering
 \includegraphics[width=0.96\textwidth]{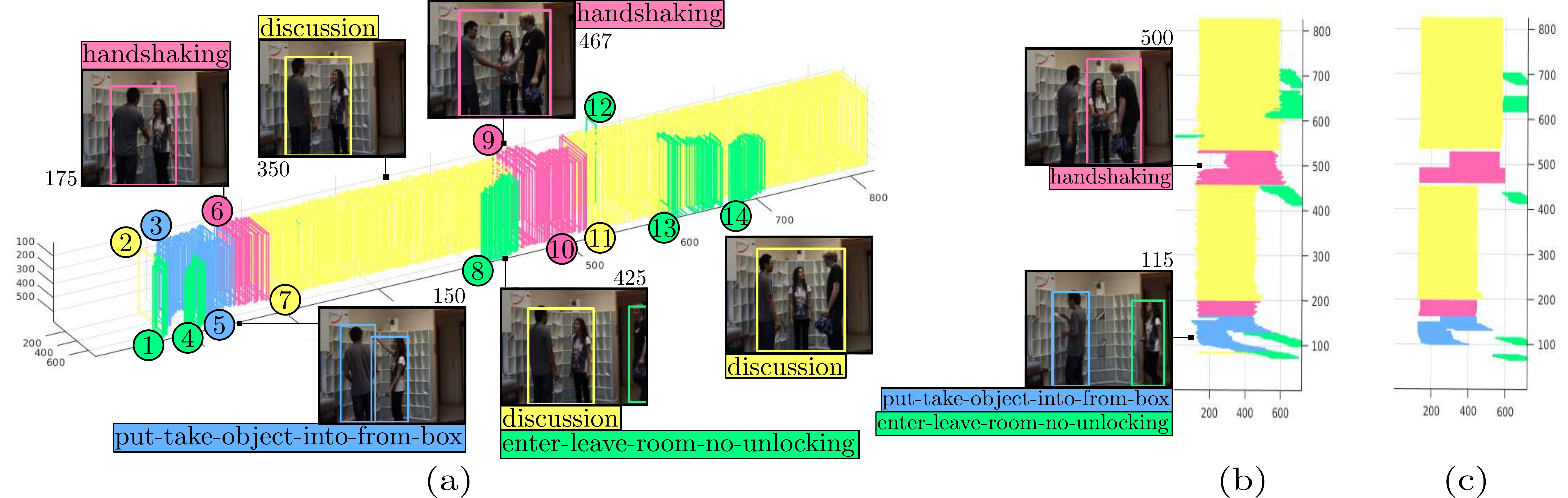}
 \caption[Visualisation co-occurring spatiotemporal action detection problem in a test video.]{
 Co-occurring spatio-temporal action detection 
 in a test video from LIRIS-HARL~\cite{liris-harl-2012}.
 \textbf{(a)} A 3D volumetric view of the video showing detection boxes and selected frames with their labels.
 Two people enter a room and put/take an object from a box (frame 150).
    They then shake hands (frame 175) and start having a discussion (frame 350).
    In frame 450, another person enters the room, shakes hands, and then joins the discussion.
    Each action tube instance is numbered and coloured according to its action category.
    \textbf{(b)}~Action tubes drawn as viewed from above, compared to \textbf{(c)}~the ground truth action tubes. 
 }
 \label{intro:fig:introTeaserCooccur}
\end{figure} 

It is important to note that the action tube detection problem involves the detection of \emph{co-occurring} action instances. These action instances could belong to the same action class (see, Figure~\ref{intro:fig:introTeaser}) or to different action classes (see frame 115 or 425 in Figure~\ref{intro:fig:introTeaserCooccur}). One class of action/interaction could turn into a different class (see instances number 6 and 7 in Figure~\ref{intro:fig:introTeaserCooccur}).

We argue that an instance-based solution to the action detection problem may provide a better understanding of a video's content, and of the complex interactions and activities taking place there.
Moreover, we strive to address this problem in a causal/online and real-time fashion, in order for any solution to be applicable to real-world problems.

\section{Contributions of the thesis}~\label{intro:sec:whatsnew}

In this thesis, we describe the first online and real-time action detection system, able to detect multiple co-occurring actions in untrimmed videos, unlike previous works~\cite{weinzaepfel2015learning,georgia2015tube} which were offline in nature. 
We establish that supervised proposals are important for action detection~\cite{saha2016deep}, and that temporal localisation can be formulated as an efficient dynamic programming-based solution, in opposition to earlier sliding window approaches. 
Next, we show how we can extend the approach to action detection based on individually processing each video frame to methods that can process multiple frames at the same time. In particular, we introduce a new way of generating cross-frame flexible anchor proposals, and compare it against the cuboid anchor proposals in~\cite{kalogeiton2017action,hou2017tube,saha2016deep}.
Also, for the first time, we introduce the future action tube prediction problem as an extension of online action tube detection problem. 
Finally, we introduce an online 3D spatiotemporal video representation based on a causal 3D CNN architecture which, while being causal, is competitive with the previous a-causal 3D representations~\cite{carreira2017quo,xie2018rethinking,tran2018closer}.

In this thesis, each contribution is separately presented in each chapter in the context of the previous related work at the end of each chapter. 
We run a comparison with the relevant previous state of the art in a ``related work'' chapter (Chapter~\ref{chapter:related_work}).
Finally, we summarise these contributions again in Chapter~\ref{chapter:conclusion}, in which the contributions of each chapter are consolidated in Section~\ref{chapter:conclusion:sec:summary_of_contribs}. 

\section{Overview of the thesis}

First and foremost, all related work is reviewed in chapter (Chapter~\ref{chapter:related_work}).
The five main contribution chapters of this thesis follow. These chapters are grouped into three parts, Part I, Part II and Part III. Finally, our contributions and possible future extensions are summarised in Chapter~\ref{chapter:conclusion}.

Part I~\ref{partI} includes two chapters (Chapter~\ref{chapter:offline}, and Chapter~\ref{chapter:online}). The main theme of Part I is online and real-time action detection.
Chapter~\ref{chapter:offline} describes an offline action detection approach based on previous works~\cite{saha2016deep,singh2016untrimmed}.
Here, we lay the groundwork for the next chapters, and a possible extension towards an online and real-time approach.
We show that supervised proposals are important for action detection~\cite{saha2016deep}, and temporal localisation can be efficiently formulated in a dynamic programming setting, in contrast with the earlier sliding window approaches. 
\\
In Chapter~\ref{chapter:online} we present what was, at the time of publication, the first online and real-time action detection system, based on our published work~\cite{singh2017online}.
We bring forward a host of changes to make the action detection system online and real-time, while being extremely competitive with other, offline systems in terms of performance.
In particular, we: (i) propose an online tube generation algorithm; (ii) exploit the real-time network architecture of single-shot-detector (SSD)~\cite{liu2016ssd}; (iii) ablate the replacement of expensive optical flow calculation approaches~\cite{brox2004high} with real-time optical flow method~\cite{kroeger2016fast}.

Part II~\ref{partII} also includes two chapters (Chapter~\ref{chapter:tramnet}, and Chapter~\ref{chapter:tpnet}).
Its main theme is the notion of action `micro-tube' for action detection and future prediction.
Chapter~\ref{chapter:tramnet} presents two approaches for extending single frame-based action detectors to multi-frame action detectors, based on two other published works of ours~\cite{saha2017amtnet,singh2018tramnet}.
The first approach (based on~\cite{saha2016deep}) involves generalising a frame-based object/action detector to a multi-frame detector by extending conventional frame-level anchor proposals in the temporal direction to generate anchor-cuboids. The main idea is that predictions made in reference to an anchor cuboid are considered linked (to form micro-tubes, as pair of bounding box detections with an attached label). This provides us with pairs of boxes which are inherently linked across frames, a step towards doing away with the temporal linking in a post-processing stage. We will call such a detector an \emph{action micro-tube network} (AMTNet).
The next approach introduces flexible anchor proposals across frames, which allow the anchor in one frame to shift to another location in the next frame, in order to handle ``dynamic'' actions.
This is done by learning a transition probability matrix from the training data in a hidden Markov model formulation, leading to an original configurable layer architecture.
We call such an action detector a \emph{transition matrix network} (TraMNet) ~\cite{singh2018tramnet}.
\\
Building on these results, 
Chapter~\ref{chapter:tpnet} shows that a micro-tube prediction network can be extended to predict the future of each action tube individually. 
We formulate this as a multi-tasking problem. A new task head is added to the micro-tube prediction architecture to predict the future locations of each micro-tube. The resulting network is called a \emph{tube predictor network} (TPNet)~\cite{singh2018predicting}. Further, we propose a tube prediction framework which completes the future of each tube by taking the outputs of the individual micro-tubes.
We show that TPNet can cope with the intrinsic uncertainty about the future better than the considered baseline methods, while remaining state-of-the-art in the action detection task.

Lastly, we have a single chapter (Chapter~\ref{chapter:rcn}) in Part III,
where we take steps to transform~\sota~three dimensional (3D) convolutional neural networks (CNNs) to causal 3D CNNs, in line with the main objective of this PhD work, which is online action detection.
The fundamental observation is that 3D CNNs are anti-causal (i.e., they exploit information from both the past and the future to produce feature representations, thus preventing their use in online settings). They also constrain the temporal reasoning horizon to the size of the temporal convolution kernel, and are not temporal resolution-preserving for video sequence-to-sequence modelling, as, e.g., in action detection. To address these serious limitations, in Chapter~\ref{chapter:rcn} we present a new architecture for the causal/online spatiotemporal representation of videos. Namely, we propose a novel \emph{recurrent convolutional network} (RCN)~\cite{singh2019recurrence} which relies on recurrence to capture the temporal context across frames at every level of network depth. We show in our experiments on the large-scale large ``Kinetics'' and ``MultiThumos'' datasets that the proposed method achieves superior performance even when compared with anti-causal 3D CNNs while being causal and using fewer parameters.

\section{Structure of the chapters}

All the main contribution chapters (Chapter~\ref{chapter:offline} to Chapter~\ref{chapter:rcn}) follows the same overall structure.
Firstly, we have an `introduction' section, where we present motivation in terms of applications and previous works. It also contains an `outline' of that particular chapter.
Next, we have `overview of the approach' section, where we highlight the main steps included in our approach. This section also includes a subsection for `resulting contributions' from our approach.
Following that, various sections describe the approach. 
An experiment section follows in which we evaluate and ablate our approach in the context of the claims made in the approach and introduction sections.
Finally, each chapter has a `summary and limitations' section, which includes a short summary of the chapter, the limitations of the presented approach, and a paragraph where we look ahead to the next chapter in anticipation.

\section{Resulting publications and resources}
\subsection{List of publications}

\paragraph{Part of thesis:} Following is the list of publications that make the major part of this PhD thesis.

\begin{enumerate}
 \item \textbf{Gurkirt Singh} and Fabio Cuzzolin, 
 ``Recurrent Convolutions for Causal 3D CNNs'', 
 in Proceedings of International Conference on Computer Vision Workshop (ICCVW) on Large Scale Holistic Video Understanding, 2019.
 
 \item \textbf{Gurkirt Singh}, Suman Saha and Fabio Cuzzolin, 
 ``TraMNet - Transition Matrix Network for Efficient Action Tube Proposals'', 
 in Proceedings of Asian Conference on Computer Vision (ACCV), 2018.

 \item \textbf{Gurkirt Singh}, Suman Saha and Fabio Cuzzolin,
 ``Predicting Action Tubes'', in Proceedings of European Conference on Computer Vision Workshops (ECCVW) on Anticipating Human Behaviour Workshop, 2018.
 
 \item \textbf{Gurkirt Singh}, Suman Saha, and Fabio Cuzzolin, 
 ``Online Real-time Multiple Spatio-temporal Action Localisation and Prediction'', 
 in proceedings of International Conference on Computer Vision (ICCV) 2017.
 
 \item Suman Saha, \textbf{Gurkirt Singh} and Fabio Cuzzolin,
 ``AMTnet: Action-Micro-Tube Regression by end-to-end Trainable Deep Architecture'', 
 in proceedings of International Conference on Computer Vision (ICCV) 2017.

 \item Suman Saha, \textbf{Gurkirt Singh}, Michael Sapienza, Philip Torr and Fabio Cuzzolin,
 Deep Learning for Detecting Multiple Space-time Action Tubes in Videos,
 proceedings of British Machine Vision Conference (BMVC) 2016.

 \item \textbf{Gurkirt Singh} and Fabio Cuzzolin,
 ``Untrimmed Video Classification for Activity Detection: Submission to ActivityNet Challenge'', 
 in Tech-report arXiv:1607.01979, presented at ActivityNet Challenge Workshop in CVPR 2016.

\end{enumerate}

\paragraph{Other works:} Below is the list of other works which were conducted during PhD but these are not part of this thesis.

\begin{enumerate}
\item Harkirat Behl, Michael Sapienza \textbf{Gurkirt Singh}, Suman Saha, Fabio Cuzzolin and Philip Torr, ``Incremental Tube Construction for Human Action Detection'', British Machine Vision Conference (BMVC), 2018. 
\item \textbf{Gurkirt Singh}${}^*$, Stephen Akrigg${}^*$, Valentina Fontana${}^*$, Manuele Di Maio, Suman Saha, Fabio Cuzzolin,``Action Detection from a Robot-Car Perspective'', Preprint arXiv: 1807.11332, 2018.
\item Silvio Olivastri, \textbf{Gurkirt Singh} and Fabio Cuzzolin,``An End-to-End Baseline for Video Captioning'', in Proceedings of International Conference on Computer Vision Workshop (ICCVW) on Large Scale Holistic Video Understanding, 2019.
\item Suman Saha, \textbf{Gurkirt Singh}, Michael Sapienza, Philip Torr and Fabio Cuzzolin,``Spatio-temporal Human Action Localisation and Instance Segmentation in Temporally Untrimmed Videos'', arXiv preprint arXiv:1707.07213, 2017. 

\end{enumerate}

\subsection{List of software packages and other resources}
 The following software packages from this thesis are available online.
 \begin{itemize}
 
 \item Source code for our ICCV 2017 ~\cite{singh2017online} work is publicly available online at: \\
 \url{https://github.com/gurkirt/realtime-action-detection}.\\
 CNN training and evaluation code are developed using Pytorch and Python. The tube construction algorithm is available in both Matlab and Python. 
 
 \item ICCV 2017 work~\cite{singh2017online} - YouTube demo video link\\
 \url{https://www.youtube.com/watch?v=e6r_39ETe-g}.

 \item Annotation of UCF101-24~\cite{ucf101} were corrected during for our ICCV 2017 ~\cite{singh2017online} work.
 The corrected annotations at available at \url{https://github.com/gurkirt/corrected-UCF101-Annots}.
 
 \item Source code for our ActivityNet challenge submission during CVPR 2016~\cite{singh2016untrimmed} work is publicly available online at \\
 \url{https://github.com/gurkirt/actNet-inAct}.\\
 The source code is developed in Python. 

 \item Source code for our BMVC 2016 ~\cite{saha2016deep} work is publicly available online at: \\
 \url{https://bitbucket.org/sahasuman/bmvc2016_code}.\\
 Source code developed using MatCaffe (the Matlab wrapper for Caffe deep learning toolbox). 
 
 \item BMVC 2016 work~\cite{saha2016deep} - YouTube demo video link\\
 \url{https://youtu.be/vBZsTgjhWaQ}.
 \end{itemize}

%% file: chapters/2_related_work.tex
\chapter{Related Work}
\label{chapter:related_work}

In this chapter, we present a literature review from various perspectives concerning the problems and techniques mentioned in the Introduction. 
We review the work most related to the concepts and contributions of this thesis, as described in the remainder of the thesis.
\\
The terms ``previous work'' and ``recent work'' are used in the context of the work presented in this thesis -- for instance,  
the paper~\cite{caba2015activitynet} on action detection was published before our relevant publication~\cite{singh2016untrimmed}, and is therefore
considered `previous' work. However, \cite{rene2017temporal} appeared after~\cite{singh2016untrimmed}, therefore it is consider `recent' work.

We first briefly review the most prominent work in action recognition in Section~\ref{related_work:sec:actionrecognition}, as the most basic video clip understanding problem.
We then outline the recent advances in temporal action detection. 
Finally, we review the state of the art in spatiotemporal action detection.

\section{Action recognition}

\emph{Action recognition}, also called \emph{video classification}, is most widely studied of all video action understanding problems (\S~\ref{intro:sec:problemtypes}).

\subsection{Traditional approaches}
\label{related_work:sec:actionrecognition}

Action recognition has been studied for a long time\cite{bobick2001recognition,zhang2008motion,laptev2005space}. 
Seminal work by I. Laptev~\cite{laptev2005space} provided a solid platform for subsequent important papers~\cite{laptev2007retrieving,laptev2008learning} in action understanding.
As a result, image-based feature descriptors (such as HOG \cite{dalal2005histograms} and SIFT \cite{lowe-2004})) were adopted or suitably extended to process videos, 
e.g. as in motion boundary histograms (MBH)~\cite{dalal2006human}. 
3D descriptors~\cite{klaser2008spatio} and dense trajectory features~\etal~\cite{wang2011densetraj,wang2013improvedtraj} were later proposed, which
would exploit Bag-of-Visual-Words (BoVW) \cite{csurka2004visual} or Fisher vector (FV) \cite{perronnin2007fisher} representations.

\subsection{2D CNNs}
\label{related_work:subsec:2DCNNs}

After the advent of deep learning~\cite{schmidhuber2015deep,lecun2015deep,krizhevsky2012imagenet}, 
image-based convolutional neural networks (2D-CNN) were adopted to tackle the action recognition problem. 
Simonyan and Zisserman~\cite{simonyan2014twostream} proposed to train two separate 2D-CNNs on RGB and optical flow images as inputs:
their approach was able to produce comparable to local feature-based approaches~\cite{wang2013improvedtraj}. 
Feichtenhofer~\etal~\cite{feichtenhofer2016convolutional} showed that fusing the two streams can improve the performance further.
Efforts were also made to better capture temporal information with 2D CNNs.
For instance, Donahue used LSTMs~\cite{donahue2015lrcn} 
on top of 2D CNN features. Wang~\etal~\cite{wang2016temporal} proposed instead to train 2D CNNs with segment-level inputs. 
Other approaches include, among others, CNN features in combination with LSTMs~\cite{ma2016learning} for temporal action detection,
2D features used in an encoder-decoder setup along with temporal convolutions \cite{rene2017temporal}, and 
conditional random fields on series of 2D features~\cite{sigurdsson2017asynchronous} for temporal action detection and recognition.
All these methods showed promising results. In all these architectures, however, 
the optical flow stream and the few layers on the top of the 2D features are the only sources of temporal reasoning.

\subsection{3D CNNs}
\label{related_work:subsec:3DCNNs}

Initial attempts to apply three dimensional (3D) CNNs models~\cite{ji20133D,tran2014learning}, 
which promised to be able to perform spatial and temporal reasoning in parallel, met limited success.
Later, Carreira~\etal~\cite{carreira2017quo} improved the existing 3D CNNs by employing ImageNet-based initialisation and by training their models on 
the large-scale Kinetics dataset~\cite{kay2017kinetics}. 
The resulting models outperformed older 2D ones.
In spite of this, 3D CNNs remain heavy and very expensive to train -- e.g., 64 GPUs were used in~\cite{carreira2017quo}.

In alternative, the notion of {factorising 3D convolutional networks} was explored by Sun~\etal~\cite{sun2015human}. 
This inspired~\cite{xie2018rethinking,qiu2017learning,tran2018closer} 
to decompose 3D convolutions into 2D (spatial) and 1D (temporal) convolutions.
Recent work by Xie~\etal~\cite{xie2018rethinking} has further promised to reduce complexity (in terms of number of parameters) while making up for the performance lost via a gating mechanism. 
Tran~\etal~\cite{tran2018closer} would keep the number of parameters equal to that of 3D convolutions, 
but boost performance by increasing the number of kernels in the 2D layer.
The size of the temporal convolution kernel needs to be fixed to a relatively small number (e.g., 3 in ~\cite{carreira2017quo,xie2018rethinking,tran2018closer}). 
Varol~\etal~\cite{varol2018long} have thus proposed the use of long-term convolutions to capture long-range dependencies in the data.
Wang~\etal~\cite{nonlocal2018wang}, instead, have introduced non-local blocks in existing 3D CNN architectures, 
to capture the non-local context in both the spatial and the temporal (present, past and future) dimensions.
Feichtenhofer~\etal~\cite{feichtenhofer2018slowfast} have proposed to combine the information coming from two branches of 3D CNNs which operate at different frame rates to boost the performance of action recognition even further. Similarly, Diba~\etal~\cite{diba2019holistic} have suggested to combine information at multiple stages of a 2D CNN with that of a 3D CNN. 

\section{Temporal action detection}~\label{action_cls:subsec:deep_rep}

In the past, the temporal action detection problem was mostly tackled using expensive sliding window approaches
\cite{laptev2007retrieving,gaidon2013temporal,tian2013spatiotemporal,oneata2014efficient,wang2015action}. 
These can deliver good results~\cite{oneata2014efficient,gaidon2013temporal,tian2013spatiotemporal}, 
but are too inefficient to work in real-time.
Recently, deep learning-based methods have led to significant advances in this area as well.
For instance, Shou~\etal~\cite{shou2016temporal} have employed 3D CNNs~\cite{ji20133D,tran2014learning} to
address temporal action detection in long videos. 
Recurrent neural networks (RNN) ~\cite{elman1990finding} and LSTMs~\cite{hochreiter1997long} are also increasingly being used~\cite{de2016online,singh2016bidirectional,yeungcvpr2016,montes2016temporal} to address the problem, as well as temporal convolutions~\cite{lea2017temporal}.
In past, dynamic programming has been employed to solve the problem efficiently 
\cite{kulkarni2015continuous,evangelidis2014continuous,singh2016untrimmed}.
Some of the above works~\cite{yeung2015every,de2016online,evangelidis2014continuous} can perform action detection in an online fashion.
In our work, specifically in Chapter~\ref{chapter:offline}, we adopt~\cite{evangelidis2014continuous} for the temporal detection of action tubes.

More recently, temporal proposal-based methods~\cite{escorcia2016daps,gao2017turn,zhao2017temporal,xu2017r,buch2017sst,chao2018rethinking,lin2018bsn} have gained much traction and have proven to be very effective~\cite{zhao2017temporal,chao2018rethinking,lin2018bsn}.
The philosophy of Faster R-CNN~\cite{ren2015faster} has been extended to videos to predict the start and end time of action instances in untrimmed videos by~\cite{xu2017r,chao2018rethinking}. 
Zhao~\etal~\cite{zhao2017temporal} would generate proposal by applying a watershed algorithm~\cite{roerdink2000watershed} on the frame-level actionness (presence of action) of video frames.
Similarly, Lin~\etal \cite{lin2018bsn} would generate proposals by predicting a score for the likelihood of each frame to belong to the start or the end of an action instance.

Currently, proposal-based methods~\cite{lin2018bsn,zhao2017temporal} are state-of-the-art in temporal action detection.

\section{Multi-label temporal action segmentation}~\label{action_cls:sec:segmentation}

Although relatively new, the task of dense temporal prediction has been studied by several authors~\cite{shou2016temporal,yeung2015every,rene2017temporal,sigurdsson2017asynchronous,piergiovanni2018learning} in a multi-label setting, i.e., in which each video may contain multiple action labels.
Two major datasets related to this task are Charades~\cite{sigurdsson2016hollywood} and MultiThumos~\cite{yeung2015every}.
Yeung~\etal~\cite{yeung2015every} has recently proposed to this purpose a multiple-output version of LSTMs, named multi-LTSM~\cite{yeung2015every}. 
The authors of \cite{shou2017cdc,rene2017temporal} use temporal deconvolution layers on top of the C3D network~\cite{tran2014learning} to recover the loss of temporal resolution due to the temporal convolutions in the base C3D network~\cite{tran2014learning}.
Piergiovanni~\etal~\cite{piergiovanni2018learning} have proposed an attention mechanism in the form of temporal structure filters that enable the
model to focus on particular sub-intervals of a video. However, their model requires the pre-computation of 3D CNN features for the entire video. 

In this thesis we show an online method able to solve the above problem more efficiently in an online/causal/streaming manner in Chapter~\ref{chapter:rcn}.

\section{Spatiotemporal action detection} 

Spatiotemporal action detection is also a widely studied problem. Later in this thesis
we provide a more extensive overview of the methods related to it.
Here, we first quickly review frame-level action detection methods in Section~\ref{related_work:subsec:singleframe}, to later move on to 
multi-frame methods in Section~\ref{related_work:subsec:multiframe}.
Lastly, we consider methods based on 3D spatiotemporal representations in Section~\ref{related_work:subsec:3d}.

\subsection{Single-frame based methods}~\label{related_work:subsec:singleframe}

Initial proposals for human action detection included human location detection-based approaches such as~\cite{prest2013explicit,lan2011discriminative,tian2013spatiotemporal,klaser2010human,wang2014video}. There, after a human was detected in the video, 3D features~\cite{klaser2008spatio} and dense trajectories~\etal~\cite{wang2011densetraj} were pooled into a feature vector representation~\cite{csurka2004visual,perronnin2007fisher}. 
Weinzaepfel~\etal's work \cite{weinzaepfel2015learning} performed both temporal and
spatial detections by coupling frame-level EdgeBoxes \cite{zitnick2014edge} region proposals with a tracking-by-detection framework. 
In their work, however, temporal trimming was still achieved via a multi-scale sliding window over each track, making the approach inefficient for longer video sequences.
\\
More recently, Saha~\etal~\cite{saha2016deep} and Peng~\etal \cite{peng2016eccv} made use of supervised region proposal networks (RPNs) \cite{ren2015faster} 
to generate region proposals for actions at the frame level, and solved the space-time association problem 
via 2 recursive passes over frame-level detections for the entire video by dynamic programming.
The use of a non-real-time and 2-pass tube generation approach, however, made their methods intrinsically offline and inefficient. 

In this thesis, we extend our previous offline frame-level work~\cite{saha2016deep} to an online and real-time action detection method~\cite{singh2017online} in Chapter~\ref{chapter:online}.
Our framework employs a real-time optical flow (OF) algorithm \cite{kroeger2016fast} and 
a single shot SSD detector \cite{liu2016ssd} to build multiple action tubes in a fully incremental way, and in real-time, unlike previous methods.
After \cite{singh2017online} was published,
several efforts~\cite{saha2017amtnet,kalogeiton2017action,hou2017tube,he2017generic} were directed towards exploiting the spatiotemporal information encoded by multiple frames.
The reason is that, in frame-based methods, the deep network is not in a condition to learn the temporal features essential for accurate action detection, and temporal reasoning is performed by some tube construction stage, in a sub-optimal post-processing step. 

\subsection{Multi-frame based methods}~\label{related_work:subsec:multiframe}

Most early attempts to solve space-time action detection using information coming from multiple frames were based on action cuboid hypotheses 
and sliding-window based approaches~\cite{laptev2007retrieving,cao2010cross,tian2013spatiotemporal,sapienza2014learning,sapienza2014phdthesis}.
Their main limitation was the assumption that an action can be localised using a cuboidal video sub-volume, together with the fact that a sliding window-based approach is very expensive.

Very recently, video-based action representation approaches have appeared~\cite{kalogeiton2017action,hou2017tube,saha2017amtnet,he2017generic} which
address this issue by learning complex non-linear functions (in the form of CNNs) which map video data (instead of image data) to a high dimensional latent feature space.
Kalogeiton~\etal~\cite{kalogeiton2017action} and Hou~\etal~\cite{hou2017tube}'s models map $K$ video frames to such a latent space, 
whereas Saha~\etal~\cite{saha2017amtnet} only require $2$ successive frames to learn spatio-temporal feature embedding.
The main advantage of~\cite{saha2017amtnet} over~\cite{kalogeiton2017action,hou2017tube} is that its framework is not constrained to process a fixed set of $K$ frames, but is
flexible enough to select training frame pairs $(f_{t}, f_{t+\Delta})$ at various intervals by varying the $\Delta$ value according to the requirements of a particular dataset
(i.e., a larger $\Delta$ for longer sequences, or a relatively smaller one for shorter video clips).
As in older approaches~\cite{laptev2007retrieving,cao2010cross,tian2013spatiotemporal}, 
these more recent methods~\cite{kalogeiton2017action,hou2017tube,saha2017amtnet,he2017generic} assume that 3D cuboidal anchors can be used to localise action instances in $K$ frames.
\\
In Chapter~\ref{chapter:tramnet}, we show that such an assumption is unrealistic for ``dynamic'' actions. 
We then present a solution able to generate flexible anchor proposals to counter this problem, based on our work in~\cite{singh2018tramnet}.
We show that cuboidal anchor-based methods are just a special case of our more general solution~\cite{singh2018tramnet}.
\\
More recently, Lin~\cite{li2018recurrent} have proposed a two-stage approach for action detection. Firstly, action tubes are linked using frame-level detections; then, tube classification is performed using a recurrent network. The resulting performance improvement, however, is marginal, and even lower when compared with the base same network (VGG~\cite{simonyan2015very}) used in other works~\cite{kalogeiton2017action,singh2018tramnet}.

\subsection{3D representations for action detection}~\label{related_work:subsec:3d}

In Section~\ref{related_work:subsec:3DCNNs}, we learned that 3D CNNs are state-of-the-art in action recognition. In fact,  
spatiotemporal 3D representations~\cite{carreira2017quo,tran2018closer,xie2018rethinking,nonlocal2018wang,feichtenhofer2018slowfast,wu2019long} 
have recently emerged as a dominant force in action detection as well~\cite{ava2017gu,girdhar2018video,duarte2018videocapsulenet}.
Gu~\etal~\cite{ava2017gu}, for instance, combine an inflated 3D (I3D) network~\cite{carreira2017quo} with 2D proposals as in~\cite{peng2016eccv} 
to exploit the representational power of I3D. A similar notion is proposed in~\cite{xie2018rethinking,girdhar2018video,li2018recurrent}. 
Duarte~\etal~\cite{duarte2018videocapsulenet}, on their part, have proposed an interesting 3D video-capsule network for frame-level spatiotemporal action segmentation.
Wu~\etal~\cite{wu2019long} combine a 3D representation with temporal filter banks to improve action detection performance. 
It is important to note that the above-mentioned methods are not specifically designed for action detection -- most of them (except~\cite{duarte2018videocapsulenet,wu2019long,girdhar2018video}) exhibit superior performance due to the high representational power that comes with a 3D representation.


\section{Online and real-time methods} 

Relatively few efforts have been directed at 
simultaneous real-time action detection and classification.
Zhang~\etal~\cite{zhangcvpr2016}, for example, would accelerate the two-stream CNN architecture of~\cite{simonyan2014twostream}, 
performing action classification at 400 frames per second.
Yu~\etal~\cite{yu2010bmvc} evaluated their real-time continuous action classification approach 
on the relatively simpler KTH~\cite{schuldt2004recognizing} and 
UT-interaction~\cite{ryoo2010ut} datasets. 
The above methods, however, were not able to perform spatiotemporal action detection. 

To the best of our knowledge, the method we propose in Chapter~\ref{chapter:online} is the first to address the spatiotemporal action detection problem in real time.

\section{Early action prediction and detection}~\label{related_work:sec:early}

Prior to 2016 early action prediction was studied by a relatively small number of authors \cite{ryoo2011human,hoai2014max,lan2014hierarchical,ma2016learning,aliakbarian2017encouraging}.
Ryoo \etal \cite{ryoo2011human} used a dynamic bag of words approach to speed up their action prediction pipeline based on handcrafted features.
Hoai \cite{hoai2014max} proposed their max-Margin early event detectors based on structured output SVM. 
In \cite{lan2014hierarchical}, Lan~\etal  proposed a way to capture human movement 
before the action actually happens, at a different level of temporal granularity, and were able to 
make predictions by observing the action for 50\% of its duration.
Different types of loss function were proposed to encourage LSTMs~\cite{hochreiter1997long} to predict the correct label as early possible in~\cite{ma2016learning,aliakbarian2017encouraging}.
None of these approaches, however, perform online spatial-temporal action detection.

More recently,
Soomro~\etal~\cite{soomro2016predicting} have proposed an online method which can predict an action's label and location by observing a relatively smaller portion of the entire video sequence.
However, \cite{soomro2016predicting}'s method works only on temporally trimmed videos and not in real-time, due to the expensive segmentation stage required.
Singh~\etal~\cite{singh2017online}, on the other hand, have brought forward an online action detection and early label prediction approach which works on temporally untrimmed videos.
Similarly, Behl \etal~\cite{behl2017incremental} solve online detection by formulating it as a multi-target tracking problem.
Subsequent works~\cite{kalogeiton2017action,singh2018tramnet,singh2018predicting} have adopted Singh~\etal~\cite{singh2017online}'s online tube generation algorithm for online action detection.
As mentioned, this approach~\cite{singh2017online} is presented in Chapter~\ref{chapter:online}.

\section{Future prediction problems}

The computer vision community is in fact witnessing rising interest in problems such as 
future action label 
prediction~\cite{nazerfard2013using,hoai2014max,aliakbarian2017encouraging,yeung2015every,yeungcvpr2016,kong2017deep,zunino2017predicting,ryoo2011human,lan2014hierarchical,koppula2013learning}, 
online temporal action detection ~\cite{ma2016learning,yeung2015every,de2016online,tahmida2017joint},
online spatio-temporal action detection ~\cite{singh2017online,soomro2016predicting,yang2017spatio}, 
future representation prediction~\cite{vondrick2015anticipating,kong2017deep} or 
trajectory prediction ~\cite{alahi2016social,kitani2012activity,lee2017desire}. 
Although all these problems are interesting, and definitely encompass a broad scope of applications, 
they do not entirely capture the complexity involved by many critical scenarios including, e.g., surgical robotics or autonomous driving. 
In opposition to \cite{singh2017online,soomro2016predicting} which can only perform early label prediction and online action detection, 
we will combine early label prediction, online action detection, and trajectory prediction into one problem in Chapter~\ref{chapter:tpnet}.

\section{Causal representations}

The use of temporal convolutions in all the above methods is, however, inherently anti-causal~\cite{carreira2018massively}, as it requires information from both past and future.
Any online or future prediction method needs to be causal in nature. 
All 2D representations-based approaches~\cite{simonyan2014twostream,singh2017online,feichtenhofer2016convolutional,ma2015hierarchical,singh2016bidirectional} are causal as they do not require future frames to make a prediction about the present.
However, as mentioned, the optical stream and the few layers on the top of the 2D features are the only sources of temporal reasoning. 
Which is why using a 3D representation is highly desirable to improve performance -- unfortunately, all existing such representations are not causal.
Relevantly, Carreira \etal~\cite{carreira2018massively} have recently proposed to address the anti-causal nature of 3D CNNs 
by predicting the future and utilising the flow of information in the network.
They train their causal network to mimic a 3D network -- the resulting performance drop, however, is significant.

In Chapter~\ref{chapter:rcn} we propose a Recurrent Convolutional Network (RCN) designed to solve exactly the above problem.

\section{Related datasets}~\label{soa:datasets}
Action detection and recogntion research is evaluated with various different datasets. 
Here, we describe the properties of the datasets used in this thesis.

\subsection{UCF101-24}~\label{dataset:ucf24}
UCF101-24 is a subset of UCF101~\cite{ucf101}, 
one of the largest and most diversified and challenging action datasets. 
A subset of 24 classes out of 101 comes with spatiotemporal detection annotation, 
released as bounding box annotations of humans 
for the THUMOS-2013 challenge\footnote{http://crcv.ucf.edu/ICCV13-Action-Workshop/download.html}.
Although each video only contains a single action category, 
it may contain multiple action instances (up to 12 in a video) of the same action class, 
with different spatial and temporal boundaries.
On average there are 1.5 action instances per video, 
each action instance covering 70\% of the duration of the video. 
For some classes, instances average duration can be as low as 30\%.
We will use this dataset to evaluate the spatiotemporal action detection problem, similar to previous works \cite{yu2015fast,weinzaepfel2015learning}, we test our method on split 1.

Note that, while conducting our tests, we noted and corrected some issues we found with the original annotation of UCF101-24~\cite{ucf101} released for the THUMOS-2013~\cite{idrees2017thumos} challenge.
The corrected annotation is available on Github\footnote{https://github.com/gurkirt/corrected-UCF101-Annots}.
These annotations were realised together with our ICCV 2017 paper~\cite{singh2017online}.
UCF101-24 dataset is widely used in this thesis to evaluate spatiotemporal action detection performance. 


\subsection{ActivityNet}~\label{dataset:activitynet}
ActivityNet~\cite{activitynet2015heilbron} dataset is a large scale 
temporal activity detection dataset with 200 activity classes and 100 videos per class, on average. In practice, a large proportion of videos have a duration between 5 and 10 minutes~\cite{activitynet2015heilbron}.
The dataset is divided into three disjoint subsets: training, validation and testing, following a ratio of 2:1:1. 
The authors of the dataset organised a challenge in 2016 and we participated with our temporal detection formulation, securing second place. We used ActivityNet dataset to evaluate temporal action detection performance in Chapter~\ref{chapter:offline}.

\subsection{J-HMDB-21}~\label{dataset:jhmdb21}
{J-HMDB-21}~\cite{jhmdbj2013towards} is a subset of the HMDB-51 dataset~\cite{hmdbkuehne2011hmdb} with 21 action categories and 928 videos, each containing a single instance of action tube, and each video trimmed to the action's duration. Hence, it does not require temporal trimming step (\S~\ref{offline:subsec:actionpathtrimming}). As in \cite{georgia2015tube} we pick the top 3 action paths from each class to be evaluated against the ground truth tube in each video. We use J-HMDB-21 dataset in this thesis to evaluate trimmed spatiotemporal action detection performance in various chapters. 

\subsection{LIRIS-HARL}~\label{dataset:lirisharl}
LIRIS-HARL is a human activity detection dataset~\cite{liris-harl-2012} 
with $107$ training and $58$ testing video sequences.
The videos in LIRIS-HARL have a relatively longer duration
than the videos in UCF101-24 and J-HMDB-21 dataset.
Videos contain multiple concurrent action instances from different action classes.
The average duration of action instances is much shorter than the duration of the videos: hence, this dataset is well suited to evaluate the temporal trimming capabilities of our method.
We use LIRIS-HARL dataset in Chapter~\ref{chapter:offline} in this thesis to evaluate the spatiotemporal action detection performance of multiple 
different co-occurring actions.

\subsection{DALY}~\label{dataset:daly}
The \textbf{DALY} dataset was released by Weinzaepfel \etal ~\cite{daly2016weinzaepfel} 
for 10 daily activities and contains 520 videos (200 for test and the rest for training)
with 3.3 million frames. 
Videos in DALY are much longer, and the action duration to video duration ratio is only 4\%, compared to UCF101-24's 70\%,
making the temporal labelling of action tubes very challenging.
The most interesting aspect of this dataset is that it is not densely annotated,
as at max 5 frames are annotated per action instance, and 
12\% of the action instances only have one annotated frame.
As a result, annotated frames are 2.2 seconds apart on average ($\Delta = 59$).
We use this dataset to evaluate our sparse annotation handling models in Chapter~\ref{chapter:tramnet}.

\subsection{Kinetics}~\label{dataset:Kinetics}
The {Kinetics} dataset comprises $400$ classes and $260K$ videos; each video contains a single atomic action. 
Kinetics has become a \emph{de facto} benchmark for recent action recognition works~\cite{carreira2018massively,xie2018rethinking,tran2018closer,carreira2017quo,nonlocal2018wang}. 
The average duration of a video clip in Kinetics is $10$ seconds.
It is used for the action classification task, we will use this dataset in Chapter~\ref{chapter:rcn} to evaluate our causal 3D CNN model.

\subsection{MultiThumos}~\label{dataset:MultiThumos}
The {MultiThumos}~\cite{yeung2015every} dataset is a multilabel extension of THUMOS\cite{jiang2014thumos}. 
It features $65$ classes and $400$ videos, with a total duration of 30 hours. On average, it
provides 1.5 labels per frame, 10.5 action classes per video.
Videos are densely labelled, as opposed to those in THUMOS~\cite{jiang2014thumos} or ActivityNet~\cite{caba2015activitynet}.
MultiThumos allows us to show the dense prediction capabilities of RCN on long, real-world videos.
We use this dataset in Chapter~\ref{chapter:rcn} to evaluate our causal 3D CNN model for the task of action segmentation.

\section{Evaluation metrics}~\label{soa:metrics}
In this thesis, we use standard evaluation metrics following the previous works or variation of these metric. 
Next, we describe the evaluation metrics used in this thesis. 

\subsection{Classification accuracy}
The classification accuracy is used to evaluate tasks of action recognition or prediction. 
Given an observation video clip or frames, we predict one class label with the highest score. 
The classification accuracy measures what percentage of our predictions are correct of all the test videos.
We use classification accuracy as a metric for Kinetics dataset in Chapter~\ref{chapter:rcn}.
Also, we use it in other chapters for UCF101-24 dataset for the tasks of action recognition, early label prediction, and future label prediction.

\subsection{Mean average precision (mAP)}
Similar to other detection tasks (object detection and human detection),
we use mean average-precision(mAP) as the main evaluation metric for both temporal or spatiotemporal detection tasks.
Average precision is the average of the maximum precision at different recall values.
The recall for a ground truth instance is measured against an overlap threshold.
In our case, the overlap is measured as the intersection-over-union (IoU) between ground truth instances and detections produced by the detection framework. 
Temporal-intersection-over-union (T-IoU) in temporal detection is defined over the temporal extent. In the case of spatial detection, instead, we measure the spatial IoU between ground truth bounding boxes and detected bounding boxes in a detected instance over the temporal duration of an action instance. We then take the average of the spatial IoUs to obtain an averaged spatial-IoU (aS-IoU) measure.
Finally, we multiply S-IoU and TIoU to get a measure of spatiotemporal-intersection-over-union (ST-IoU).
If a detected tube has the same label as the ground truth, and ST-IoU is greater than the chosen threshold ($\delta$), then it is classified as true-positive; otherwise, it is considered a false positive. 
All the true positives and false positives are then sorted in decreasing ordered based on the score of each detection to compute the average precision for each class.
The difficulty of detection increases with any increase in the detection threshold ($\delta$) -- a value of $0.5$ (i.e., a $50\%$ overlap with ground truth) is standard. 

\subsection{Completion-mAP}
Completion-mAP is similar to mAP, but here tubes are generated by only observed part of the video.
The entire tube predicted (by observing only a small portion (\%) of the video) is compared against the ground truth tube for the whole video.
Based on the detection threshold we can compute mean-average-precision for the complete tubes, we call this metric \emph{completion-mAP} (c-mAP). We will use this metric in Chapter~\ref{chapter:tpnet}.

\subsection{Prediction-mAP}
Similar to c-mAP, prediction-mAP is used to evaluate the prediction made about the future of the action tubes.
We measure how well the future predicted part of the tube localises.
In this measure, we compare the predicted tube with the corresponding ground truth future tube segment. 
Given the ground truth and the predicted future tubes, we can compute the mean-average precision for the predicted tubes, we call this metric \emph{prediction-mAP} (p-mAP). We will use this metric in Chapter~\ref{chapter:tpnet}.

%% file: pads/partI.tex
\cleardoublepage
\phantomsection
\newcommand{\pname}{Part I : Towards Online Action Detection}
\addcontentsline{toc}{chapter}{\pname}\label{partI}

\pagebreak
\hspace{14pt}
\vfill
\begin{center}
\textbf{\large{\pname}}
\end{center}
\vfill
\hspace{0pt}
\pagebreak

%% file: chapters/3_offline_detection.tex
\chapter{Deep learning for Spatiotemporal Detection of Multiple Action Tubes}
\renewcommand{\imagepath}{figures/offline/} 
\label{chapter:offline}
\input{chapters/partI/offline_intro}
\input{chapters/partI/offline_approach}
\input{chapters/partI/offline_exp}

\section{Summary and limitations}\label{offline:summary}
\paragraph{Summary:}
The superior performance of the proposed method is due to a number of reasons.
1) Instead of using unsupervised region proposal algorithms as in~\cite{uijlings2013selective,zitnick2014edge},
our pipeline takes advantage of a supervised RPN-based region proposal approach 
which exhibits better recall values than~\cite{uijlings2013selective}
~(see Section~\ref{offline:exp:subsec:recall2iou}). 
2) Our fusion technique~(Section~\ref{offline:subsec:fusion}) improves the mAP score (over the individual appearance or motion models) by $9.4\%$ on the UCF101-24, 
J-HMDB-21 and LIRIS HARL datasets respectively, as shown in Section~\ref{offline:exp:subsec:st-detection} and Table~\ref{offline:table:st-results}.
3) Temporal detection via label consistency is responsible for significant improvements in mAP, quantifiable at around $20\%$ on LIRIS HARL and at $6\%$ on UCF-101: see Section~\ref{offline:exp:subsec:labelsmooth}. 

Qualitative results are also provided in the supplementary video \footnote{\url{https://www.youtube.com/embed/vBZsTgjhWaQ}}
and on the project's web page\footnote{\url{http://sahasuman.bitbucket.io/bmvc2016}}, where the code has also been made available.

\paragraph{Limitations}

Although the proposed method produces remarkable results compared to the previous state-of-the-art, while also being more efficient, it still suffers from
three major limitations, which are described below. 

Firstly, dynamic programming is used to solve the optimisation problem formulation of both action path building (\S~\ref{offline:sec:buildingpaths}) and temporal trimming (\S~\ref{offline:sec:temporaltrimming}) of action paths into action tubes. 
A dynamic programming-based solution requires various iterative passes. Hence, it requires observations from the entire video to be at hand, and as a result is intrinsically offline in nature, i.e., it cannot process videos in a streaming fashion.

Secondly, even though the proposed method is more efficient than other approaches, it cannot still perform in real time at test time. One of the bottleneck is the two-step (RPN and detection) frame-level detection framework (\S~\ref{offline:sec:detectionframwork}), which requires 200ms (\S~\ref{offline:subsec:time_analysis}) to process a single frame. 
Another bottleneck is the computation time required for dense optical flow estimation~\cite{brox2004high}, which again does not allow real time processing.

Thirdly, in the approach presented in this chapter the temporal reasoning is limited to the tube construction method (\S~\ref{offline:sec:buildingpaths}) and the optical flow stream (\S~\ref{online:sec:opticalflow}).
The detection network, on the other hand, does not involve any temporal reasoning either at training or at testing time. 
We think it is a limitation of similar frame-level methods not being able to learn to perform temporal reasoning implicitly.

\paragraph{Looking ahead}

The limitations explained above provides us with an opportunity to devise a new way to tackle action detection problems. 
In the next chapter, we propose to overcome the first two limitations with the help of an online, real-time action detection method.
The online, real-time processing in an action detection system makes it applicable to a wide variety of problems and applications, such as human-robot interaction, self-driving vehicles, autonomous robotic surgery. We will cover these aspects in the next chapter. 
Later on, in Chapter~\ref{chapter:tramnet}, we will propose solutions to enhance the temporal reasoning for predicting longer tubes directly from the set of input frames.

%% file: chapters/partI/offline_intro.tex
\section{Introduction} 
\label{offline:intro}

\begin{figure}[ht!]
  \centering
  \includegraphics[width=0.95\textwidth]{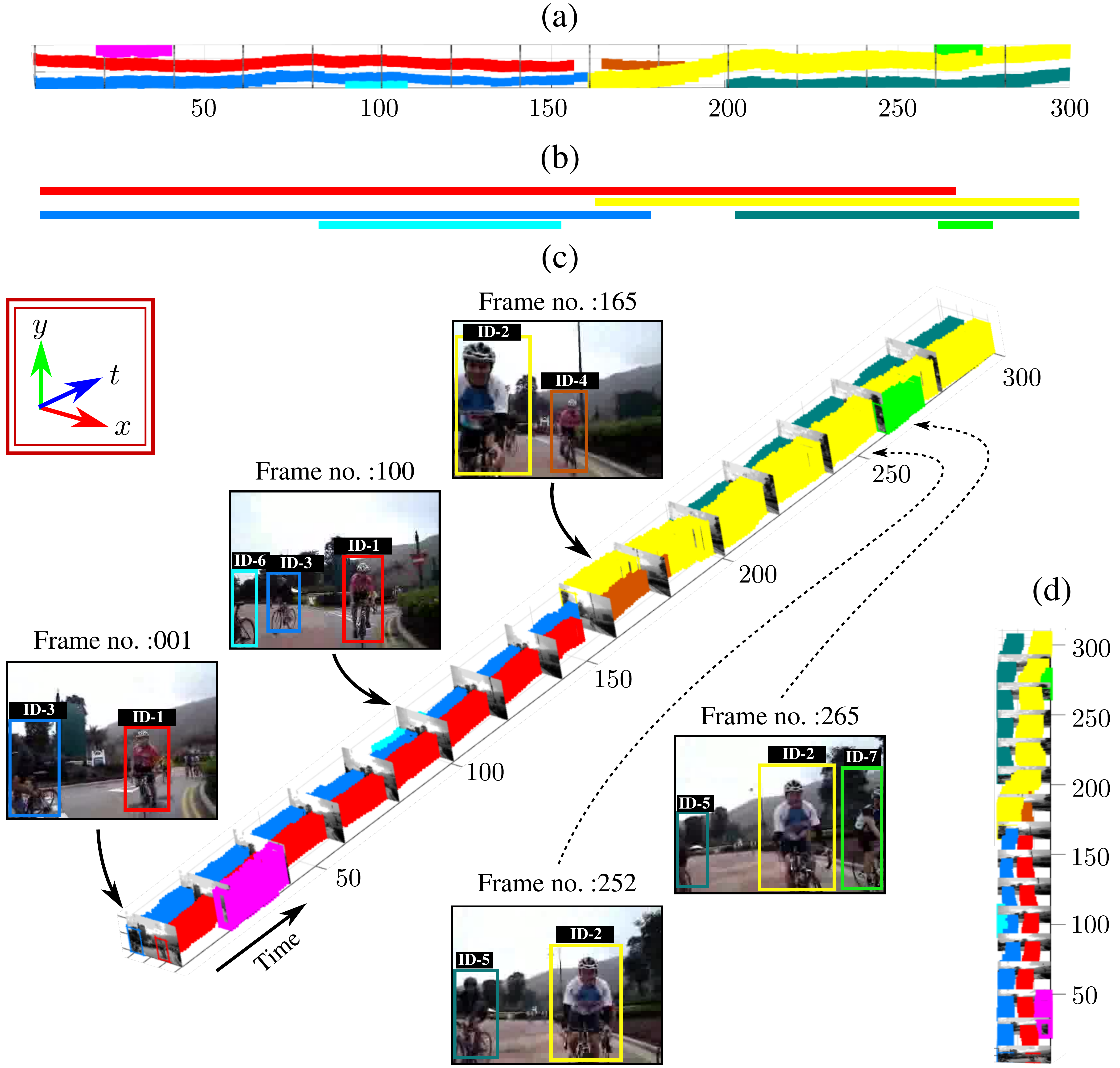}
  \caption[Visualisation of action tube detections in a test video.]
  {Visualisation of action tube detections in a test `biking' video taken from UCF101-24~\cite{ucf101}.
  \textbf{(a)} Side view of the detected action tubes, in which each colour represents a particular instance.
  The detection boxes in each frame are linked up to form space-time action tubes.
  \textbf{(b)} Illustration of the ground truth temporal duration for comparison.
  \textbf{(c)} Viewing the video as a 3D volume with selected image frames;
  \textbf{(d)} Top-down view.
  }
  \label{offline:fig:introTeaser}
\end{figure}

We maintained in Chapter~\ref{chapter:intro} that 
`action tube'~\cite{georgia2015tube} detection provides a holistic approach to 
the human action understanding problem.  We defined 
an action tube to be a set of connected bounding boxes, covering the temporal extent of an action instance.
A video can contain multiple instances of the same action, as well as instances of different actions.
A pictorial representation of the action tube detection problem is shown in Figure~\ref{offline:fig:introTeaser}. 
In this chapter, we propose to tackle this problem with the help of what were in 2016 the latest CNN architectures.
The following chapters will build upon these ideas to achieve better, faster, and more elegant solutions to action tube detection and future prediction of action tubes. 

Advances in object detection via CNNs which took place in 2014-16~\cite{girshick2014rcnn} have triggered a significant performance improvement 
in the state-of-the-art action detectors~\cite{georgia2015tube,weinzaepfel2015learning}. 
However, these approaches had two significant drawbacks. 
First, their accuracy was limited by their reliance on unsupervised region proposal 
algorithms such as Selective Search~\cite{georgia2015tube} or EdgeBoxes~\cite{weinzaepfel2015learning} 
which, besides being resource-demanding, cannot be trained for a specific detection task and are disconnected from the overall detection objective.
Second, temporal detection was either achieved by expensive sliding window approach in~\cite{weinzaepfel2015learning} or is not taken care as in~\cite{georgia2015tube}.

To overcome these issues we propose 
a novel action detection framework which,
instead of adopting an expensive multi-stage detection frameworks~\cite{girshick2014rcnn},
takes advantage of the most recent end-to-end trainable architectures for object detection~\cite{ren2015faster},
in which a single CNN is trained for both detecting and classifying frame-level region proposals in an end-to-end fashion.
Detected frame-level proposals are subsequently linked and trimmed to form space-time action tubes by solving two optimisation problems via dynamic programming.

\subsection{Action tube detection approach}\label{offline:subsec:stintro}

We propose to solve the action tube detection problem by dividing it into three sub-problems. 
Firstly, we employ supervised proposal generation network to detect the action bounding boxes within each video frame. 
Secondly, we connect the detected bounding boxes within a suitable optimisation framework~\cite{georgia2015tube} to form video-long action paths, using dynamic programming. 
Lastly, we trim the resulting action paths to the temporal extent of action instances via another dynamic programming-based solution~\cite{evangelidis2014continuous}.

We demonstrate that the proposed action detection pipeline is at least $5\times$ faster in test time detection speeds as compared to previous works~\cite{georgia2015tube,weinzaepfel2015learning}, and it can be trained efficiently as well for the same reason.
In Section~\ref{offline:subsec:time_analysis}, 
we compare the computing time requirements during testing of our approach with~\cite{georgia2015tube,weinzaepfel2015learning}
on the J-HMDB-21~\cite{jhmdbj2013towards} dataset.
Moreover, our pipeline consistently outperforms
\footnote{The model presented in this chapter appeared in BMVC 2016 and it outperformed the state-of-the-art results at that time.
}
previous state-of-the-art results 
(Section~\ref{offline:sec:exp}). 

\subsection{Temporal action detection approach}\label{offline:subsec:temp-intro}

Inspired by~\cite{kulkarni2015continuous}, Evangelidis~\etal~\cite{evangelidis2014continuous} proposed a simple label consistency constraint over frame-level gesture class scores, and solved the optimisation with dynamic programming in multi-class setting. 
We propose to use a similar formulation for temporal action detection. The difference, however, is that we tackle temporal consistency for $k$ classes independently. 
This class-wise temporal detection method is used to find the temporal bounds (start and end time) of each action tube --
label consistency for all constituting detection boxes is ensured. 
In contrast with other approaches~\cite{weinzaepfel2015learning,laptev2007retrieving,tian2013spatiotemporal} which achieve temporal trimming via an inefficient, multi-scale sliding window, we rely on a simple and efficient dynamic programming formulation.

We demonstrate that our temporal trimming solution provides performance improvement in spatiotemporal action detection on the LIRIS-HARL and UCF101-24 datasets, as well as on the temporal activity detection task part of the 2016 Activitynet~\cite{caba2015activitynet} challenge~\footnote{We compare ourselves with the methods published before Activitynet 2016 \url{http://activity-net.org/challenges/2016/data/anet\_challenge\_summary.pdf}}. 

\paragraph{\textbf{Related publications:}}
The work presented in this chapter appeared in BMVC 2016~\cite{saha2016deep} and ~\cite{singh2016untrimmed} as part of our Activitynet~\cite{caba2015activitynet} challenge 2016\footnote{http://activity-net.org/challenges/2016/} submission. Our entry secured second place in the activity detection task at a detection threshold of $0.5$, and first place when detection performance was averaged across multiple detection thresholds, from $0.5$ to $0.95$. This dissertation's author was the primary investigator in~\cite{singh2016untrimmed} and co-author in ~\cite{saha2016deep}, to which he contributed the dynamic programming-based solution~\cite{singh2016untrimmed} to the temporal trimming of the action tubes, explained in Section~\ref{offline:sec:temporaltrimming}.

%% file: chapters/partI/offline_approach.tex
\paragraph{Outline:} 
the rest of the chapter is organised as follows.
We start by presenting an overview of our action tube detection approach in Section~\ref{offline:sec:overview}.
Next, we describe the three stages of our action tube detection pipeline. 
Stage one (\S~\ref{offline:sec:detectionframwork}) includes an end-to-end trainable frame-level detector used to produce bounding box detections around humans.
Stage two (\S~\ref{offline:sec:buildingpaths}) is the process of connecting detection bounding boxes into video long \emph{action paths}.
Stage three (\S~\ref{offline:sec:temporaltrimming}) adopts a video-level temporal action detection formulation (\S~\ref{offline:subsec:frameleveltrimming}) for the temporal trimming of action paths (\S~\ref{offline:subsec:actionpathtrimming}).
Next, we report the experimental validation of the proposed model in Section~\ref{offline:sec:exp}, from both a qualitative and a quantitative perspective.
In Section~\ref{offline:sec:impl_detail}, the implementation details are presented.
Finally, we present the method's summary and limitations in Section~\ref{offline:summary}. 

\section{Overview of action tube detection approach}
\label{offline:sec:overview}
\begin{figure}[ht!]
    \centering
    \includegraphics[width=0.95\textwidth]{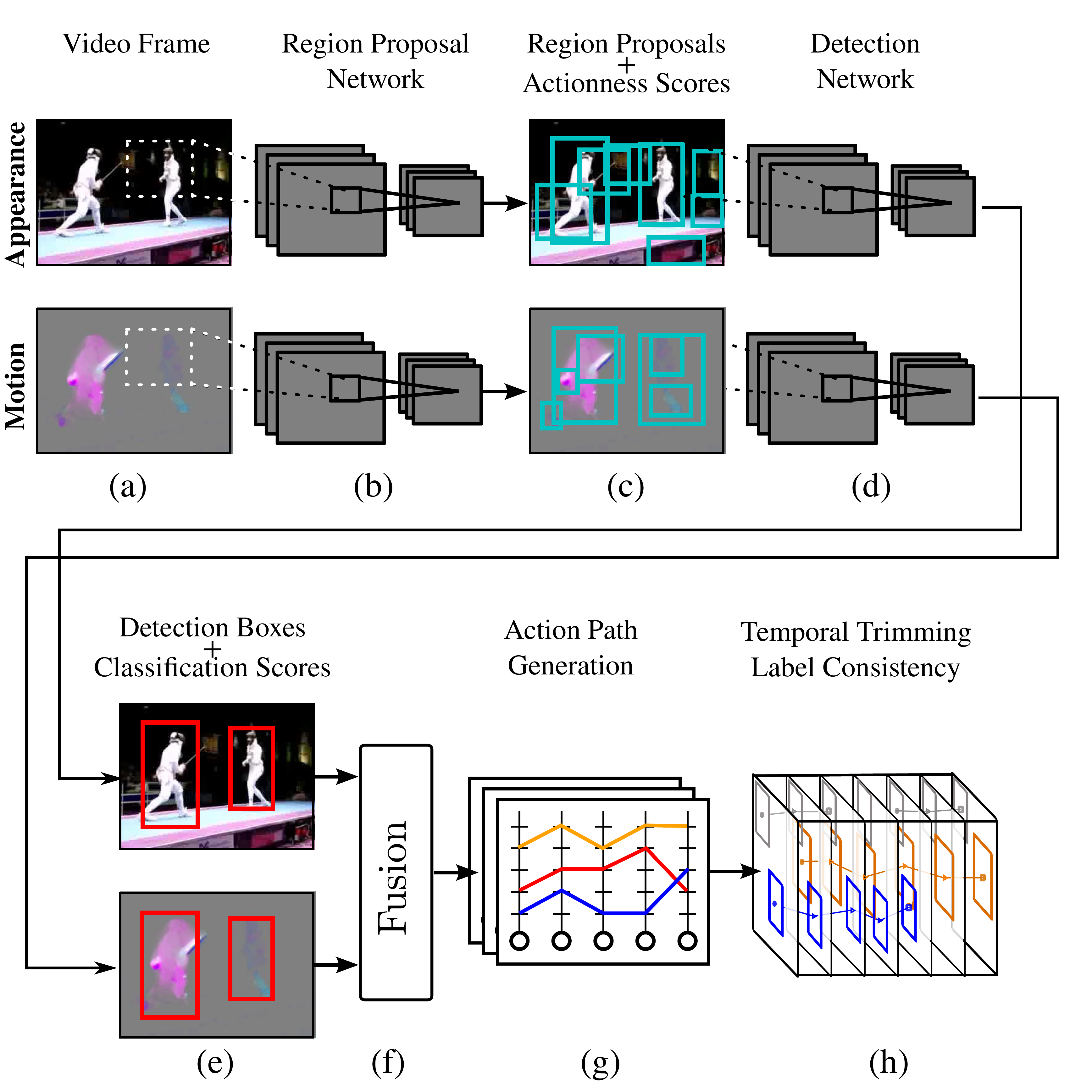}
    \caption[Overview of two-stream based deep action detection pipeline.]
    {
    Overview of our two-stream deep action detection pipeline at test time.
    \textbf{(a)}~RGB and optical-flow images are passed to
    \textbf{(b)}~two separate region proposal networks (RPNs) (\S~\ref{offline:subsec:rpn}).
    \textbf{(c)}~Each network outputs its own region proposals with their associated ``actionness'' scores .
    \textbf{(d)}~Each appearance/motion detection network (\S~\ref{offline:subsec:frcnn}) takes as input the relevant image and RPN-generated region proposals, and
    \textbf{(e)}~outputs detection boxes and softmax probability scores.
    \textbf{(f)}~Appearance and motion based detections are fused (\S~\ref{offline:subsec:fusion}) and
    \textbf{(g)}~linked up to generate class-specific action paths spanning the whole video (\S~\ref{offline:sec:buildingpaths}).
    \textbf{(h)}~Finally the action paths are temporally trimmed to form action tubes (\S~\ref{offline:subsec:actionpathtrimming}).
    }
    \label{offline:fig:overview}
\end{figure}   
    
Our approach is summarised in Figure~\ref{offline:fig:overview}. 
We train two pairs of Region Proposal Networks (RPN)~\cite{ren2015faster} and Fast R-CNN~\cite{girshick2015fast} detection networks
- one on RGB and the other on optical flow images~\cite{georgia2015tube}.
For each pipeline, the RPN \textbf{(b)}, takes as input a video frame \textbf{(a)},
and generates a set of region proposals \textbf{(c)},
and their associated ``actionness'' \footnote{The term \emph{actionness}~\cite{chen2014actionness} is used to denote the possibility of an action being present within a region proposal.}
scores\footnote{A normalised score with softmax operator.}. 
Next, a Fast R-CNN~\cite{ren2015faster} detection network \textbf{(d)} 
takes as input the original video frame and a fraction of the region proposals generated by the RPN,
and outputs a `regressed' detection box and a softmax classification score for each input proposal,
indicating the probability of an action class being present within the box.
To merge appearance and motion cues,
we fuse \textbf{(f)} the softmax scores from the appearance- and motion-based detection boxes \textbf{(e)} (\S~\ref{offline:subsec:fusion}).
We find that this strategy significantly boosts detection accuracy.

After fusing the set of detections over the entire video,
we identify sequences of frame regions most likely to be associated with a single action tube. 
Detection boxes in a tube need to display a high score for the considered action class,
as well as a significant spatial overlap for consecutive detections.
Class-specific action paths \textbf{(g)} spanning the whole video duration are generated via a {Viterbi} forward-backward pass (as  in~\cite{georgia2015tube}). 
An additional second pass of dynamic programming is introduced to take care of temporal detection \textbf{(h)}. 
As a result, our action tubes are not constrained to span the entire video duration, as in~\cite{georgia2015tube}.
Furthermore, extracting multiple paths allows our algorithm to account for multiple co-occurring instances of the same action class. 




\section{Frame-level detection framework} \label{offline:sec:detectionframwork} 

As outlined in Figure \ref{offline:fig:overview}  
our approach combines a region-proposal network (\S~\ref{offline:subsec:rpn}-Fig.~\ref{offline:fig:overview}(b)) with
a detection network (\S~\ref{offline:subsec:frcnn}-Fig.~\ref{offline:fig:overview}(d)),
and fuses the resulting outputs (\S~\ref{offline:subsec:fusion}-Fig.~\ref{offline:fig:overview}(f)).
These two components provide each a set of detection bounding boxes for each video frame, where each bounding box is associated with a classification score vector of length $C+1$, including one for a `background' class.

Each component of our system is described in detail below.

\subsection{Region proposal network} \label{offline:subsec:rpn}

To generate rectangular action region hypotheses in each video frame we adopt the Region Proposal Network (RPN) first published in~\cite{ren2015faster}, 
which is built on top of the last convolutional layer of the VGG-16 architecture by Simonyan and Zisserman~\cite{simonyan2015very}. 
To generate region proposals, this mini-network slides over the convolutional feature map outputted by the last layer, 
processing at each location a $n\times n$ spatial window and mapping it to a lower-dimensional feature vector (512-d for VGG-16).
The feature vector is then passed to two fully connected layers: a box-regression layer and a box-classification layer.

During training, for each image location,
$k$ region proposals (also called `anchors') \cite{ren2015faster} are generated.
We consider those anchors having a high spatial intersection-over-union (IoU) with the ground truth boxes (IoU $> 0.7$) as positive examples, whilst those with IoU $ < 0.3$ are considered as negative examples.
Based on these training examples, the network's objective function is minimised using stochastic gradient descent (SGD), encouraging the prediction of both the probability of an anchor belonging to action or no-action category (a binary classification) and the 4~coordinates of the bounding box.
Similar to \cite{ren2015faster}, smooth L1 loss function is used for regression and crossentropy for classification objective.  

\subsection{Detection network}  \label{offline:subsec:frcnn}

As the detection network we use a Fast R-CNN net~\cite{girshick2015fast} with a VGG-16 base architecture~\cite{simonyan2015very}. 
This takes the RPN-based region proposals (\S~\ref{offline:subsec:rpn}) and regresses a new set of bounding boxes for each action class with the associated classification scores.
Each RPN-generated region proposal leads to $C$ (number of classes) regressed bounding boxes with corresponding class scores.

Analogously to the RPN component, the detection network is also built upon the convolutional feature map outputted by the last layer of the VGG-16 network.
It generates a feature vector for each proposal generated by RPN, which is again fed to two sibling fully-connected layers: a box-regression layer and a box-classification layer.
Unlike what happens in RPNs, these layers produce $C$ multi-class softmax scores and refined boxes (one for each action category) for each input region proposal.

\subsection{CNN training strategy} \label{offline:subsec:training_rpn_frcnn}

We employ a variation on the training strategy of~\cite{ren2015faster} to train both the RPN and the Fast R-CNN networks.
Shaoqing~\etal\ ~\cite{ren2015faster} suggested a 4-steps `alternating training' algorithm in which in the first 2~steps, 
an RPN and Fast R-CNN nets are trained independently, while in the 3$^{\textnormal{rd}}$ and 4$^{\textnormal{th}}$ steps the two networks are fine-tuned with shared convolutional layers.
In practice, we found empirically that the detection accuracy on UCF101-24 slightly decreases when using shared convolutional features, i.e.,
when fine-tuning the RPN and Fast-RCNN trained models obtained after the first two steps.
As a result, we train the RPN and the Fast R-CNN networks independently following only the 1$^{\textnormal{st}}$ and 2$^{\textnormal{nd}}$ steps of~\cite{ren2015faster},
while neglecting the 3$^{\textnormal{rd}}$ and 4$^{\textnormal{th}}$ steps suggested by~\cite{ren2015faster}.

\subsection{Fusion of appearance and motion cues} \label{offline:subsec:fusion}

In a work by Redmon~\etal~\cite{redmon2015you},
the authors combine the outputs from Fast R-CNN\cite{girshick2015fast} and YOLO~\cite{redmon2015you} (You Only Look Once) object detection networks to reduce background detections and improve the overall detection quality.
Inspired by their work, we use our motion-based detection network to improve the scores of the appearance-based detection net (see Figure~\ref{offline:fig:overview}(f)).

Let $\{\Vector{b}^{s}_i\}$ and $\{\Vector{b}^{f}_j\}$ denote the sets of detection boxes generated by the appearance- and motion-based detection networks, respectively, on a given test frame and for a specific action class $c$.
Let $\Vector{b}^{f}_{max}$ be the motion-based detection box with maximum overlap with a given appearance-based detection box $\Vector{b}^{s}_{i}$. 
If this maximum overlap, quantified using the IoU, is above a given threshold $\Scalar{\tau}$,
we augment the softmax score $s_{c}(\Vector{b}^{s}_{i})$ of the appearance-based box 
as follows: 

\begin{equation}
\label{offline:eq:fusion}
s_{c}^{*}(\Vector{b}^{s}_{i}) = s_{c}(\Vector{b}^{s}_{i}) + s_{c}(\Vector{b}^{f}_{max})\times IoU(\Vector{b}^{s}_{i}, \Vector{b}^{f}_{max}).
\end{equation} 

The second term adds to the existing score of the appearance-based detection box a proportion, equal to the amount of overlap, of the motion-based detection score.
We set $\Scalar{\tau}=0.3$ by cross-validation on the training set.
We also tried augmenting the softmax scores of the motion-based detection boxes as per their
maximum IoU overlaps with the appearance-based detections, but this did not produce better results.

\section{Building action paths.}\label{offline:sec:buildingpaths} 

\subsection{Problem formulation} \label{offline:sec:buildingpathsFormulation} 

The output of our fusion stage (\S~\ref{offline:subsec:fusion}) is, for each video frame,
a collection of detection boxes for each action category, together with their associated augmented classification scores~(Equation~\ref{offline:eq:fusion}).
Detection boxes can then be linked up in time to identify video regions most likely to be associated with single or multiple action instances, which we term \emph{action paths}.

Action paths are connected sequences of detection boxes in time, without interruptions, 
which span the entire video duration.
They are obtained as solutions to an energy maximisation problem.
A number of action-specific paths $\Vector{p}_{c}= [\Vector{b}_{\Index{1}}, \dots,  \Vector{b}_{T}]$,
spanning the entire video length, are constructed by linking detection boxes over time in virtue of their class-specific scores and their spatial overlap. 
\\
For example, consider a video composed of $200$ frames, in which an action is present between frame $52$ and $87$.
The generated action paths, however, will span the entire video duration --
i.e., each path starts at frame $1$ and ends at frame $200$.
For those frames where the action is not present,
the action path generation algorithm 
(\S~\ref{offline:subsec:buildingpathsSoultion})
picks up suitable detections according to the energy defined for a particular path 
(Equation~\ref{offline:eq:actionPathOptimisation}).
Subsequently, these noisy detections are expected to be removed from each path during 
the temporal trimming step (\S~\ref{offline:subsec:actionpathtrimming}).

\begin{figure}[ht!]
  \centering
  \includegraphics[width=0.99\textwidth]{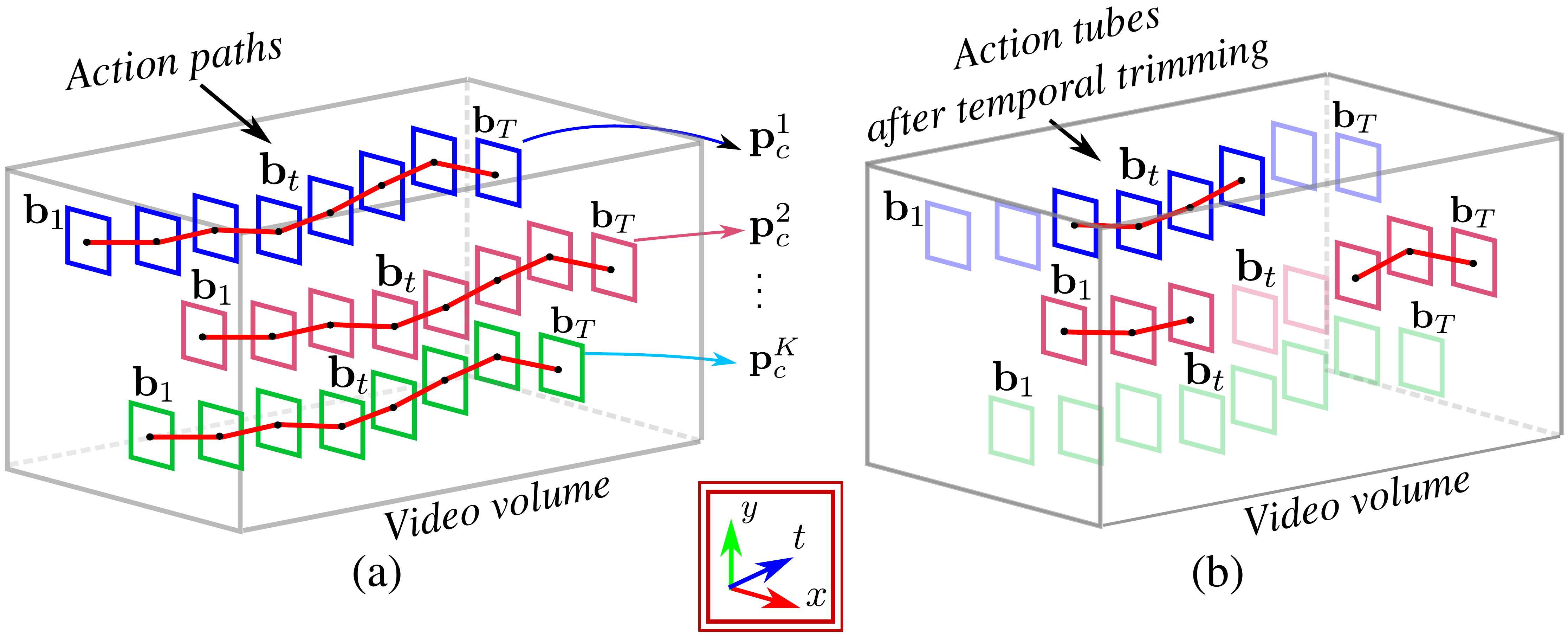}
  \caption[Class-specific $K$ action paths and tubes.]
  {
  \textbf{(a)} $K$ class-specific action paths $\Vector{p}_{c}$ are built by temporally linking a number of frame-level detections $\Vector{b}_t$.
  \textbf{(b)} Action tubes are generated after applying temporal trimming on the $K$ class-specific action paths. In this example, $\Vector{p}_c^1$ is trimmed down to one action tube, $\Vector{p}_c^2$ is trimmed to two action tubes, whereas $\Vector{p}_c^K$ does not contain any actual action tube, hence it is entirely discarded.
  }
  \label{offline:fig:action_paths_tubes}
\end{figure}

Video-long action paths are illustrated in Figure~\ref{offline:fig:action_paths_tubes}(a).

\subsection{Optimisation formulation}\label{offline:subsec:buildingpathsSoultion}

We define the energy $E(\Vector{p}_c)$ for a particular path $\Vector{p}_c$ linking up detection boxes for class $c$ across time to be a sum of unary and pairwise potentials:
\begin{equation}
  \label{offline:eq:actionPathOptimisation}
  E(\Vector{p}_c) = \sum_{t=1}^T s^*_c(\Vector{b}_t) + \lambda_{o} \sum_{t=2}^T \psi_{o} \left( \Vector{b}_t, \Vector{b}_{t-1} \right),
\end{equation}
where $s^*_c(\Vector{b}_t)$ denotes the augmented score (Equation~\ref{offline:eq:fusion}) of detection $\Vector{b}_t$,
the overlap potential $\psi_{o}(\Vector{b}_{\Index{t}}, \Vector{b}_{\Index{t}-1})$ is the IoU of the two boxes $\Vector{b}_t$ and $\Vector{b}_{t-1}$, and $\lambda_{o}$ is a scalar parameter weighting the relative importance of the pairwise term, $\lambda_{o}$ is set to 1 after cross-validation.
The value of the energy (Equation~\ref{offline:eq:actionPathOptimisation}) 
is high for paths whose detection boxes score highly for the particular action category $c$, and for which consecutive detection boxes overlap significantly.
We can find the path which maximises the energy,

\begin{equation}
 \label{offline:eq:region_max}
 \Vector{p}_{c}^* = \textnormal{argmax}_{\Vector{p}_{c}} \ E(\Vector{p}_c)
\end{equation}
by simply applying the Viterbi algorithm used by~\cite{georgia2015tube}.

Once an optimal path has been found, we remove all the detection boxes associated with it and recursively seek the next best action path.
Extracting multiple paths allows our algorithm to account for multiple co-occurring instances of the same action class.

\section{Temporal action detection} \label{offline:sec:temporaltrimming}

In this section we describe how a label consistency formulation can be applied to convert frame-level score vectors into a temporal action detection method (Section~\ref{offline:subsec:frameleveltrimming}), as well as 
into a method designed to trim action paths to the actual temporal extent of the action instances there contained (Section~\ref{offline:subsec:actionpathtrimming}), as illustrated in Figure~\ref{offline:fig:action_paths_tubes}(b).

\subsection{Frame-level temporal action detection for the ActivityNet challenge}
\label{offline:subsec:frameleveltrimming}

The optimisation formulation we used to perform temporal action detection in the context of the 2016 ActivityNet challenge is described here.

For that competition, we used the frame-level and video-level pre-computed CNN features provided by the organisers. 
Each video was first classified into one of the $C$ classes using an ensemble of classifiers trained~\cite{singh2016untrimmed} on video-level features. 
We would only perform temporal trimming for the top-5 video-level labels.
More details about video classification and features are given in Section~\ref{offline:sec:impl_detail} --
here we are interested, in particular, in the label consistency formulation designed to solve the temporal action detection problem starting from the score vectors of the observed video frames.
We trained a random forest~\cite{breiman2001random} classifier on the frame-level features, which outputted a classification score of length $C$ for each frame.

Formally,
given the frame-level scores $\{s_t, t=1,...,T\}$ for a video of length $T$, we want to assign to each frame a binary label \mbox{${l}_{t}$ $\in$ $\{1,0\}$} (where zero represents the `background' or `no-activity' class), which maximises:
\begin{equation}~\label{offline:eqn:secondpassenergy}
E(L) =  \sum_{t=1}^T s_t - \lambda \sum_{t=2}^T \psi_l \left( \Scalar{l}_{t}, 
\Scalar{l}_{t-1} \right) ,
\end{equation}
where 
$\lambda$ is a scalar parameter weighting the relative importance of the pairwise term. The pairwise potential $\psi_l$ is defined as: 

\begin{equation}
  \label{offline:eqn:pottsframelevel}
  \psi_l(l_t,l_{t-1}) = \begin{cases} 0 &\mbox{if } l_t=l_{t-1}\\
  \alpha & \mbox{otherwise} , \end{cases}
\end{equation}

where $\alpha$ is a parameter which we set by cross-validation. $\lambda_l$ set to 1 after cross validation.
This penalises labelling $L = \{l_1,...,l_T\}$ which are not smooth, thus enforcing a piece-wise constant solution.
All contiguous sub-sequences form the desired activity proposal (which can be as many as there are instances of activities). 
Each activity proposal is assigned a global score of $S_a$ equal to the mean of the scores of its constituting frames. 
This optimisation problem can be efficiently solved by dynamic programming~\cite{evangelidis2014continuous}. 

\subsection{Temporal trimming in action paths}\label{offline:subsec:actionpathtrimming}

We can address the temporal trimming of action paths with the same formulation as in Section~\ref{offline:subsec:frameleveltrimming}.

As a result of the action path building formulation of Section~\ref{offline:sec:buildingpaths}, action-paths span the entire video duration, whereas human actions typically only occupy a fraction of it. Temporal trimming, therefore, becomes necessary.
The dynamic programming solution of Equation~\ref{offline:eq:actionPathOptimisation} aims at extracting connected paths by penalising regions which do not overlap in time.
As a result, however, not all detection boxes within a path exhibit strong action-class scores.

The goal here is to assign to every box $\Vector{b}_t \in \Vector{p}_{c}$ in an action path $\Vector{p}_{c}$ a binary label \mbox{${l}_{t}$ $\in$ $\{c,0\}$}
(where zero represents the `background' or `no-action' class),
subject to the conditions that  the path's labelling ${\Vector{L}_{\Vector{p}_{c}}} = [ l_1, l_2, \dots, l_T]'$:
i) is consistent with the unary scores~(Equation~\ref{offline:eq:fusion});
and ii) is smooth (no sudden jumps).\\
As in the previous pass, we may solve for the best labelling by maximising:
\begin{equation}
  \label{offline:eq:pathtemporaltrimming}
  {E(\Vector{L})}_{\Vector{p}_{c}} = \underset{\Vector{L_{\Vector{p}_{c}}}}{\textnormal{argmax}} \ \left( \sum_{t=1}^T s_{l_t}(\Vector{b}_t) - \lambda_{l} \sum_{t=2}^T \psi_l \left( \Scalar{l}_{t}, \Scalar{l}_{t-1} \right) \right),
\end{equation}
where 
$\lambda_{l}$ is a scalar parameter weighting the relative importance of the pairwise term. The pairwise potential $\psi_l$ is defined to be:
\begin{equation}
  \label{offline:eqn:potts}
  \psi_l(l_t,l_{t-1}) = \begin{cases} 0 &\mbox{if } l_t=l_{t-1}\\
  \alpha_c & \mbox{otherwise} , \end{cases}
\end{equation}
where $\alpha_c$ is a class-specific constant parameter which we set by cross validation. We set $\lambda_l$ to 1 after cross validation.
As each action category has its own short-term and long-range motion dynamics,
cross-validating $\alpha_c$ for each action class separately 
improves the temporal detection accuracy (\S~\ref{offline:exp:subsec:labelsmooth}).
Equation~\ref{offline:eqn:potts} is the standard Potts model which penalises labelling that are not smooth, thus enforcing a piece-wise constant solution.
Again, we solve (Equation~\ref{offline:eq:actionPathOptimisation}) using the Viterbi algorithm. 
 
All contiguous sub-sequences of the retained action paths $\Vector{p}_c$ associated with category label $c$ constitute our action tubes. 
As a result, one or more distinct action tubes spanning arbitrary temporal intervals may be found in each video for each action class $c$.
Finally, 
each action tube is assigned a global score equal to the mean of the top $k$ augmented class scores~(Equation~\ref{offline:subsec:fusion}) of its constituting detection boxes. 

If an action occurs multiple times in a video, then each occurrence of that action represents a different action instance which is encoded by a different action tube.
Figure~\ref{offline:fig:action_paths_tubes}(b)
illustrates the action tube generation step in which $K$ action tubes are obtained
by applying temporal trimming on the $K$ class-specific action paths.

%% file: chapters/partI/offline_exp.tex
\section{Experimental validation} \label{offline:sec:exp}  

We evaluate our method on two separate challenging problems: 
i) the temporal action detection task on the ActivityNet dataset in the context of the 2016 challenge (\S~\ref{offline:subsec:exp:temporalActivitnet}), 
in order to assess the strength of the label consistency formulation introduced in Section~\ref{offline:sec:temporaltrimming};
ii) the spatiotemporal action detection task (\S~\ref{offline:exp:subsec:st-detection})
on the UCF101-24, J-HMDB-21 and LIRIS-HARL datasets, which evaluates our overall action tube detection framework (\S~\ref{offline:sec:overview}).


The detailed information about evaluation benchmarks is given in Section~\ref{soa:datasets} in Chapter~\ref{chapter:related_work}. 
In the following we show the results of our method on different thresholds to understand the performance under gradually more difficult tasks.
The mAP metric, when used to evaluate the detection of tubes in videos, is called video-mAP. In our case, video-mAP and mAP are used interchangeably. More information about evaluation metric can be found in Section~\ref{soa:metrics}.

\begin{table}[ht!]
  \centering
  \footnotesize
  \scalebox{1.0}{ 
  \begin{tabular}{l|cccccc}
  \toprule
  T-IoU threshold $\delta$ = & 0.1 & 0.2 & 0.3 & 0.4 & 0.5 \\
  \hline
  Validation-Set Caba \etal~\cite{caba2015activitynet}& 12.5\% & 11.9\% & 11.1\% & 10.4\% & 09.7\% \\
  Validation-Set Montes \etal~\cite{caba2015activitynet}& -- & -- & -- & -- & 22.5 \\
  Validation-Set proposed  & 52.1\% & 47.9\% & 43.5\% & 39.2\% & 34.5\% \\
  \hline
  Testing-Set Singh~\etal~\cite{singh2016bidirectional}  & -- & -- & -- & -- & 28.8\% \\ 
  Testing-Set proposed  & -- & -- & -- & -- & 36.4\% \\ 
  \bottomrule
  \end{tabular}
  }
  \caption[Activity detection performance on validation and testing set of ActivityNet dataset.]
  {Activity detection performance results~\cite{singh2016untrimmed} on validation and testing set. The quantity $\delta$ is the Temporal Intersection over Union  (T-IoU) threshold.}
  \label{offline:table:temporaldetection}
\end{table}

\subsection{Temporal detection performance comparison}\label{offline:subsec:exp:temporalActivitnet}

Table~\ref{offline:table:temporaldetection} show the results obtained using 
the temporal label smoothing framework proposed in Section~\ref{offline:sec:temporaltrimming}. 
Our method show superior performance than the baseline method of Caba~\etal\cite{caba2015activitynet}. Note that we used the same pre-computed features as Caba~\etal\cite{caba2015activitynet} to build our temporal action detection framework. 
It clearly shows that temporal detection formulation presented in Section~\ref{offline:sec:temporaltrimming} is effective, hence, its adoption for trimming the action paths in Section~\ref{offline:subsec:actionpathtrimming} is natural. 

\begin{table}[ht!]
  \vspace{2mm}
  \centering
  \footnotesize
  \scalebox{1.0}{
  \begin{tabular}{l|ccc|cc|cc}
  \toprule
   Dataset & \multicolumn{3}{c}{UCF101-24} & \multicolumn{2}{c}{J-HMDB-21} & \multicolumn{2}{c}{LIRIS-HARL}\\
   \hline
  ST-IoU threshold $\delta$ & 0.05 & 0.2 & 0.5 & 0.2 & 0.5 & 0.2 & 0.5 \\ \midrule
  FAP~\cite{yu2015fast} & 42.80 & -- & -- & -- & -- & -- & --\\
  ActionTube~\cite{georgia2015tube}  & -- & -- & -- & -- & 53.3 & -- & -- \\
  Wang~\etal~\cite{wangcvpr2016actionness}  & -- & -- & -- & -- & 56.4 & -- & --\\
  STMH~\cite{weinzaepfel2015learning} & 54.28 & 46.77 & -- & 63.1 & 60.7 & -- & -- \\ \midrule
  Our (appearance stream)       & 67.56  & 56.55  & 30.64 & 52.94 & 51.34 & 41.94 &  20.43 \\    
  Our (optical flow stream)       & 65.19 &  55.68  &  27.84 & 69.59 & 67.90 & 46.58 & 19.28\\                
  \textbf{Our (fused)} &    \textbf{78.85}  &  \textbf{66.36}   &   \textbf{34.82} & \textbf{72.63} & \textbf{71.50} & \textbf{49.10} & \textbf{21.36}\\ 
  \bottomrule 
  \end{tabular}
  }
  \caption[Quantitative spatialtemporal action detection results (mAP).]
  {Quantitative action detection results (mAP)~\cite{saha2016deep} on UCF101-24, J-HMDB-21 and LIRIS-HARL datasets.}
  \label{offline:table:st-results}
\end{table} 

\subsection{Spatiotemporal detection performance comparison}
\label{offline:exp:subsec:st-detection}

Table~\ref{offline:table:st-results} presents the results we obtained on UCF101-24, J-HMDB-21, LIRIS-HARL datasets. 
It contains a comparison with the previous state-of-the-art paper~\cite{weinzaepfel2015learning,yu2015fast,wangcvpr2016actionness,georgia2015tube} on each dataset.
Next, we discuss the performance of our proposed method on each individual dataset.
It is important to note that our fusion strategy (\S~\ref{offline:subsec:fusion}) 
helps improve the performance of our method consistently on all datasets -- see the last three rows in Table~\ref{offline:table:st-results}.

\paragraph{Performance comparison on UCF101-24}

We achieve an mAP of $66.36\%$ compared to the $46.77\%$ reported by~\cite{weinzaepfel2015learning} (a $20\%$ gain), at the standard threshold of $\delta~=~0.2$.
At a threshold of $\delta=0.4$ we still get a high score of $45.24\%$,
(comparable to $46.77\%$~\cite{weinzaepfel2015learning} at $\delta=0.2$).
Note that we are the first to report results on UCF101-24 up to $\delta=0.6$,
which attests the robustness of our approach to more accurate detection requirements.
Although our separate appearance- and motion-based detection pipelines already outperform the state-of-the-art (Table~\ref{offline:table:st-results}), 
their combination (\S~\ref{offline:subsec:fusion}) delivers a significant performance increase.

\begin{figure}[ht]
  \centering
  \includegraphics[width=0.98\textwidth]{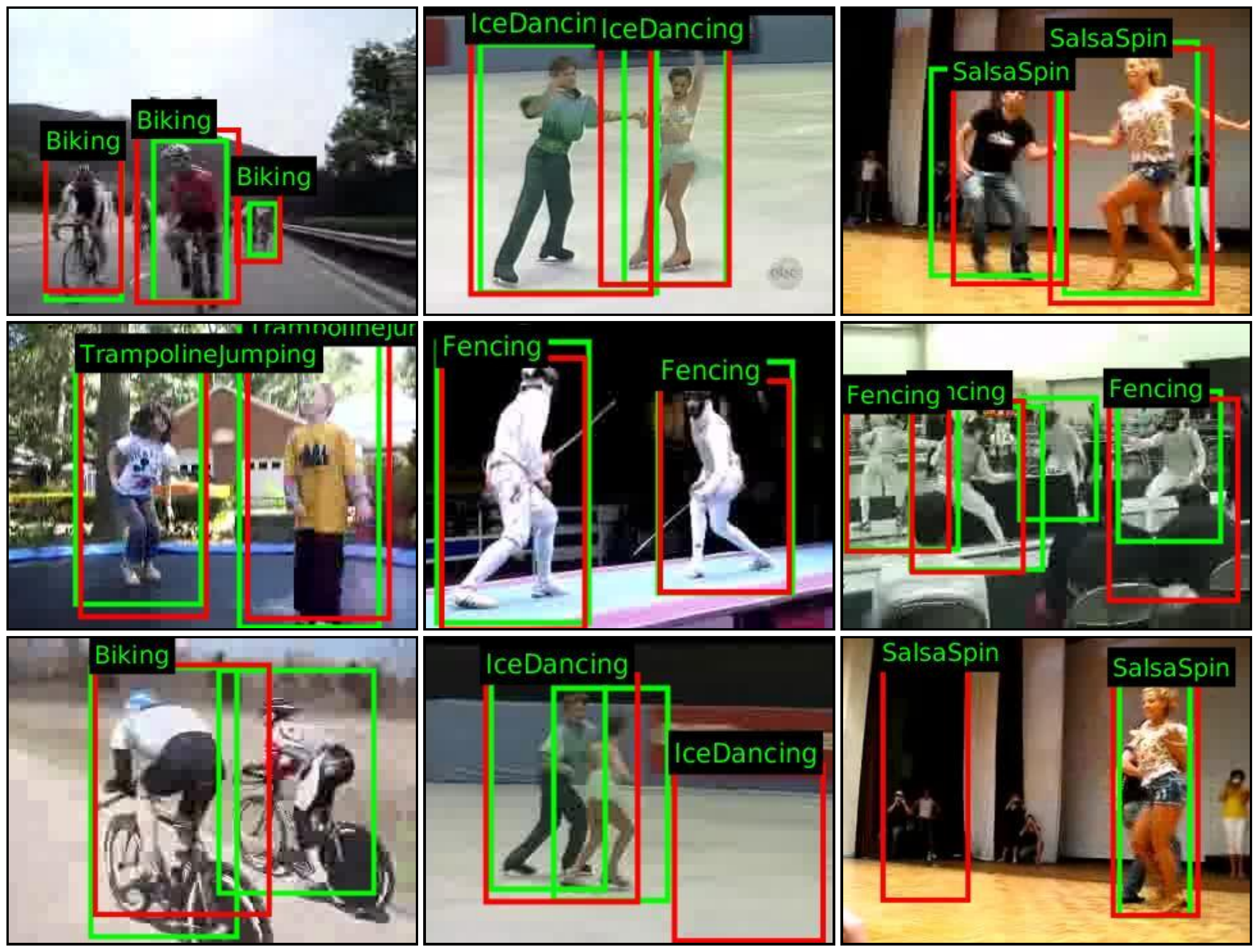}
  \caption[Sample action detection results on UCF101-24.]
  { Sample action detection results~\cite{saha2016deep} on UCF101-24.
    Ground-truth boxes are in green, detection boxes in red.
    The top two rows show correct detections,
    the bottom one contains examples of more mixed results. 
    In the second row right most frame, 3 out of 4 `Fencing' instances are nevertheless correctly detected.
  }
  \label{offline:exp:fig:ucf101-results}
\end{figure}

Some representative example results from UCF101-24 are shown in Figure~\ref{offline:exp:fig:ucf101-results}.
Our method can detect several (more than $2$) action instances concurrently,
as shown in Figure~\ref{offline:fig:introTeaser}, 
in which three concurrent instances and in 
total of six action instances is detected correctly.
Quantitatively, our method achieves class-specific video AP (average precision in $\%$) of $88.0$, $83.0$ and $62.5$ on the UCF101-24 action categories \emph{`fencing'}, \emph{`salsa spin'} and 
\emph{`ice dancing'}, 
respectively, which all concern multiple inherently co-occurring action instances.

\begin{figure}[h]
  \centering
  \includegraphics[width=0.98\textwidth]{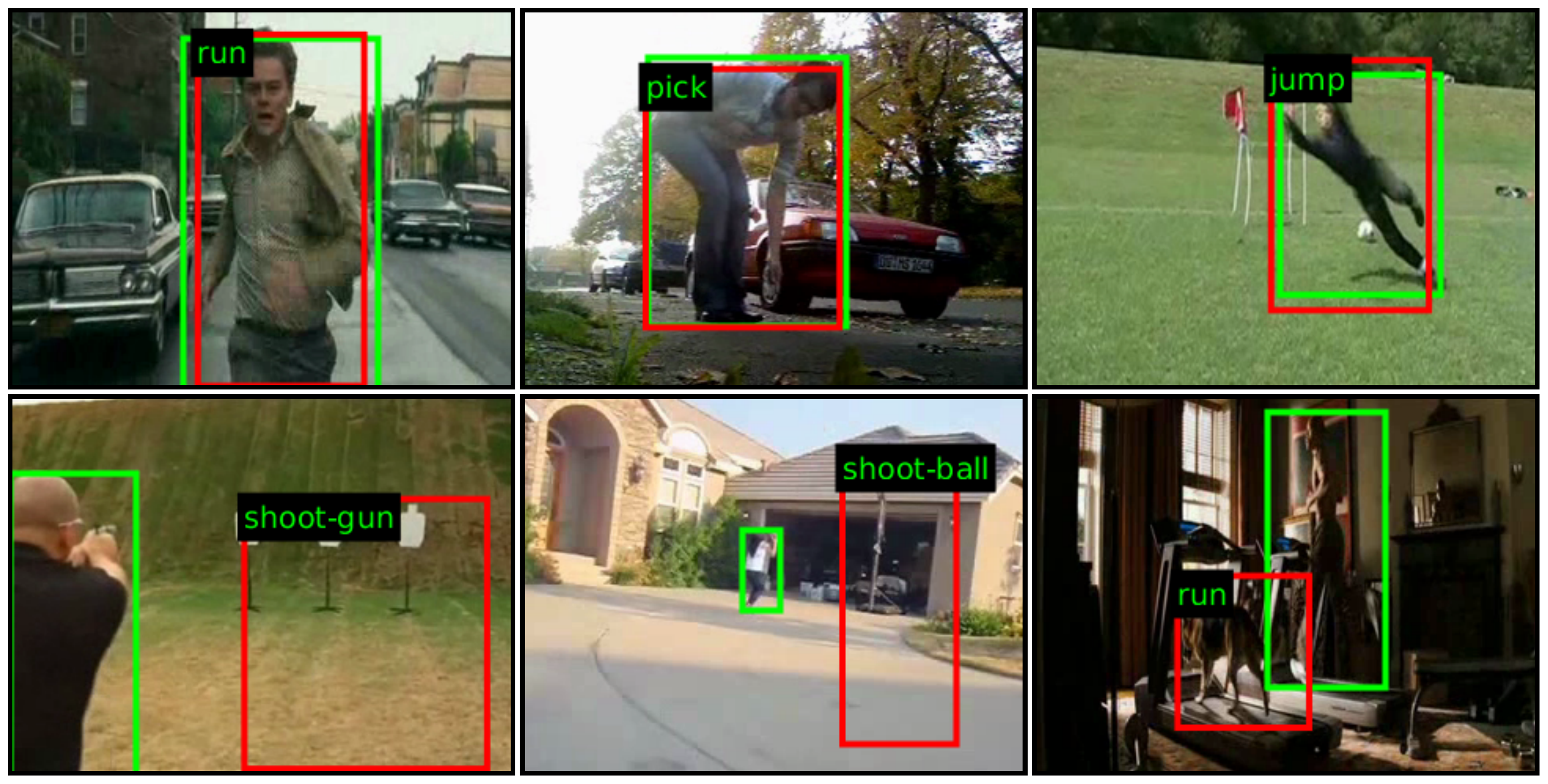}
  \caption[Sample action detection results on J-HMDB-21.]
  {Sample action detection results~\cite{saha2016deep} on J-HMDB-21. Top row: accurate detection examples. Bottom row: mis-detection examples.}
  \label{offline:exp:fig:JHMDB-21-results}
\end{figure}

\paragraph{Performance comparison on J-HMDB-21}
The results we obtained on J-HMDB-21 are presented in the second part of Table~\ref{offline:table:st-results}.
Our method again outperforms the state-of-the-art, with an mAP increase of $18\%$ and $11\%$ at $\delta=0.5$ as compared to \cite{georgia2015tube} and \cite{weinzaepfel2015learning}, respectively. 
Note that our motion-based detection pipeline alone exhibits superior results,
and when combined with appearance-based detections leads to a further improvement of $4\%$ at $\delta=0.5$.
\\
These results confirm the high precision of the detections -
a large portion of the detection boxes have high IoU overlap with the ground truth boxes,
a feature due to the superior quality of RPN-based region proposals as opposed to Selective Search's
(a direct comparison is provided in Section~\ref{offline:exp:subsec:recall2iou}).
Sample detections on J-HMDB-21 are shown in Figure \ref{offline:exp:fig:JHMDB-21-results}.

\begin{figure}[h]
  \centering
  \includegraphics[width=0.98\textwidth]{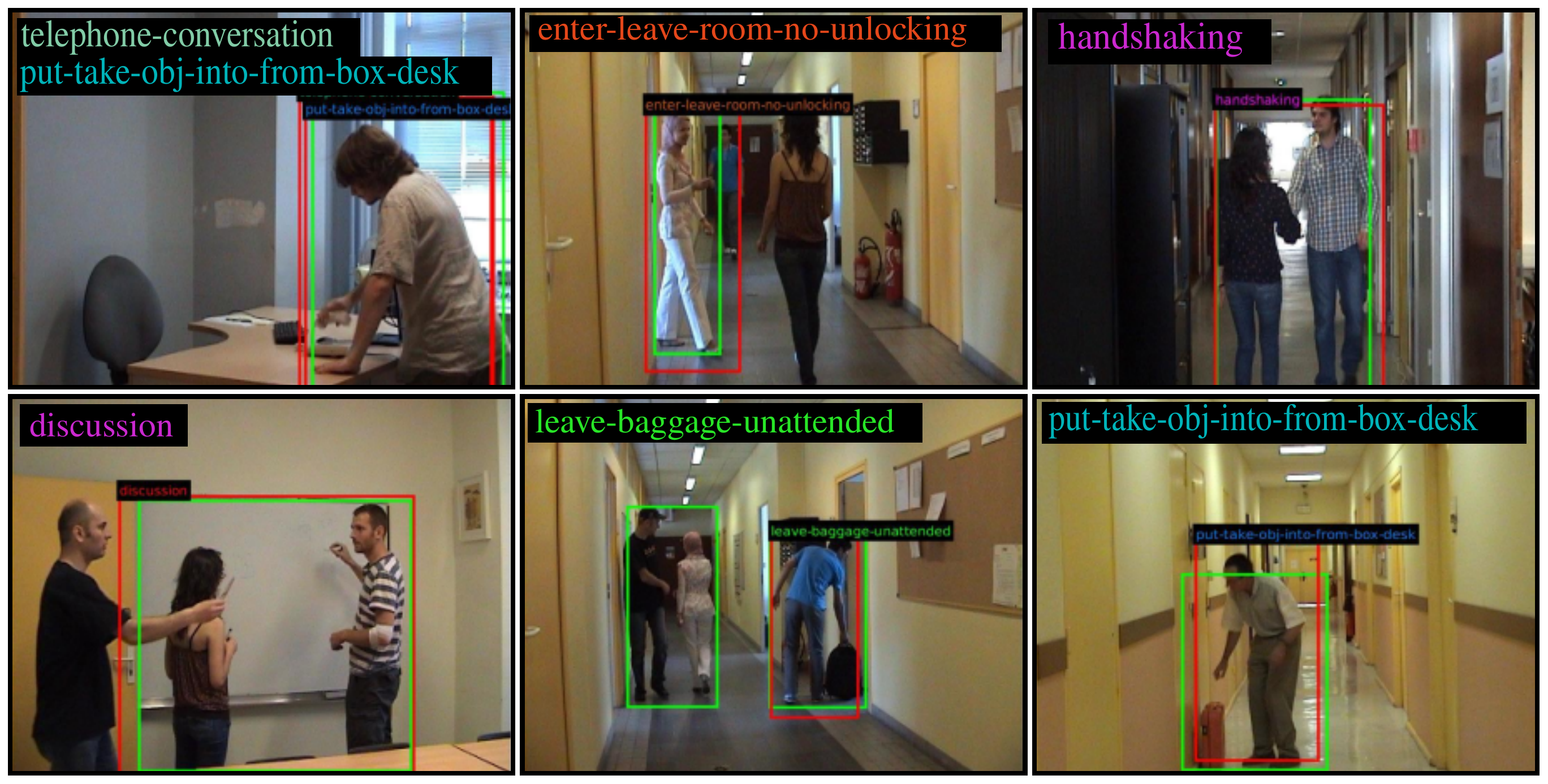}
  \caption[Action detection results on LIRIS-HARL.]
    {Sample action detection results~\cite{saha2016deep} on LIRIS-HARL,
    some of which include single actions involving more than one person like `handshaking' and `discussion'.
    Top row: accurate detection examples. Bottom row: mis-detection examples.
    }
    \label{offline:figure:liris_results}
\end{figure}

\paragraph{Performance comparison on LIRIS-HARL}

LIRIS HARL allows us to demonstrate the efficacy of our approach on 
temporally long untrimmed videos with co-occurring actions. 
We also report in Table~\ref{offline:table:st-results} 
the mAP scores obtained by the appearance, motion and the fusion detection models, 
respectively (note that there is no prior state of the art to report in this case).
Again, we can observe an improvement of $7\%$ mAP at $\delta=0.2$ associated with our fusion strategy. 
Qualitative results on LIRIS reported in Figure~\ref{offline:figure:liris_results} show how our method can handle co-occurring action instances of different action classes.

Similarly to what happens with the mAP metric, the
F1 score can be computed using precision and recall at a particular threshold. 
The F1 score was used as an evaluation metric for the LIRIS-HARL dataset to set the benchmark.
Our method achieves an F1 score of 0.58 as compared with 0.05 and 0.03 obtained by the two benchmark approaches of VPULABUAM-13~\cite{liris-teamp-13} and IACAS-51~\cite{liris-teamp-51} respectively. 
 
\begin{figure}[h!]
  \centering
  \includegraphics[width=0.9\textwidth]{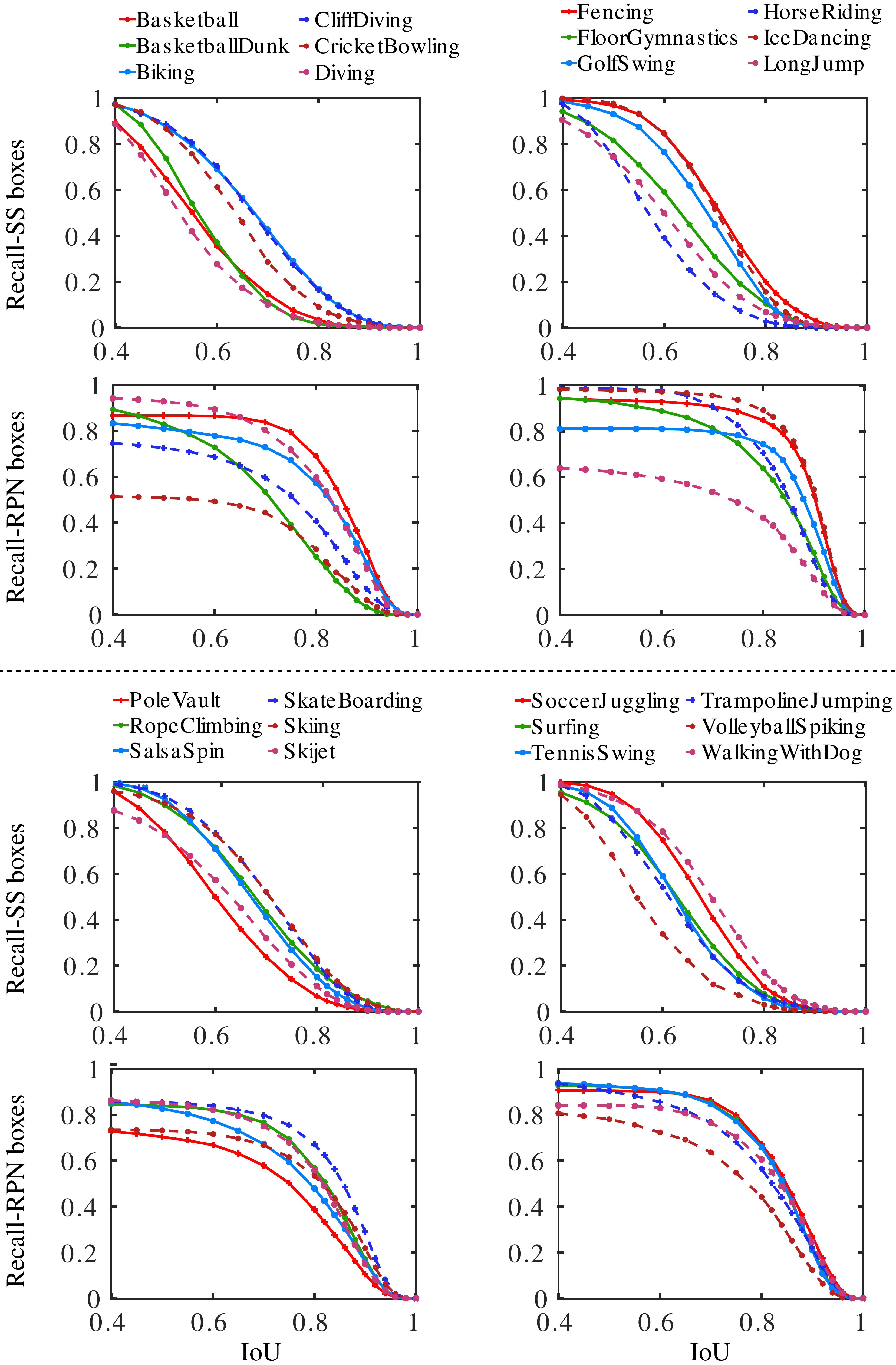}
  \caption[Performance comparison between Selective Search and RPN-based region proposals.]
  {Performance comparison between Selective Search (SS) and RPN-based region proposals   
    on 24 action classes of UCF101-24 dataset.
    $1^{st}$ \& $3^{rd}$ rows: recall vs. IoU curve for Selective Search based region proposals; 
    $2^{nd}$ \& $4^{th}$ rows: recall vs. IoU curve for RPN-based region proposals.} 
  \label{offline:exp:fig:comp_ss_rpn}
\end{figure}

\subsection{Comparative analysis of region proposal quality} 
\label{offline:exp:subsec:recall2iou}

To understand the importance of region proposal generation,
we also analyse the quality of Selective Search vs that of RPN (\S~\ref{offline:subsec:rpn}) region proposals using the Recall-to-IoU measure \cite{ren2015faster}.
We first extract Selective Search (SS) boxes (approximately 1000 boxes/frame) and RPN-based detection boxes  (300 boxes/frame) 
from our detection network on UCF101-24 test-split-1.
We also apply a constraint on the RPN-based proposals by applying a 
threshold to their class-specific softmax probability scores $s_c$, and only considering those proposals with $s_c\geq 0.2$.
For each UCF101-24 action category we compute the recall of these proposals at different threshold values.
Even with a relatively smaller number of proposals and the additional constraint on the classification 
probability score, RPN-based proposals exhibit much better recall values than SS-based boxes as depicted in Figure~\ref{offline:exp:fig:comp_ss_rpn}.

\subsection{Impact of label smoothing}
\label{offline:exp:subsec:labelsmooth}

Along with temporal action detection results shown on ActivityNet dataset in Section
~\ref{offline:subsec:exp:temporalActivitnet}, 
we demonstrate the advantage of our temporal label smoothing step (\S~\ref{offline:subsec:actionpathtrimming}) on UCF101-24 dataset and LIRIS-HARL datasets. 
We shown results (mAP) using the action tubes generated by trimming action paths (\S~\ref{offline:subsec:buildingpathsSoultion}) with our label smoothing method (\S~\ref{offline:subsec:actionpathtrimming}) for action paths.
Without the 2$^\textnormal{nd}$ (trimming) pass, performance decreases by $20\%$, highlighting the importance of temporal trimming in the construction of accurate action tubes.

First, we conduct an experiment where we evaluate the performance difference observed when we perform a temporal label smoothing (as in Section~\ref{offline:subsec:actionpathtrimming}) of the action paths.
We evaluate the action paths generated on LIRIS-HARL against the ground truth action tubes, i.e., without temporal trimming them using the method of Section~\ref{offline:subsec:actionpathtrimming}: see the first row in Table~\ref{offline:table:liris-temporal-trimming}.
Similarly, we evaluate on LIRIS-HARL the action tubes generated using the label temporal smoothing formulation of Section~\ref{offline:subsec:actionpathtrimming}: see the second row in Table~\ref{offline:table:liris-temporal-trimming}.
We can clearly see from Table~\ref{offline:table:liris-temporal-trimming} that proposed temporal label smoothing formulation helps significantly. 

\begin{table}[ht!]
  \centering
  \footnotesize
  \scalebox{0.85}{
  \begin{tabular}{lccccc}
  \toprule
  Proposed - no temporal smoothing (i.e., without \S~\ref{offline:sec:temporaltrimming}) & 38.1 &  29.46 & 23.58  &  14.54 & 9.59 \\
  Proposed - with temporal smoothing (i.e., with \S~\ref{offline:sec:temporaltrimming}) & \textbf{54.18} &  \textbf{49.10} & \textbf{35.91}  &  \textbf{28.03} & \textbf{21.36} \\
  \bottomrule
  \end{tabular}
  }
  \caption[Action detection results (mAP) on LIRIS-HARL.]
  {Action detection results (mAP) on LIRIS-HARL with and without temporal trimming (\S~\ref{offline:sec:temporaltrimming}).}
  \label{offline:table:liris-temporal-trimming}
\end{table} 

Next, we also conduct experiments to show the significance of the label smoothing step with the help of parameter class-specific $\alpha_c$.  
More specifically, we show that class-specific $\alpha_c$ values help
smoothing the action paths 
for each action category independently, resulting in an overall performance boost in the spatiotemporal detection accuracy.
First, we generate detection results on UCF101-24 test set (split-1) by setting the constant parameter $\alpha_c=0$ for each action category. Then, we use a fixed value of $\alpha_c=3$.
Lastly, we use the cross-validated class-specific $\alpha_c$ values and again generated detection results.
In our experiment, we set the spatiotemporal IoU threshold to $\delta=0.2$.
\\
Table~\ref{offline:table:res_ucf101_diff_alphas_stes} presents the results for  
the following three cases: 
\textbf{(a)} detection results obtained by setting $\alpha_c=0$ for all action classes; 
\textbf{(b)} detection results generated using a fixed value of $\alpha_c$ for all classes;
\textbf{(c)} detection results obtained using the cross validated class-specific $\alpha_c$ values.
It shows that $\alpha_c$ should be greater than 0 to force piece-wise constant results, otherwise the resulting labelling path would just track the score values over time. 

\begin{table}[ht!]
  \centering
  \footnotesize

  \scalebox{0.9}{
  \begin{tabular}{l|c}
  \toprule
  & mAP \\ \midrule
  $\alpha_c=0$ & 60.77 \\
  $\alpha_c=3$ & 66.03 \\
  class-specific $\alpha_c$ &  \textbf{66.36} \\
  \bottomrule
  \end{tabular}
  }
  \caption[Spatiotemporal action detection results (mAP) on UCF101-24.]
  {Spatiotemporal action detection results (mAP) on UCF101-24 using three different sets of $\alpha_c$ values.}
  \label{offline:table:res_ucf101_diff_alphas_stes}
\end{table} 

\subsection{Computing time analysis for testing} \label{offline:subsec:time_analysis}

We think useful to compare the detection speed at test time of the combined region proposal generation and CNN feature extraction approach used in~(\cite{georgia2015tube,weinzaepfel2015learning})
with our neural-net based, single stage action proposal and classification pipeline on the J-HMDB-21 dataset: see Table \ref{offline:table:detection_time_analysis}.
We find our method to be $10\times$ faster than~\cite{georgia2015tube} and $5\times$ faster than~\cite{weinzaepfel2015learning},
with a mean of 113.52 \cite{georgia2015tube}, 52.23 \cite{weinzaepfel2015learning} and 10.89 (ours) seconds per video, averaged over all the videos in J-HMDB-21 split1. 

We exclude our $2^{nd}$ label smoothing step (\S~\ref{offline:subsec:actionpathtrimming}) due to the fact that J-HMDB-21 video clips do not require temporal trimming.
The action path building (\S~\ref{offline:subsec:buildingpathsSoultion}) step and optical flow based `motion frame' generation steps are common to~\cite{georgia2015tube} and our pipeline -- we thus
exclude these steps as well in our comparison.
We compare the computation times required for the region proposal generation and CNN feature extraction steps of~\cite{georgia2015tube,weinzaepfel2015learning} 
with our RPN and detection nets computation times.
Table~\ref{offline:table:detection_time_analysis} shows the time required for each step.
The reported computation time is averaged over all the videos in the J-HMDB-21 test split1.
The time is in second per video clip. 
It takes 10 seconds on an average to process a video by detection framework.
On the other side of computation, we can say that it takes 200 milliseconds per frame to run detection framework.
All the Experimental results were generated using a desktop computer
with an Intel Xeon CPU@3.20GHz and NVIDIA Titan X GPU.

Our method is at least $10\times$ faster than~\cite{georgia2015tube} and $5\times$ than~\cite{weinzaepfel2015learning} for those steps we compared with (see Table~\ref{offline:table:detection_time_analysis}).

It is important to note that our adoption of Faster-RCNN~\cite{girshick2015fast} for action tube detection does not only provide improvement in performance (see Table~\ref{offline:table:st-results}) but also a simpler and faster solution to the problem. 
We do away with the expensive sliding window-based approach in \cite{weinzaepfel2015learning} for temporal detection in action paths, which is replaced by our efficient~\cite{evangelidis2014continuous} label smoothing formulation. 

When combined, our efficient label smoothing step and the use of a faster detection network produce an elegant, simple, and efficient framework for action tube detection.

\begin{table}[t!]
  \centering
  \caption[Test time detection speed comparison on J-HMDB-21.]{Test time detection speed comparison on J-HMDB-21 with~\cite{georgia2015tube,weinzaepfel2015learning}.}
  {\footnotesize
  \scalebox{0.99}{
  \begin{tabular}{lc}
  \toprule
  ActionTube~\cite{georgia2015tube}, STMH~\cite{weinzaepfel2015learning} & Average time (Sec./video)  \\ \midrule
  Selective Search~\cite{georgia2015tube} / EdgeBoxes~\cite{weinzaepfel2015learning}  & 68.10 / 6.81  \\
  CNN feature extraction & 45.42   \\ \midrule
  \textbf{Average detection time} & 113.52~\cite{georgia2015tube} / 52.23~\cite{weinzaepfel2015learning}  \\
  \bottomrule  \\ \\ 
  \toprule
  Ours & Average time(Sec./video) \\ \midrule
  RPN proposal generation & 4.08 \\
  Detection network &  6.81 \\ \midrule
  \textbf{Average detection time} & \textbf{10.89} \\
  \bottomrule 
  \end{tabular}
  }
  }
  \label{offline:table:detection_time_analysis} 
\end{table} 

\subsection{Implementation details}
\label{offline:sec:impl_detail}

\paragraph{Training data preparation} 

We divided the UCF101-24 train split one into two subsets.
The first subset consists of $70\%$ ($1605$ videos $\sim$ $240k$ frames) of the training videos from UCF101-24 train split one, whereas 
the second subset contains the remaining $30\%$ ($688$ videos) .
We selected the videos uniformly at random for each action class, and
trained the RPN and Fast R-CNN networks using the first subset, while 
the second subset was used as a validation set for CNN training.
For the J-HMDB-21 and LIRIS HARL D2 datasets we used the original training sets provided by the authors~\cite{jhmdbj2013towards,liris-harl-2014}.

\paragraph{CNN weight initialisation}
The RPN and Fast R-CNN networks~\cite{ren2015faster} were initialised with weights from a model pre-trained on ImageNet dataset~\cite{deng2009imagenet}.

\paragraph{CNN solver configuration setting.}
For UCF101-24, we trained both RPN and Fast R-CNN for $320k$ iterations.
For the first $240k$  iterations we used a learning rate $0.001$, 
while for the remaining $80k$ iterations a learning rate of $0.0001$ was set.
For both the J-HMDB-21 and the LIRIS-HARL datasets, we trained both RPN and Fast R-CNN networks for $180k$ iterations.
For the first $120k$ iterations a learning rate of $0.001$ was used - for the remaining $60k$ iterations,
we set the learning rate to $0.0001$.
The momentum was set to a constant value of $0.9$, while weight decay was fixed to $0.0005$.

\paragraph{Stochastic Gradient Descent mini-batch size.}
We selected an SGD mini-batch size of 256 for RPN, and of 128 for Fast R-CNN training.

\paragraph{CNN training.}
First, we trained an RPN network with either a set of RGB or optical flow-based training video frames.
At each training iteration, the RPN would take as input a video frame and its associated ground truth bounding boxes.
Once the RPN net was trained, we used the trained model to extract frame-level region proposals.
A trained RPN net outputs a set of region proposals (around $16k$ to $17k$) per frame
and their associated ``actionness'' scores.
We then filtered these region proposals using non-maximal suppression (NMS) and 
selected top $2k$ proposals based on their ``actionness'' scores.
These top $2k$ region proposals along with the frame and its ground truth boxes were then passed to a Fast R-CNN for training.

\paragraph{CNN testing.}
Once training both RPN and Fast R-CNN networks, we extracted region proposals
from test video frames using the learnt RPN model.
Similarly to what done in the training stage, we filtered the region proposals using NMS - 
however, at test time, we chose the top $300$ region proposals and passed them to the Fast R-CNN network to obtain
the final detection boxes: a set of $300 \times C$ regressed boxes and their associated softmax probability scores (where 
$C$ is the number of ground truth action categories in a given dataset).
For each action category, we first filtered the detection boxes using NMS and then selected the top $5$ boxes per frame based on their 
softmax probability scores.
We used an NMS threshold of $0.7$ to filter the RPN-generated region proposals, and a threshold of $0.3$ when filtering the Fast R-CNN detection boxes.

\paragraph{Implementation details for temporal detection on Activitynet.}
In this case we used the features provided on ActivityNet's~\cite{caba2015activitynet} 
web page\footnote{\url{http://activity-net.org/challenges/2016/download.html}}. 
More specifically, these video-level features comprised of two sets of features.
First, ImageNetShuffle features, which are video-level features generated by~\cite{mettesicmr16} using a Google inception net (GoogLeNet~\cite{szegedy2015going}).
Second, Motion Boundary Histogram (MBH) features, which are generated with the aid of the 
improved trajectories~\cite{wang2013improvedtraj} executable\footnote{\url{http://lear.inrialpes.fr/people/wang/improved\_trajectories}}.
Video level features were passed to an SVM classifier for video classification.
Finally, C3D features were generated at 2 frames per second using a C3D network~\cite{tran2014learning} with a temporal resolution of 16 frames.
The frame-level features were then used for frame-level classification with a random forest classifier~\cite{breiman2001random} to produce frame-level scores. 
The frame-level scores of top 5 video-level categories were fed to the temporal detection formulation described in Section~\ref{offline:sec:temporaltrimming} to obtain temporal action detection results.
We refer to~\cite{singh2016untrimmed} for more details. The working code is available on our public GitHub repository\footnote{\url{https://github.com/gurkirt/actNet-inAct}}.

%% file: chapters/4_online_detection.tex
\chapter{Online and Real-time Action Tube Detection}
\renewcommand{\imagepath}{figures/online/} 
\label{chapter:online}

\input{chapters/partI/online_intro}
\input{chapters/partI/online_approach}
\input{chapters/partI/online_exp}

\section{Summary and limitations}\label{summary:online}
\paragraph{Summary: }
We presented a first and novel online framework for action detection and prediction able to 
address the challenges involved in concurrent multiple human action recognition, spatial detection and temporal detection, in real-time.
Thanks to an efficient deep learning strategy for the simultaneous detection and 
classification of region proposals and a new incremental action tube generation approach,
our method achieves superior performances compared to the previous state-of-the-art 
on early action prediction and online detection, 
while outperforming the top offline competitors, in particular at high detection overlap.
Its combination of high accuracy and fast detection speed at test time paves 
the way for its application to real-time applications such as autonomous driving, 
human-robot interaction and surgical robotics, among others.




In summary, we present a holistic framework for the real-time, online spatial 
and temporal detection of multiple action instances in videos which:
\begin{itemize}[leftmargin=0.5cm,noitemsep,nolistsep]
  \item incorporates the SSD~\cite{liu2016ssd} neural architecture to improve the speed to predict frame-level detection boxes and the associated action class-specific
  confidence scores, in a single-stage regression and classification approach (\S~\ref{online:sec:intg_det_net});
  \item devises an original, greedy online algorithm capable of generating multiple action tubes incrementally (\S~\ref{online:sec:tubeconstruction} - Fig.~\ref{online:fig:algorithmOverview}(g));
  \item provides early action class label predictions (\S~\ref{online:sec:action_prediction}) and online spatiotemporal detection (\S~\ref{online:sec:tubeconstruction})
  from partially observed action instances in untrimmed videos;%

  \item functions in real-time, while outperforming the previous (offline) state of the art on the untrimmed videos of UCF101-24 dataset. 
\end{itemize}

\paragraph{Limitations:}\label{online:limitations}
as for the previously illustrated offline action tubes detection method, the proposed online method retains the same limitations in terms of temporal reasoning, for the latter is limited to the tube construction method (\S~\ref{online:sec:tubeconstruction}) and the optical flow stream (\S~\ref{online:subsec:fusion}).

\paragraph{Looking ahead:}
in the next chapter, we propose to solve these limitations by devising a closer-to-optimal formulation for the entire action tube detection problem. 
We propose to address tube detection by dividing it into a set of sub-problems, in which the detection of a `micro-tube' from a small set of consecutive frames is one of the sub-problems.
As shown in Chapter \ref{chapter:tramnet}, the proposed solution also helps with the linking formulation.

%% file: chapters/partI/online_intro.tex
\section{Introduction} \label{online:intro}

It has been observed, both in the previous chapter (Chapter~\ref{chapter:offline}) and in
other works~\cite{weinzaepfel2015learning,saha2016deep,peng2016eccv}, that detecting human action tubes in unconstrained videos is challenging,
owing to their deformable, spatiotemporal nature. 
The problem is made even harder if the detection is to be performed in an online setting and at real-time speed.
Despite the performance of state-of-the-art spatiotemporal action detection systems (as described in the previous chapter and in~\cite{saha2016deep,peng2016eccv}) being far from real-time,
current systems also assume that the entire video (taken as a 3D block of pixels) is available ahead of time to detect action instances.
As previously explained in the limitation Section~\ref{offline:summary} of the previous chapter, in order for such a detector to be applicable in real-world scenarios, such as video surveillance and human-robot interaction,
video frames need to be processed in real-time and in an online fashion.
Moreover, the action detection system needs to generate action tubes in an incremental and online fashion, as each new frame is captured.

The results from Chapter~\ref{chapter:offline} show that supervised region proposal generation is crucial to improve detection accuracy.
Although, \cite{saha2016deep,peng2016eccv} exhibit the excellent spatiotemporal action detection accuracy (\S~\ref{offline:exp:subsec:st-detection}), 
their test time detection strategy involves using computationally expensive optical flow frames~\cite{brox2004high}. In addition, these approaches require a two step process, composed by
a region proposal network (RPN)~\cite{ren2015faster} and an R-CNN formulation~\cite{girshick2014rcnn}, limiting real-time deployment.
Also, \cite{saha2016deep,peng2016eccv} both employ offline tube generation methods which process the entire video in two passes:
the first pass links up detection boxes into paths which stretch from start to end of the video;
the second pass is there to temporally trim the video-long constructed paths down to proper action tubes.
Finally, these methods' mAP at high detection overlap values remains disappointingly low, and incompatible with many real-world applications.

\subsection{An online, real-time framework}

\begin{figure}[t]
  \centering
  \includegraphics[scale=0.25]{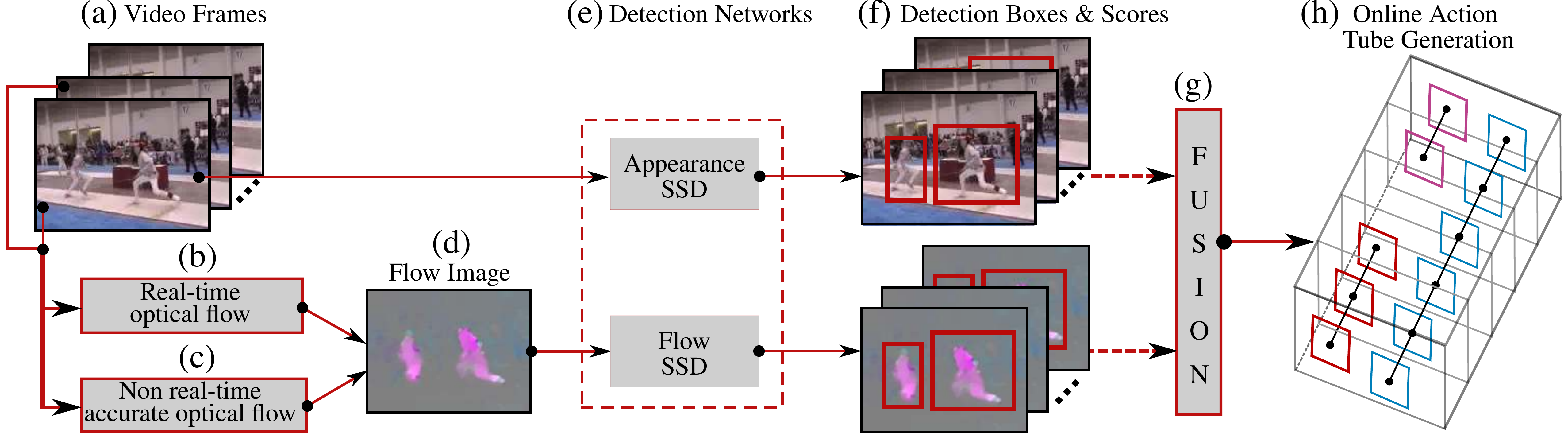}
  \caption[Overview of online and real-time action detection framework.]{
        At test time, the input to the framework is a sequence of RGB video frames \textbf{(a)}.
        A real-time optical flow (OF) algorithm \textbf{(b)} \cite{kroeger2016fast} takes the consecutive RGB frames as input to produce flow images \textbf{(d)}.
        As an option, \textbf{(c)} a more accurate optical flow algorithm 
        \cite{brox2004high} can be used
        (although not in real time).
        \textbf{(e)}~RGB and OF images are fed to two separate SSD detection~\cite{liu2016ssd} networks (\S~\ref{online:sec:intg_det_net}).
        \textbf{(f)}~Each network outputs a set of detection boxes along with their class-specific confidence scores.
        \textbf{(g)}~Appearance and flow detections are fused (\S~\ref{online:subsec:fusion}). 
        Finally \textbf{(h)}, multiple action tubes are built up in an online fashion 
        by associating current detections with partial tubes (\S~\ref{online:sec:tubeconstruction}).
        }
  \label{online:fig:algorithmOverview} 
\end{figure}

In this chapter, we illustrate an online framework, originally proposed in~\cite{singh2017online} and outlined in Figure~\ref{online:fig:algorithmOverview}, which overcomes all the above limitations. 
More specifically we propose three fundamental changes compared to the (then) existing action tube detection pipelines~\cite{saha2016deep,peng2016eccv} to achieve real-time processing and overcome the offline nature of the tube building process.

Firstly, the pipeline takes advantage of the SSD (Single Shot MultiBox Detector) object detector~\cite{liu2016ssd} to address issues with accuracy and speed in previous~\cite{ren2015faster} frame-level detection networks.
This is possible as SSD eliminates the region proposal generation step and is a single-stage, end-to-end trainable detection network. 

Secondly, to leverage the performance of  SSD in an online fashion, we design a novel single-pass online tube building method which leads to both superior accuracy (compared to~\cite{weinzaepfel2015learning,saha2016deep,peng2016eccv}), especially at realistic detection precision, and better real-time detection speed.
Unlike previous tube-generation approaches~\cite{georgia2015tube,saha2016deep,peng2016eccv,weinzaepfel2015learning} our algorithm works in an online fashion as tubes are updated frame by frame, together with their overall action-specific scores and labels.
As soon as non-real-time optical flow ~\cite{brox2004high} is replaced by a
less accurate (but real-time) optical flow algorithm \cite{kroeger2016fast}, the resulting system performs in real-time (28fps), with just a little performance degradation, 
an essential feature for real-world applications.

Lastly, we show that we can replace non-real-time optical flow ~\cite{brox2004high} with 
less accurate but real-time optical flow proposed by \cite{kroeger2016fast} 
in two stream setting (shown in ~\ref{online:fig:algorithmOverview}) without sacrificing much performance.
Although a CNN trained on real-time optical flow~\cite{kroeger2016fast} does not lead to the same performance as to the network trained on more accurate optical flow~\cite{brox2004high} by itself, but when used in a two-stream setting (in which detections from the RGB and the OF streams are fused) it can provide a similar boost as provided by an accurate optical flow~\cite{brox2004high} network in parallel to the appearance stream.

\subsection{Extension to early label predictions}

The incremental nature of our system makes it possible to accurately foresee the class label of an entire test video and detect action instances within it by just observing a small fraction of frames (the \emph{early action prediction and detection} problem, which we described in the Introduction).
Such a system was proposed by
Soomro~\etal~\cite{soomro2016predicting}, who showed that both action prediction and online detection performance improve over time as more and more video frames become available.
Using \cite{soomro2016predicting} as a baseline, we report here new state-of-the-art results
on the temporally trimmed J-HMDB-21 videos.
Furthermore, compared to Soomro~\etal~\cite{soomro2016predicting}, we are able to
demonstrate action prediction and detection capabilities 
from partially observed \emph{untrimmed} streaming videos on the challenging UCF101-24 dataset,
while retaining real-time detection speeds. 

In summary, we demonstrate that our online approach:
\begin{itemize}[leftmargin=0.5cm,noitemsep,nolistsep]
    \item significantly outperforms current offline methods, especially on realistic detection thresholds of $0.5$ or greater;
    \item is capable of superior early action prediction performance compared to the existing baseline~\cite{soomro2016predicting};
    \item achieves a real-time detection speed (up to 40fps), that is 5 to 6 times faster than previous works. 
\end{itemize}

\paragraph{Related publications:}
The work presented in this chapter appeared in ICCV 2017~\cite{singh2017online}. 
The dissertation author was the primary investigator in~\cite{singh2017online}.
Our code is also available online\footnote{\url{https://github.com/gurkirt/realtime-action-detection}}. Our statistics show that the code has been forked 60 times and stared 188 times in GitHub.

\paragraph{Outline:}\label{online:outline}
This chapter is organised as follows.
We start by presenting an overview of our action tube detection approach in Section~\ref{online:sec:overview}. Next, we describe in detail the three steps we introduced in this new online action tube detection framework.
We first describe the integration of the Single-Shot Detector network~\cite{liu2016ssd} in Section~\ref{online:sec:intg_det_net}, which is used to predict detection boxes and class-specific confidence scores for appearance and flow frames independently. 
Secondly, we describe our proposed novel online tube construction algorithm in Section~\ref{online:sec:tubeconstruction}.
The third step (\S~\ref{online:sec:opticalflow}) is where we describe the efficient computation of optical flow. Section~\ref{online:sec:action_prediction} illustrates, in particular, how the online tube construction algorithm can be used for early action label prediction.
Next, we report the experimental validation of the proposed model 
(Section~\ref{online:sec:exp}) from both a qualitative and a quantitative perspective.
Finally, we present a summary and the approach's limitations in Section~\ref{summary:online}.

%% file: chapters/partI/online_approach.tex
\section{Overview of the approach}\label{online:sec:overview}

The proposed framework is outlined in 
Figure~\ref{online:fig:algorithmOverview}. 
We train two SSD detection networks~\cite{liu2016ssd}:
one on RGB frames (appearance), the other on optical-flow images~\cite{georgia2015tube}.
At test time, the input to the framework is a sequence of RGB video frames (Fig.~\ref{online:fig:algorithmOverview}(a)).
A real-time optical flow algorithm ~\cite{kroeger2016fast} takes the RGB frames as input to produce flow images (Fig.~\ref{online:fig:algorithmOverview}(d)).
As an option, more accurate OF (Fig.~\ref{online:fig:algorithmOverview}(c)) images can be computed in a non real-time setting~\cite{brox2004high} (but still incrementally).
For each pipeline, the related detection network (Fig.~\ref{online:fig:algorithmOverview}(b)) takes as input
an RGB (or OF) frame, and outputs a set of `regressed' detection boxes (Fig.~\ref{online:fig:algorithmOverview}(f)), 
together with a set of class-specific confidence scores for each box 
indicating the likelihood of an action class being present there. 
We then we merge (Fig.~\ref{online:fig:algorithmOverview}(g)) the class-specific confidence scores associated with the flow-based and the appearance-based detection boxes,
as this significantly boosts detection accuracy.
Finally, sequences of detection boxes are linked up to form action-specific tubes in a completely online, incremental fashion (Fig.~\ref{online:fig:algorithmOverview}(h)). 

The resulting framework is the first method to perform action tube detection in an online and real-time fashion. 
Unlike those in \cite{georgia2015tube}, the resulting action tubes are not constrained to span the entire video duration. 
Unlike what happens in \cite{weinzaepfel2015learning,saha2016deep,peng2016eccv}, tubes are built in an online fashion allowing 
for early action label prediction and detection.
Usage of real-time optical flow~\cite{kroeger2016fast}, and SSD detection networks~\cite{liu2016ssd} allows us ot process videos in streaming (online and real-time) fashion. 

Our framework is the first with a demonstrated ability to perform online spatiotemporal action detection.

\section{Integrated detection network} \label{online:sec:intg_det_net}

We use a single-stage convolutional neural network (Fig.~\ref{online:fig:algorithmOverview}(e)) for bounding box prediction and classification,
which follows an end-to-end trainable architecture proposed in~\cite{liu2016ssd}.
The architecture unifies a number of functionalities in single CNN 
which are, in other action and object detectors, performed
by separate components~\cite{georgia2015tube,weinzaepfel2015learning,ren2015faster,saha2016deep}, namely: 
{(i)} region proposal generation, 
{(ii)} bounding box prediction and 
{(iii)} estimation of class-specific confidence scores for the predicted boxes.
This allows for relatively faster training and higher test time detection speeds.

\subsection{Detection network design and training.}\label{online:subsec:training}

For our integrated detection network we adopt the network design and architecture of the SSD~\cite{liu2016ssd} object detector, 
with an input image size of $300\times300$.
We do not use the $512\times512$ SSD architecture~\cite{liu2016ssd}, as detection speed is much slower~\cite{liu2016ssd}.
As in~\cite{liu2016ssd}, we also use an ImageNet pretrained VGG16 network 
provided by~\cite{liu2016ssd} (\url{https://gist.github.com/weiliu89/2ed6e13bfd5b57cf81d6}).
We adopt the training procedure described by ~\cite{liu2016ssd} along with their
publicly available code for network training (\url{https://github.com/weiliu89/caffe/tree/ssd}).
We use a learning rate of $0.0005$ for the appearance stream and of $0.0001$ for the flow stream on UCF101-24, 
whereas that for J-HMDB-21 is set to $0.0001$ for both appearance and flow. 

\subsection{Fusion of appearance and flow cues}\label{online:subsec:fusion}

The detection boxes generated by the appearance and flow detection networks (Fig.~\ref{online:fig:algorithmOverview}(f))
need to be merged to improve robustness and accuracy (Fig.~\ref{online:fig:algorithmOverview}(g)).
We conducted experiments using two distinct fusion strategies.

\noindent
\textbf{Boost-fusion}. As in the previous chapter, here we follow the approach in~\cite{saha2016deep}, with a minor modification.
Firstly, we perform L-1 normalisation on the detection boxes' scores after fusion.
Secondly, we retain any flow detection boxes for which an associated appearance based box was not found,
as we found that discarding the boxes lowers the overall recall.

\noindent
\textbf{Fusion by taking the union-set.}\label{subsec:fusion_by_union}
 A different, effective fusion strategy consists in retaining the union  
$\{b^{a}_i\}\cup\{b^{f}_j\}$ of the two sets of appearance $\{b^{a}_i\}$ and flow $\{b^{f}_j\}$ detection boxes, respectively.
The rationale is that in UCF101-24, for instance, several action classes
(such as `Biking', `IceDancing', or `SalsaSpin') have concurrent action
instances in the majority of 
video clips: an increased number of detection boxes may so
help to detect concurrent action instances.

\section{Online action tube generation} \label{online:sec:tubeconstruction}

\subsection{Problem formulation}
Given a set of detections at time $t=1, \ldots,T$, for each given action class $c$, we seek the sets of consecutive detections (or \emph{action tubes}) $\mathcal{T}_c = \{ b_{t_s}, \ldots,b_{t_e} \}$ which, among all possible such collections, are more likely to constitute an action instance. This is done separately for each class, so that the results for class $c$ do not influence those for other classes.
We allow the number of tubes $n_c(t)$ to vary in time, within the constraint given by the number of available input detections. We allow action tubes to start or end at any given time.
Finally, we require: (i) consecutive detections part of an action tube to have spatial overlap above a threshold $\lambda$; (ii) each class-specific detection to belong to a single action tube; (iii) the online update of the tubes' temporal labels.
\\
Previous approaches to the problem \cite{georgia2015tube,saha2016deep} would constrain tubes to span the entire video duration. In both \cite{saha2016deep} and \cite{peng2016eccv}, in addition, action paths are temporally trimmed to proper action tubes using a second pass of dynamic programming.

In opposition, we propose a simple but efficient online action tube generation algorithm which incrementally (frame by frame) builds multiple action tubes for each action class in parallel.  
Action tubes are treated as `tracklets', as in multi-target tracking approaches~\cite{milan2016multi}. 
We propose a greedy algorithm (\ref{online:subssec:algo}) 
similar to \cite{majecka2009statistical,singh2010categorising} 
for associating detection boxes in the upcoming frame with the current set of (partial) action tubes. 
Concurrently, each tube is temporally trimmed in an online temporal labelling (\ref{online:subssec:temporallabelling}) setting.


\subsection{A novel greedy algorithm}\label{online:subssec:algo}

The input to the algorithm is the fused frame-level detection boxes
with their class-specific scores (Sec. \ref{online:subsec:fusion}). 
{At each time step $t$, the top $n$ class-specific detection boxes $\{b_{c}\}$ are selected by
applying non-maximum suppression on a per-class basis.
At the first frame of the video, $n_c(1) = n$ action tubes per class $c$ are initialised using the $n$ detection boxes at $t=1$.
The algorithm incrementally grows the tubes over time by adding one box at a time.
The number of tubes $n_c(t)$ varies with time, as new tubes are added and/or old tubes are terminated.

At each time step, we sort the existing partial tubes so that the best tube can potentially match the best box from the set of detection boxes in the next frame $t$.
Also, for each partial tube $\mathcal{T}_{c}^i$ at time $t-1$, we restrict the potential matches to detection boxes at time $t$ whose IoU (Intersection over Union) 
with the last box of $\mathcal{T}_{c}^i$ is above a threshold $\lambda$. In this way, tubes cannot simply drift off, and they can be terminated if no matches are found for $k$ consecutive frames.  
Finally, each newly updated tube is temporally trimmed by
performing a binary labelling 

Summarising, 
action tubes are constructed by applying the following 7 steps to every new frame at time~$t$:}
\begin{enumerate} [topsep=0pt,itemsep=-1ex,partopsep=1ex,parsep=1ex]
 \item Execute steps 2 to 7 for each class $c$.
 \item Sort the action tubes generated up to time $t-1$ in decreasing order, based on 
 the mean of the class scores of the tube's member detection boxes.
 \item {\footnotesize \textbf{LOOP START: }} $i = 1 $ to $n_c(t-1)$ - traverse the sorted tube list.
 \item Pick tube $\mathcal{T}_{c}^i$ from the list and find a matching box for it
 among the $n$ class-specific detection boxes $\{b_{c}^j, j=1,...,n\}$ at frame $t$ based on the following conditions:
 \begin{enumerate} [topsep=0pt,itemsep=-1ex,partopsep=1ex,parsep=1ex]
  \item {for all $j=1,...,n$, if the IoU between the last box of tube $\mathcal{T}_c^{i}$ and the detection box $b_{c}^{j}$
   is greater than $\lambda$, then add it to a potential match list $\mathcal{B}^i$;}
 \item if the list of potential matches is not empty, $\mathcal{B}^i \neq \emptyset$, select the box $b_{c}^{max}$ from $\mathcal{B}^i$ 
  with the highest score for class $c$ as the match, and remove it from the set of available detection boxes at time $t$;
 \item if $\mathcal{B}^i = \emptyset$, retain the tube anyway, without adding any new detection box, unless more than $k$ frames have passed with no match found for it.
 \end{enumerate}
 \item Update the temporal labelling 
 for tube $\mathcal{T}_c^{i}$ using the score $s(b_{c}^{max})$ of the selected box $b_{c}^{max}$ (see \S~\ref{online:subssec:temporallabelling}).
 \item {\footnotesize\textbf{LOOP END }}
 \item If any detection box is left unassigned, start a new tube at time $t$ using this box.
\end{enumerate}
In all our experiments, we set $\lambda = 0.1$, $n = 10$, and $k=5$.

\subsection{Temporal labelling} \label{online:subssec:temporallabelling}

Although $n$ action tubes per class are initialised at frame $t=1$, we want
all action specific tubes to be allowed to start and end at any arbitrary time points $t_s$ and $t_e$.
The online temporal relabelling step 5 in the above algorithm is designed to take care of this.

Similarly to \cite{saha2016deep,evangelidis2014continuous} and to Chapter~\ref{chapter:offline}'s approach, 
each detection box $b_r$, $r=1,...,T$ in a tube $\mathcal{T}_c$, 
where $T$ is the current duration of the tube and $r$ is its temporal position within it, is assigned
a binary label \mbox{${l}_{r} \in \{c,0\}$}, 
where $c$ is the tube's class label and $0$ denotes the background class.
The temporal trimming of an action tube thus reduces to finding an optimal binary labelling $\mathbf{l} = \{ l_1,...,l_T \}$ for all the constituting bounding boxes.
This can be achieved by maximising for each tube $\mathcal{T}_c$ the energy:
\begin{equation} \label{eqn:secondpassenergy}
E(\mathbf{l}) =  \sum_{r=1}^T s_{l_r}({b}_r) - \alpha_{l} \sum_{r=2}^T \psi_l \left(\Scalar{l}_{r}, \Scalar{l}_{r-1} \right) ,
\end{equation}
where $s_{l_r}(b_r) = s_{c}(b_r)$ if $l_r=c$, $1-s_{c}(b_r)$ if $l_r=0$,
$\alpha_{l}$ is a scalar parameter, and
the pairwise potential $\psi_l$ is defined as: $\psi_l(l_r,l_{r-1}) = 0$ if $l_r = l_{r-1}$, $\psi_l(l_r, l_{r-1}) =\alpha_c$ otherwise.

\subsubsection{Online Viterbi}

The maximisation problem~(\ref{eqn:secondpassenergy}) can be solved by Viterbi dynamic programming~\cite{saha2016deep,evangelidis2014continuous}.
An optimal labelling $\hat{\mathbf{l}}$ for a tube $\mathcal{T}_c$ can be generated 
by a Viterbi backward pass at any arbitrary time instant $t$ in linear time.
That, however, would require a backward pass all the way back to the start of each action path, which is expensive to perform for every action path at each time step. 
Instead, we keep track of the minimum cost edges from the past box-to-tube associations 
from the start of the tube up to $t-m$,
which eliminates the requirement for an entire backward pass at each time step but only up to the previous m steps. 
In the following we used $m$ equal to 5.
This makes temporal labelling very efficient, and suitable to be used in an online fashion. 
This can be further optimised for much longer videos by finding the coalescence point \cite{vsramek2007line}.
As stated in step 5 above, the {temporal labelling} of each tube is updated at each time step whenever a new box is added.

\section{Optical flow computation} \label{online:sec:opticalflow}

The input to our framework is a sequence of RGB images. 
As in prior work in action detection~\cite{saha2016deep,georgia2015tube,weinzaepfel2015learning},
we use a two-stream CNN approach \cite{simonyan2014twostream} in which optical flow and appearance are processed in two parallel, distinct streams.
As our aim is to perform action detection in real-time, 
we employ real-time optical flow (Fig.~\ref{online:fig:algorithmOverview}(b))
\cite{kroeger2016fast} to generate the flow images (Fig.~\ref{online:fig:algorithmOverview}(d)).
As an option, one can compute optical flow more accurately (Fig.~\ref{online:fig:algorithmOverview}(c)),
using Brox \etal's \cite{brox2004high} method.
We thus train two different networks for the two OF algorithms, 
while at test time only one network is used 
depending on whether the focus is on speed rather than accuracy.
Following the transfer learning approach on motion vectors of \cite{zhangcvpr2016}, 
we first train the SSD network on accurate flow results, 
to later transfer the learned weights to initialise those of the real-time OF network. 
Performance would degrade whenever transfer learning was not used.

\section{Early action prediction} \label{online:sec:action_prediction}

As for each test video multiple tubes are built incrementally at each time step $t$ (\S \ref{online:sec:tubeconstruction}),
we can predict at any time instant the label of the whole video as the label of the current highest-scoring tube, where
the score of a tube is defined as the mean of the tube boxes' individual detection scores:

\begin{equation}
  \hat{c}(t) = \arg \max_{c} \left ( \max_{\mathcal{T}_c} \frac{1}{T}\sum_{r=1}^T s(b_r) \right ).
\end{equation}

%% file: chapters/partI/online_exp.tex
\section{Experiments} \label{online:sec:exp}

We test our online framework 
on three separate challenging setups: 

\noindent
i) early action label prediction (\S~\ref{online:exp:subsec:prediction});

\noindent
ii) online spatiotemporal action detection over time (\S~\ref{sec:online-st-detection});

\noindent
iii) an offline setting in which entire video is observed, so that the global spatiotemporal action detection performance can be assessed (\S~\ref{sec:offline-st-detection}), including a comparison with offline action detection methods.


\vskip 2mm
\noindent
In all settings we generate results by running our framework in five different `modes':
\begin{enumerate}
  \item \emph{Appearance (A)} -- only RGB video frames are processed by a single SSD;
  \item \emph{Real-time flow (RTF)} -- optical flow images are computed in real-time~\cite{kroeger2016fast} and fed to a single SSD;
  \item \emph{A$+$RTF}: both RGB and real-time optical flow images are processed by  SSD in two separate streams;
\item \emph{Accurate flow (AF)} optical flow images are computed as in~\cite{brox2004high}, and
\item \emph{A$+$AF}: both RGB and non-real-time optical flow frames \cite{brox2004high} are used.  
\end{enumerate}

Modes 1), 2) and 3) run in real-time whereas modes 4) and 5)'s performances are non-real-time (while still working incrementally), 
due to the relatively higher computation time needed to generate accurate optical flow.
Note that all experiments are conducted in a fully online fashion, using the proposed tube generation algorithm (\S~\ref{online:sec:tubeconstruction}).

In Section~\ref{online:exp:contribution-of-tubeconstruction}, we see how our online tube construction method contributes to the overall performance of our method.
Next, we ablate the contribution of the different optical methods in Section~\ref{online:exp:flow-contribution}.
Lastly, we provide evidence of real-time capability in Section~\ref{sec:detection-speed}.


As in the previous chapter and in previous related works~\cite{saha2016deep,yu2015fast,peng2016eccv,weinzaepfel2015learning}
we evaluate our model on the UCF101-24~\cite{ucf101} and 
J-HMDB-21~\cite{jhmdbj2013towards} benchmarks.
Please refer to Section~\ref{soa:datasets} in Chapter~\ref{chapter:related_work} for more details on these dataset.
We also used ActivityNet~\cite{activitynet2015heilbron} dataset to evaluate temporal action detection performance, more details about the dataset are available in Section~\ref{soa:datasets} in Chapter~\ref{chapter:related_work}.

\textbf{Evaluation metrics.}
For the early action label prediction (\S~\ref{online:exp:subsec:prediction}) and 
the online action detection (\S~\ref{sec:online-st-detection}) tasks we follow
the experimental setup of~\cite{soomro2016predicting}, and use  the
traditional detection metrics, such as AUC (area under the curve) and mAP (mean average precision).
We report performance as a function of \emph{Video Observation Percentage}, i.e.,
with respect to the portion (\%) of the entire video observed before predicting action label and
location.

We also report a performance comparison with the relevant offline methods~\cite{saha2016deep,yu2015fast,peng2016eccv,weinzaepfel2015learning} 
using the mean-average-precision (mAP) at various detection thresholds ($\delta$), as explained in Section~\ref{soa:metrics} in Chapter~\ref{chapter:related_work}.
Also, similarly to the evaluation protocol for object detection in the Microsoft COCO~\cite{lin2014microsoft} dataset, we propose to use the \emph{``average detection performance''} (avg-mAP) to compare our performance with that of the previous state-of-the-art.
To obtain the latter, we first compute the video-mAPs at higher IoU thresholds ($\delta$) ranging $[0.5:0.05:0.95]$, and then take the average of these video-mAPs.

\begin{figure}[h]
  \centering
  \includegraphics[width=0.7\textwidth]{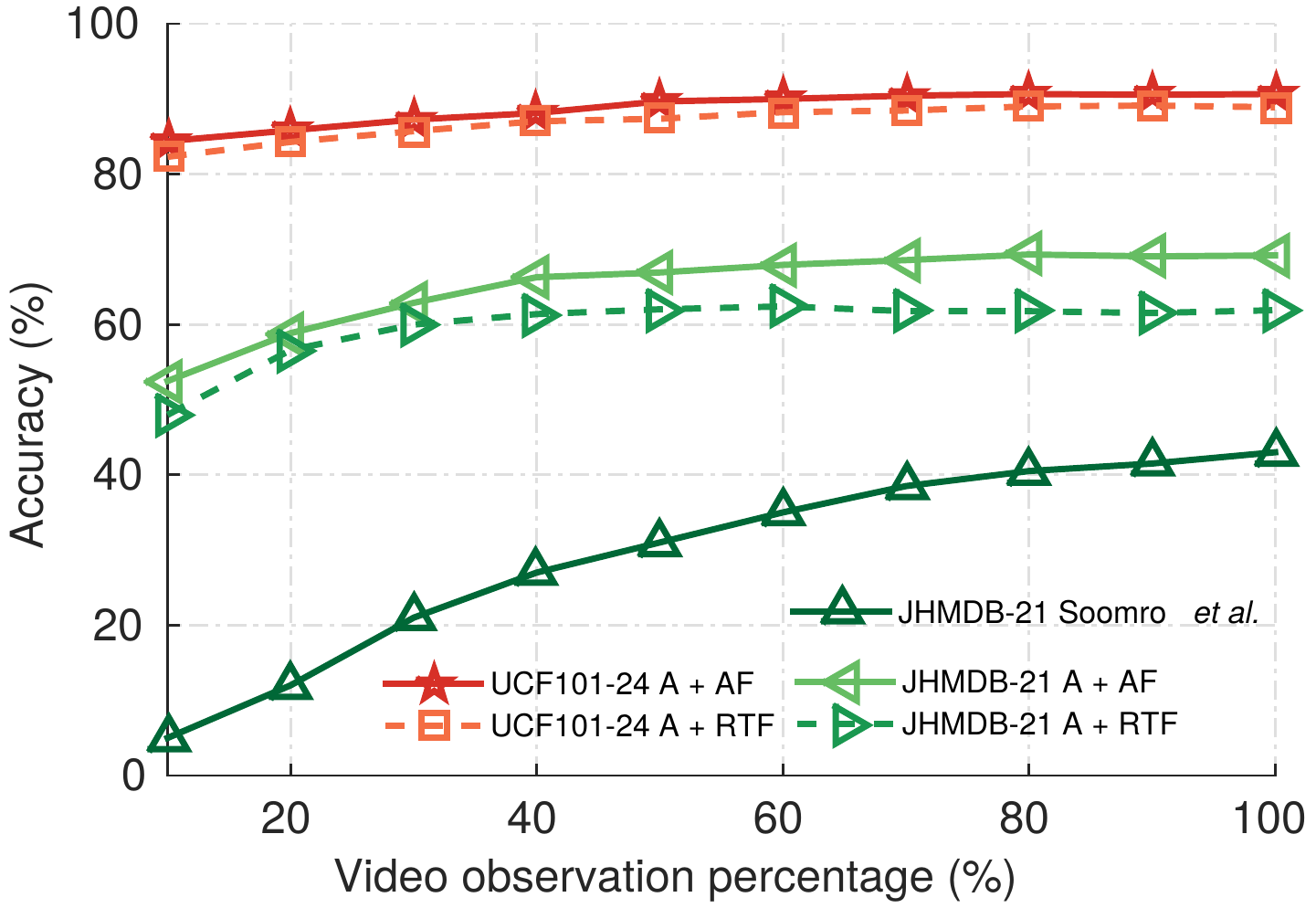}  
  \caption[Early action label prediction results.]{
        Early action label prediction results (accuracy \%) over the UCF101-24 and J-HMDB-21 datasets.
  }
  \label{online:fig:prediction} 
\end{figure}

\subsection{Early action label prediction} \label{online:exp:subsec:prediction}

Here, we need to drive the inference (obtain the video action label) before the action is over (see Section~\ref{online:sec:action_prediction} for video-level score computation).
Although action tubes are computed by our framework frame by frame, we sample them
at 10 different time `check-points' on each video,
starting at $10\%$ of the total number of video frames and with a step size of $10\%$.
We use the union-set and boost fusion strategies~(\S~\ref{online:subsec:fusion}) for UCF101-24 and J-HMDB-21, respectively.
Fig.~\ref{online:fig:prediction} 
compares the early action prediction accuracy of our approach with that of~\cite{soomro2016predicting}, 
as a function of the portion (\%) of video observed. 
Our method clearly demonstrates superior performance, 
as it is able to predict the actual video label by observing a very small portion of the entire video at a very initial stage. 
For instance,  by observing only the initial $10\%$ of the videos in J-HMDB-21, 
we are able to achieve a prediction accuracy of $48\%$ as compared to $5\%$ by Soomro~\etal~\cite{soomro2016predicting}, 
which is, in fact, higher than the $43\%$ accuracy achieved by \cite{soomro2016predicting} after observing the \emph{entire} video.
{We do not run comparisons with the early action prediction work by Ma \etal \cite{ma2016learning} for they only show results on ActivityNet~\cite{caba2015activitynet}, as dataset which has only temporal annotations. The early prediction capability of our approach is a sub-product of its being online, as in~\cite{soomro2016predicting}: thus, we only compare ourselves with Soomro~\etal~\cite{soomro2016predicting} re early action prediction results.}

Compared to \cite{soomro2016predicting} we take one step further, 
and perform early label prediction on the untrimmed videos of UCF101-24 as well (see Fig.~\ref{online:fig:prediction}).
It can be noted that our method performs much better on UCF101-24 than on J-HMBD-21 at the prediction task.
This relatively higher performance may be attributed to the larger number of training examples, subject to more modes of variations, 
present in UCF101-24, which improves the generalisation ability of the model and prevents it from overfitting.
Interestingly, we can observe that the performances of the real-time (A $+$ RTF) 
and non-real-time (A $+$ AF) modalities 
are quite similar, which suggests that accurate optical flow might be not so crucial for action classification on UCF101-24 dataset,
which strongly supports our framework's ambition to accurately solve the problem in 
real-world situations in which real-time speed is required. 

\begin{figure}[h]
  \centering
  \includegraphics[width=0.7\textwidth]{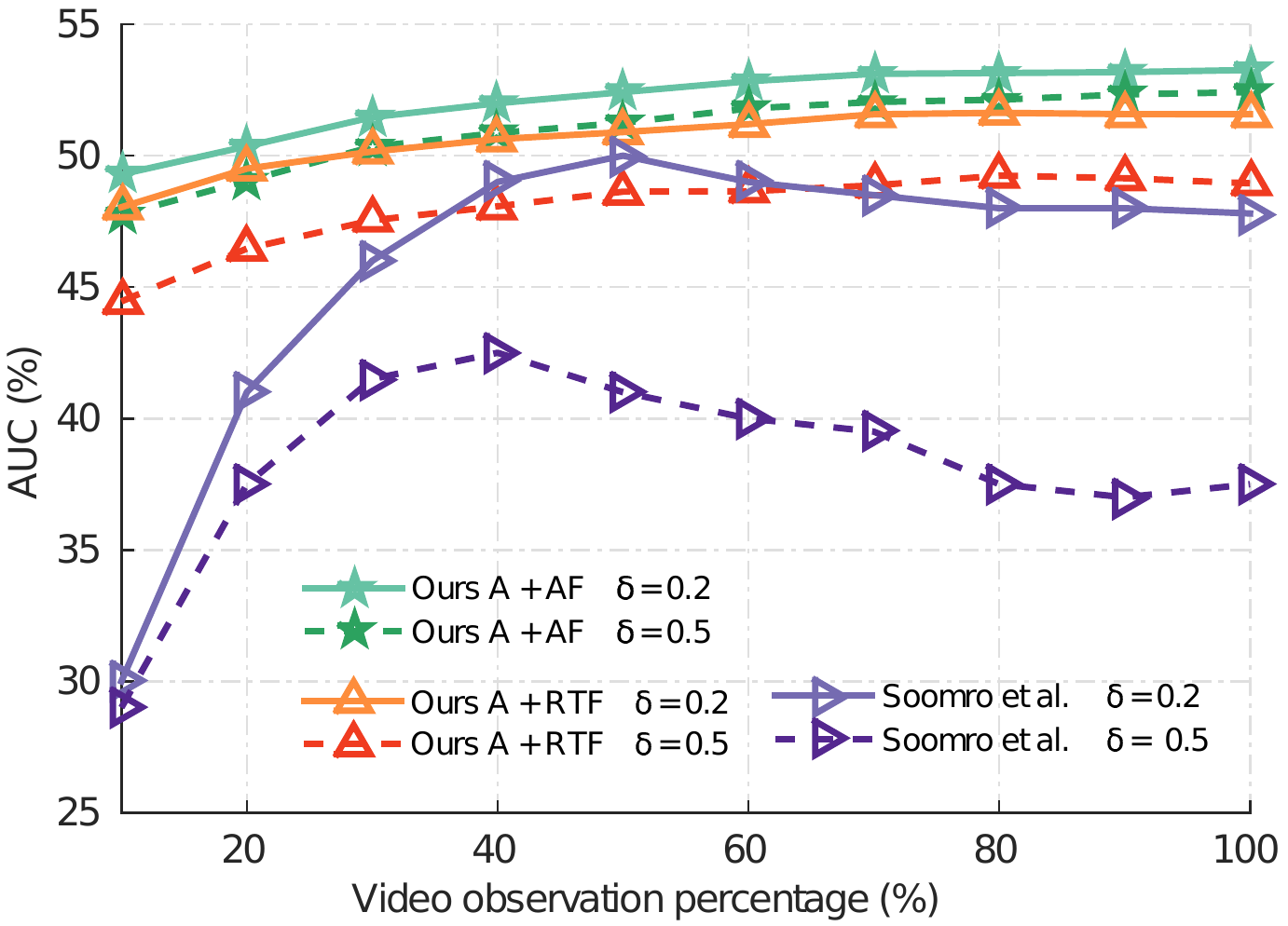}  
  \caption[Online action detection results using the AUC (\%) metric on J-HMDB-21.]{
        Online action detection results using the AUC (\%) metric on J-HMDB-21, at IoU thresholds of $\delta=0.2$, $0.5$.
  }
  \label{online:fig:detection_plot_jhmdb} 
\end{figure}

\subsection{Spatiotemporal performance over time}\label{sec:online-st-detection}

Our action tubes are built incrementally and carry associated labels and scores at each time step.
At any arbitrary time $t$, we can thus compute the spatiotemporal IoU between the tubes generated by our online algorithm
and the ground truth tubes, up to time $t$.


Figure~\ref{online:fig:detection_plot_jhmdb} plots the AUC curves 
against the observed portion of the video at different IoU thresholds ($\delta = 0.2$  and $0.5$)
for the proposed approach versus our competitor~\cite{soomro2016predicting}.
Our method outperforms~\cite{soomro2016predicting} on online action detection
by a large margin at all the IoU thresholds and video observation percentage.
Notice that our online detection performance (Fig.~\ref{online:fig:detection_plot_jhmdb}) is a stable 
function of the video observation percentage, 
whereas Soomro~\etal~\cite{soomro2016predicting}'s method needs some `warm-up' time to 
reach stability, and its accuracy slightly decreases at the end.
In addition, \cite{soomro2016predicting} only reports online spatial detection results
on the temporally trimmed J-HMDB-21 test videos, and their approach lacks  
temporal detection capabilities.

\begin{figure}[h]
  \centering
  \includegraphics[width=0.7\textwidth]{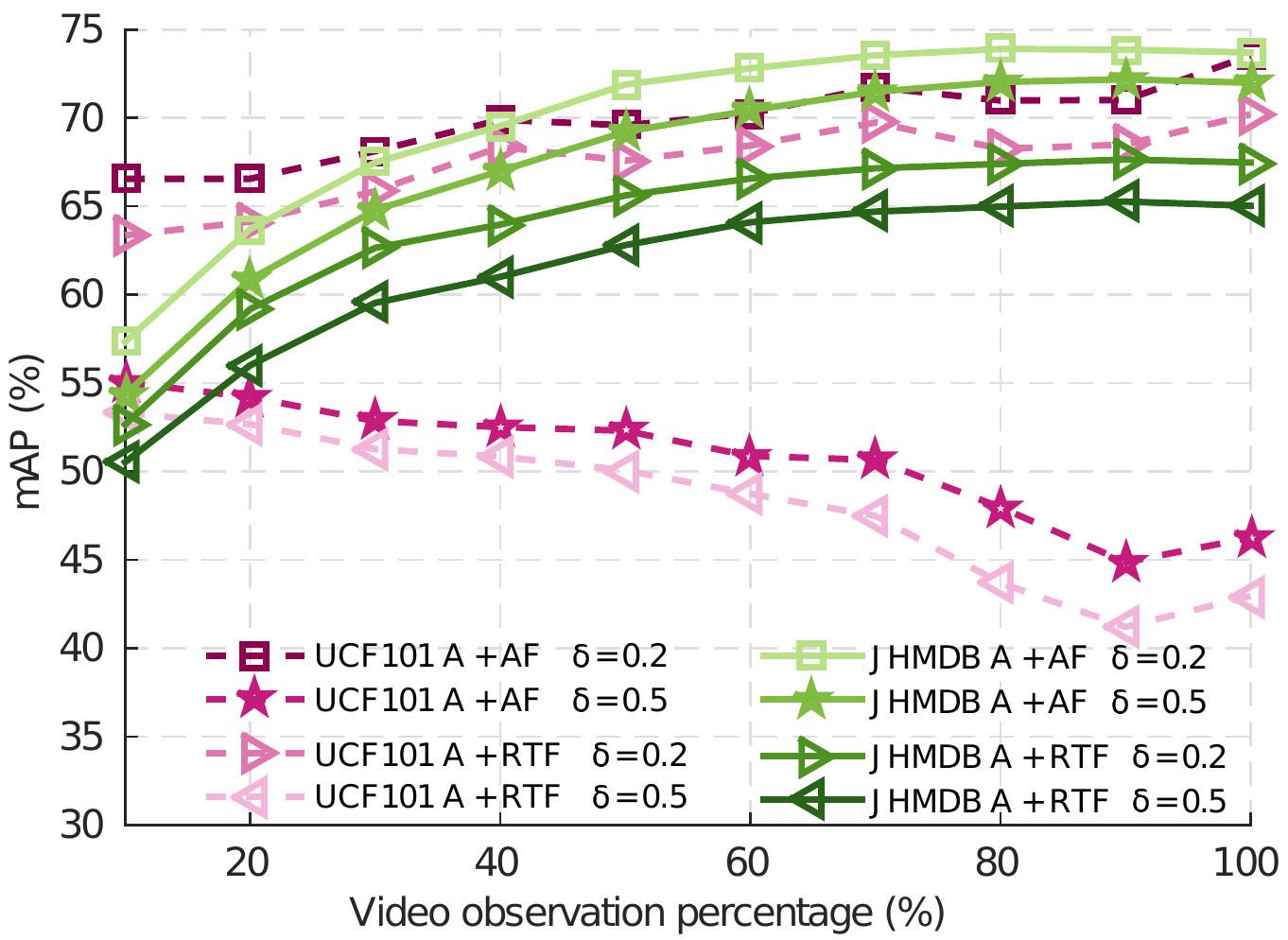}  
  \caption[Action detection results using the mAP (\%) metric.]{
        Action detection results using the mAP (\%) metric on UCF101-24 and J-HMDB-21, at IoU thresholds of $\delta=0.2,0.5$.
  }
  \label{online:fig:detection_plot_mAP} 
\end{figure}

Our framework, instead, can perform online spatiotemporal detection:
to demonstrate this, we present results on the temporally untrimmed UCF101-24 test videos as well. In Figure~\ref{online:fig:detection_plot_mAP} we report
online spatial-temporal detection results
on UCF101-24 and JHMBD-21 using the standard mAP metric (not reported in \cite{soomro2016predicting}).
Interestingly, for UCF101-24, at a relatively smaller IoU threshold ($\delta = 0.2$) the performance gradually increases over time as more video frames are observed,
whereas at a higher IoU threshold ($\delta = 0.5$) it slightly degrades over time.
A reason for this could be that
UCF101-24 videos are temporally untrimmed and contain multiple action instances, 
so that accurate detection may be challenging at higher detection thresholds (e.g. $\delta = 0.5$).
If temporal labelling is not very accurate, 
as required at high thresholds ($\delta = 0.5$), this might result in more false positives as the video progress, 
hence the observed drop in performance over time. 

\begin{table}[h]
  \centering
  \caption[Action detection results (mAP) on UCF101-24 dataset.]{Action detection results (mAP) on UCF101-24 dataset in split1.}
  {\footnotesize
  \scalebox{0.96}{
  \begin{tabular}{lcccc}
  \toprule
  IoU threshold $\delta$                    & 0.2   & 0.5 & 0.75 & avg-mAP \\ \midrule
  Yu \etal~\cite{yu2015fast}${}^{\ddagger}$                        & 26.5  & --   & -- & --\\
  Weinzaepfel~\etal~\cite{weinzaepfel2015learning}${}^{\ddagger}$  & 46.8  & --   & -- & -- \\   
  Peng and Schmid~\cite{peng2016eccv}${}^{\dagger}$     & \textbf{73.5}  & 32.1 & 02.7 & 07.3 \\
  Saha ~\etal~\cite{saha2016deep}${}^{\dagger}$                       & 66.6  & 36.4 & 07.9 & 14.4 \\\midrule
  Ours: $\textnormal{Appearance (A)}^{*}$          & 69.8  & 40.9 & \textbf{15.5} & 18.7 \\
  Ours: $\textnormal{Real-time flow (RTF)}^{*}$       & 42.5  & 13.9 & 00.5 & 03.3  \\ 
  Ours: $\textnormal{A $+$ RTF (boost-fusion)}^{*}$ & 69.7  & 41.9 & 14.1 & 18.4 \\
  Ours: $\textnormal{A $+$ RTF (union-set)}^{*}$      & 70.2  & 43.0 & 14.5 & 19.2 \\\midrule
  Ours: $\textnormal{Accurate - flow (AF)}^{**}$   & 63.7  & 30.8 & 02.8 & 11.0 \\
  Ours: $\textnormal{A $+$ AF (boost-fusion)}^{**}$ & 73.0  & 44.0 & 14.1 & 19.2 \\
  Ours: $\textnormal{A $+$ AF (union-set)}^{**}$      & \textbf{73.5}  & \textbf{46.3} & 15.0 & \textbf{20.4} \\\midrule
  SSD$+$~\cite{saha2016deep} A $+$ AF (union-set)${}^{\dagger}$ & 71.7  & 43.3 & 13.2 & 18.6 \\
  \bottomrule 
\multicolumn{5}{l}{${}^{\ddagger}$ These methods were using different annotations to~\cite{peng2016eccv,saha2016deep} and ours.} \\
\multicolumn{5}{l}{New annotations are available at \tiny{\url{https://github.com/gurkirt/corrected-UCF101-Annots}}} \\
\multicolumn{5}{l}{${}^{*}$ Incremental \& real-time \ \ ${}^{**}$ Incremental, non real-time \ \ ${}^{\dagger}$ Offline} 
  \end{tabular}
  }
  }
  \label{online:table:ucf101_results} 
\end{table} 

\subsection{Global spatiotemporal performance}\label{sec:offline-st-detection}

To demonstrate the strength of our online framework,
 we compare as well its absolute detection performances (avg-mAP after the video is fully observed) 
 to those of the previous offline competitors~\cite{saha2016deep,weinzaepfel2015learning,peng2016eccv,yu2015fast}. 
To ensure a fair comparison with~\cite{saha2016deep} (tube construction method presented in the previous chapter), 
we evaluate their offline tube generation method using the detection of bounding boxes produced by the SSD net.

\textbf{Improvement over the top performers}.
Results on {UCF101-24} are reported in Table~\ref{online:table:ucf101_results}.
In an online real-time setting we achieve an mAP of $70.2\%$ compared to $66.6\%$ reported by~\cite{saha2016deep} 
at the standard IoU threshold of $\delta=0.2$.
{In non-real time mode, we observe a further performance improvement of around $3.3\%$, leading to a $73.5\%$ mAP, comparable
to the $73.5$ reported by the current top performer~\cite{peng2016eccv}. 
The similar performance of our method (A$+$AF) to~\cite{peng2016eccv} at $\delta=0.2$ suggests that 
SSD and the multi-region adaptation of Faster-RCNN by ~\cite{peng2016eccv} produce similar quality frame level detection boxes.}

\textbf{Performance under more realistic requirements}.
Our method significantly outperforms ~\cite{saha2016deep,peng2016eccv} at more meaningful detection thresholds $\delta=0.5$ or higher. 
For instance, we achieve a $46.2\%$ mAP at $\delta=0.5$ as opposed to $32.1\%$ by ~\cite{peng2016eccv} and $36.4\%$ by ~\cite{saha2016deep}, 
an improvement of \textbf{14\%} and \textbf{9.8\%}, respectively.
This attests the superiority of our tube building algorithm when compared to those of ~\cite{peng2016eccv,saha2016deep}.
In fact, even in real time mode our pipeline (A $+$ RTF) still performs better than both ~\cite{saha2016deep,peng2016eccv} at $\delta=0.5$ or higher.


It is important to note that, our proposed fusion method (\textit{union-set-fusion}) 
significantly outperforms \textit{boost-fusion} proposed by~\cite{saha2016deep} on UCF101-24 dataset (see Table~\ref{online:table:ucf101_results}).
UCF101-24 includes many co-occurring action instances, we can infer that the union-set fusion strategy 
improves the performance by providing a larger number of high confidence boxes from either the
the appearance or the flow network.
When a single action is present in each video, as in 
J-HMDB, \textit{boost-fusion} perform better (Table~\ref{online:table:jhmdb21_results}).
In the Table~\ref{online:table:ablation-study-ucf10105}, we present a complete class-wise performance comparison of the two fusion strategies on both datasets.

\begin{table}[h] 
  \centering
  \footnotesize
  \caption[Action detection results (mAP) on J-HMDB-21.]{Action detection results (mAP) on J-HMDB-21.} 
  \scalebox{0.96}{
  \begin{tabular}{lcccc}
  \toprule
   IoU threshold $\delta$                                 & 0.2    & 0.5 & 0.75 & avg-mAP \\ \midrule
   Gkioxari and Malik~\cite{georgia2015tube}${}^{\dagger}$             & --   & 53.3  & --   & -- \\
   Wang~\etal~\cite{wangcvpr2016actionness}${}^{\dagger}$                       & --   & 56.4  & --   & -- \\
   Weinzaepfel~\etal~\cite{weinzaepfel2015learning}${}^{\dagger}$          & 63.1 & 60.7  & --   & -- \\
   Saha ~\etal~\cite{saha2016deep}${}^{\dagger}$                             & 72.6 & 71.5  & 43.3 & 40.0 \\
   Peng and Schmid~\cite{peng2016eccv}${}^{\dagger}$                  & \textbf{74.1} & \textbf{73.1}  & --   & -- \\\midrule
   Ours: $\textnormal{Appearance (A)}^{*}$                & 60.8 & 59.7  & 37.5 & 33.9 \\
   Ours: $\textnormal{Real-time flow (RTF)}^{*}$            & 56.9 & 47.4  & 20.2 & 19.3 \\ 
   Ours: $\textnormal{A $+$ RTF (union-set)}^{*}$             & 66.0 & 63.9  & 35.1 & 34.4 \\
   Ours: $\textnormal{A $+$ RTF (boost-fusion)}^{*}$          & 67.5 & 65.0  & 36.7 & 38.8 \\\midrule
   Ours: $\textnormal{Accurate - flow (AF)}^{**}$            & 68.5 & 67.0  & 38.7 & 36.1 \\
   Ours: $\textnormal{A $+$ AF (union-set)}^{**}$             & 70.8 & 70.1  & 43.7 & 39.7 \\
   Ours: $\textnormal{A $+$ AF (boost-fusion)}^{**}$          & 73.8 & 72.0  & \textbf{44.5} & \textbf{41.6} \\\midrule
   SSD$+$~\cite{saha2016deep} A $+$ AF (boost-fusion)${}^{\dagger}$   & 73.2 & 71.1  & 40.5 & 38.0 \\
  \bottomrule 
  \multicolumn{5}{l}{${}^{*}$ Incremental \& real-time \ \ ${}^{**}$ Incremental, non real-time \ \ ${}^{\dagger}$ Offline} 
  \end{tabular}
  }
  \label{online:table:jhmdb21_results}
\end{table} 

\textbf{Evaluation on J-HMDB-21}.
Table~\ref{online:table:jhmdb21_results} reports action detection results
averaged over the three splits of \textit{J-HMDB-21}, 
and compares them with those to our closest (offline) competitors.
\noindent Our framework outperforms the multi-stage approaches 
of ~\cite{georgia2015tube,wangcvpr2016actionness,weinzaepfel2015learning} in non-real-time mode at the
standard IoU threshold of $0.5$, while it attains figures very close to those of
\cite{saha2016deep,peng2016eccv} (73.8 versus 74.1 and 72.6, respectively) approaches, 
which make use of a two-stage Faster-RCNN.

Once again it is very important to point out that \cite{peng2016eccv} employs a battery of frame-level detectors, 
among which one based on strong priors on human body parts. 
Our approach does not make any prior assumption on the object(s)/actor(s) performing the action of interest, and is thus arguably more general-purpose.


\subsection{Relative contribution of tube generation and SSD}\label{online:exp:contribution-of-tubeconstruction}

As anticipated we evaluated the offline tube generation method of~\cite{saha2016deep} 
using the detection bounding boxes produced by the SSD network,
to both provide a fair comparison and to understand each component's influence on performance.
The related results appear in the last row of Table~\ref{online:table:ucf101_results} and Table \ref{online:table:jhmdb21_results}.
Similar comparison is shown in Table~\ref{online:table:ablation-study-ucf10105}, it shows that most of the classes benifit from proposed tube construction method~\ref{online:sec:tubeconstruction}.

From comparing the figures in the last two rows of both tables
it is apparent that our online tube generation performs better than the
offline tube generation of \cite{saha2016deep}, especially providing significant improvements at higher detection thresholds for both datasets.
We can infer that the increase in performance comes from both 
the higher-quality detections generated by SSD, 
as well as our new online tube generation method.
The fact that our tube generation is online, greedy and outperforms offline methods, 
so it suggests that offline approaches have a big room for improvements.

The reason for not observing a big boost with use of SSD on J-HMDB-21 maybe due to
its relatively smaller size, which does not allow us to leverage on the expressive power of SSD models.
Nevertheless, cross-validating the CNNs' hyper-parameters (e.g. learning rate), 
might lead to further improvements there as well.


\subsection{Contribution of the flow stream}\label{online:exp:flow-contribution}

The optical flow stream (\S~\ref{online:sec:opticalflow}) is an essential part of the framework. 
Fusing the real-time flow stream with the appearance stream (A$+$RTF mode) on UCF101-24 leads to a $2.1\%$ improvement at $\delta=0.5$. Accurate flow adds a further $3.3\%$.
A similar trend can be observed on J-HMDB-21, where A$+$RTF gives a $5\%$ boost at $\delta=0.5$, 
and the A$+$RTF mode takes it further to $72\%$. 
It is clear from Table~\ref{online:table:ucf101_results} and Table~\ref{online:table:jhmdb21_results} 
that optical flow plays a much bigger role on the J-HMDB-21 dataset as compared to UCF101-24.
Real-time OF does not provide as big a boost as accurate flow, but still 
pushes the overall performance towards that of the top competitors, 
with the invaluable addition of real-time speed.

\begin{table}[h!]
  \centering
  \footnotesize
  \caption[Class-wise action detection results on UCF101-24.]{Spatiotemporal action detection results (video APs in $\%$) on UCF101-24 at $\delta=0.2$ along with class-wise statistics about UCF101-24 dataset in first two rows 
  (number of action instance per video and action instance duration compared to video duration). All 24 classes are presented in following order: 
  Basketball (1-BB), BasketballDunk (2-BD), Biking (3-BK), CliffDiving (4-CD), CricketBowling (5-CB), Diving (6-DV), Fencing (7-FC), FloorGymnastics (8-FG), 
  GolfSwing (9-GS), HorseRiding (10-HR), IceDancing (11-ID), LongJump (12-LJ), PoleVault (13-PV), RopeClimbing (14-RC), SalsaSpin (15-SS), SkateBoarding (16-SB),
  Skiing (17-SK), Skijet (18-SI), SoccerJuggling (19-SJ), Surfing (20-SF), 
  TennisSwing (21-TS), TrampolineJumping (22-TJ), VolleyballSpiking (23-VS), 
  and WalkingWithDog (24-WD)}
  \scalebox{0.85}{
  \begin{tabular}{lcccccccc}
  \toprule
  Actions                     & 1-BB  &   2-BD &  3-BK &  4-CD &  5-CB &  6-DV & 7-FC & 8-FG  \\ \midrule 
  Number of Actions/Video        & 1.0 & 1.0 & 1.8 & 1.0 & 1.1 & 1.0 & 2.4 & 1.0  \\ 
  Action/Video duration (ratio)    & 0.34 & 0.59 & 0.70 & 0.64 & 0.36 & 0.65 & 0.89 & 1.00 \\ \midrule
  Saha \etal \cite{saha2016deep}      & 39.6 & 49.7 & 66.9 & 73.2 & 14.1 & 93.6 & 85.9 & 99.8 \\ \midrule
  Ours-Appearance (A)               & 43.0 & 67.4 & 75.8 & 67.2 & 50.5 & 100.0 & 88.5 & 97.9 \\
  Ours-A + RTF (boost-fusion)        & 42.2 & 69.0 & 71.7 & 73.1 & 41.3 & 100.0 & 87.6 & 99.1 \\
  Ours-A + RTF (union-set)        & 42.0 & 64.6 & 73.7 & 75.2 & 41.5 & 100.0 & 86.5 & 97.9 \\ 
  Ours-A + AF (boost-fusion)        & 45.0 & 86.4 & 67.6 & 78.2 & 44.2 & 100.0 & 89.8 & 99.9 \\ 
  Ours-A + AF (union-set)        & 43.9 & 81.6 & 73.6 & 73.7 & 49.0 & 100.0 & 90.2 & 97.9 \\ \midrule 
  SSD+\cite{saha2016deep} A + AF (union-set) & 43.2 & 78.5 & 65.8 & 72.0 & 43.6 & 100.0 & 86.1 & 98.1 \\ \midrule \\ \midrule
  
  Actions             & 9-GS   &  10-HR & 11-ID & 12-LJ & 13-PV &  14-RC &  15-SS & 16-SB \\ \midrule
  Number of Actions/Video        & 1.0 & 1.0 & 2.3 & 1.0 & 1.1 & 1.0 & 4.9 & 1.0 \\
  Action/Video duration (ratio) & 1.00 & 0.95 & 0.86 & 0.53 & 0.30 & 0.65 & 0.31 & 0.84 \\ \midrule
  Saha \etal \cite{saha2016deep}     & 68.3 & 94.1 & 63.1 & 57.2 & 75.1 & 89.6 & 31.1 & 85.1 \\ \midrule
  Ours-Appearance (A)               & 59.9 & 95.9 & 56.5 & 59.7 & 80.8 & 93.4 & 36.9 & 86.1 \\
  Ours-A + RTF (boost-fusion)        & 64.3 & 96.0 & 72.8 & 68.8 & 72.7 & 94.5 & 19.7 & 85.6 \\ 
  Ours-A + RTF (union-set)        & 62.1 & 96.0 & 77.6 & 69.7 & 76.1 & 96.1 & 22.2 & 87.4 \\ 
  Ours-A + AF (boost-fusion)        & 65.8 & 96.0 & 74.0 & 81.4 & 80.3 & 95.8 & 23.1 & 88.3 \\ 
  Ours-A + AF (union-set)        & 62.0 & 96.0 & 76.3 & 82.9 & 82.7 & 98.1 & 25.7 & 87.8 \\ \midrule 
  SSD+\cite{saha2016deep} A + AF (union-set) & 61.3 & 96.0 & 60.6 & 83.4 & 84.6 & 98.6 & 20.1 & 88.4 \\ \midrule \midrule 
  Actions                & 17-SK & 18-SI & 19-SJ & 20-SF & 21-TS & 22-TJ & 23-VS & 24-WD \\ \midrule
  Number of Actions/Video        & 1.0 & 1.0 & 1.1 & 1.6 & 1.3 & 2.5 & 1.1 & 1.1 \\
  Action/Video duration (ratio) & 0.67 & 0.95 & 0.85 & 0.98 & 0.81 & 1.00 & 0.34 & 1.00  \\ \midrule
  Saha \etal \cite{saha2016deep}    & 79.6 & 96.1 & 89.1 & 63.2 & 33.6 & 52.7 & 20.9 & 75.6 \\ \midrule
  Ours-Appearance (A)              & 78.6 & 94.6 & 71.0 & 65.4 & 37.6 & 59.4 & 24.8 & 83.7 \\
  Ours-A + RTF (boost-fusion)        & 78.8 & 89.9 & 82.1 & 62.1 & 31.7 & 57.7 & 27.0 & 83.2 \\ 
  Ours-A + RTF (union-set)        & 81.0 & 87.1 & 82.4 & 62.1 & 37.4 & 59.4 & 21.7 & 85.1 \\ 
  Ours-A + AF (boost-fusion)        & 80.0 & 91.1 & 93.1 & 65.4 & 38.7 & 57.4 & 28.3 & 84.3 \\ 
  Ours-A + AF (union-set)        & 81.8 & 93.2 & 93.0 & 65.8 & 38.2 & 58.5 & 26.1 & 86.9 \\ \midrule
  SSD+\cite{saha2016deep} A + AF (union-set) & 82.0 & 93.5 & 92.7 & 61.1 & 38.8 & 56.8 & 30.7 & 86.6 \\ 
  \bottomrule
  \end{tabular}
  }
  \label{online:table:ablation-study-ucf10105}  
\end{table}

We also report an ablation study of the online spatiotemporal action detection performance on UCF101-24 dataset.
Table~\ref{online:table:ablation-study-ucf10105} shows the class-specific video AP (average precision in $\%$) for each action category of UCF101-24 generated by
the appearance (A) model, appearance plus real-time flow (A \& RTF ) fusion model and appearance plus accurate flow (A \& AF ) fusion model.
Results are generated at a spatiotemporal overlap threshold of $\delta=0.2$ in Table~\ref{online:table:ablation-study-ucf10105}.
For $12$ out of $24$ action classes, our \emph{appearance plus accurate flow} fusion technique gives the best APs.
The appearance-based detection network alone achieves the best APs for three classes: ``HorseRiding'', ``Surfing'' and ``Fencing''.
It is worth noting that for action classes ``HorseRiding'' and ``Fencing'', appearance cues such as ``horse'' and ``fencing player with white dress'' are the most discriminative features.
Whereas, in action class ``Surfing'', sea motion might be preventing the flow networks from learning discriminative features.

A difference between the two fusion strategies can be observed (Table~\ref{online:table:ablation-study-ucf10105}) in classes with multiple actors, such as ``IceDancing'', ``SalsaSpin''  and ``Biking''. There, union-set fusion exhibits a significant improvement when compared with boost-fusion strategy.

\textbf{Difficult classes.} ``Basketball'', ``CricketBowling'', ``VolleyballSpiking'' and ``TennisSwing'' are the most difficult classes.
Most of the ``Basketball'' training videos have at least one actor (basketball player) present in the video. The ``Basketball'' action, however, is performed within
a small temporal extent. As temporal detection is difficult, action categories with a relatively large number of temporally untrimmed  test videos such as
``CricketBowling'' , ``VolleyballSpiking'' and ``TennisSwing'' show lower APs.
Further, running (during the ``CricketBowling'' action) is not considered as a part of the action which makes it even more difficult to detect.
``VolleyballSpiking'' videos contain many potential actors (Volleyball players) which are difficult to distinguish. It is clear from the above pieces of evidence that, it is necessary to re-train network using a background frame, which is not done. 
To achieve that we would need to modify the cost function of SSD to accept an image without any positive instance. At the moment it requires at least one positive instance present in any video frame.

\textbf{Easy classes.} ``FloorGymnastics'', ``HorseRiding'' and ``SoccerJuggling'' are the easiest classes to detect, possibly because these classes contain mostly one actor at a time and have salient appearance features. An example, consider the presence of a horse in the ``HorseRiding'' class.

\textbf{Dataset statistics of UCF101-24 dataset.} The first two rows of Table~\ref{online:table:ablation-study-ucf10105} shows class-wise statistics on the UCF101-24 dataset. 
The number of action instances per video is shown in the first row, averaged over test-list of split 1. 
The duration of action instance compared with the duration of the video is shown in the second row, averaged over test-list of split 1. 
It is clear that the most difficult classes are those which span a shorter fraction of the entire video duration.

\subsection{Test time detection speed} 
\label{sec:detection-speed}
To support our claim to real time capability, 
we report the test time detection speed of our pipeline 
under all three types of input A (RGB), A$+$RTF (real-time flow), A $+$ AF (accurate flow) in
Table~\ref{online:table:detection_time_analysis}.
These figures
were generated using a desktop computer with an Intel Xeon CPU@2.80GHz (8 cores)
and two NVIDIA Titan X GPUs.
Real-time capabilities can be achieved by either not using optical flow (using only appearance (A) stream on one GPU)
or by computing real-time optical flow~\cite{kroeger2016fast} on a CPU in parallel
with two CNN forward passes on two GPUs.
For action tube generation (\S~\ref{online:sec:tubeconstruction}) we ran 8 CPU threads in parallel for each class.
We used the real-time optical flow algorithm ~\cite{kroeger2016fast} in a customised setting, 
with a minimum number of pyramid levels set to $2$ instead of $3$, and 
patch overlap $0.6$ rather than $0.4$.
OF computation averages $\sim7$ ms per image. 

Table~\ref{online:table:detection_time_analysis} also compares our detection speed to that reported by Saha \etal \cite{saha2016deep} (see Chapter \ref{chapter:offline}).
With an overall detection speed of 40 fps (when using RGB only) and 28 fps (when using also real-time OF), 
our framework is able to detect multiple co-occurring action instances in real-time while retaining very competitive performance.

\begin{table}[h]
\centering
\footnotesize
\caption[Test time detection speed.]{Test time detection speed.}
\scalebox{0.96}{
\begin{tabular}{l  c  c  c c}
\toprule
Framework modules & A & A$+$RTF & A$+$AF & \cite{saha2016deep} \\\midrule
Flow computation (ms$^*$)  & -- & 7.0 & 110 & 110\\
Detection network time (ms$^*$)  & 21.8 & 21.8 & 21.8 & 145 \\
Tube generation time (ms$^*$)  & 2.5 & 3.0 & 3.0 & 10.0\\\midrule
Overall speed (fps$^{**}$ )  & 40 & 28 & 7 & 4 \\
\bottomrule 
\multicolumn{3}{l}{$^*$ ms - milliseconds \ \ \ $^{**}$ fps - frame per second.} 
\end{tabular}
}
\label{online:table:detection_time_analysis} 
\end{table}

\subsection{Qualitative results} \label{subsec:qty-res}
\begin{figure}[h]
  \centering
  \includegraphics[width=0.7\textwidth]{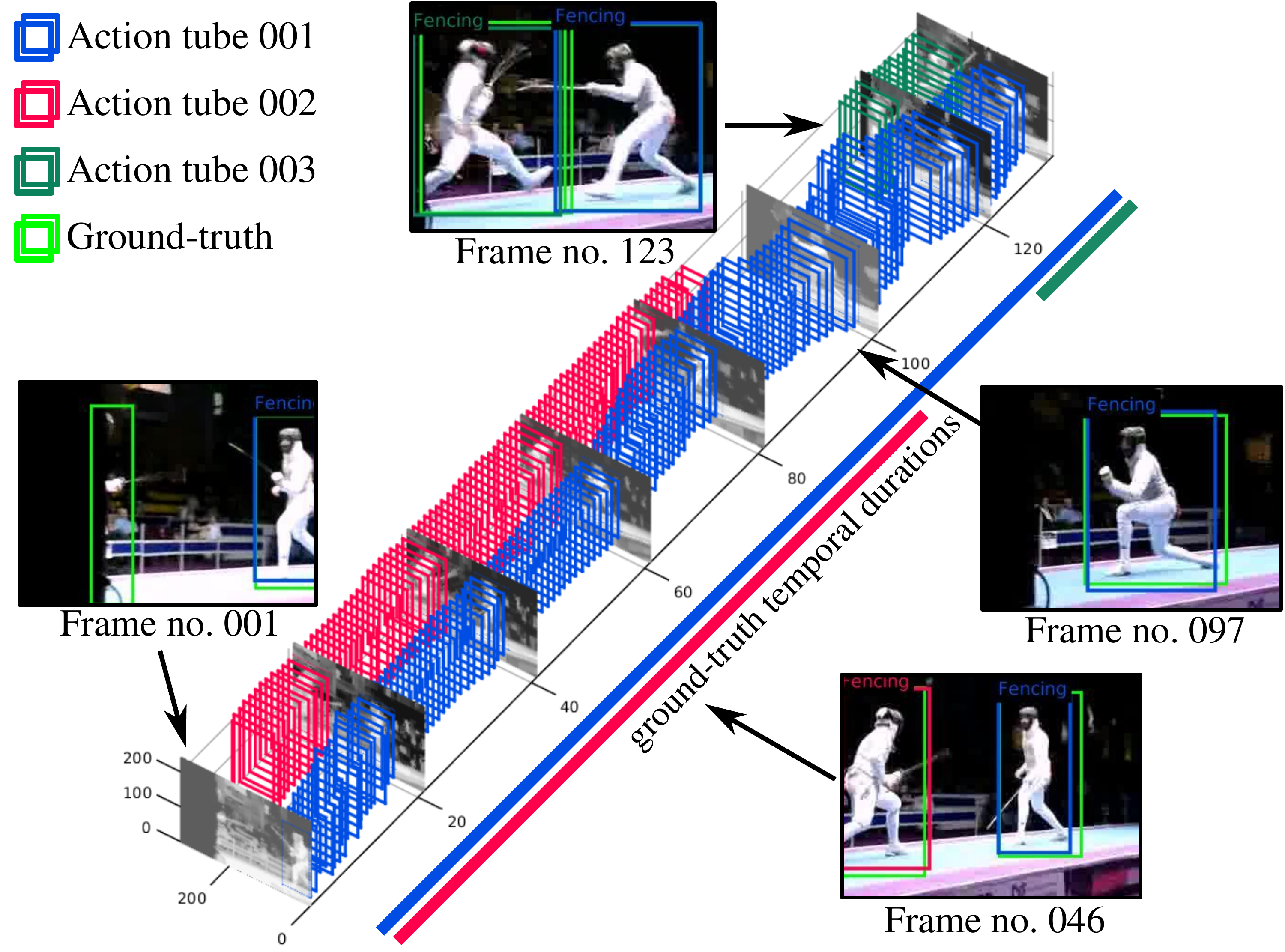}
  \caption[Sample action detection results in ``Fencing'' video.]{Sample spatiotemporal action detection results on a ``Fencing'' action sequence taken from UCF101-24 test-set.
     The detected action tubes are plotted in 3D and drawn in different colour indicating $3$ different action instances.
     The ground-truth temporal duration of each action instance is shown by the coloured bars. Note that the temporal duration 
     of the detected and the ground-truth tubes are closely matched (with good temporal overlaps).   
  }
  \label{online:fig:temporal-trimming-1}
\end{figure}

\begin{figure}[h]
  \centering
  \includegraphics[width=0.7\textwidth]{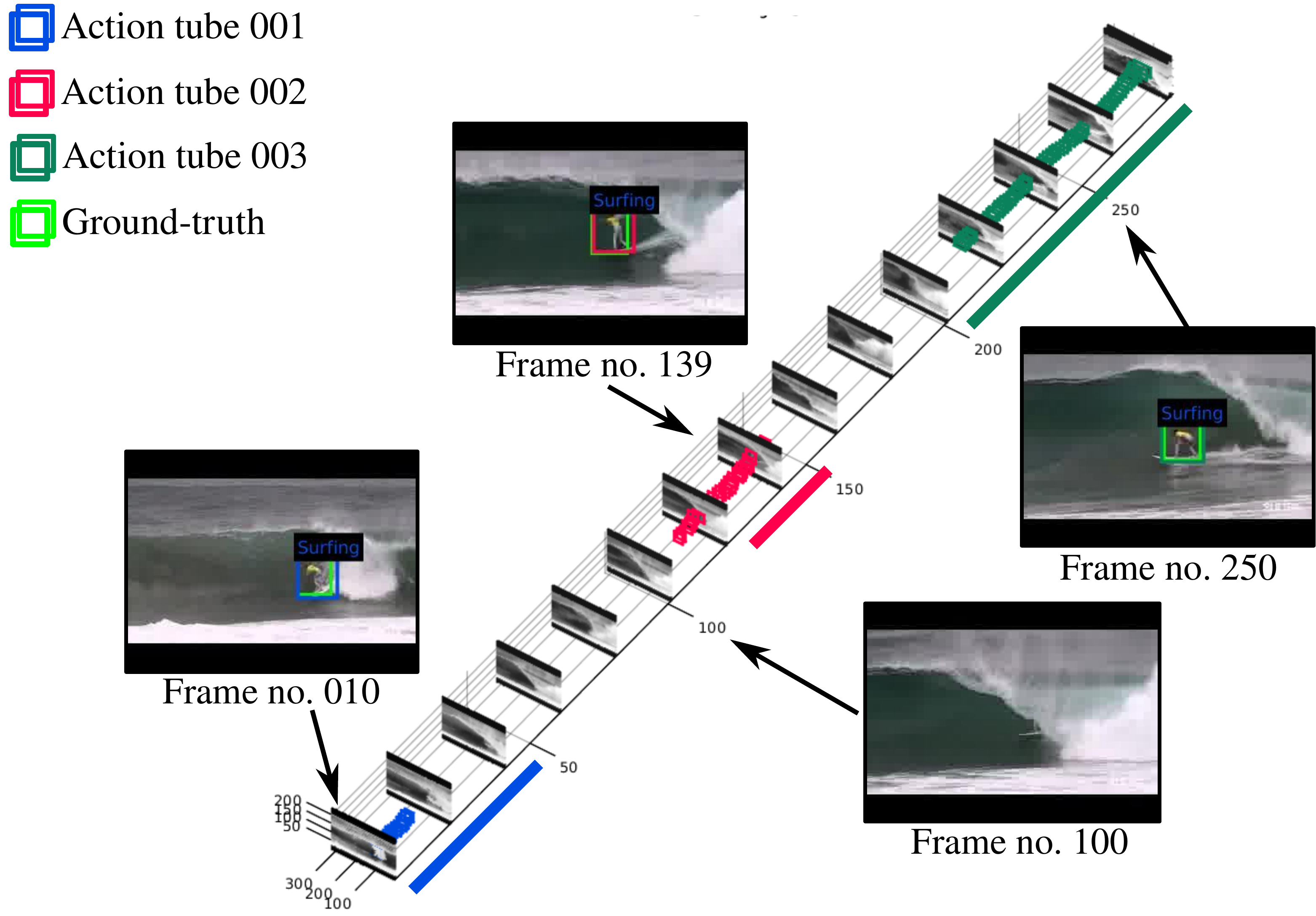}
  \caption[Sample action detection results in ``Surfing'' video.]{Sample spatiotemporal action detection results on a ``Surfing'' action sequence taken from UCF101-24 test-set.
     The detected action tubes are plotted in 3D and drawn in different colour indicating $3$ different action instances.
     The ground-truth temporal duration of each action instance is shown by the coloured bars. Note that the temporal duration 
     of the detection and the ground-truth tubes are closely matched.}
  \label{online:fig:temporal-trimming-2}  
\end{figure}

\begin{figure}[h]
  \centering
  \includegraphics[width=0.87\textwidth]{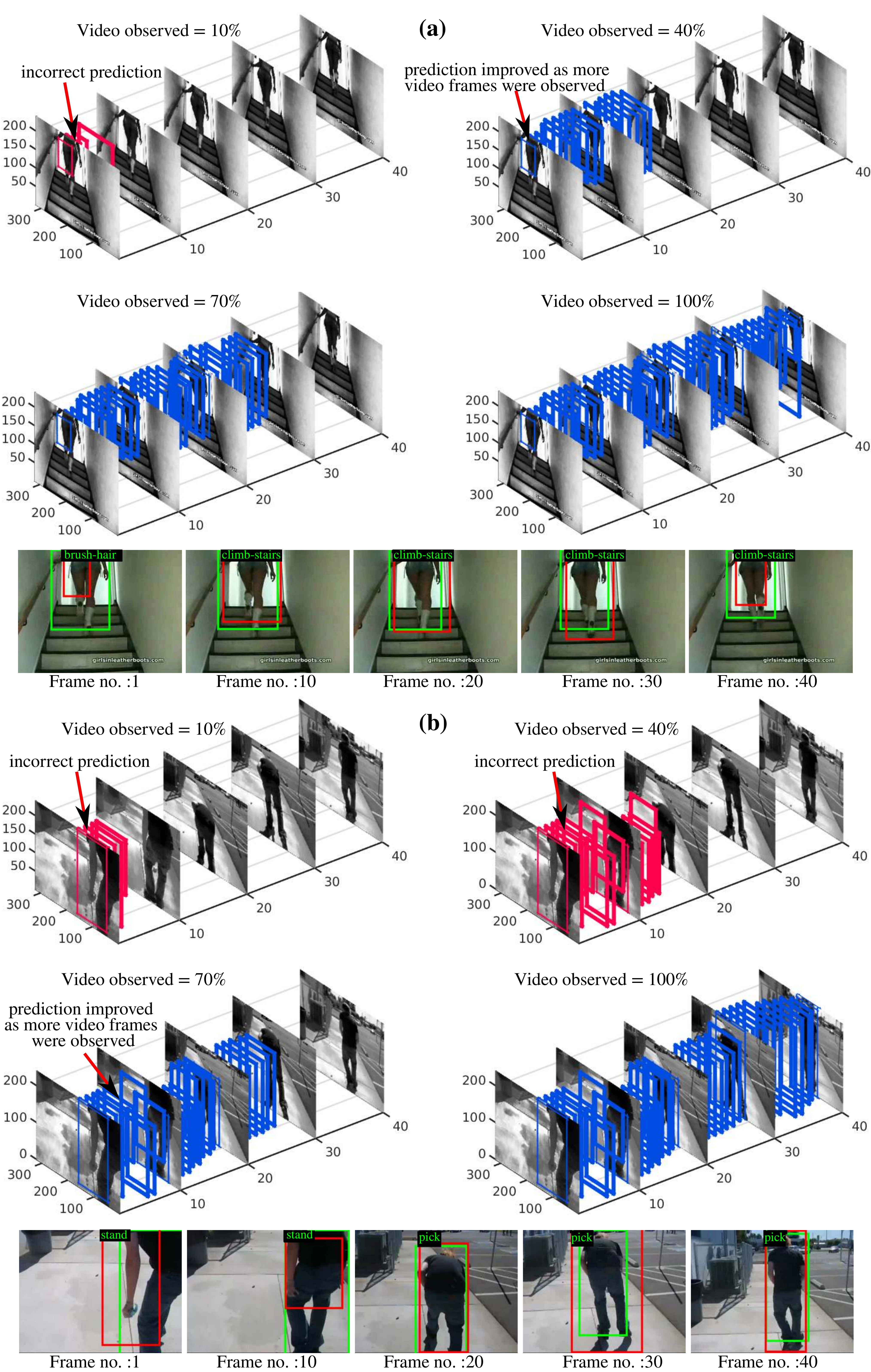}
  \caption[Sample early action label prediction and online action detection results on J-HMDB-21.]
  {Sample early action label prediction and online action detection results on J-HMDB-21 dataset.
      \textbf{(a)} and \textbf{(b)} show prediction results of 2 different test videos with 
      ground-truth action labels `climb-stairs' and `pick' respectively.
      Each video and its corresponding space-time detection tube were plotted in 3D  at different time points (i.e., \% of video observed).   
      Detection tubes are drawn in two different colours to indicate the wrong early label prediction and the improved prediction as more video frames were observed in time.      
      Below the 3D plot,  the predicted action labels for the same video at different time points are overlaid on the corresponding video frames
      in which the green box depicts the ground-truth and red depicts the predicted bounding box.}
  \label{online:fig:jhmdb_results_detection_sup}
\end{figure}

\begin{figure}[h]
  \centering
  \includegraphics[width=0.98\textwidth]{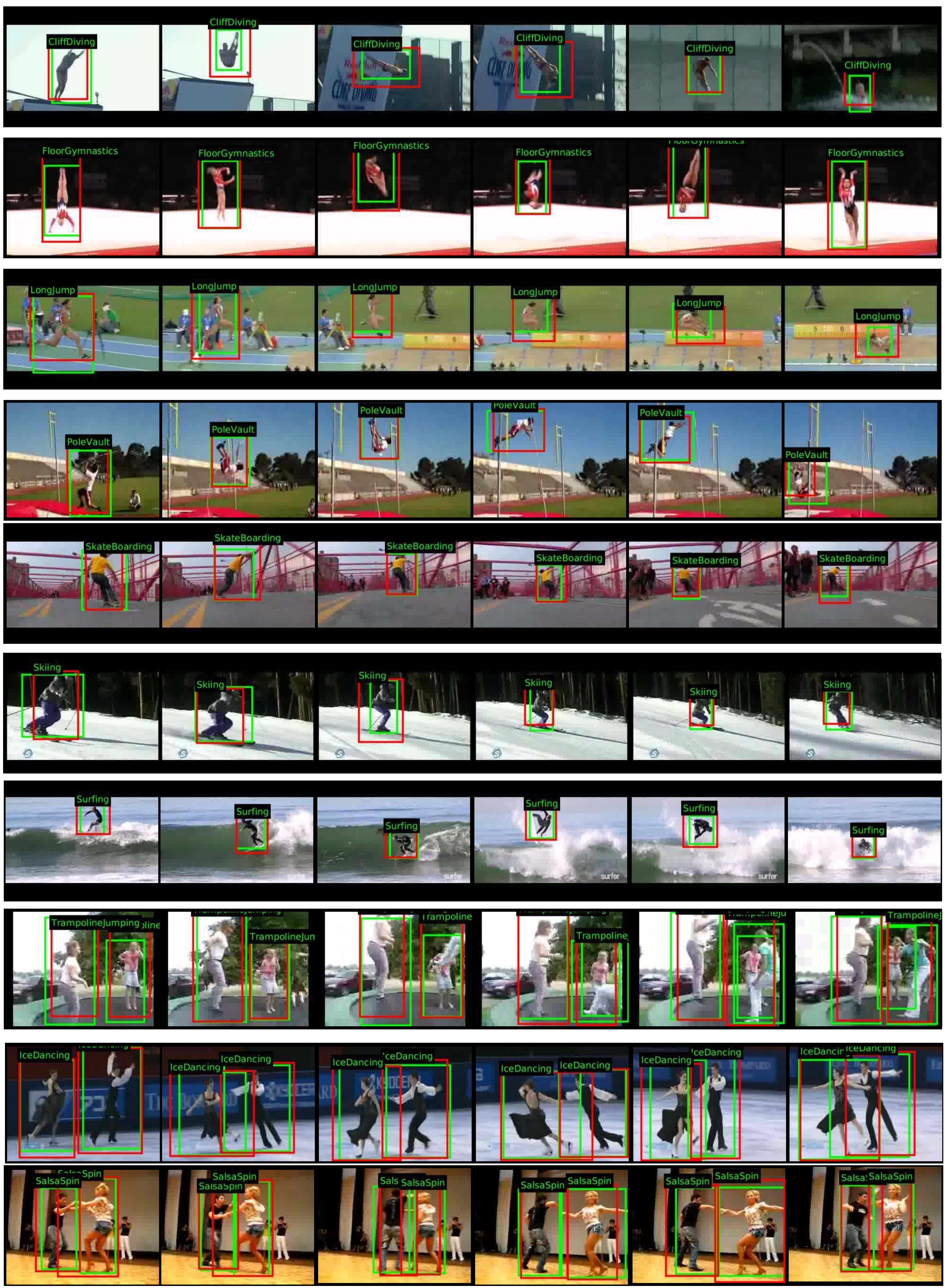}
  \caption[Sample action detection results on UCF101-24.]
  {Sample action detection results on UCF101-24. 
  Each row represents a UCF101-24 test video clip.
  Ground-truth bounding boxes are drawn in green and detection boxes are in red.} 
  \label{online:fig:spatiotemporal-res-images}
\end{figure}

Fig.~\ref{online:fig:temporal-trimming-1} and ~\ref{online:fig:temporal-trimming-2} show qualitative results of spatiotemporal action detection on 
temporally untrimmed ``Fencing'' and ``Surfing'' action sequences taken from UCF101-24 test-set.
Fig.~\ref{online:fig:jhmdb_results_detection_sup} shows sample early action label prediction and online action detection qualitative results on the J-HMDB-21 dataset.
Fig.~\ref{online:fig:spatiotemporal-res-images} provides additional evidence on the action detection performance of our method on UCF101-24 dataset.

%% file: pads/partII.tex
\cleardoublepage
\phantomsection
\renewcommand{\pname}{Part II : Micro-tubes for Action Detection and Future Prediction of Tubes}
\addcontentsline{toc}{chapter}{\pname}\label{partII}

\pagebreak
\hspace{14pt}
\vfill
\begin{center}
\textbf{\pname}
\end{center}
\vfill
\hspace{0pt}
\pagebreak

%% file: chapters/5_tramnet.tex
\chapter{Transition Matrix Network for Flexible Micro-tube Proposal Generation}
\label{chapter:tramnet}
\renewcommand{\imagepath}{figures/tramnet} 
\input{chapters/partII/tramnet_intro}
\input{chapters/partII/tramnet_approach}
\input{chapters/partII/tramnet_exp}

\section{Summary and limitations}\label{summary:tramnet}
\paragraph{Summary: }
We presented AMTNet and TraMNet two deep learning framework for action detection in videos.
While AMTNet remains simple and efficient as it generates action cuboid proposals, 
whereas, TraMNet can cope with real-world videos containing ``dynamic'' actions whose location significantly changes over time. 
This is done by learning a transition probability matrix for each feature pyramid layer from the training data in a hidden Markov model formulation, 
leading to an original configurable layer architecture.
Furthermore, unlike their competitors
\cite{kalogeiton2017action,hou2017tube}, 
which require dense frame-level bounding box annotation,
AMTNet and TraMNet builds on a similar network architecture
in which action representations are learnt from pairs of frames rather than chunks of consecutive frames, 
thus eliminating the need for dense annotation.
An extensive experimental analysis supports TraMNet's action detection capabilities, especially
under dynamic actions and sparse annotations. 

In summary, we present a novel deep learning architecture for spatiotemporal action detection which:
\begin{itemize}
     \item on the methodological side, constitutes a key conceptual step forward from action detection paradigms relying on frame-level region proposals towards networks able
     to regress optimal solutions to the problem in the form of complete action tubes;
     \item a novel, end-to-end trainable deep network architecture which addresses
     the spatiotemporal action detection and classification task jointly using a single round of optimisation;
     \item introduces an efficient and flexible anchor micro-tube hypothesis generation framework to generate high-quality action proposals;
     \item can handle significant spatial movement in dynamic actions without penalising more static ones;
     \item is a scalable solution for training models on both sparse and dense annotations.
\end{itemize}

\paragraph{Limitations of TraMNet as oppose to AMTNet:}
We can argue that TraMNet is more generic and provides better spatiotemporal localisation performance.
However there are few advantages to AMTNet, i) AMTNet is simple and does not require pre-computation of the transition matrix, 
ii) it has fewer hyperparameters, as it does not require a threshold to sample micro-tubes proposals from training data, iii) the time complexity of AMTNet is constant iv) in cases where there is not much movement, AMTNet might still be a more reliable option.
We recommend trying AMTNet first on a new dataset before moving on to more complex TraMNet.
As a result, we recommend using both methods in tandem.

\paragraph{Looking ahead:} 
So far in this dissertation, we covered different methods of action tube detection in untrimmed videos in online fashion.
Interestingly, similar to early action prediction problem observed in chapter~\ref{chapter:online} as a consequence of the online nature of action tube detection solution,
we can look to predict the future of action tubes themselves. 
Any future prediction approach needs to online in nature, hence, we can use our existing online action tube detection methods for predicting the future of action tubes.
In the next chapter, we will see how online tube construction along with future regression help in the future prediction of action tubes.

%% file: chapters/partII/tramnet_intro.tex
\section{Introduction} 
\label{tramnet:intro}

A frame-level action detection approach, such as those presented in the previous two chapters, are sub-optimal for long term action tube detection, 
for they do not allow any feature-level interaction across frames.

In this chapter we take two steps towards a truly optimal solution of the action detection problem by considering video-level region proposals formed 
by a pair of bounding boxes spanning two successive video frames at an arbitrary temporal interval $\Delta$. 
This allows us to learn a feature representation from pairs of video frames, unlike the single frame-based approaches of Chapter~\ref{chapter:offline} and Chapter~\ref{chapter:online}.
It is quite easy to stack multiple frames and pass them through a CNN, but we also need to solve the feature-level correspondence problem, 
so that we can take features from both frames into account with respect to the reference anchor boxes.
Unlike frame-based actor/object detectors, we need reference anchor boxes to span multiple frames.

\subsection{AMTNet - 3D proposals for micro-tube detection}

3D proposals can be created by extending 2D frame-level anchors/proposals in the temporal dimension. The networks will then learn a feature representation from pairs of video frames, 
allowing them to implicitly learn the temporal correspondence between inter-frame action regions (bounding boxes).
As a result, they can predict action \emph{micro-tubes}~\cite{saha2017amtnet}, i.e., temporally linked frame-level detections for short sub-sequences of a test video clip.
We call the resulting architecture an \emph{action micro-tube network} (AMTNet)~\cite{saha2017amtnet}. 
In the following we demonstrate that AMTNet does improve performance in comparison with a frame-based approach by a large margin.

\begin{figure}[h]
  \centering
  \includegraphics[scale=0.22]{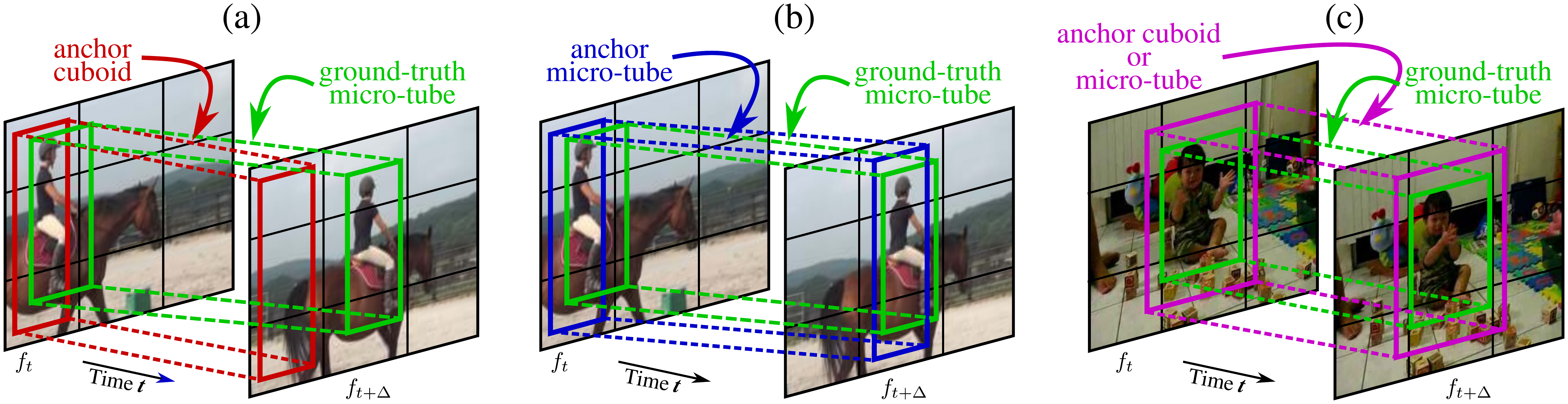}
  \caption[Illustrating the key limitation of anchor cuboids.]
  {
      Illustrating the key limitation of anchor cuboids using a ``dynamic'' action such as
      ``horse riding''. \textbf{(a)} A horse rider changes its location from frame $f_t$ to $f_{t+\Delta}$
      as shown by the ground truth bounding boxes (in green).
      As anchor cuboid generation~\cite{saha2017amtnet,kalogeiton2017action} is constrained by the spatial location of the anchor box in the first frame $f_t$, the overall spatiotemporal IoU overlap between the ground-truth micro-tube and the anchor cuboid is relatively low.
      \textbf{(b)} In contrast, anchor micro-tube proposal generator is much more flexible,
      as it efficiently explores the video search space via an approximate transition matrix estimated according to a hidden Markov model (HMM) formulation. As a result,
      the anchor micro-tube proposal (in blue) generated by the proposed model exhibits higher overlap with the ground-truth.
      \textbf{(c)} For ``static'' actions (such as ``clap'') in which the actor does not change location over time, 
      anchor cuboid and anchor micro-tubes have the same 
        }
\label{tramnet:fig:opt_prop} 
\end{figure}

A major concern with AMTNet is that action proposals are generated by extending
2D object proposals (anchor/prior boxes for images)~\cite{liu2016ssd,ren2015faster} 
to 3D proposals (anchor cuboids spanning multiple frames) (see Fig.~\ref{tramnet:fig:opt_prop}(a)).
These 3D anchor cuboids may be suitable for ``static'' actions (e.g. ``handshake'' or ``clap'', 
in which the spatial location of the actor(s) does not vary over time, see Figure~\ref{tramnet:fig:opt_prop}(c)), but
are most inappropriate for ``dynamic'' ones (e.g. ``horse riding'', ``skiing'').
Figure~\ref{tramnet:fig:opt_prop}(a) 
underscores this issue.
In the remainder of the chapter we thus illustrate a way to handle this problem. However, towards the end of this chapter, we show evidence that, even considering these issues,
the anchor cuboid-based AMTNet possesses multiple advantages, and 
improves on the single frame-based approaches presented in the previous chapters. 


\subsection{TraMNet - Flexible 3D proposals}

We observed in Figure~\ref{tramnet:fig:opt_prop} that anchor cuboids are limiting for dynamic actions.
However, the video proposal search space ($\mathcal{O}(n^f)$) is much larger than the image proposal search space ($\mathcal{O}(n)$), 
where $n$ is the number of anchor boxes per frame and $f$ is the number of video frames considered.

As a next step, we thus propose a flexible micro-tube proposal generation method which allows us to  explore the video search space efficiently via an approximate transition matrix.
Such transition matrix encodes the probability of a temporal link between
an anchor box at time $t$ and one at $t+\Delta$,
and is estimated within the framework of discrete state/continuous observation
hidden Markov models (HMMs, see Section~\ref{tramnet:sec:hmm})~\cite{elliott2008hidden}. 
Here, the hidden states are the 
2D bounding-box coordinates $[x_{min}, y_{min}, x_{max}, y_{max}]'$
of each anchor box from a (finite) hierarchy of fixed grids at different scales.
Anchor micro-tubes are not bound to be of strictly cuboidal shape (as in~\cite{saha2017amtnet,kalogeiton2017action,hou2017tube}), 
thus resulting in a higher IoU overlap with the ground-truth (see the anchor micro-tubes in Figure~\ref{tramnet:fig:opt_prop}(b)), specifically
for instances where the spatial location of the actor changes significantly 
from $f_t$ to $ f_{t+\Delta}$ in a training pair.
We thus propose a novel configurable deep neural network architecture
(see Fig.~\ref{tramnet:fig:algorithmOverview}), which leverages high-quality micro-tubes shaped by learnt anchor transition probabilities.
We call it the \emph{transition matrix network} (TraMNet).

In Section~\ref{tramnet:subsec:amntnet_matrix}
we show that AMTNet is special case of TraMNet , i.e., a TraMNet can be converted into an AMTNet by replacing the learnt probabilistic transition matrix with the identity matrix.

To validate these notions we produced a new action detection dataset which is a ``transformed'' version of UCF101-24~\cite{ucf101},
in which we force action instances to be dynamic (i.e., to change their spatial location significantly over time) by introducing random translations in the 2d spatial domain.
We show that our proposed action detection approach outperforms the baseline AMTNet method when trained and tested on this transformed dataset.

\subsection{Handling sparse annotations}

The previous action detection methods such as as~\cite{kalogeiton2017action,hou2017tube} 
require dense ground-truth annotation for network training:
bounding-box annotation is required for $k$ consecutive video frames,
where $k$ is the number of frames in a training example.
Kalogeiton~\etal~\cite{kalogeiton2017action} use $k=6$ whereas 
for Hou~\etal~\cite{hou2017tube} $k=8$.
Generating such dense bounding box annotation for long video sequences is 
highly expensive and impractical ~\cite{daly2016weinzaepfel,ava2017gu}.
The latest generation action detection benchmarks DALY~\cite{daly2016weinzaepfel} 
and AVA~\cite{ava2017gu}, in contrast, provide sparse bounding-box annotations.
More specifically,  DALY has $1$ to $5$ frames bounding box annotation per action instance, irrespective of the duration of an instance, 
whereas AVA only has one frame annotation per second.
{This motivates the design of a deep network capable of handling sparse annotations, 
while still being able to predict micro-tubes over multiple frames}.

Unlike~\cite{kalogeiton2017action,hou2017tube}, 
AMTNet uses \emph{pairs} of successive frames
$(f_t, f_{t+\Delta})$, thus eliminating the need for dense training annotation when 
$\Delta$ is large (e.g., $\Delta=\{ 5,10,21\}$) or arbitrary as in DALY~\cite{daly2016weinzaepfel}.
If the spatiotemporal IoU (Intersection over Union) overlap between ground-truth micro-tube 
and action proposal could be improved (cf. Fig.~\ref{tramnet:fig:opt_prop}),
such a network would be able to handle sparse annotation
(e.g., pairs of frames which are $\Delta=21$ apart). Indeed, the use of 
pairs of successive frames $(f_t, f_{t+\Delta})$, in combination with the flexible anchor proposals introduced here, is arguably more efficient than any other state-of-the-art method~\cite{saha2017amtnet,kay2017kinetics,hou2017tube} in handling sparse annotations (e.g. DALY~\cite{daly2016weinzaepfel} and AVA~\cite{ava2017gu}),
which we quantitatively demonstrate later in the chapter.



\paragraph{Related publications:}
The AMTNet work presented in this chapter appeared in ICCV 2017~\cite{saha2017amtnet}. 
The dissertation author was the main co-investigator in ~\cite{singh2018tramnet}. 
Here, we present multiple improvements over the original work~\cite{saha2017amtnet}, namely:
i) we incorporate the SSD architecture to keep the pipeline real-time, as compared to the Faster R-CNN used in~\cite{saha2017amtnet};
ii) we adopt the online tube construction method~\cite{singh2017online} of Chapter \ref{chapter:online} because of its performance and its online processing capabilities.
The TraMNet work presented in this chapter appeared in ACCV 2018~\cite{singh2018tramnet}. 
The dissertation author was the primary investigator in ~\cite{singh2018tramnet}.

\paragraph{Outline:}\label{tramnet:outline}
The remainder of the chapter is organised as follows.
We start by presenting an overview of our TraMNet based action detection approach in Section~\ref{tramnet:sec:overview}. 
In Section~\ref{tramnet:sec:base_network},
we introduce the base network architecture used for feature learning.
We cast the action proposal generation problem
in a hidden Markov model (HMM) formulation (\S~Section~\ref{tramnet:sec:hmm}),
and introduce an approximate estimation
of the HMM transition probability matrix using
a heuristic approach (\S~Section~\ref{tramnet:subsec:transition_matrix}).
In Section~\ref{tramnet:sec:reconfig_pooling_layer},
a configurable pooling layer architecture is presented
which pools convolutional features from the regions in the two frames linked by the estimated transition probabilities.
Finally, the output layers of the network (i.e., the micro-tube
regression and classification layers) are described in Section~\ref{tramnet:subsec:conv_2_linear}.
Next, we report the experimental validation of the proposed model 
in Section~\ref{tramnet:sec:exp} from both qualitative and quantitative perspective.
Finally, we present a summary and comparison among AMTNet and TraMNet in Section~\ref{summary:tramnet}. 

%% file: chapters/partII/tramnet_approach.tex

\begin{figure}[h]
  \centering
  \includegraphics[scale=0.23]{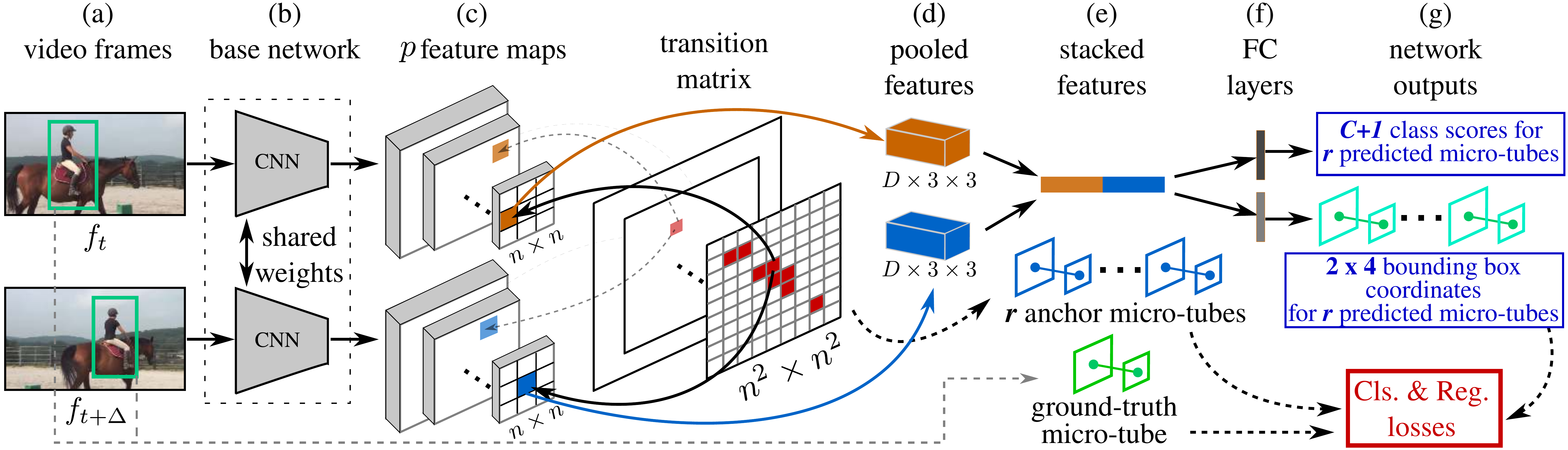}
  \caption[Overview of transition matrix network (TraMNet).]
  {Overview of our proposed TraMNet at training time.}
\label{tramnet:fig:algorithmOverview} 
\end{figure}

\section{Overview of the approach.}\label{tramnet:sec:overview}

The network architecture of AMTNet and TraMNet 
builds on the architectural components of SSD~\cite{liu2016ssd} (Fig.~\ref{tramnet:fig:algorithmOverview}).
The proposed network takes as input a pair of successive video frames $f_t, f_{t+\Delta}$ 
(where $\Delta$ is the inter-frame distance) (Fig.~\ref{tramnet:fig:algorithmOverview}~{(a)})
and propagates these frames through 
a base network comprised of two parallel CNN networks
(\S~\ref{tramnet:sec:base_network} Fig.~\ref{tramnet:fig:algorithmOverview}~{(b)}),
which produce two sets of $p$ conv feature maps
$K^{t}_{p}$ and $K^{t+\Delta}_{p}$ forming a pyramid~(Fig.~\ref{tramnet:fig:algorithmOverview}~{(c)}).
These feature pyramids are used by a configurable pooling layer 
(\S~\ref{tramnet:sec:reconfig_pooling_layer} and Fig.~\ref{tramnet:fig:algorithmOverview}~{(d)})
to pool features based on the transition probabilities defined by a 
transition matrix $\mathbf{A}$, probabilistic matrix for TraMNet (\S~\ref{tramnet:subsec:transition_matrix})
and identity matrix for AMTNet (\S~\ref{tramnet:subsec:amntnet_matrix}).
The pooled convolutional features are then stacked 
(\S~\ref{tramnet:sec:reconfig_pooling_layer} and Fig.~\ref{tramnet:fig:algorithmOverview}~{(e)}),
and the resulting feature vector is passed to two parallel fully connected (linear) layers
(one for classification and another for micro-tube regression, see \S~\ref{tramnet:subsec:conv_2_linear} and Fig.~\ref{tramnet:fig:algorithmOverview}~{(f)}),
which predict the output micro-tube  and its classification scores for each class $C$.

Each training mini-batch is used to compute the classification and
micro-tube regression losses given 
the output predictions, 
ground truth and anchor micro-tubes.
We call our network \emph{``configurable''} because 
the configuration of the pooling layer (see Fig.~\ref{tramnet:fig:algorithmOverview}~{(d)})
depends on the transition matrices $\mathbf{A}$, 
and can be changed by altering the threshold applied to $\mathbf{A}$ 
(see Section~\ref{tramnet:subsec:sampling})
or by replacing the transition matrices with a new one for another dataset or having a different transition matrices at test time (see \S~\ref{tramnet:subsec:generality}).




\section{Base network}\label{tramnet:sec:base_network}

\begin{figure}[h!]
  \centering
  \includegraphics[scale=0.23]{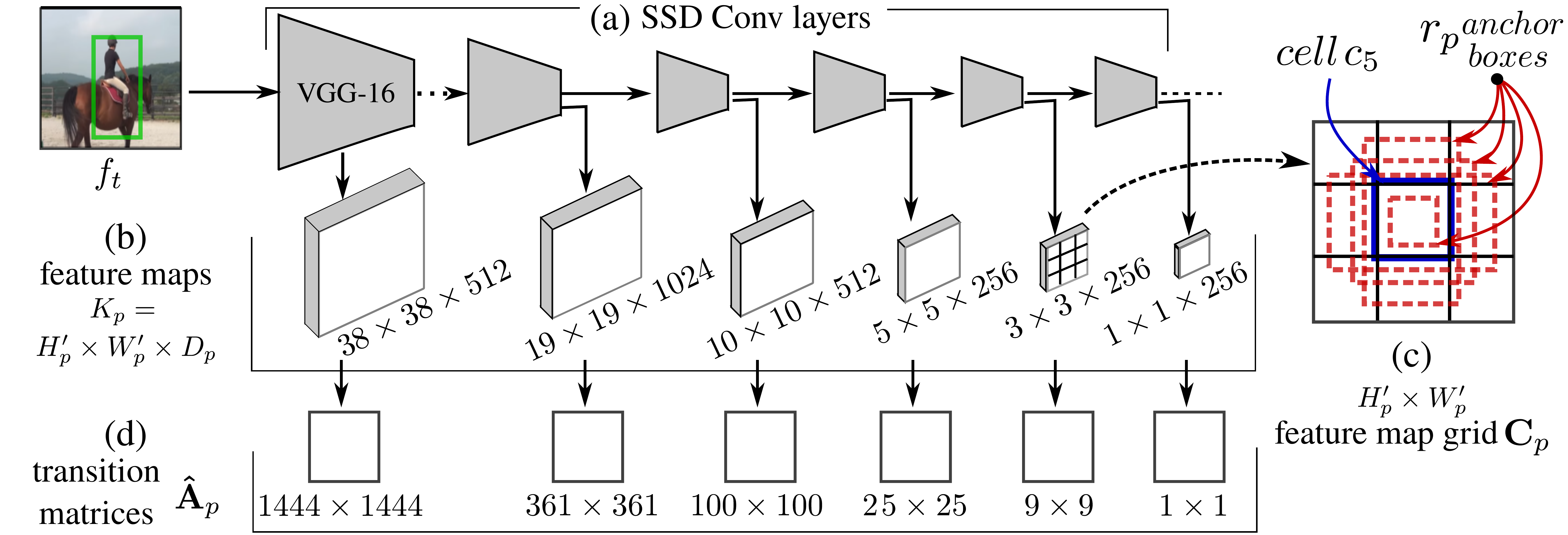}
  \caption[Base network architecture for AMTNet and TraMNet.]
      {Base network architecture.
      \textbf{(a)} SSD convolutional layers;
      \textbf{(b)} the corresponding convolutional feature maps outputted by each convolution layer;
      \textbf{(c)} $r$ anchor boxes with different aspect ratios
      assigned to cell location $c_5$ of the $3 \times 3$ feature map grid;
      \textbf{(d)} transition matrices for the $P$ feature map grids in the pyramid, where $P=6$.}
\label{tramnet:fig:base_net}
\end{figure}

The base network takes as inputs a pair of video frames
$(f_t,f_{t+\Delta})$ and propagates them through two parallel CNN streams
(cf. Fig.~\ref{tramnet:fig:algorithmOverview}~{(b)}).
In Fig.~\ref{tramnet:fig:base_net}~{(a)},
we show the network diagram of one of the CNN streams; the other follows the same design.

The network architecture is based on that of the Single-Shot-Detector (SSD)~\cite{liu2016ssd}.
The CNN stream outputs a set of $P$ convolutional feature maps
$K_{p}$, $p = \{ 1,2, ..., P = 6 \}$
(feature pyramid, cfr. Fig.~\ref{tramnet:fig:base_net}~{(b)})
of shape $[H'_{p} \times W'_{p} \times D_p]$,
where $H'_{p}$, $W'_{p}$ and $D_p$ are the height, width and depth
of the feature map at network depth $p$, respectively.
The feature maps are extracted at different depths $p = \{ 1,2, ..., 6 \}$ of the network.
For $P=6$, the convolutional feature map spatial dimensions are $H' = W' = \{ 38, 19, 10, 5, 3, 1\}$, respectively.
The feature maps at the lower depth levels (i.e., $p=1,2$ or 3)
are responsible for encoding smaller objects/actions,
whereas feature maps at higher depth levels encode
larger actions/objects.
For each cell location $c_{ij}$ of $[H'_{p} \times W'_{p}]$ feature map grid $\mathbf{C_p}$,
$r$ anchor boxes (each with different aspect ratio or scale)
are assigned where $r_{p} = \{ 4, 6, 6, 6, 4, 4 \}$.
E.g. at each cell location of the $3 \times 3$ grid in the pyramid,
$4$ anchor boxes are produced (Fig.~\ref{tramnet:fig:base_net}~{(c)}),
resulting in a total of $3 \times 3  \times 4 = 36$ anchor boxes.
These anchor boxes, assigned for all $P=6$ distinct feature map grids,
are then used to generate action proposal hypotheses based on the
transition probability matrix, as explained below.

Note that the proposed framework is not limited to any particular base network architecture,
and is flexible enough to accommodate any latest 3D networks~\cite{lin2017focal,carreira2017quo,nonlocal2018wang} or single-stage object detectors~\cite{lin2017focal,lin2017feature}. 

\section{HMM-based action proposal generation} \label{tramnet:sec:hmm}
A \emph{hidden Markov model} (HMM) models a time series of (directly measurable) \emph{observations} $\mathbf{O} = \{ \mathbf{o}_1, \mathbf{o}_2, ..., \mathbf{o}_T\}$, either discrete or continuous, as randomly generated at each time instant $t$ by a \emph{hidden state} $\mathbf{q}_t \in \mathbf{Q} = \{ \mathbf{q}_1, \mathbf{q}_2, ..., \mathbf{q}_{N} \}$, whose series form a Markov chain, i.e., the conditional probability of the state at time $t$ given $\mathbf{q}_1,...,\mathbf{q}_{t-1}$ only depends on the value of the state $\mathbf{q}_{t-1}$ at time $t-1$. The whole information on the time series' dynamics is thus contained in a \emph{transition probability matrix}
$\mathbf{A} = [p_{ij}; i,j=1,..,n]$,
where $p_{ij} = P(\mathbf{q}_j|\mathbf{q}_i)$ is the probability of moving from state $i$ to state $j$, and $\sum^{N}_{j=1} p_{ij} = 1$ $\forall i$.
\\
In our setting, a state $\mathbf{q}_n$ is a vector containing the 2D bounding-box coordinates
of one of the anchor boxes
$[x^{a}_{min}, y^{a}_{min}, x^{a}_{max}, y^{a}_{max}]'$ 
in one of the grids forming the pyramid (\S~\ref{tramnet:sec:base_network}).
The transition matrix encodes the probabilities of a temporal link existing between
an anchor box (indexed by $i$) at time $t$ and another anchor box (indexed by $j$) at time $t+\Delta$. The continuous observations $\mathbf{o}_t$, $t=1,...,T$ are the ground-truth bounding boxes, so that $\mathbf{O}$ corresponds to a ground-truth action tube.

In hidden Markov models, observations are assumed to be Gaussian distributed given a state $\mathbf{q}_i$, with mean $\mathbf{o}_\mu^i$ and covariance $\mathbf{Q}_\Sigma^i$.
After assuming an appropriate distribution for the initial state, e.g. $P(\mathbf{q}_0)\sim \mathcal{N}(0,I)$, the transition model $A = [P(\mathbf{q}_j|\mathbf{q}_i)]$ allows us to predict at each time $t$ the probability $P(\mathbf{q}_t | \mathbf{O}_{1:t})$ of the current state given the history of previous observations, i.e.,
the probability of each anchor box at time $t$ given the observed (partial)
ground-truth action tube.
Given a training set, the optimal HMM parameters ($A$, $\mathbf{o}_\mu^i$ and $\mathbf{Q}_\Sigma^i$ for $i=1,...,N$) can be learned using standard expectation maximisation (EM) or the Baum-Welch algorithm, by optimising the likelihood of the predictions $P(\mathbf{q}_t | \mathbf{O}_{1:t})$ produced by the model.

Once training is done, at test time, the mean
$\mathbf{o}_\mu^{\hat{\mathbf{q}}_t}$ of the conditional distribution of the observations given the state associated with the predicted state $\hat{\mathbf{q}}_t \doteq \arg \max_i P(\mathbf{q}_i|\mathbf{O}_{1:t})$ at time $t$
can be used to initialise the anchor boxes for each of the $P$ CNN feature map grids (\S~\ref{tramnet:sec:base_network}).
The learnt transition matrix $\mathbf{A}$ can be used to
generate a set of training action proposals hypotheses (i.e., anchor micro-tubes, Fig.~\ref{tramnet:fig:opt_prop}).
As in our case the mean vectors $\mathbf{o}_\mu^i$, $i=1,..., N$ are known a-priori
(as the coordinates of the anchor boxes are predefined
for each feature map grid, \S~\ref{tramnet:sec:base_network}),
we do not allow the M-step of EM algorithm to update $\mathbf{Q}_{\mu} = [\mathbf{o}_\mu^i, i=1,...,N]$. Only the covariance matrix $\mathbf{Q}_{\Sigma}$ is updated.

\subsection{Approximation of the HMM transition matrix}\label{tramnet:subsec:transition_matrix}

Although the above setting perfectly formalises the anchor box-ground truth detection relation over the time series of training frames,
a number of computational issues arise. At training time,
some states (anchor boxes)
may not be associated with any of the observations (ground-truth boxes) in the E-step, leading to
zero covariance for those states. 
Furthermore, for a large number of states (in our case $N = 8732$ anchor boxes),
it takes around $4$ days to complete a single HMM training iteration.

In response, we propose to approximate the HMM's transition probability matrix $\mathbf{A}$ with
a matrix $\mathbf{\hat{A}}$ generated by a heuristic approach explained below.
The problem is to learn a transition probability, i.e.,
the probability of a temporal link (edge)
between two anchor boxes $\{b^{a}_{t}, b^{a}_{t+\Delta}\}$
belonging to two feature map grids
$\mathbf{C}^{t}_{p}$ and $\mathbf{C}^{t+\Delta}_{p'}$.
If we assume that transitions only take place between states at the same level $p = p'$ of the feature pyramid, the two sets of anchor boxes
$\mathcal{B}^{t}_{p} = \{ b^{a}_{t_{1}}, ..., b^{a}_{t_{N}} \}$ and
$\mathcal{B}^{t+\Delta}_{p} = \{b^{a}_{{(t+\Delta)}_{1}}, ..., b^{a}_{{(t+\Delta)}_{N}}\}$
belonging to a pair of grids
$\{\mathbf{C}^{t}_{p},\mathbf{C}^{t+\Delta}_{p}\}$
are identical, namely:
$\mathcal{B}^{t}_{p} = \mathcal{B}^{t+\Delta}_{p} \doteq \mathcal{B}_{p} = \{b^a_i, i=1,...,N\}$,
allowing us to remove the time superscript. 
Recall that each feature map grid $C_{p}$ has spatial dimension $[H'_{p} \times W'_{p}]$.

We compute a transition probability matrix $\mathbf{\hat{A}}_{p}$ \emph{individually for each grid level} $p$, 
resulting in $p$ such matrices of shape $[(H'_{p})^2 \times (W'_{p})^2]$ (see Fig.~\ref{tramnet:fig:base_net}~{(d)}).
For example, at level $p=5$ we have a $3 \times 3$ feature map grids, so that
the transition matrix $\mathbf{\hat{A}}_{p}$ will be $[3^2 \times 3^2]$.
Each cell in the grid is assigned to $r_{p}$ anchor boxes, resulting in
$n = H'_{p} \times W'_{p} \times r_{p}$ total anchor boxes per grid
(\S~\ref{tramnet:sec:base_network}).
Note that here we are interested in learning a transition probability between
two anchor boxes 
belonging to two different cell locations
in the grid, 
thus further reducing the search space complexity from
$\mathcal{O}(n^f)$ to $\mathcal{O}(L^f)$
where, $L$ is the number of cell locations in a grid.

\subsection{Transition matrix computation for TraMNet}\label{tramnet:subsec:tm_computation}
Initially, all entries of the transition matrix are 
set to zero: $\mathbf{\hat{A}}[i,j] = 0$.
Given a ground-truth micro-tube $\mathbf{m}^{g} = \{b^{g}_{t}, b^{g}_{t+\Delta}\}$
(a pair of temporally linked ground-truth boxes~\cite{saha2017amtnet}),
we compute the IoU overlap for each ground-truth box
with all the anchor boxes $\mathcal{B}_{p}$ in the considered grid, namely:
$IoU(b^{g}_{t}, \mathcal{B}_{p})$ and $IoU(b^{g}_{t+\Delta}, \mathcal{B}_{p})$.
We select the pair of anchor boxes
$\mathbf{m}^{a} = \{ b^{a}_{i}, b^{a}_{j} \}$
(which we term \emph{anchor micro-tube})
having the maximum IoU overlap with $\mathbf{m}^{g}$,
where $i$ and $j$ are two cell locations.
If $i = j$ (the resulting anchor boxes are in the same location) we get an anchor cuboid, otherwise a general anchor micro-tube.

It is repeated for all $P$ feature map grids $\mathbf{C}_{p}$
to select the anchor micro-tube $\mathbf{m}^{a}_{p}$ with the highest overlap.
The best match anchor micro-tube $\mathbf{m}^{a}_{\hat{p}}$
for a given ground-truth micro-tube $\mathbf{m}^{g}$
is selected among those $P$, and
the transition matrix is updated as follows:
$\mathbf{\hat{A}}[i, j] = \mathbf{\hat{A}}[i, j] + 1$.
The above steps are repeated for all the ground-truth micro-tubes in a training set.
Finally, each row of the transition matrix $\mathbf{\hat{A}}$ is normalised by dividing
each entry by the sum of that row.

\begin{figure*}[ht!]
  \centering
  \includegraphics[scale=0.18]{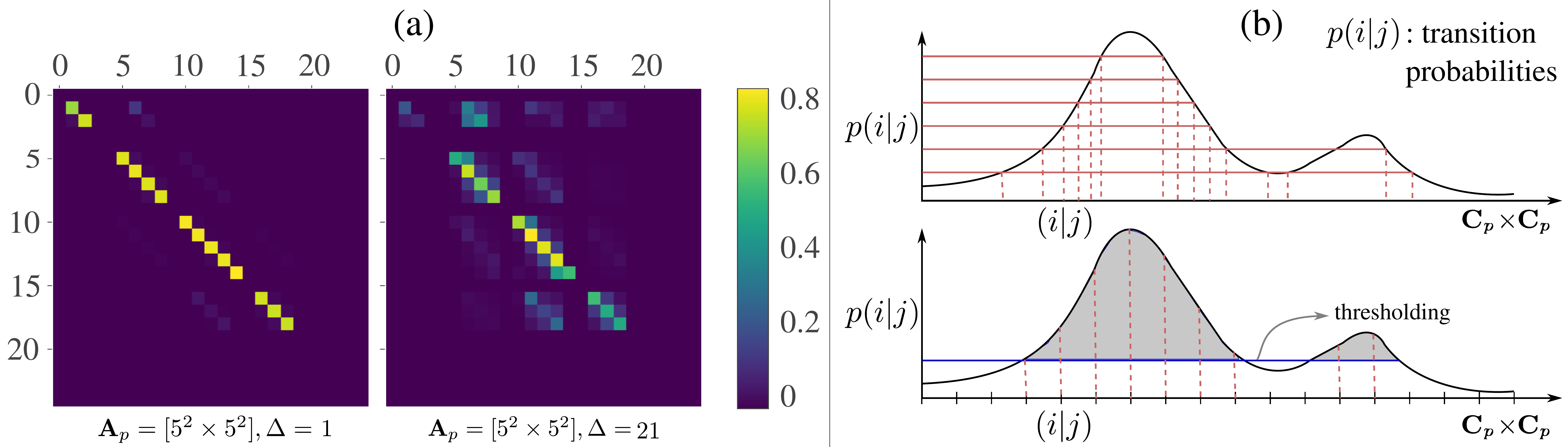}
   \vskip -0cm
  \caption[Transition matrix and sampling of transition hypotheses.]{
      \textbf{(a)}
      Transition matrix for a $5 \times 5$ feature map grid ($p=4$)
      for different $\Delta$ values.
      As $\Delta$ increases,
      off-diagonal probability values also increase,
      indicating a need for anchor micro-tubes rather than anchor-cuboids.
      \textbf{(b)} {Top - Monte Carlo sampling of transition hypotheses $(i,j) \in \mathbf{C}_{p} \times \mathbf{C}_{p}$ based on uniformly sampling the $[0,1]$ range. Bottom - our anchor micro-tube sampling scheme, based on thresholding the transition probabilities $p(i|j)$, is also stochastic in nature and emulates Monte Carlo sampling. The blue line denotes the threshold and the shaded area above the threshold line shows the sampling region, a subset of the product grid $\mathbf{C}_{p} \times \mathbf{C}_{p}$.}
 }
\label{tramnet:fig:trans_mat}
\end{figure*}
\noindent
Fig.~\ref{tramnet:fig:trans_mat} plots the transition matrix $\mathbf{\hat{A}}_p$ for $p=4$ (a feature map grid $5 \times 5$), for different values of $\Delta$.
As explained in the following, the configurable pooling layer
employs these matrices to pool convolutional features for action proposal classification and regression.

Although our approach learns transition probabilities for anchor boxes
belonging to the same feature map grid $\mathbf{C}_{p}$, we realise that
the quality of the resulting action proposals could be further
improved by learning transitions between anchors
across different levels of the pyramid. 
As the feature dimension of each map varies in SSD, e.g. 1024 for $p=2$ and 512 for $p=1$, 
a more consistent network such as FPN~\cite{lin2017focal} with ResNet~\cite{he2016deep} would be a better choice as base architecture. 
Here we stick to SSD to produce a fair comparison with~\cite{kalogeiton2017action,singh2017online,saha2017amtnet}, and
leave this extension to future work.

\subsection{Identity matrix for AMTNet}\label{tramnet:subsec:amntnet_matrix}
As explained in the introduction and overview, the AMTNet is a sub-case of TraMNet. We can implement AMTNet by replacing the transition matrix in TraMNet by an identity, i.e.

\begin{equation}
  \mathbf{\hat{A}}[i, j] = \begin{cases} 1 &\mbox{if} \; i==j \\
  0 &\mbox{otherwise}. \end{cases}
\end{equation}
In all of our experiment on AMTNet~\cite{saha2017amtnet}, we use similar implementation and hyperparameters as for TraMNet.

\section{Configurable pooling layer} \label{tramnet:sec:reconfig_pooling_layer}

The SSD~\cite{liu2016ssd} network uses convolutional kernels of dimension
$[3 \times 3 \times D]$ as classification and regression layers (called \emph{classification} and \emph{regression heads}).
More specifically, SSD uses $r \times 4$ kernels for bounding box regression
(recall $r$ anchor boxes with different aspect ratios are assigned to each cell location (\S~\ref{tramnet:sec:base_network}))
and $(C+1) \times r$ kernels for classification over the $p$ convolutional feature maps
(\S~\ref{tramnet:sec:base_network}). This is fine when the number of proposal hypotheses
is fixed (e.g., for object detection in images, the number of anchor boxes is set to $8732$).
In our setting, however,
the number of proposals varies depending upon the cardinality $| \mathbf{\hat{A}}_{p} |$ of the transition matrix (\S~\ref{tramnet:subsec:transition_matrix}).
Consequently, it is more principled to implement the classification and regression heads
as fully connected layers (see Fig.~\ref{tramnet:fig:algorithmOverview}~{(f)}).
If we observe consistent off-diagonal entries in the transition matrices (e.g. lots of cells moving one step in the same direction),
we could perform pooling as convolution feature map stacking with padding to allow spatial movement.
However, transition matrices are empirically extremely sparse 
(e.g., there are only 25 and 1908 off-diagonal non-zero entries in the transition matrices at $\Delta$ equal to 4 and 20, respectively, on the UCF101-24 dataset).

\subsection{Anchor micro-tube sampling}\label{tramnet:subsec:sampling}

Each transition matrix is converted into a binary one by a threshold so that the cardinality of the matrix depends not only on the data but also on the transition probability threshold. 
Our transition matrix-based anchor micro-tube sampling scheme is stochastic in nature and emulates Monte Carlo sampling techniques (Fig.~\ref{tramnet:fig:trans_mat} (b)). 
A threshold on the transition matrix allows us to sample a variable number of anchors rather than a fixed one.  
We empirically found that a 10\% threshold gives the best results in all of our tests. We discuss the threshold and its effect on performance in \S~\ref{tramnet:subsec:transition_matrix}. 

The pooling layer (see Fig.~\ref{tramnet:fig:algorithmOverview}~{(d)}) is configured to
pool features from a pair of convolutional feature maps
$\{K^{t}_{p}$, $K^{t+\Delta}_{p}\}$, each of size
$[H'_{p} \times W'_{p} \times D]$.
The pooling is done at cell locations 
$i$ and $j$,
specified by the estimated (thresholded) transition matrix $\mathbf{\hat{A}}_p$ 
(\S~\ref{tramnet:subsec:transition_matrix}).
The pooling kernel has dimension $[3 \times 3 \times D]$.
Pooled features are subsequently stacked
(see Fig.~\ref{tramnet:fig:algorithmOverview}~{(e)})
to get a single feature representation of size
$[2 \times 3 \times 3 \times D]$ per anchor micro-tube.

\subsection{Classification and regression layers}\label{tramnet:subsec:conv_2_linear}
After pooling and stacking, we get $M$ convolutional features of size
$[2 \times 3 \times 3 \times D]$, for each $M$ anchor micro-tube cell regions,
where $M = \sum_{p=1}^{P=6} |\mathbf{\hat{A}}_p|$
is the sum of the  cardinalities of the $P$ transition matrices.
We pass these $M$ features to a classification layer
$( (18 \times D)$, $((C+1) \times r) )$,
and a regression layer
$( (18 \times D)$, $((2 \times 4) \times r) )$
(see Fig.~\ref{tramnet:fig:algorithmOverview}~{(f)}).
The classification layer outputs $C+1$ class scores
and the regression layer outputs $2 \times 4$ bounding-box coordinates
for $r$ anchor micro-tubes per anchor micro-tube cell region
(see Fig.~\ref{tramnet:fig:algorithmOverview}~{(g)}).
The linear classification and regression layers have the
the same number of parameters as the convolutional heads in the SSD network~\cite{liu2016ssd}.

\subsection{Generality at test time}\label{tramnet:subsec:generality}

Our approach allows the pooling layer to be \emph{``configurable''}
for any transition matrix,
while keeping the network parameters intact.
This configurable nature provides us with the advantage of using different transition matrices
at test time, we can generate a more generic set of anchor micro-tubes
by setting the transition probabilities to $1$ for those diagonal or off-diagonal entries
which were earlier $0$ for the training set, or accommodating all anchor micro-tubes transitions relative to each cell.
At test time, in cases in which test $\Delta$ is larger than train $\Delta$, 
we can use the Markov chain rule to generate the new transition matrix. 

\begin{figure}[h]
  \centering
  \includegraphics[scale=0.32]{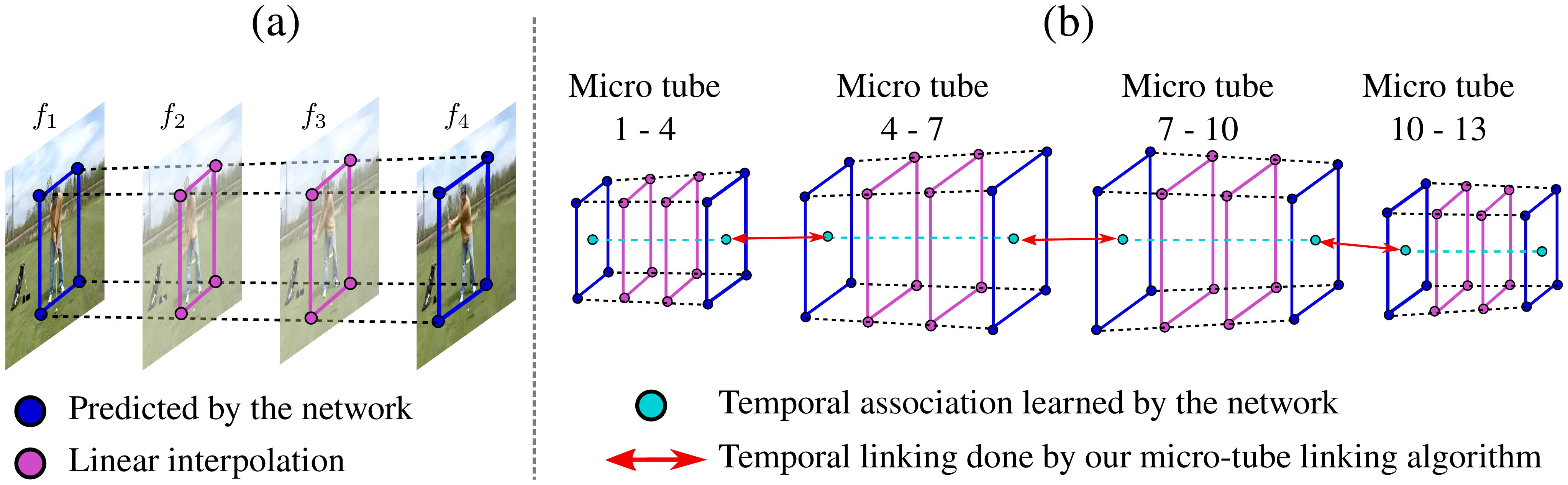}
  \caption[Micro-tube interpolation and linking illustration.]{
    \textbf{(a)} Action micro-tube generation by linear interpolation of coordinates of predicted bounding boxes. 
    The blue circles denote the $8$ $(x,y)$ coordinate values of the bounding boxes predicted by our network for a pair of successive (but not necessarily consecutive) test video frames.
    The first frame in the pair is indexed by $t=1$, the second by $t=4$.
    We generate the coordinates of the detection bounding boxes (pink circles) for intermediate frames by linear interpolation of the predicted coordinates.
    \textbf{(b)} Action tube generation takes place via the linking of micro-tubes generated as in \textbf{(a)}. 
    Note that, for a video sequence with $T$ frames, our model only needs to perform $T/\Delta$ forward passes -- as a result, 
    the micro-tube linking algorithm (\S~\ref{tramnet:sec:tube-gen-algo}) only needs to connect $T/\Delta -1 $ of these frames.}
  \label{tramnet:fig:intrpl-mt-linking}
\end{figure}

\section{Online action tube generation}\label{tramnet:sec:tube-gen-algo}
The output of the proposed network is a set of detection micro-tubes
and their class confidence scores (see Fig.~\ref{tramnet:fig:algorithmOverview}~{(g)}).
Based on the previous chapter, we adapt the online action tube generation algorithm
proposed by Singh~\etal~\cite{singh2017online} (presented in Section~\ref{online:subssec:algo} of Chapter~\ref{chapter:online})
to compose these detection micro-tubes into complete action paths (tracklets)
spanning the entire video.
Note that, Singh~\etal~\cite{singh2017online} use their tube generation
algorithm to temporally connect frame-level detection bounding-boxes,
whereas our modified version of the algorithm connects video-level detection
micro-tubes. 
Similarly to~\cite{singh2017online},
we build action paths incrementally by connecting micro-tubes across time.

Let $B_t$ be the set of detection bounding boxes from a
frame $f_t$. 
Since their method is based on a frame-level detector, Singh \etal~\cite{singh2017online} associate boxes in $B_t$ to boxes in $B_{t+1}$. 
In our case, as TraMNet/AMTnet generates 
micro-tubes $m_t \in M_t \doteq B_t^{1} \times B_{t+\Delta}^{2}$ each spanning
a pair of frames $\{f_t,f_{t+\Delta}\}$ separated by an interval of $\Delta$,
the algorithm needs to link the micro-tubes associated with the pair $t,t+\Delta$ with those
$m_{t+\Delta} \in M_{t+\Delta} \doteq B_{t+\Delta}^1 \times B_{t+2\Delta}^2$ associated with the following pair of frames 
$\{f_{t+\Delta},f_{t+2\Delta}\}$.
Note that the set of detections $B_{t+\Delta}^{2}$ generated for time $t+\Delta$ by the network when processing the first pair of frames, 
will in general differ from the set of detections $B_{t+\Delta}^{1}$ for the \emph{same} frame $t+\Delta$ generated by 
the network when processing $\{f_{t+\Delta},f_{t+2\Delta}\}$ frames.

Micro-tube linking happens by associating elements of $B_{t+\Delta}^2$, coming
from $M_t$, with elements of $B_{t+\Delta}^1$, coming from $M_{t+\Delta}$, as shown in Figure~\ref{tramnet:fig:intrpl-mt-linking}(b).
Interestingly, the latter is a relatively easier sub-problem, 
as all such detections are generated based on the same frame, 
unlike the across-frame association problem considered in~\cite{singh2017online}. 
Association is achieved based on both Intersection over Union (IoU) and class score, as 
the tubes are built separately for each class in a multi-label scenario.
For more details, we refer the reader to~\cite{singh2017online}.

\subsection{Bounding box interpolation}\label{tramnet:subsec:bbox-interp}

At test time, for a input pair of successive video frames ($f_{t}$ and $f_{t+\Delta}$), the final outputs of 
AMTnet (\S~\ref{tramnet:fig:algorithmOverview}) are the top $k$ detection micro-tubes and their class confidence scores. 
For example, in Figure~\ref{tramnet:fig:intrpl-mt-linking}~{(a)}
an action micro-tube predicted by the network for an input pair $(f_{1},f_{4})$ ($\Delta = 3$), 
composed by the two blue bounding boxes highlighted and uniquely determined by 8 coordinate values (blue circles)
is depicted. In practice, the network predicts $M$ such micro-tubes for each input pair, where $M$ is a parameter which depends on the number of anchor locations. 
The predicted micro-tube, however, does not provide bounding boxes for the intermediate frames (in the example $f_{2}$ and $f_{3}$). 
We thus generate detection boxes for all intermediate frames using a simple coordinate wise linear interpolation. 

\subsection{Fusion of appearance and motion cues}

We follow a late fusion strategy~\cite{kalogeiton2017action}
to fuse appearance and motion cues, performed at test time after all the detections
are extracted from the two streams.
Kalogeiton~\etal~\cite{kalogeiton2017action}
demonstrated that
\textit{mean} fusion works better than both 
\textit{boost} fusion~\cite{saha2016deep} and
\textit{union-set} fusion~\cite{singh2017online}.
Thus, in this chapter we produce all results 
using \textit{mean} fusion~\cite{kalogeiton2017action}.
We also report an ablation study of the appearance and flow stream performance
in experiment Section (\S~\ref{tramnet:sec:exp}).

%% file: chapters/partII/tramnet_exp.tex
\section{Experiments} \label{tramnet:sec:exp}
We first present datasets, evaluation setup, fair comparison and 
implementation details used in Section~\ref{tramnet:subsec:datasets}.
Here, we show how AMTNet and TraMNet is able to improve spatial-temporal action detection in Section~\ref{tramnet:subsec:st_performance}.
Thirdly, in Section~\ref{subsec:generality}, we discuss how a network learned using transition matrices is able to generalise at test time, when more general anchor-micro-tubes are used to evaluate the network. 
Finally, in Section~\ref{subsec:sparsity}, we quantitatively demonstrate that TraMNet can effectively handle sparse annotation as in the DALY dataset, and generalise well on the various train and test $\Delta$'s on the UCF101-24 dataset.

\subsection{Datasets and evaluation setup}\label{tramnet:subsec:datasets}
Similar to previous chapter and previous works~\cite{kalogeiton2017action,singh2017online}. 
we evaluate our models on the UCF101-24~\cite{ucf101} to validate spatiotemporal detection performance.
We use DALY~\cite{daly2016weinzaepfel} and UCF101-24~\cite{ucf101} to evaluate the method on sparse annotations.

\textbf{Transformed-UCF101-24} was created by us by padding all images along both the horizontal and the vertical dimension. 
We set the maximum padding values to
32 and 20 pixels, respectively, as $40\%$ of the average width (80) and height (52) of bounding box annotations.
A uniformly sampled random fraction of 32 pixels is padded on the left edge of the image, 
the remaining is padded on the right edge of the image. Similar random padding is performed at the top and bottom of each frame. 
The padding itself is obtained by mirroring the adjacent portion of the image through the edge.
The same offset is applied to the bounding box annotations.

\textbf{Evaluation setup.} \label{para:eval_metric}
Similar to previous chapter, we use \emph{``average detection performance''} (avg-mAP) to compare AMTNet's and TraMNet's performance 
with the previous state-of-the-art approaches~\cite{peng2016eccv,yu2015fast,singh2017online,kalogeiton2017action} on UCF101-24 dataset.

On the DALY dataset, we also evaluate at various thresholds in both an untrimmed and a trimmed setting. 
The latter is achieved by trimming the action paths generated by the boundaries of the ground truth~\cite{daly2016weinzaepfel}. 
We further report the video classification accuracy using the predicted tubes as in~\cite{singh2017online}, 
in which videos are assigned the label of the highest-scoring tube.
One could improve classification performance on DALY by 
taking into consideration other tube scores.
Nevertheless, in our tests, we adopt the existing protocol.

\noindent
For \textbf{\textit{fair comparison}} \label{para:fair_compar}
we re-implemented the methods of our previous\footnote{This work was published in ACCV 2018, we compare with state-of-the-art approaches at the time of submission.} 
competitors~\cite{kalogeiton2017action,singh2017online}
with SSD as the base network. As in our AMTNet/TraMNet networks, 
we also replaced SSD's convolutional heads with new linear layers. 
The same tube generation~\cite{singh2017online} and data augmentation~\cite{liu2016ssd} methods were adopted,
and the same hyperparameters were used for training all the networks, including TraMNet and AMTNet.
The only difference is that the anchor micro-tubes used in~\cite{kalogeiton2017action} were cuboidal for six frames, 
whereas AMTNet's and TraMNet's anchor micro-tubes are generated using transition matrices for two consecutive frames separated by $\Delta$.
We refer to these approaches as SSD-L (SSD-linear-heads)~\cite{singh2017online}, and as ACT-L (ACT-detector-linear-heads)~\cite{kalogeiton2017action}.

\noindent
\subsection{Network training and implementation details.}
We used well-established training settings from previous chapters for all the above methods.
While training on the UCF101-24 dataset,
we used a batch size of $16$ and an initial learning rate of $0.0005$, with the
learning rate dropping after $100K$ iterations for the appearance stream and $140K$ for the flow stream.
Whereas the appearance stream is only trained for $180K$ iterations,
the flow stream is trained for $200K$ iterations.
In all cases, the input image size was $3\times 300\times 300$ for the appearance stream, while a
stack of five optical flow images~\cite{brox2004high} ($15\times 300\times 300$) was used as input for flow-stream.
Each network was trained on $2$ 1080Ti GPUs.
SSD-L and ACT-L are also trained using the same training setup.

\begin{table}[h]
  \centering
  \setlength{\tabcolsep}{3.5pt}
  \caption[Action detection results on UCF101-24.]{Action detection results on untrimmed videos from UCF101-24 split1.
  The table is divided into 5 parts. The first part lists approaches which have single frames as input;
  the second part approaches which take multiple frames as input;
  the third part contemplates the re-implemented versions of previous approaches~\cite{singh2017online,kalogeiton2017action};
  the fourth part reports the performance of AMTNet; 
  lastly, we report TraMNet's performance.}
  \vspace{-0.01cm}
  {\footnotesize
  \scalebox{0.95}{
  \begin{tabular}{lccccccc}
  \toprule
  Methods & Train $\Delta$ & Test $\Delta$    &$\delta$ = 0.2   &$\delta$ = 0.5 & $\delta$ = 0.75 & $\delta$ = .5:.95 & Acc \% \\ \midrule
  T-CNN~\cite{hou2017tube} & NA &NA & 47.1  & --   & -- & -- & --\\
  MR-TS~\cite{peng2016eccv}     & NA &NA & 73.5  & 32.1 & 02.7 & 07.3 & --\\
  Saha~\etal~\cite{saha2016deep}    & NA &NA & 66.6  & 36.4 & 07.9 & 14.4 & --\\
  SSD~\cite{singh2017online}    & NA & NA & 73.2 & 46.3 & 15.0 & 20.4 & --\\\midrule
  AMTnet~\cite{saha2017amtnet} rgb-only  & 1,2,3 & 1 & 63.0  & 33.1 & 00.5 & 10.7 & --\\
  ACT~\cite{kalogeiton2017action}   & 1 & 1 & 76.2  & 49.2 & 19.7 & 23.4 & --\\
  RTPR (VGG)~\cite{li2018recurrent}   & 1 & 1 & 76.3  & -- & -- & -- & --\\
  Gu \etal~\cite{ava2017gu} (\cite{peng2016eccv} + \cite{carreira2017quo})   & NA & NA & --  & \textbf{59.9} & -- & -- & --\\
  \midrule
  SSD-L with-trimming  & NA & NA & 76.2 & 45.5 & 16.4 & 20.6  & 92.0 \\
  SSD-L     & NA & NA & 76.8 & 48.2 & 17.0 & 21.7  & 92.1 \\
  ACT-L     & 1 & 1 & 77.9 & 50.8 & 19.8 & \textbf{23.9}  & 91.4 \\ \midrule
  AMTnet (ours) & 1 & 1 & \textbf{79.4} & \textbf{51.2} & 19.0 & 23.4  & \textbf{92.9} \\
  AMTnet (ours) & 5 & 5 & 77.5 & 49.5 & 17.3 & 22.5  & 91.6 \\
  AMTnet (ours) & 21 & 5 & 76.2 & 47.6 & 16.5 & 21.6  & 90.0 \\\midrule
  TraMNet (ours)& 1 & 1 & 79.0 & 50.9 & \textbf{20.1} & \textbf{23.9}  & 92.4 \\
  TraMNet (ours)& 5 & 5 & 77.6 & 49.7 & 18.4 & 22.8  & 91.3 \\
  TraMNet (ours)& 21 & 5 & 75.2 & 47.8 & 17.4 & 22.3  & 90.7 \\ \bottomrule
  \end{tabular}
  }
  }
  \label{tramnet:table:ucf101_results} \vspace{-6mm}
\end{table}
\subsection{Action detection performance}\label{tramnet:subsec:st_performance}

Table~\ref{tramnet:table:ucf101_results} shows the resulting performance on UCF101-24 at multiple train and test $\Delta$s for TraMNet versus other competitors~\cite{saha2016deep,kalogeiton2017action,singh2017online,peng2016eccv,hou2017tube}.
Note that Gu~\etal~\cite{ava2017gu} build upon MS-TS~\cite{peng2016eccv} by adding a strong I3D~\cite{carreira2017quo} base network, making it unfair to compare \cite{ava2017gu} to SSD-L, AMTnet, ACT-L and TraMNet, which all use 2D VGG~\cite{simonyan2015very} as a base network.

The ACT is a dense network (processes $6$ consecutive frames), which
shows the best performance at high overlap (an avg-mAP of 23.9\%).
AMTnet is slightly inferior ($23.4$\%), 
most likely due to its learning representations from only a pair of consecutive frames at its best training and test settings ($\Delta = 1$).
TraMNet is able to match ACT-L's performance at high overlap ($23.9$\%) while being comparatively more efficient (only using a pair of frames as compared to 6 used by ACT-L).

We cross-validated different transition probability thresholds on transition matrices. Thresholds of $2\%$, $5\%$, $10\%$, $15\%$ and $20\%$ yielded an avg-mAP of $21.6\%$, $22.0\%$, $22.4\%$, $21.9\%$ and $21.2\%$, respectively, on the appearance stream. Given such evidence, we concluded that a $10\%$ transition probability threshold was to be adopted throughout all our experiments.

\subsection{AMTNet comparison with 2D methods}\label{tramnet:subsec:amtnet_comparision}
We SSD as base network for AMTNet and TraMNet, it is only fair to compare AMTNet's performance to that of SSD-L in Table~\ref{tramnet:table:ucf101_results}. 
We can see that AMTNet performs better that the frame-level SSD-L method presented in the previous chapter.

\subsection{TraMNet on Transformed-UCF101-24}
The evaluation of AMTNet on Transformed-UCF101-24 (\S ~\ref{tramnet:subsec:datasets}) shows an avg-mAP of $19.3\%$ using the appearance stream only, 
whereas TraMNet records an avg-mAP of $20.5\%$, a gain of $1.2\%$ that can be attributed to its estimating grid location transition probabilities. 
It shows that TraMNet is more suited to action instances involving substantial shifts from one frame to the next. 
A similar phenomenon can be observed on the standard UCF101-24 when the train or test $\Delta$ is greater than $1$ in Table~\ref{tramnet:table:ucf101_results}.

\subsection{Location invariance at test time}\label{subsec:generality}
Anchor micro-tubes are sampled based on the transition probabilities 
from specific cells (at frame $f_t$) to other specific 
cells (at frame $f_{t+\Delta}$) (\S~\ref{tramnet:subsec:tm_computation}) based on the training data.
As a result, anchor miro-tubes are location dependent.
However, as at test time action instances of the same class may appear 
in other regions of the image plane than those observed at training time,
it is desirable to generate additional anchor micro-tubes proposals than those produced by the learnt transition matrices.
Such \emph{location invariance} property can be achieved at test time by augmenting 
the binary transition matrix (\S~\ref{tramnet:subsec:sampling}) 
with likely transitions from other grid locations.

Each row/column of the transition matrix $\mathbf{\hat{A}}$ (\S~\ref{tramnet:subsec:tm_computation}) 
corresponds to a cell location in the grid.
One augmentation technique is to set all the diagonal entries to 1 (i.e., $\mathbf{\hat{A}}[i,j] = 1$, where $i==j$). 
This amounts to generating anchor cuboids 
which may have been missing at training time (cfr. Fig.~\ref{tramnet:fig:trans_mat} (a)).
The network can then be evaluated using this new set of anchor micro-tubes 
by configuring the pooling layer (\S~\ref{tramnet:sec:reconfig_pooling_layer})) accordingly.
When doing so, however, we observed only a very minor difference in avg-mAP at the second decimal point for TraMNet. 
Secondly, we also evaluated TraMNet by incorporating the transitions from each cell to its 8 \emph{neighbouring} cells (also at test time), 
but observed no significant change in avg-mAP.

A third approach, given a pyramid level $p$, and the initial binary transition matrix for that level, 
consists of computing the relative transition offsets for all grid cells (offset $= i-j \ \forall i,j$ where  $\mathbf{\hat{A}}[i,j] = 1$).
All such transition offsets correspond to different spatial translation patterns (of action instances) present in the dataset at different locations in the given video.
Augmenting all the rows with these spatial translation patterns, by taking each diagonal entry in the transition matrix as a reference point, yields a more dense transition matrix whose anchor micro-tubes are translation invariant, i.e., spatial location invariant.
However, after training TraMNet at train $\Delta = 1$ we observed that the final avg-mAP at test $\Delta = 1$ was $22.6\%$ as compared to $23.9\%$ when using the original (sparse) transition matrix. 
As in the experiments (i.e., added diagonal and neighbour transitions) explained above, we evaluated the network that was trained on the original transition matrices at train $\Delta=1$  by using the transition matrix generated via relative offsets, observing an avg-mAP consistent (i.e., $23.9\%$) with the original results. 

This shows that the system should be trained using the original transition matrices learned from the data, whereas more anchor micro-tube proposals can be assessed at test time without loss of generality.
Hence, we can train our network efficiently and evaluate with more anchors as the evaluation is much faster than training.
It also shows that UCF101-24 is not sufficiently realistic a dataset from the point of view of translation invariance, which is why we conducted tests on Transformed-UCF101-24 (\S ~\ref{tramnet:subsec:datasets}) to highlight this issue. 

\begin{table}[t]
  \vskip 2mm
  \centering
  \caption[Action detection results (video-mAP) on the DALY dataset.]{Action detection results (video-mAP) on the DALY dataset. SSD-L without trimming refers to when action paths are not trimmed and the network is SSD.}
  {\footnotesize
  \scalebox{0.82}{
  \begin{tabular}{lc|ccc|cccc}
 
  \toprule
  \multicolumn{2}{c}{} &\multicolumn{3}{|c|}{Untrimmed Videos} & \multicolumn{4}{c}{Trimmed Videos} \\
  Methods & Test $\Delta$    &  $\delta$=0.2   & $\delta$=0.5 & Acc\% & $\delta$=0.5 & $\delta$=.5:.95 & Acc\% & \small{CleaningFloor} \\\midrule
  Weinzaepfel \etal ~\cite{daly2016weinzaepfel} & NA & 13.9 & -- & -- & 63.9   & -- & -- & --\\
  SSD-L without-trimming & NA & 06.1 & 01.1 & 61.5 & \multicolumn{4}{c}{}\\
  SSD-L & NA & \textbf{14.6} & \textbf{05.7} & 58.5 & 63.9 & 38.2  & 75.5 & 80.2\\
  AMTnet (ours) & 3 & 12.1 & 04.3 & 62.0  & 63.7 & 39.3 & 76.5 & 83.4\\
  TraMNet (ours) & 3 & 13.4 & 04.6 & \textbf{67.0} & \textbf{64.2}  & \textbf{41.4}  & \textbf{78.5} & \textbf{86.6}\\ \bottomrule
  \end{tabular}
  }
  }
  \label{table:daly_results} \vspace{-6mm}
\end{table}

\subsection{Handling sparse annotations}\label{subsec:sparsity}

Table~\ref{table:daly_results} shows the results on the DALY dataset.
We can see that TraMNet significantly improves on AMTnet and AMTNet on SSD-L in the trimmed video setting, 
with an average video-mAP of $41.4$\%.
TraMNet reaches top classification accuracy in both the trimmed and the untrimmed cases.
As we would expect, TraMNet improves the temporal linking via better micro-tubes and classification, 
as clearly indicated in the trimmed videos setting.
Nevertheless, SSD-L is the best when it comes to temporal trimming.
We think this is because of each micro-tube in 
our case is 4 frames long as the test $\Delta$ is equal to 3,
and each micro-tube only has one score vector rather than 4 score vectors for each frame,
which might render temporal segmentation ineffective (loss of discriminative power across frames).
Given the above evidence and in Section~\ref{tramnet:subsec:amtnet_comparision}, we can say that AMTNet and TraMNet are better that spatial detection than temporal trimming as compared to single-frame approach defined by SSD-L from the previous chapter.

DALY allows us to show how TraMNet is able to handle 
sparse annotations better than AMTNet, which uses anchor cuboids, 
strengthening the argument that learning transition matrices help generate better micro-tubes.

TramNet's performance on `CleaningFloor' at $\delta$ equal to 0.5 in 
the trimmed case highlights the effectiveness of general anchor micro-tubes for \textbf{dynamic classes}. 
`CleaningFloor' is one of DALY's classes in which the actor moves spatially while the camera is mostly static. 
To further strengthen the argument, 
we picked classes showing fast spatial movements across frames in the UCF101-24 dataset
and observed the class-wise average-precision (AP) at $\delta$ equal to $0.2$.
For `BasketballDunk', `Skiing' and `VolleyballSpiking' 
TraMNet performs significantly better than both AMTnet and ACT-L; e.g. 
on `Skiing', the performance of TraMNet, 
AMTNet and ACT-L is $85.2$, $82.4$ and $81.1$, respectively.


\subsection{Training and testing at multiple $\Delta$'s}~\label{subsec:deltas}
\begin{figure}[t]
    \centering
    \includegraphics[width=0.96\textwidth]{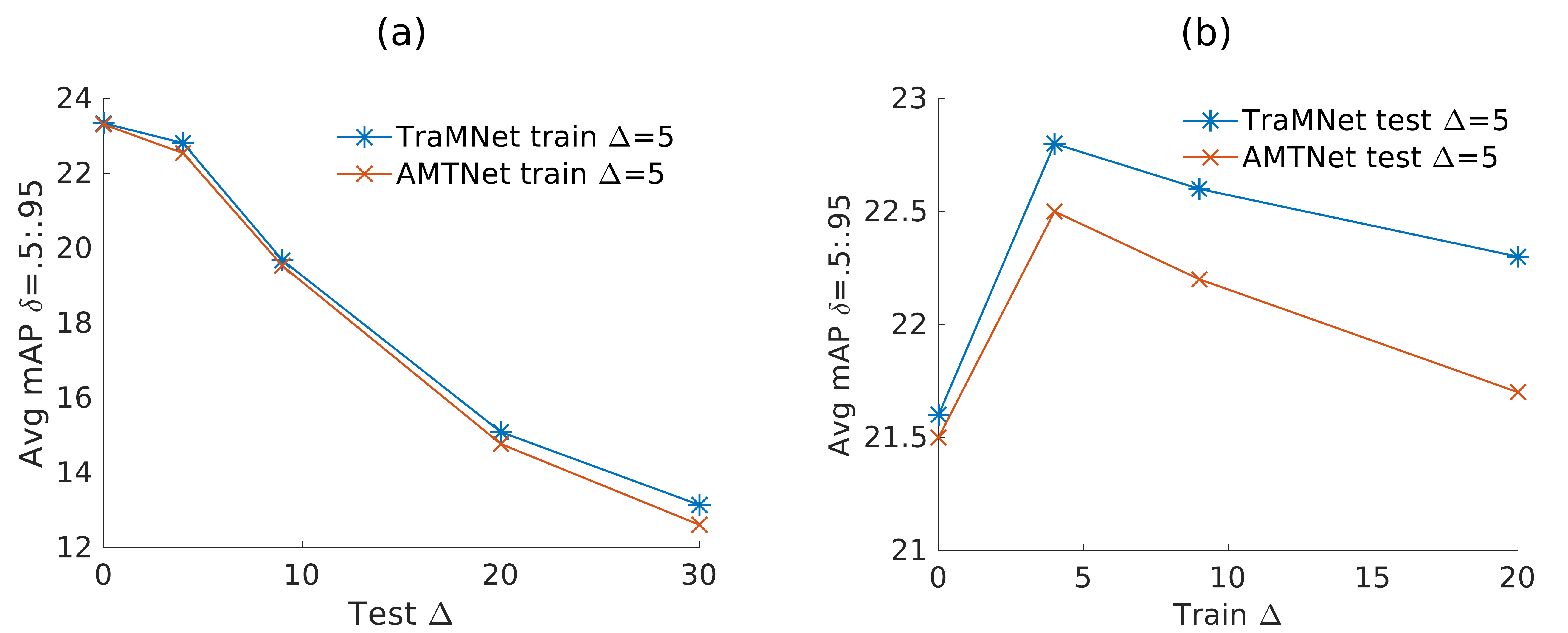} 
    \caption[Performance of TraMNet with various train and test $\Delta$.]
    {Avg mAP ($\delta=0.5:0.95$) performance of TraMNet and AMTNet,
    (a) when tested at increasing $\Delta$ from 1 to 30 and trained at constant $\Delta$ equal to 5,
    (b) when tested at  constant $\Delta$ equal to 5 and trained on increasing $\Delta$ from 1 to 20.}
    \label{tramnet:fig:deltas}
\end{figure}

To further test whether TraMNet can handle sparse annotation 
we introduced an artificial gap ($\Delta$) in UCF101-24's training examples, 
while testing on frames that are far away (e.g. $\Delta = 30$).
We can observe in Figure \ref{tramnet:fig:deltas}(b) 
that performance is preserved when increasing the training $\Delta$ 
while keeping the test $\Delta$ small (e.g. equal to 5, as shown in plot (a)).
One could think of increasing $\Delta$ at test time to improve run-time efficiency: 
we can observe from Figure \ref{tramnet:fig:deltas}(a) that performance drops linearly as speed linearly increases.
In both cases TraMNet consistently outperforms AMTNet.
When $\Delta$ is large TraMNet's improvement is large as well. 

\subsection{Temporal labelling} 

Temporal labelling is performed using the labelling formulation presented in \cite{singh2017online} (similar to Chapter~\ref{chapter:offline} and Chapter~\ref{chapter:online}).
Actually, temporal labelling hurts the performance on UCF101-24, 
as shown in Table \ref{tramnet:table:ucf101_results} where `SSD-L-with-trimming' 
uses \cite{singh2017online}'s temporal labelling technique, whereas `SSD-L' and 
the other methods below that do not.
In contrast, on DALY the results are quite the opposite:
the same temporal labelling framework improves the performance from $6.1\%$ to $14.9\%$ at $\delta = 0.2$.
We think that these (superficially) contradictory results relate to 
the fact that action instances cover on average a very different 
fraction (70\% versus 4\%) of the video duration in UCF101-24 and DALY, respectively.
Hence, we can conclude that temporal trimming does not have major impact on overall performance on UCF101-24 (\S~\ref{tramnet:subsec:st_performance}) dataset except few difficult classes pointed out in Section~\ref{online:exp:flow-contribution} of Chapter~\ref{chapter:online}.

\subsection{Detection speed} 
Similar to previous chapters, we continue to keep our models real-time and online. 
We measured the average time taken for a forward pass for a batch size of 1 as compared to 8 by ~\cite{singh2017online}. 
A single-stream forward pass takes 29.8 milliseconds (i.e. 33fps) on a single 1080Ti GPU. 
One can improve speed even further by evaluating TraMNet with $\Delta$ equal 
to 2 or 4,  obtaining a 2$\times$ or 4$\times$ speed improvement 
while paying very little in terms of performance, as shown in Figure \ref{tramnet:fig:deltas}(a), as explained in Section~\ref{tramnet:subsec:bbox-interp} and Fig.\ref{tramnet:fig:intrpl-mt-linking} .

%% file: chapters/6_tpnet.tex
\chapter{Predicting Future of Action Tubes}
\label{chapter:tpnet}
\renewcommand{\imagepath}{figures/tpnet} 
\input{chapters/partII/tpnet_intro}
\input{chapters/partII/tpnet_approach}

\input{chapters/partII/tpnet_exp}

\section{Summary and limitations}\label{summary:tpnet}
\paragraph{Summary:} 
We presented TPnet, a deep learning framework for future action tube prediction in videos which, 
unlike previous online tube detection methods~\cite{singh2017online,soomro2016predicting}, 
generates future of action tubes as early as when $10\%$ of the video is observed.
It can cope with future uncertainty better than 
the baseline methods while remaining state-of-the-art in action detection task.
Hence, we provide a scalable platform to push the boundaries of action tube prediction research; 
it is implicitly scalable to multiple action tube instances in the video as future prediction is made for each action tube separately. 

In summary, we present a Tube Predictor network (TPNet) which: 
\begin{itemize}
    \item given a partially observed video, can (early) predict video long action tubes in terms of both their classes and the constituting bounding boxes;
    \item demonstrates that training a network to make predictions also helps in improving action detection performance;
    \item demonstrates that feature-based fusion works better than late fusion in the context of spatiotemporal action detection. 
\end{itemize}

\paragraph{Limitations:} 
The current tube prediction method is limited in the following ways: 
i) it work on small videos which contain only one action instance each, 
ii) training is sub-optimal when prediction needs to be made for the long term because videos are small.
iii) feature representation only depends on two frames separated by $\Delta$ steps.

\paragraph{Looking ahead: }
We presented action detection and prediction framework in previous chapters which solely rely on frame-level 2D CNNs for feature representation, which is required for speed purpose.
However, 3D representation from a continuous set of frames is state-of-the-art in many video understanding tasks, as discussed in the related work chapter (Chapter~\ref{chapter:related_work}).
In the next chapter, we will present a 3D representation method which is causal/online in nature. 
These online methods are important for any online action detection/prediction approach to be useful in real-world applications.

%% file: chapters/partII/tpnet_intro.tex
\section{Introduction} 
\label{tpnet:intro}

Imagine a pedestrian on the sidewalk, and an autonomous car cruising on the nearby road. 
If the pedestrian stays on the sidewalk and continues walking, they are of no concern for an autonomous vehicle.
What, instead, if they start approaching the road, in a possible attempt to cross it?
Any future prediction about the pedestrian's action and their possible position on/off the road would crucially help the autonomous car avoid any potential incident. 
It would suffice to foresee the pedestrian's action label and position half a second early to avoid a major accident.
As a result, awareness about surrounding human actions and their future location are essential for the robot-car.

\begin{figure}[t]
  \centering
  \includegraphics[scale=0.7]{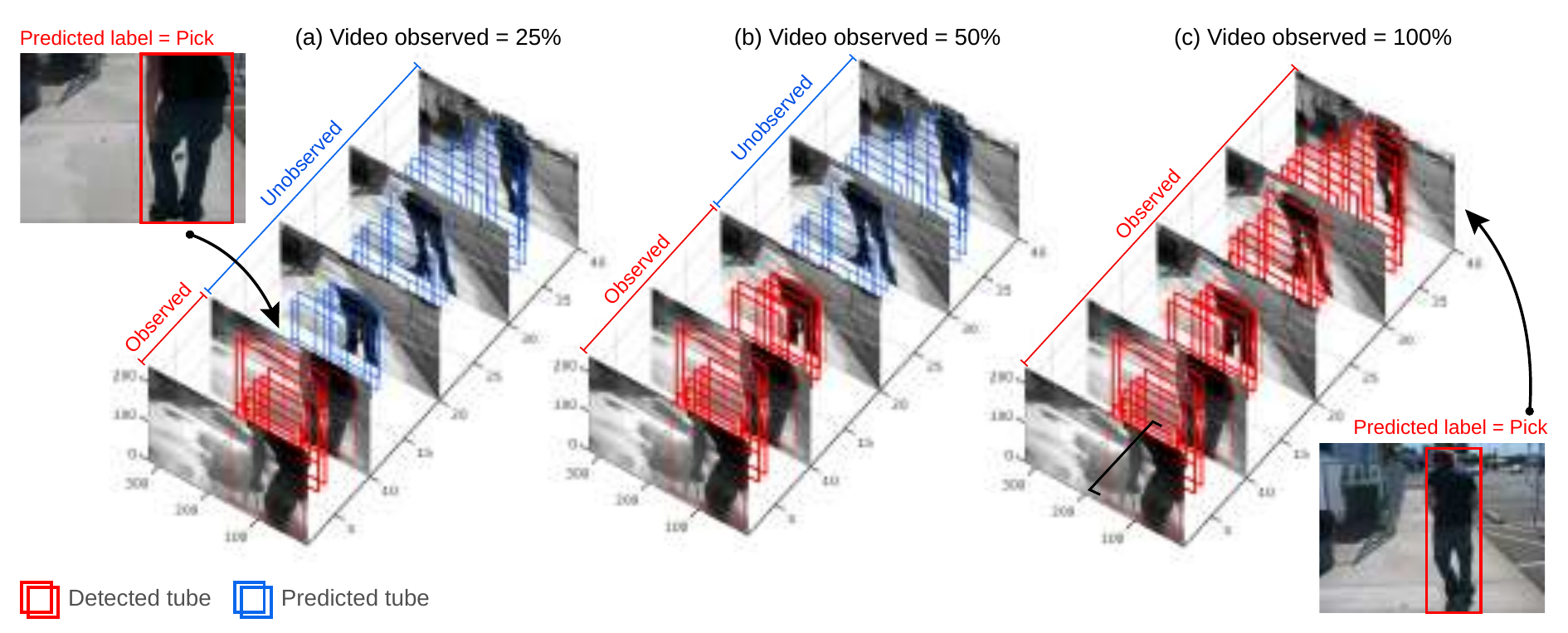}
  \caption[Future prediction of tubes, a problem statement.]{
      An Illustration of the action tube prediction problem using an example in which a ``pickup'' action is being performed on a sidewalk.
      As an ideal case, we want the system to predict an action tube as shown in (c) (i.e. when 100\% of the video has been processed) just by observing 25\% of the entire clip (a).       
      We want the tube predictor to predict the action class label (shown in red) alongside predicting the spatial location of the tube.
        The red shaded bounding boxes denote the detected tube in the observed portion of the input video, whereas, the blue coloured bounding boxes represent the future predicted action tube for the unobserved part of the clip.        
    }
\label{tpnet:fig:problem-statement}
\end{figure}

We can formalise the above problem as follows. 
We seek to predict both the class label and the future spatial location(s) of an action instance as early as possible, 
as shown in Figure~\ref{tpnet:fig:problem-statement}. 
Basically, it translates into early spatiotemporal action detection (similar to~\cite{singh2017online} and 
Chapter~\ref{chapter:online}) and future location prediction of tubes, 
achieved by completing action instance(s) for the unobserved part of the video.
In earlier Chapter~\ref{chapter:online}, we learned to predict the action class label early for an input video by observing a smaller portion (a few frames) of it, 
whilst the system incrementally builds action tubes in an online fashion. 
In contrast, here, we propose to predict both the class label of an action and its future location(s) (i.e., the future shape of an action tube). 
In this chapter, by \emph{`prediction'} we refer to the estimation of both an action's label and location in \emph{future}, unobserved video segments. 
An illustration of above is shown in Figure~\ref{tpnet:fig:problem-statement}.

As observed in Chapter~\ref{chapter:tramnet}, the ability to predict~\emph{action micro-tubes} 
(sets of temporally connected bounding boxes spanning $k$ video frames) 
from pairs of frames~\cite{saha2017amtnet} or sets of $k$ frames~\cite{kalogeiton2017action,hou2017tube} 
provides a powerful tool to extend the single frame-based online localisation approach discussed in Chapter~\ref{chapter:online} (i.e.~\cite{singh2017online}) to a action location prediction approach, 
while retaining its incremental nature.
Combining the basic philosophies of~\cite{singh2017online} (Chapter~\ref{chapter:online}) and 
\cite{saha2017amtnet} (Chapter~\ref{chapter:tramnet}) thus has the potential to provide an interesting and scalable approach to action prediction.
As a result, we will build our tube prediction approach on two previous chapters.
We can extend AMTNet by adding future prediction output heads to AMTNet to predict the future and past of each micro-tube.
We call this new deep network a Tube Predictor Network (`TPNet'). 
We will show that such a network not only able to perform future tube prediction but also improve detection performance.


\paragraph{Related publications.} 
The work presented in this chapter has appeared in 
ECCV workshop on anticipating human behaviour, 2018~\cite{singh2018predicting}.
The dissertation author is the primary investigator in~\cite{singh2018predicting}.
Note, that this publication~\cite{singh2018predicting} and chapter built upon our previous works described in~\cite{singh2017online,saha2017amtnet} and two previous chapters.

\paragraph{Outline:}\label{tramnet:outline}
The remaining chapter is organised as follows.
We start by presenting an overview of our TPNet based tube prediction approach in Section~\ref{tpnet:sec:overview}. 
Next, we describe the modifications in AMTNet as compared to the previous chapter in Section~\ref{tpnet:sec:base_net}.
In Section~\ref{tpnet:sec:training}, we describe the training procedure of our TPNet along with multi-task learning objective in Section~\ref{tpnet:subsec:multitask}.
Next in Section~\ref{tpnet:sec:predict}, we present how to use TPNet in a sliding window fashion in the temporal direction while generating micro-tubes and corresponding future predictions. 
These, eventually, are fed to a tube prediction framework (\S~\ref{tpnet:sec:predict}) 
to generate the future of any current action tube being built using micro-tubes.
Then, we present the evaluation of our TPNet based approach in Section~\ref{tpnet:sec:exp}.
Finally, we present a summary in Section~\ref{summary:tpnet}.

%% file: chapters/partII/tpnet_approach.tex
\begin{figure}[t]
  \centering
  \includegraphics[scale=0.60]{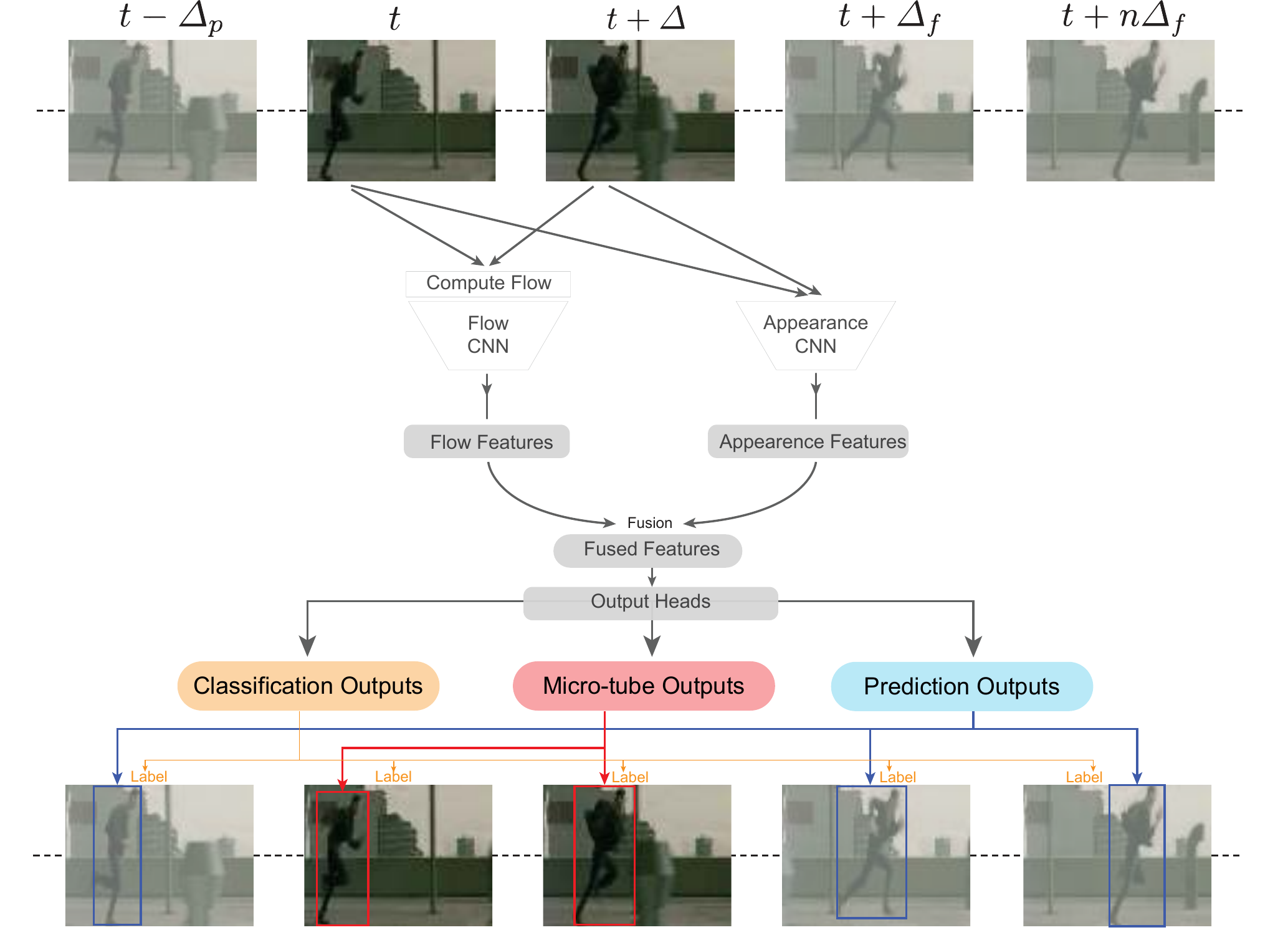}
  \caption[Tube predictor network at test time.]
    {Workflow illustrating the application of TPNet to a test video at a time instant $t$. 
      The network takes frames $f_t$ and $f_{t+\Delta}$ as input and 
      generates classification scores, the
      micro-tube (in red) for frames $f_t$ and $f_{t+\Delta}$, and
      prediction bounding boxes (in blue) for frames $f_{t-\Delta_p}$, $f_{t+\Delta_f}$ up to $f_{t+n\Delta_f}$. 
      All bounding boxes are considered to be linked to the micro-tube.
      Note that predictions also span the past: a setting called \emph{smoothing} in the estimation literature. 
      $n$, $\Delta_f$, and $\Delta_p$ are network parameters that we cross-validate during training.}
\label{tpnet:fig:testing}
\end{figure}
\noindent

\section{Overview of the approach}
\label{tpnet:sec:overview}
We propose to extend the action micro-tube detection 
architecture presented in previous chapter to produce, at any time $t$, past ($\tau<t$), present, and future ($\tau>t$) detection bounding boxes, 
so that each (extended) micro-tube contains bounding boxes for both observed and not yet observed frames. 
All bounding boxes, spanning presently observed frames as well as past and future ones (in which case we call them predicted bounding boxes), 
are considered to be linked, as shown in blue in Figure~\ref{tpnet:fig:testing}.

Further, in virtue of TPNet and online tube construction, 
the temporally linked micro-tubes forming each 
currently detected action tube 
(spanning the observed segment of the video) 
also, contain past and future estimated bounding boxes. 
As these predicted boxes are implicitly linked to the micro-tubes 
which compose the presently detected tube, the problem of linking to the latter the future bounding boxes, leading to a whole action tube, 
is automatically addressed.


\begin{figure}[t]
  \centering
  \includegraphics[scale=0.3]{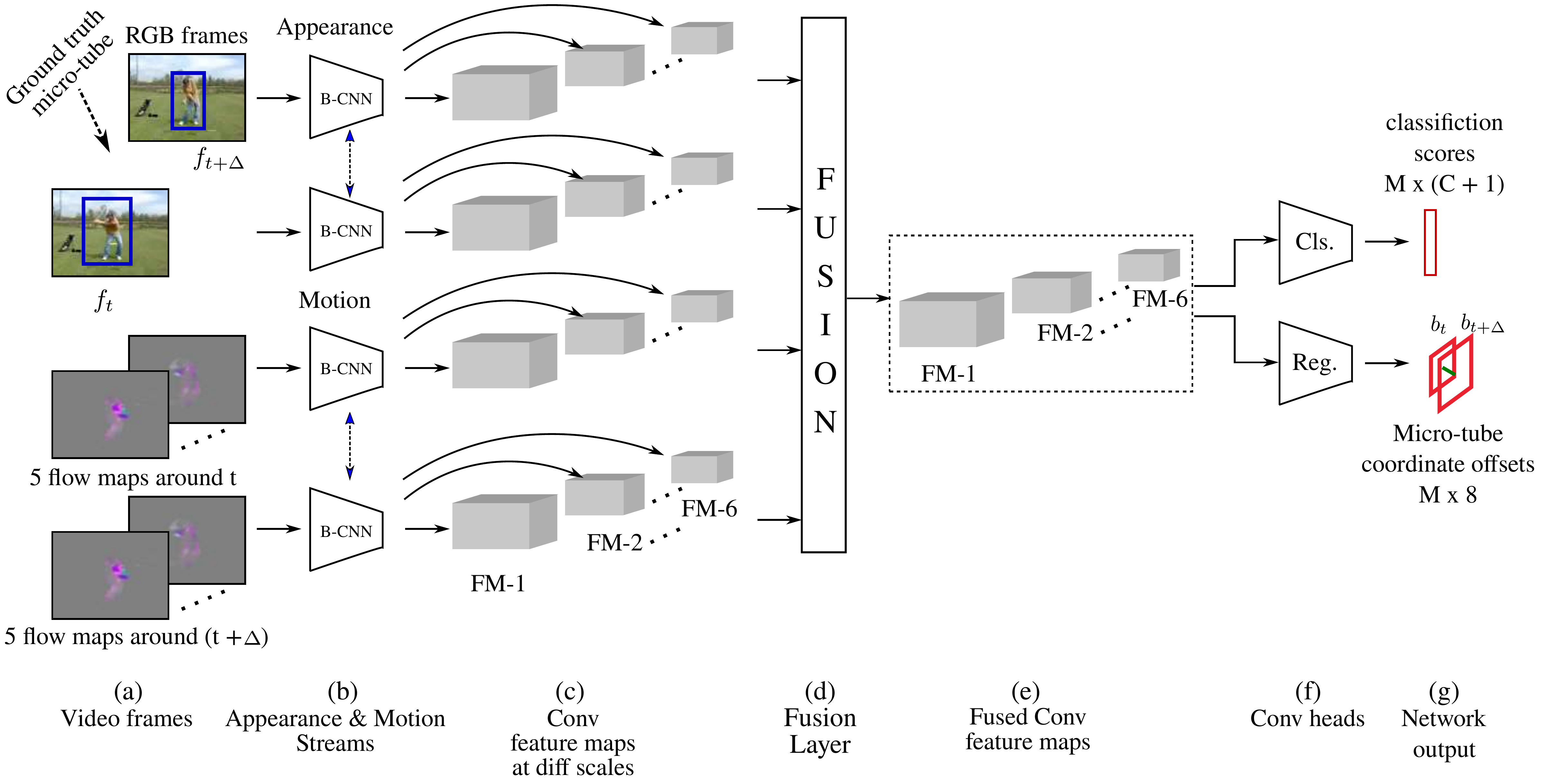}
  \caption[Overview of the action micro-tube detection network (AMTnet).]
  {Overview of the action micro-tube detection network (AMTnet). Here we introduce feature-level fusion (d).} 
\label{tpnet:fig:amnet-fused}
\end{figure}

\section{Base network}\label{tpnet:sec:base_net}
Our TPNet build on AMTNet in previous chapter. 
However, we make some improvement on the fusion side, earlier we performed a late fusion of appearance and motion cues. 
Here,  in this chapter, we introduce a feature-level fusion of appearance and motion cues for AMTNet network.
An overview of AMTNet with fusion is shown in Figure~\ref{tpnet:fig:amnet-fused}
We try two different types (summation and concatenation) of fusion methods. We show that both methods perform similarly, 
however, frame-level fusion provides a big performance improvement over late-fusion techniques used in previous chapters on J-HMDB-21 dataset.

\subsection{Converting AMTNet into TPNet}
AMTNet allows us to generate micro-tubes which has to connected boxes from a pair of frames at time $t$ and ${t+\Delta}$.
While observing the same pair frames, we can extend the idea of connected boxes further by predicting a box for frame time $t+\Delta_f$.
We can take it even further, we can predict connected boxes for $n$ future frames which we haven't observed, also even for a past frame $t-\Delta_p$.
Note that predictions also span the past: a setting called \emph{smoothing} in the estimation literature.
We only observe a pair of frames from time $t$ and ${t+\Delta}$, and $\Delta_p$ measure how far in the past we are predicting into,
whereas $\Delta_f$ is a future step size, and $n$ is the number of future steps.
Figure~\ref{tpnet:fig:training} show an overview of tube predictor network (TPNet), which incorporate above prediction ideas.
Based on this idea we will now explain how to train such a network in the next Section~\ref{tpnet:sec:training}, and for prediction in Section~\ref{tpnet:sec:predict}.


\section{Training TPNet}\label{tpnet:sec:training}

\begin{figure}[t]
  \centering
  \includegraphics[scale=0.60]{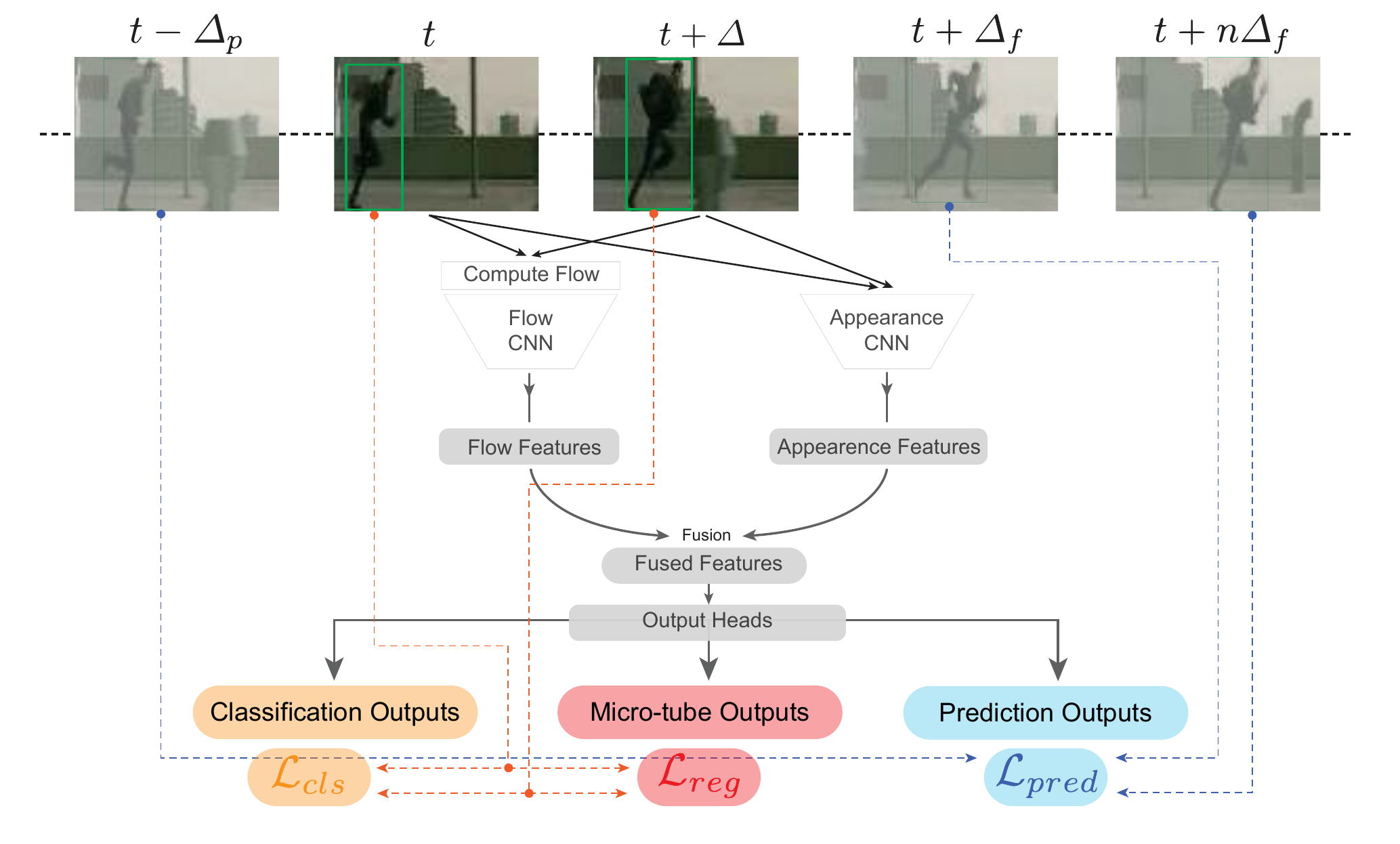}
  \caption[Overview of the tube predictor network (TPNet) at training time.]
  {Overview of the tube predictor network architecture at training time.}
\label{tpnet:fig:training}
\end{figure}

The underlying architecture of TPNet is shown in Figure~\ref{tpnet:fig:training}.
TPNet takes two successive frames from time $t$ and $t+\Delta$ as input.
The two input frames are fed to two parallel CNN streams, 
one for appearance and one for optical flow.
The resulting feature maps are fused, either by summation or concatenation of the given feature maps.
Finally, three types of convolutional output heads are used for 
$P$ prior boxes as shown in Figure~\ref{tpnet:fig:training}.
The first one produces the $P\times(C+1)$ classification outputs; the
the second one regresses the $P\times 8$ coordinates of the micro-tubes, as in AMTnet; the 
last one regresses $P\times(4(1+n))$ coordinates, 
where $4$ coordinates correspond to the frame at $t-\Delta_p$, and the remaining $4n$ are associated with the $n$ future steps. 
The training procedure for the new architecture is illustrated below.

\subsection{Multi-task learning}\label{tpnet:subsec:multitask}
TPNet is designed to strive for three objectives, for each prior box $p$. 
The first task (i) is to classify the $P$ prior boxes; 
the second task (ii) is to regress the coordinates of the micro-tubes;  
the last (iii) is to regress the coordinates of the past and future detections associated with each micro-tube.

Given a set of $P$ anchor boxes and the respective outputs 
we compute a loss following the training objective of SSD~\cite{liu2016ssd}. Let $x_{i,j}^c = \{0,1\}$ be the indicator  
for matching the $i$-th prior box to the $j$-th ground truth box of category $c$. 
We use the bipartite matching procedure described in~\cite{liu2016ssd} 
for matching the ground truth micro-tubes $G = \{g_t,g_{t+\Delta}\}$ 
to the prior boxes, where $g_t$ is a ground truth box at time $t$.
The overlap is computed between a prior box $p$ and micro-tube $G$ as the 
mean IoU between $p$ and the ground truth boxes in $G$.
A match is defined as positive ($x_{i,j}^c = 1$) if the overlap is more than or equal to $0.5$.  

The overall loss function $\mathcal{L}$ is 
the following weighted sum of classification loss ($\mathcal{L}_{cls}$), 
micro-tube regression loss ($\mathcal{L}_{reg}$) and prediction loss ($\mathcal{L}_{pred}$):
\begin{equation}
\mathcal{L}(x,c,m,G,z,Y) = \frac{1}{N} \big ( \mathcal{L}_{cls}(x,c) + \alpha\mathcal{L}_{reg}(x,m,G) + \beta\mathcal{L}_{pred}(x,z,Y) \big ),
\end{equation}
where $N$ is the number of matches, $c$ is the ground truth class, 
$m$ is the predicted micro-tube, $G$ is the ground truth micro-tube,
$z$ assembles the predictions for the future and the past, and $Y$ is the ground truth of future and past bounding boxes associated with the ground truth micro-tube $G$. 
The values of $\alpha$ and $\beta$ are both set to $1$ in all of our experiments:
different values might result in better performance.

The classification loss $\mathcal{L}_{cls}$ is a softmax cross-entropy loss; 
a hard negative mining strategy is also employed, as proposed in~\cite{liu2016ssd}.
The micro-tube loss $\mathcal{L}_{reg}$ is a smooth L1 loss~\cite{ren2015faster} 
between the predicted ($m$) 
and the ground truth ($G$) micro-tube. 
Similarly, the prediction loss $\mathcal{L}_{pred}$ is also a smooth L1 loss between the predicted boxes ($z$) and the ground truth boxes ($Y$). 
As in \cite{liu2016ssd,ren2015faster}, 
we regress the offsets with respect to the coordinates of 
matched prior box $p$ matched to $G$ for both $m$ and $z$.
We use the same offset encoding scheme as used in~\cite{liu2016ssd}. 

\section{Tube prediction framework}\label{tpnet:sec:predict}

\subsection{Problem statement}\label{sec:problem_statement}
Similar to previous chapters, we define an \emph{action tube} as a connected sequence of detection boxes 
in time without interruptions and associated with a same action class $c$, 
starting at first frame $f_1$ and ending last frame $f_T$, 
in trimmed video: $\mathcal{T}_c = \{ {b}_{1}, ... {b}_{t}, ... {b}_{T}\}$.
Tubes are constrained to span the entire video duration, like in ~\cite{georgia2015tube}.
At any time point $t$, a tube is divided into two parts, one needs 
to be detected $\mathcal{T}_{c}^{d} = \{ {b}_{1}, ... {b}_{t}\}$ up to $f_t$ and 
another part needs to be predicted/estimated
$\mathcal{T}_{c}^{p} = \{{b}_{t+1}, ... {b}_{T}\}$ from frame $f_{t+1}$ to $f_{T}$ 
along with its class $c$.
The observed part of the video is responsible for generating 
$\mathcal{T}_{c}^{d}$ (red in Fig~\ref{tpnet:fig:problem-statement}), 
while we need to estimate the future section of the tube $\mathcal{T}_{c}^{p}$ 
(blue in Fig~\ref{tpnet:fig:problem-statement}) for the unobserved segment of the video.
The first sub-problem, the online detection of $\mathcal{T}_{c}^{d}$, 
is explained in Section~\ref{tramnet:sec:tube-gen-algo} of Chapter~\ref{chapter:tramnet}. 
The second sub-problem (the estimation of the future tube segment $\mathcal{T}_{c}^{p}$) 
is tackled by a tube predictor network (TPNet, \S~\ref{tpnet:sec:training}) 
in a novel tube prediction framework describe in next Section~\ref{tpnet:subsec:tubepred}.

\subsection{Tube prediction using TPNet}\label{tpnet:subsec:tubepred}
TPNet is shown in Figure~\ref{tpnet:fig:testing} at test time. 
Similar to the training setting, TPNet observes only two frames that are $\Delta$ apart at any time point $t$. 
The outputs of TPNet at any time $t$ are linked to a micro-tube,
each micro-tube containing a set of bounding boxes 
$\{m_t = \{b_t, b_{t+\Delta}\}; z_t = \{b_{t-\Delta_p},b_{t+\Delta_f},...,b_{t+\Delta_f}\}\}$, 
which are considered as linked together.

\begin{figure}[t]
  \centering
  \includegraphics[scale=0.55]{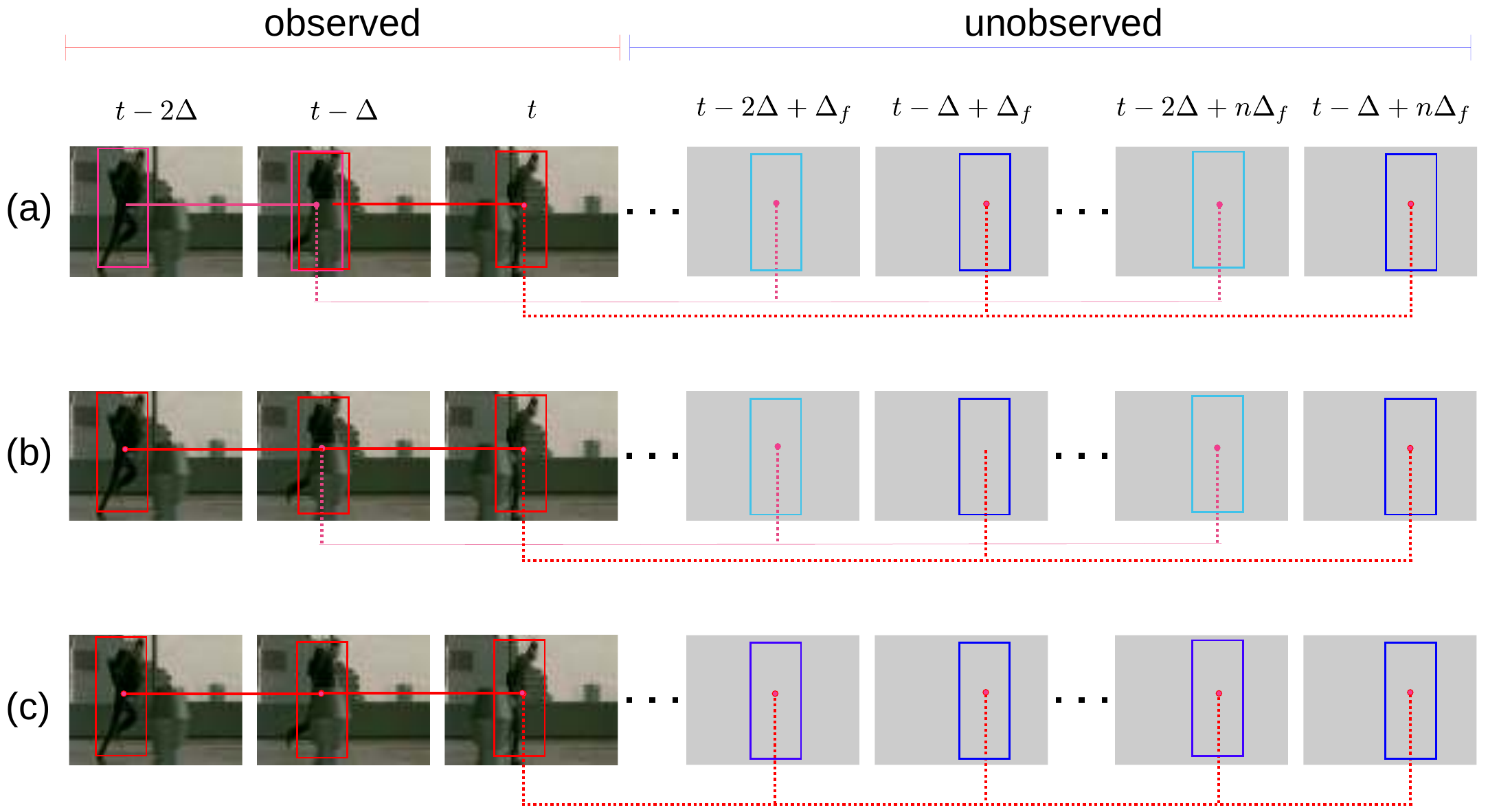}
      \caption[Overview of future tube prediction framework.]
      {Overview of future tube ($\mathcal{T}_{c}^{p}$)
      prediction using the predictions that are linked to micro-tubes.
      The first row (a) shows two output micro-tubes in light red and red and 
      their corresponding predictions in future in light blue and blue.
      In row (b) two micro-tubes are linked together, after which 
      they are shown in the same colour (red).
      By induction on the previous step, in row (c) we show that the predictions associated with two micro-tubes are linked together as well, 
      hence forming one single tube. 
      The observed segment is shown in red, while the predicted segment for the part of the video yet to observe
      is shown in blue in case of prediction.
      There may be an overlap of predictions for the same tube, precedence to use the box in final tube is defined by: detection box $>>$ most recent prediction $>>$ older prediction box. It is explained in more details at the end of Section~\ref{tpnet:subsec:tubepred}.
      }
\label{tpnet:fig:linking}
\end{figure}

As explained in Section~\ref{tramnet:sec:tube-gen-algo} of Chapter~\ref{chapter:tramnet}, given a set of 
micro-tubes $\{{m}_{1} ... {m}_{t-\Delta}\}$, we can construct $\mathcal{T}_{c}^{d}$ by online linking~\cite{singh2017online} of the micro-tubes. 
As a result, we can use predictions for $t+\Delta_f$ up to $t+n\Delta_f$ to generate the future of $\mathcal{T}_{c}^{d}$, 
thus extending it further into the future as shown in Figure~\ref{tpnet:fig:linking}. 
More specifically, as it is indicated in Figure~\ref{tpnet:fig:linking}(a), 
a micro tube at $t-2\Delta$ is composed by $n+2$ bounding 
boxes ($\{b_{t-2\Delta},b_{t-\Delta}, b_{t-2\Delta+\Delta_f}, ... b_{t-\Delta+n\Delta_f}\}$) linked together.
The last micro-tube is generated from $t-\Delta$.
In the same fashion, putting together the predictions associated with all the past micro-tubes ($\{{m}_{1} ... {m}_{t-\Delta}\}$) 
yields a set of linked future bounding boxes 
($\{b_{t+1}, ... , b_{t+\Delta+\Delta_f}, ... , b_{t-\Delta+n\Delta_f}\}$) 
for the current action tube $\mathcal{T}_{c}^{d}$, thus outputting a part of the desired future $\mathcal{T}_{c}^{p}$.

Now, we can generate future tube $\mathcal{T}_{c}^{p}$ from 
the set of linked future bounding boxes 
($\{b_{t+1}, ... b_{t-\Delta+\Delta_f},... b_{t-\Delta+n\Delta_f}\}$)
from $t+1$ to $t-\Delta+n\Delta_f$ and 
simple linear extrapolation of 
bounding boxes from $t-\Delta + n\Delta_f$ to $T$.
Linear extrapolation is performed based on the average velocity 
of the each coordinates from last $5$ frames, 
predictions outside the image coordinate are trimmed to the image edges.

There may be an overlap of predictions for the same tube, for instance, if both $\Delta_f$ and $\Delta$ equal to 1 then at time $t$ there a future predicted box from $t-1$ and a detected box from $t$, in that case, detection takes precedence over future prediction, i.e detected box is used in final tube.
In any of such case, precedence to use the box in final tube is defined by: detection box $>>$ most recent prediction $>>$ older prediction box.

%% file: chapters/partII/tpnet_exp.tex
\section{Experiments} \label{tpnet:sec:exp}
We test our action tube prediction framework (\S~\ref{tpnet:sec:predict}) on four challenging problems: 
\\
i) action detection (\S~\ref{sec:detection}),
\\
ii) early action prediction (\S~\ref{sec:prediction-online-detection}),
\\
iii) online action detection (\S~\ref{sec:prediction-online-detection}),
\\
iv) future action tube prediction (\S~\ref{sec:future-detection})
\\
Finally, evidence of real time capability is quantitatively demonstrated in (\S~\ref{sec:detection-speed}).

\textbf{Evaluation protocol.}
Now, we define the evaluation metrics used in the current chapter for the tasks mentioned above.
i) We use a standard mean-average precision metric to evaluate the detection performance when the whole video is observed.
ii) Early label prediction task is evaluated by 
video classification accuracy~\cite{soomro2016predicting,singh2017online} 
as early as when only 10\% of the video frames are observed. 
\\
iii) Online action detection (\S~\ref{sec:prediction-online-detection}) is set up based 
on the experimental setup of~\cite{singh2017online}, 
and use mAP (mean average precision) as metric for online action detection 
i.e. it evaluates present tube ($\mathcal{T}_{c}^{d}$) built-in online fashion.
\\
iv) The future tube prediction is a new task; we propose to evaluate its performance in two ways, one {completion-mAP} (c-mAP) and {prediction-mAP} (p-mAP). These metrics are define in Section~\ref{soa:metrics}.

Similar to Chapter~\ref{chapter:online}, we report the performance of previous three tasks (i.e. task ii to iv) as a function of \emph{Video Observation Percentage}, i.e., the portion (\%) of the entire video observed. 

\textbf{J-HMDB-21.}
We evaluate our model on the J-HMDB-21~\cite{jhmdbj2013towards} dataset.
Each video contains an instance of an atomic action for 20-40 frames.
Although, videos are of short duration (max $40$ frames), 
we consider this dataset because tubes belong to the same class and 
we think it is a good dataset to start with for action prediction task.
More details about the dataset can be found in Section~\ref{soa:datasets}.

\noindent
\textbf{Baseline.}
We modified AMTnet to fuse flow and appearance features~\ref{tpnet:sec:base_net}. 
We treat it as a baseline for all of our tasks.
Firstly, we show how feature fusion helps AMTnet in Section~\ref{sec:detection}, 
and compare it with other action detection methods along with our TPNet.
Secondly in Section~\ref{sec:future-detection}, 
we linearly extrapolate the detection from AMTnet to construct the future tubes and use them as a baseline for the tube prediction task. 

\noindent
\textbf{TPNet${}_{abc}$.} 
The training parameters of our TPNet are used to define the name of the setting in which we use our tube prediction network. 
The network name TPNet${}_{abc}$ represents our 
TPNet where $a = \Delta_p$, $b = \Delta_f$ and $c = n$, 
if $\Delta_p$ is set to $0$ it means network doesn't learn 
to predict the past bounding boxes.
Future prediction window is defined by $b \times c$, i.e. $\Delta_f \times n$.
In all of our settings, we use $\Delta=1$ in order to keep our experiments consistent with previous work(AMTNet ~\cite{saha2017amtnet}.

\noindent
\textbf{Implementation details.}
We train all of our networks with the same set of hyper-parameters 
to ensure fair comparison and consistency, including TPNet and AMTnet.
We use an initial learning rate of $0.0005$, 
and the learning rate drops by a factor of $10$ after $5K$ and $7K$  iterations.
All the networks are trained up to $10K$ iterations. 
We implemented AMTnet using PyTorch (\url{https://pytorch.org/}).
We initialise AMTnet and TPNet models using the pretrained SSD network on J-HMDB-21 dataset on its respective train splits. 
The SSD network training is initialised using image-net trained VGG network.
For, optical flow images, we used optical flow algorithm of Brox \etal~\cite{brox2004high}.
Optical flow output is put into a three-channel image, two channels are made of flow vector and the third channel is the magnitude of the flow vector. 

\begin{table}[t]
  \centering
  \setlength{\tabcolsep}{4pt}
  \caption[Action detection results on J-HMDB dataset.]{Action detection results on J-HMDB dataset.
  The table is divided into four parts. 
  The first part lists approaches which takes a single frame as input;
  the second part presents approaches which takes multiple frames as input;
  the third part contemplates different fusion strategies of our feature-level 
  fusion (based on AMTnet);
  lastly, we report the detection performance of our TPNet by ignoring the future and past predictions and only use the detected micro-tubes to produce the final action tubes.}
  {\footnotesize
  \scalebox{1}{
  \begin{tabular}{lccccc}
  \toprule
  Methods & $\delta$ = 0.2 &$\delta$ = 0.5 & $\delta$ = 0.75 & $\delta$ = .5:.95 & Acc \%\\\midrule
  MR-TS Peng~\etal~\cite{peng2016eccv} & 74.1  & 73.1 & -- & -- & --\\
  FasterRCNN Saha~\etal~\cite{saha2016deep}   & 72.2  & 71.5 & 43.5 & 40.0 & --\\
  OJLA Behl~\etal~\cite{behl2017incremental}${}^*$  & -- & 67.3 & -- & 36.1 & --\\
  SSD Singh~\etal~\cite{singh2017online}${}^*$  & 73.8 & 72.0 & 44.5 & 41.6 & --\\
  \midrule
  AMTnet Saha~\etal~\cite{saha2017amtnet} rgb-only  & 57.7  & 55.3 & -- & -- & --\\
  ACT Kalogeiton~\etal~\cite{kalogeiton2017action}${}^*$   & 74.2 & 73.7 & 52.1 & 44.8 & 61.7 \\
  T-CNN (offline) Hou~\etal~\cite{hou2017tube} & \textbf{78.4} & 76.9   & -- & -- & 67.2 \\
  MR-TS~\cite{peng2016eccv} + I3D~\cite{carreira2017quo} Gu \etal~\cite{ava2017gu}  
  & --  & \textbf{78.6} & -- & -- & --\\
  \midrule
  AMTnet-Late-Fusion${}^*$  & 71.7 & 71.2 & 49.7 & 42.5 & 65.8 \\
  AMTnet-Feat-Fusion-Concat${}^*$  & 73.1 & 72.6 & 59.8 & 48.3 & 68.4 \\
  AMTnet-Feat-Fusion-Sum${}^*$  & 73.5 & 72.8 & 59.7 & 48.1 & 69.6 \\
  \midrule
  Ours TPNet${}_{053}$${}^*$ & 72.6 & 72.1 & 58.0 & 46.7 & 67.5 \\
  Ours TPNet${}_{453}$${}^*$ & 73.8 & 73.0 & 59.1 & 47.3 & 68.2 \\
  Ours TPNet${}_{051}$${}^*$ & 74.6 & 73.1 & 60.5 & 49.0 & \textbf{69.8} \\
  Ours TPNet${}_{451}$${}^*$ & 74.8 & 74.1 & \textbf{61.3} & \textbf{49.1} & 68.9 \\ \bottomrule
  \multicolumn{6}{l}{ TPNet${}_{abc}$ represents configurations of TPNet where $a = \Delta_p$, $b = \Delta_f$ and $c = n$;}\\ 
  \multicolumn{6}{l}{$\Delta_p$ is past step, $\Delta_f$ is future step, and $n$ is number of future steps;}\\
  \multicolumn{6}{l}{Future prediction window is defined by $b \times c$, i.e. $\Delta_f \times n$;}\\
  \multicolumn{6}{l}{${}^*$ means online methods}\\
  \end{tabular}
  }
  }
\label{tpnet:table:detection} 
\end{table}

\subsection{Action detection performance}\label{sec:detection}
Table~\ref{tpnet:table:detection} shows the traditional action detection results for 
the whole action tube detection in the videos of J-HMBD-21 dataset.

\textbf{Feature fusion} compared to the late fusion scheme in AMTnet shows (Table~\ref{tpnet:table:detection}) remarkable improvement, at detection threshold $\delta=0.75$ the gain with feature level fusion is $10\%$, as a result, it is able to surpass the performance of ACT~\cite{kalogeiton2017action}, which relies on a set of $6$ frames as compared to AMTnet which uses only $2$ successive frames as input.
Looking at the average-mAP ($\delta=0.5:95$), we can see that the fused model improves by almost $8\%$  as compared to single frame SSD model of Singh~\etal~\cite{singh2017online}.
We can see that concatenation and sum fusion perform almost similar for AMTnet.
Sum fusion is little less memory intensive on the GPUs 
as compared to the concatenation fusion; as a result, 
we use sum fusion in our TPNet.

\textbf{TPNet for detection} is shown in the last part of the Table~\ref{tpnet:table:detection}, 
where we only use the detected micro-tubes by 
TPNet to construct the action tubes.
We train TPNet to predict future and past 
(i.e. when $\Delta_p>0$) as well as present micro-tubes.
We think that predicting bounding boxes for both the past 
and future video segments act as a regulariser and helps 
improving the representation of the whole network.
Thus, improving the detection performance 
(Table~\ref{tpnet:table:detection}~TPNet${}_{051}$ and TPNet${}_{451}$).
However, that does not mean adding extra prediction task always help when a network is asked to learn prediction in far future, 
as is the case in TPNet${}_{053}$ and TPNet${}_{453}$,
we have a drop in the detection performance. 
We think there might be two possible reasons for this,  
i) network might starts to focus more on the prediction task, and
ii) videos in J-HMDB-21 are short and the number of training samples 
decreases drastically ($19K$ for TPNet${}_{051}$ and $10K$ for TPNet${}_{453}$), 
because we can not use edge frames of the video in training samples as 
we need a ground truth bounding box which is $15$ frames in the future, 
as $\Delta_f=5$ and $n=3$ for TPNet${}_{053}$.
However, in Section~\ref{sec:future-detection}, 
we show that the TPNet${}_{053}$ model is the best to predict the future very early. 

\begin{figure}[t]
  \centering
  \includegraphics[scale=0.38]{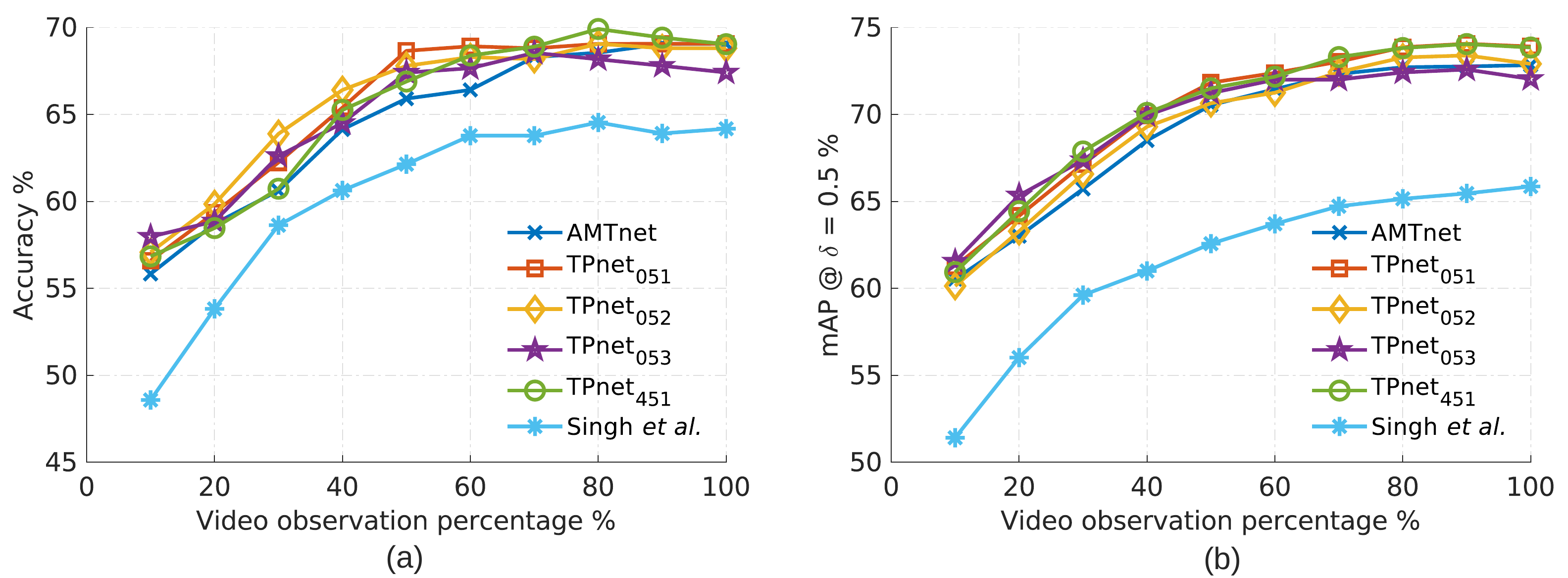}
  \caption[Early label prediction results on J-HMDB-21.]
    {Early label prediction results (video-level label prediction accuracy) 
      on J-HMDB-21 dataset in sub-figure (a). 
      Online action detection results (mAP with detection threshold $\delta = 0.5$) 
      on J-HMDB-21 dataset are shown in sub-figure (b).
      TPNet${}_{abc}$ represents our TPNet where $a = \Delta_p$, $b = \Delta_f$ and $c = n$.}
\label{tpnet:fig:onlineNlabel}
\end{figure}

\subsection{Early label prediction and online detection}\label{sec:prediction-online-detection}
Figure~\ref{tpnet:fig:onlineNlabel} (a) \& (b) show the early prediction 
and online detection capabilities of Singh~\etal~\cite{singh2017online}, 
AMTnet-Feature Fusion-sum and our TPNet.

Soomro~\etal~\cite{soomro2016predicting}'s method also perform early label prediction on J-HMDB-21; however, their performance is deficient, as a result, the plot would become skewed (Figure~\ref{tpnet:fig:onlineNlabel}(a)), so we omit theirs from the figure. 
For instance, by observing only the initial $10\%$ of the videos in J-HMDB-21,
TPNet${}_{453}$ able to achieve a prediction accuracy of $58\%$ 
as compared to $48\%$ by Singh~\etal~\cite{singh2017online} and $5\%$ by Soomro~\etal~\cite{soomro2016predicting}, 
which is in fact higher than the $43\%$ accuracy achieved by \cite{soomro2016predicting} 
after observing the \emph{entire} video. 
As more and more video observed, all the methods improve, but TPNet${}_{451}$ show the most gain, 
however, TPNet${}_{053}$ observed the least gain from all the TPNet settings shown.
Which is in-line with action detection performance discussed in the previous section~\ref{sec:detection}.
We can observe the similar trends in online action detection performance shown in Figure~\ref{tpnet:fig:onlineNlabel}(b). 
To reiterate, TPNet${}_{053}$ doesn't get to see the training samples from the end portion of the videos, 
as it needs a ground truth bounding box from $15$ frames ahead. 
So, the last frame it sees of any training video is $T-15$, 
which is almost half the length of the most extended video($40$ frames) in J-HMDB-21. 
\begin{figure*}[t]
  \centering
  \includegraphics[scale=0.38]{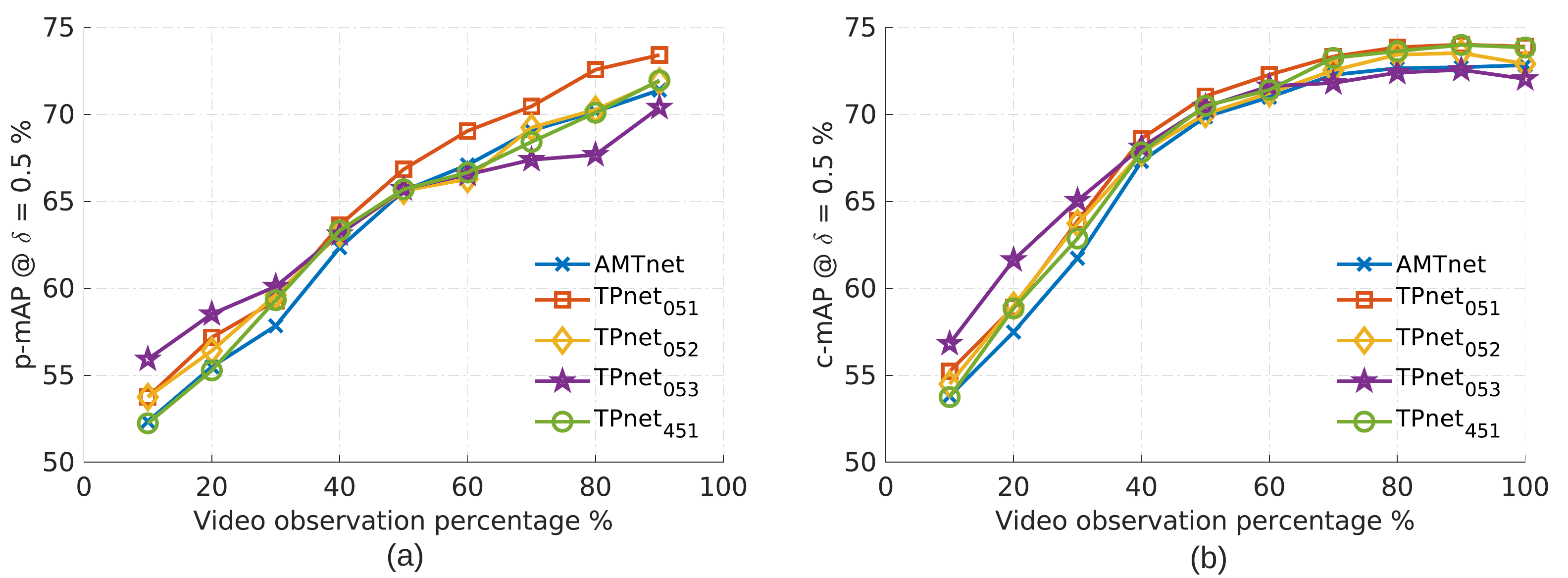}
  \caption[Future action tube prediction results.]
  {Future action tube prediction results (a) (prediction-mAP (p-mAP)) 
      for predicting the tube in unobserved part of the video.
      Action tube prediction results (b) (completion-mAP (c-mAP)) 
      for predicting video long tubes as early as possible 
      on J-HMDB-21 dataset in sub-figure (b).
      We use p-mAP (a) and c-mAP (b) with detection 
      threshold $\delta = 0.5$ as evaluation metrics on J-HMDB-21 dataset.
      TPNet${}_{abc}$ represents our TPNet where $a = \Delta_p$, $b = \Delta_f$ and $c = n$.}
\label{tpnet:fig:prediction_mAP_05}
\end{figure*}

\subsection{Future action tube prediction}\label{sec:future-detection}
Our main task of this work is to predict the future of action tubes.
We evaluate it using two newly proposed metrics (\emph{p-mAP} and \emph{c-mAP}) 
as explained earlier at the start of the experiment section~\ref{tpnet:sec:exp}.
Result are shown in Figure~\ref{tpnet:fig:prediction_mAP_05} for future tube prediction (Figure~\ref{tpnet:fig:prediction_mAP_05} (a)) with p-mAP metric and 
tube completion with c-mAP as metric.

Although, the TPNet${}_{053}$ is the worst setting of TPNet model for early label prediction (Fig.~\ref{tpnet:fig:onlineNlabel}(a)), online detection(Fig.~\ref{tpnet:fig:onlineNlabel}(b)) and action tube detection (Table~\ref{tpnet:table:detection}), but as it predicts furthest in the future (i.e. $15$ frame away from the present time), 
it is the best model for early future tube prediction (Fig.~\ref{tpnet:fig:prediction_mAP_05}(a)). 
However, it does not observe as much appreciation in performance as other settings as more and more frames are seen, owing to the reduction in the number of training samples. 
On the other hand, TPNet${}_{451}$ observed large improvement as 
compared to TPNet${}_{051}$ as more and more portion of the video is observed 
for tube completion task (Fig.\ref{tpnet:fig:prediction_mAP_05}(b)), 
which strengthen our argument that predicting not only the future but also, the past is useful to achieve more regularised predictions.

\noindent
\textbf{Comparison with the baseline.}
As explained above, we use AMTnet as a baseline, and its results can be seen in all the plots and the Table. 
We can observe that our TPNet performs better than AMTnet in almost all the cases, 
especially in our desired task of early future prediction (Fig~\ref{tpnet:fig:prediction_mAP_05}(a)) 
TPNet${}_{043}$ shows almost $4\%$ improvement in p-mAP (at $10\%$ video observation) over AMTnet.

\noindent
\textbf{Discussion.} 
Predicting further into the future is essential to produce any meaningful predictions (seen in TPNet${}_{053}$),  but at the same time, predicting past is helpful to improve overall tube completion performance. 
One of the reasons for such behaviour could be that J-HMDB-21 tubes are short (max $40$ frames long).
We think training samples for a combination of TPNet${}_{053}$ and TPNet${}_{451}$, i.e. TPNet${}_{453}$ are chosen uniformly over the whole video while taking care of the absence of ground truth in the loss function could give us better of both settings. 
The idea of regularising based on past prediction is similar to the one used by Ma~\etal~\cite{ma2016learning}, where they show that regularisation based on past can help in the prediction of the future.

\subsection{Test Time Detection Speed}
\label{sec:detection-speed}
Continue from previous chapters, here, we show real-time capabilities.
Here we use their online tube generation method from Chapter~\ref{chapter:online} for our tube prediction framework to inherit online capabilities.
The only question mark is TPNet's forward pass speed.
We thus measured the average time taken for 
a forward pass for a batch size of 1 as compared to 8 by ~\cite{singh2017online}.
A single forward pass takes 46.8 milliseconds to process one example, 
showing that it can be run in almost real-time at 21fps with two streams on a single 1080Ti GPU.
One can improve speed even further by testing TPNet with $\Delta$ equal to $2$ or $4$ and obtain a speed improvement of  $2\times$ or $2\times$.
However, use of dense optical flow~\cite{brox2004high}, which is slow, but as in ~\cite{singh2017online}, 
we can always switch to real-time optical~\cite{kroeger2016fast} with small drop in performance.

%% file: pads/partIII.tex
\cleardoublepage
\phantomsection
\renewcommand{\pname}{Part III : Online/Causal 3D Representations}
\addcontentsline{toc}{chapter}{\pname}\label{partIII}

\pagebreak
\hspace{14pt}
\vfill
\begin{center}
\textbf{\pname}
\end{center}
\vfill
\hspace{0pt}
\pagebreak

%% file: chapters/7_rcn.tex
\chapter{Recurrent Convolution for Causal 3D CNNs}
\label{chapter:rcn}
\renewcommand{\imagepath}{figures/rcn} 
\input{chapters/partIII/rcn_intro}
\input{chapters/partIII/rcn_approach}

\input{chapters/partIII/rcn_exp}

\section{Summary and limitations}\label{rcn:sec:summary}
\paragraph{Summary: }
In this work, we presented a recurrence-based convolutional network (RCN) able to generate 
causal spatiotemporal representations by converting 3D CNNs in to causal 3D CNNs
while using 2.6 times fewer parameters compared to its traditional 3D counterparts.
RCN can model long-term temporal dependencies without the need to specify temporal extents.
The proposed RCN is not only causal in nature and temporal resolution-preserving  
but was also shown to outperform the main baseline 3D networks in all the fair comparisons we ran. 
We showed that ImageNet-based initialisation is at the heart of the success of 3D CNNs. 
Indeed, although RCN is recurrent in nature, it can still utilise the weights of a pre-trained 2D network for initialisation. 
In summary, we  present a new approach to video feature representation based on an original convolutional network with recurrent hidden states at each depth level, which:
\begin{itemize}
    \item allows flexible temporal reasoning, as it exploits by design information coming from all the input sequence observed up to time $t$;
    \item generates output representations in a causal way, allowing online video processing and enabling the use of 3D networks in scenarios in which causality is key;
    \item is designed to directly benefit from model initialisation via ImageNet pre-trained weights, as opposed to state of the art approaches, 
    and in line with clear emerging trends in the field.
\end{itemize}

\paragraph{Limitations: } our proposed network has all the nice properties like causality, flexible temporal reasoning temporal resolution preservation. 
However, there some limitations as follows: 
i) it is only evaluated on small input clip lengths of 8-16 frames,
ii) it require more GPU memory to train than a 3D network because of temporal resolution preservation
iii) it relies on good initialisation for its performance. 
Despite these problems, our RCN is the best causal video representation model out there.

\paragraph{Looking ahead: }
The causal nature of our recurrent 3D convolutional network opens up manifold research directions, 
with direct and promising potential application in areas such as online action detection and future event/action prediction. 
In the next chapter, we will discuss the potential application of causal 3D CNNs along with the inspiration from previous chapters.

%% file: chapters/partIII/rcn_intro.tex
\section{Introduction} 
\label{rcn:intro}
Convolutional neural networks (CNN) are starting to exhibit gains in action recognition from videos
similar to those previously observed in image recognition \cite{krizhevsky2012imagenet,simonyan2015very}
thanks to new 3D CNNs~\cite{carreira2017quo,xie2018rethinking,tran2018closer,hara2018can,nonlocal2018wang}.  
For instance, Hare~\etal~\cite{hara2018can} have shown that that is the case for the 3D version of 2D residual networks (ResNets) \cite{he2016deep}. 
However, they have multiple problem concerning their deployment 
in online video understanding problems, 
e.g., online action detection~\cite{singh2017online,soomro2016predicting},
future action label/tube prediction~\cite{kong2017deep,singh2018predicting}, and
future representation prediction~\cite{vondrick2015anticipating}.

Firstly, temporal convolutions are inherently anti-causal, which is necessary for online video understanding.
Secondly, the temporal convolution size needs to be picked by hand at every level of network depth.
As a result, the temporal reasoning horizon or `receptive field' is effectively constrained by the size of the temporal convolution kernel(s). 
Lastly, 3D CNNs do not preserve temporal resolution, as the latter drops with network depth. 
Preserving temporal resolution, in opposition, is essential in problems such where we need predictions to be made 
on each frame of input clip while reasoning about temporal context, 
e.g. bounding box regression on each frame for action tube detection~\cite{georgia2015tube,singh2017online} 
or temporal label prediction on each frame for temporal action segmentation~\cite{sigurdsson2016hollywood,rene2017temporal}
or online video segmentation~\cite{xu2012streaming}.


\emph{Our method: combining implicit and explicit temporal modelling}. 
Inspired by the success of 3D CNNs, we propose to make 3DCNNs causal by explicit temporal modelling.
Hidden state models, such as Markov ones~\cite{baum1966statistical}, recurrent neural networks (RNN)~\cite{jordan1986attractor,elman1990finding}, and
long short-term memory (LSTM) networks~\cite{hochreiter1997long} can all be
used to model temporal dynamics in videos ~\cite{donahue2015lrcn,poppe2010survey}, allowing flexible temporal reasoning.

\begin{figure}[t]
    \centering
    \includegraphics[scale=1.0]{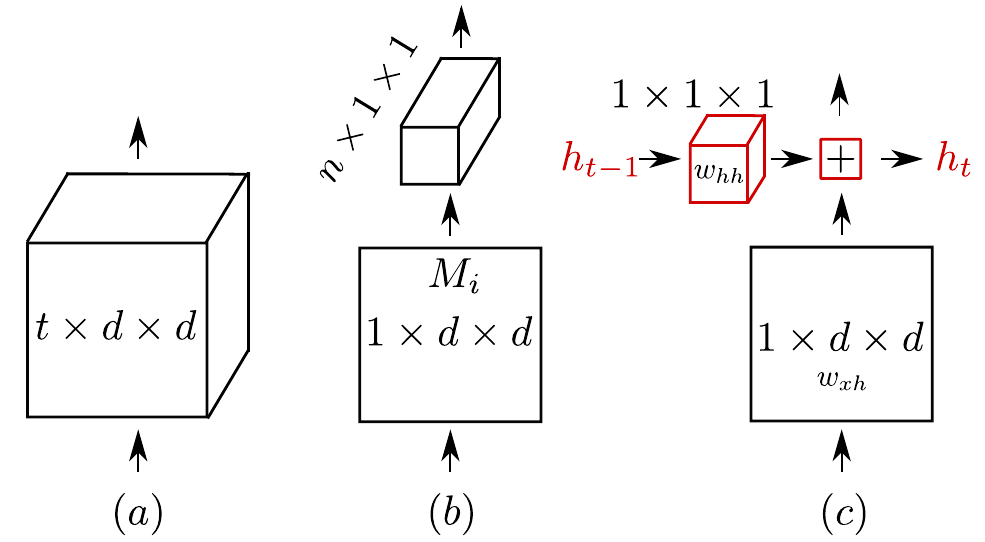}
    \caption[Illustration of basic build units of 3D CNNs.]
        {Illustration of basic 3D units used to build 3D networks to process sequences of input frames. 
        (a) Standard 3D convolution, as in I3D~\cite{carreira2017quo} or C3D~\cite{tran2014learning}.
        (b) 3D convolution decomposed into a 2D spatial convolution followed by a 1D temporal one, as in S3D~\cite{xie2018rethinking}. In R(2+1)D~\cite{tran2018closer} the number $M_i$ of middle 
        planes is increased to match the number of parameters in standard 3D convolution.
        (c) Our proposed decomposition of 3D convolution into 2D spatial convolution and 
        recurrence (in red) in the temporal direction, with a $1\times 1\times 1$ convolution as hidden state transformation.}
  \label{rcn:fig:convs} 
\end{figure}

In an approach which aims to combine the representation power of explicit dynamical models with the discriminative power of 3D networks, 
in this work we propose a recurrent alternative to 3D convolution illustrated in Figure~\ref{rcn:fig:convs}(c). 
In this new architecture, spatial reasoning, expressed in the form of a mapping from the input to a hidden state, 
is performed by spatial convolution 
(with kernel size $1 \times d \times d$),  
whereas temporal reasoning (represented by hidden state-to-hidden state transformations) is performed by a convolution (with kernel size $1\times 1 \times 1$) taking place
at every point in time and at each level (depth-wise) of the network.
In a setting which combines the effect of both operators, 
the hidden state at time $t$ (denoted by $h_t$) is a function of both the output of the spatial convolution and of the output of the temporal (hidden) convolution with $h_{t-1}$ as an input.   

As a result, the temporal reasoning horizon is effectively unconstrained, as the hidden state $h_t$ is a function of the input in the interval $[0,t]$.
Also, we are able to convert anti causal 3DCNNs into causal one while preserving the temporal resolution. 

To complete the transformation of 3DCNNs into causal 3DCNNs, we need to take care of some limitations of RNNs, such as vanishing/exploding gradients problem and activation functions (which could slow down the training).
Interestingly, Le \etal~\cite{le2015simple} show that simple RNNs can exhibit long-term memory properties if appropriately initialised, even better than LSTMs.
They use \emph{ReLU}~\cite{glorot2011deep} activation functions because of their fast convergence and sparsity properties~\cite{glorot2011deep}, 
as opposed \textit{sigmoid} or \textit{tanh} activation function used with traditional RNNs. The use of ReLU is in line with standard practice in CNNs. 
We thus follow~\cite{le2015simple} and initialise our hidden-to-hidden 
convolution (kernel size $N \times N \times 1 \times 1 \times 1$) by the identity matrix,
where $N$ is number of hidden state kernels. Spatial convolution, instead, can simply be initialised using ImageNet pre-trained weights.
We will show that such initialisation is important to achieve competitive performance to that of 3DCNNs.

\paragraph{Related publications.} 
The work presented in this chapter~\cite{singh2019recurrence} has been accepted for publication at ICCV workshop on large scale video understanding, 2019.
The dissertation author is the primary investigator in ~\cite{singh2019recurrence}.

%% file: chapters/partIII/rcn_approach.tex
\paragraph{Outline:}
The rest of the chapter is organised as follow: we start with the overview of our approach in Section~\ref{rcn:overview}.
We explain how 3D CNNs are created from 2D CNNs in Section~\ref{rcn:sec:3dcnns} for baseline purpose.
Next, we introduce our proposed casual 3D CNN architecture in Section~\ref{rcn:sec:rcn}, including its properties.
In Section~\ref{rcn:sec:exp}, we validate our approach with various studies on two large scale dataset.
Finally, we present a short summary and in Section~\ref{rcn:sec:summary}.

\section{Overview of the approach}\label{rcn:overview}
Our approach builds upon existing 3D CNNs, which are described in Section~\ref{rcn:sec:3dcnns}.
We replace every 3D convolution (Fig.~\ref{rcn:fig:convs}(a)) in 3D CNNs (\S~\ref{rcn:sec:3dcnns}) with our Recurrent Convolutional Unit (RCU) (Fig.~\ref{rcn:fig:convs}(c) - \S~\ref{rcn:subsec:rcu}). 
As, a result, a 3D CNN is converted into a casual 3D CNNs, named RCN (\S~\ref{rcn:subsec:rcn}).
Next, it is important to start from good initialisation, to this, we proposed to initialise our RCN with two separate initialisation process. 
First, each 2D convolution is initialised with pre-trained 2D network weights (\S~\ref{rcn:subsec:imagenetInit}).
Next, hidden state convolution matrix weight are initialised with identity matrix, as explained in Section~\ref{rcn:subsec:idInit}.


In our experiments, we show that our proposed RCN outperforms baseline I3D and (2+1)D models, 
while displaying all the above desirable properties.

\section{2D to 3D CNNs}~\label{rcn:sec:3dcnns}
There are two main reason why 3D CNNs~\cite{carreira2017quo,tran2018closer,xie2018rethinking,nonlocal2018wang} 
evolved from 2D CNNs~\cite{simonyan2015very,wang2016temporal} 
perform better than 3D CNNs built from scratch~\cite{tran2014learning,ji20133D}.
Firstly, 2D CNNs are well tried and tested on the problem of image recognition and a video is, after all, a sequence of images -- hence, transferability makes sense. 
Secondly, initialisation from a good starting guess leads to better convergence in videos~\cite{carreira2017quo}, since the number of parameters in 3D networks is huge. 

In this section, we recall the two basic types of 3D CNNs that can be built using a 2D CNN architecture.
We will use them as baselines in our experiments.

\subsection{Inflated 3D network (I3D)}~\label{rcn:subsec:i3d}

A 2D network can be converted/inflated into a 3D one by replacing a 2D ($d \times d$) convolution
with a 3D ($n \times d \times d$) convolution as shown in Figure~\ref{rcn:fig:convs}(a). 
Usually, the kernel's temporal dimension $n$ is set to be equal to the spatial dimension $d$, 
as in the inflated 3D network (I3D)~\cite{carreira2017quo} or the convolutional 3D network (C3D)~\cite{tran2014learning}.
\\
Here, we inflate the 18 layer ResNet~\cite{he2016deep} network into an I3D one as shown in Table~\ref{rcn:table:netArch}, 
where each 2D convolution is inflated into a 3D convolution. 
Similarly to the I3D network in~\cite{carreira2017quo}, a convolutional layer is used for classification, instead of the fully connected layer used in~\cite{tran2018closer,nonlocal2018wang}. 
A convolutional classification layer allows us to evaluate the model on sequences of variable length at test time. 
In this way, video-level results can be obtained in a seamless fashion as compared to computing clip-level outputs in sliding window fashion to obtain the video-level output. 

\begin{table}[t]
  \centering
  \caption[Base network architecture of ResNet-18 based I3D model.]
  {I3D ResNet-18 network architecture with its outputs sizes and RCN's output sizes for input with $16\times 112\times 112$ size. 
  Each Convolution layer of the network is defined by temporal ($n$) and spatial ($d$) size of kernel and number of kernels. 
  ConvC (convolutional classification) layer uses the number of classes as the number of kernels.}
  \vspace{1mm}
  {\footnotesize
  {
  \begin{tabular}{cc|cc}
    \toprule
    \multicolumn{2}{c}{Layers} & \multicolumn{2}{c}{Output Sizes} \\ 
    \midrule
    Names  & $n$, $d$, number of kernels & I3D   & RCN \\
    \midrule
    conv1 & $3,7,64$; stride $1,2,2$           & $16\times 56\times 56$  & $16\times 56\times 56$ \\
    res2 & $[3,3,64 $ \& $ 3,3,64]\times 2$   & $16\times 56\times 56$  & $16\times 56\times 56$ \\
    res3 & $[3,3,128 $ \& $ 3,3,128]\times 2$  & $8\times 28\times 28$  & $16\times 28\times 28$ \\
    res4 & $[3,3,256 $ \& $ 3,3,256]\times 2$  & $4\times 14\times 14$   & $16\times 14\times 14$ \\
    res5 & $[3,3,512 $ \& $ 3,3,512]\times 2$  & $2\times 7\times 7$     & $16\times 7\times 7$ \\
    pool & spatial pooling  & $2\times 1\times 1$     & $16\times 1\times 1$ \\
    convC & classification; $1,1,C $ & $2\times C$ & $16\times C$ \\
    mean & temporal pooling & $C$ & $C$ \\
    \bottomrule
  \end{tabular}
  }
  }
  \label{rcn:table:netArch} 
\end{table}

\subsection{Separated convolution networks}\label{rcn:subsec:s3d} 

Figure~\ref{rcn:fig:convs}(b) shows how a 3D ($t \times d \times d$) convolution can be decomposed into a ($1 \times d\times d$) spatial convolution 
and a ($t \times 1 \times 1$) temporal one. Usually, the size of the temporal kernel $t$ is set to be equal to its spatial dimension $d$, 
as in both I3D~\cite{carreira2017quo} and C3D~\cite{tran2014learning}.
\\
Such a separated convolutional network (S3D) was introduced by
Xie~\etal~\cite{xie2018rethinking}. The authors showed that such a decomposition not only reduces the number of parameters, 
but also delivers performances very much similar to those of traditional 3D CNNs.
After taking a closer look at 3D convolution separation,
Tran~\etal~\cite{tran2018closer} argued that if the number of kernels $M_i$ used in spatial convolution (Figure \ref{rcn:fig:convs}(b)) are increased
in such way that the numbers of parameters of spatial and temporal convolution combined are equal to the number of parameters in 3D convolution, 
then performance actually improves over 3D networks. 
However, such a change in the number of kernels does not allow initialisation from ImageNet pre-trained models any longer.
They refer to their model as $(2+1)D$ model. Although the latter can be considered a special case of Pseudo-3D networks (P3D) ~\cite{qiu2017learning} 
models, because of its homogeneity and simplicity the (2+1)D model performs better than P3D.

We re-implemented (2+1)D without ImageNet initialisation as an additional baseline 
as it has the most promising result without any additional trick like gating in S3Dg (S3D with gating). 

\section{3D Recurrent convolutional network}\label{rcn:sec:rcn}


We are now ready to describe the architecture of our causal \emph{Recurrent Convolutional (3D) Network} (RCN) and its properties in detail. 
Firstly, we show how Recurrent Convolutional Units (RCUs) (\S~\ref{rcn:subsec:rcu}) 
are used to replace \emph{every} 3D convolutions in the I3D network (\S~\ref{rcn:subsec:i3d}), 
resulting in our RCN model (\S~\ref{rcn:subsec:rcn}).
Next, we show how RCUs preserve temporal resolution in Section~\ref{rcn:subsec:Resolution}.
Then in Section~\ref{rcn:subsec:Causality}, we show how our network behaves in a causal manner. 
Lastly, in~\S~\ref{rcn:subsec:imagenetInit} and~\S~\ref{rcn:subsec:idInit}, we illustrate the initialisation process for RCN and RCU.

\subsection{Recurrent convolutional unit}\label{rcn:subsec:rcu}

\begin{figure}[t]
  \centering
  \includegraphics[scale=1.0]{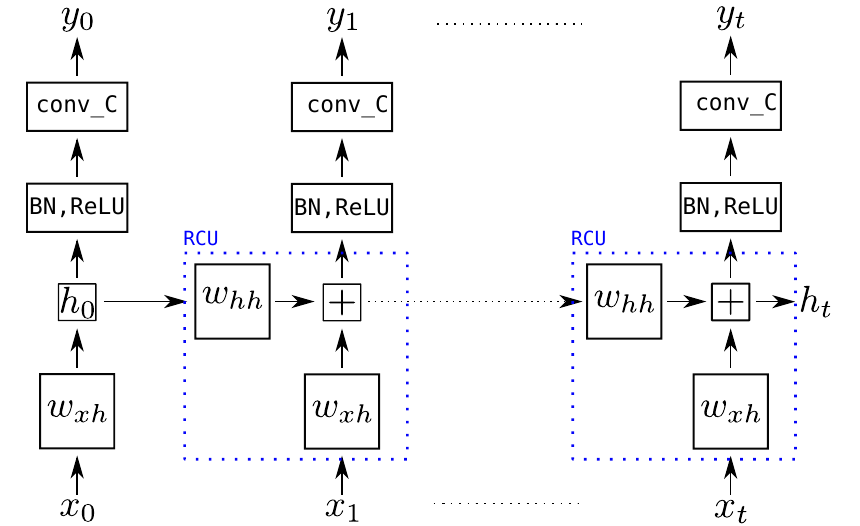}
  \caption[An unrolled version of Recurrent Convolutional Network (RCN).]{An unrolled version of Recurrent Convolutional Network (RCN) composed by a single RCU layer followed by a batch normalisation (BN) layer, 
      a ReLU activation layer, and a final convolutional layer used for classification.}
\label{rcn:fig:rcn}
\end{figure}

A pictorial illustration of our proposed \emph{Recurrent Convolutional Unit} (RCU) is given in Figure~\ref{rcn:fig:convs}(c).
The input at any time instant $t$
passes through 3D spatial convolution (with kernel of size $1 \times d \times d$, denoted by $w_{xh}$). 
The result is added to the output of a recurrent convolution operation, with kernel denoted by $w_{hh}$, of size $1 \times 1 \times 1$.
\\
The result is termed the \emph{hidden state} $h_t$ of the unit.
Analytically, a recurrent convolutional unit can be described by the following relation:
\begin{equation}\label{equ:rcu}
h(t) = (h_{t-1} * w_{hh} + x_t * w_{xh})/2.0,
\end{equation}
where $w_{hh}$ and $w_{xh}$ are parameters of the RCU, and $*$ denotes the convolution operator. 
An RCU can be seen in the context of other modules of a network in Figure~\ref{rcn:fig:rcn}.

In practise, we replace every 3D convolution in Table~\ref{rcn:table:netArch} with RCU to obtain an RCN. 
In general, one could skip/add a few replacements to obtain a mixed network.

\subsection{Unrolling a recurrent convolutional network}\label{rcn:subsec:rcn}

Figure~\ref{rcn:fig:rcn} represents a simple recurrent convolutional network (RCN) composed by a single RCU unit, unrolled up to time $t$.
At each time step $t$, an input $x_t$ 
is processed by the RCU and the other layers to produce an output $y_t$.

The unrolling principle allows us to build an RCN from 2D/3D networks, e.g. by replacing 3D convolutions with RCUs in any I3D network. 
Indeed, the network architecture of our proposed model builds on the I3D network architecture shown in Table~\ref{rcn:table:netArch}, where the same parameters ($d$, number of kernels) used for 3D convolutions are used for our RCU. 
Unlike I3D, however, our RCN does not require a temporal convolution size $n$ (cfr. Table~\ref{rcn:table:netArch}) as a parameter.
As in 2D or I3D ResNet models~\cite{he2016deep,tran2018closer,hara2018can}, 
our proposed RCN also has residual connections.
The hidden state $h_t$ at time $t$ is considered to be the output at that time instant -- as such, it acts as input to next hidden state and to the whole next depth-level layer. 
Table~\ref{rcn:table:netArch} describes the network architecture of ResNet-18~\cite{he2016deep} with 18 layers. 
Similarly, we can build upon other variants of ResNet.

\subsection{Temporal resolution preservation} \label{rcn:subsec:Resolution}

The output sizes for both I3D and our proposed RCN are shown in Table~\ref{rcn:table:netArch}. 
Our RCN only uses spatial pooling and a convolutional layer for classification, 
unlike the spatiotemporal pooling of~\cite{carreira2017quo,nonlocal2018wang,tran2018closer}.
From Table~\ref{rcn:table:netArch}, compared to I3D, RCN produces 16 classification score vectors with an input sequence length of 16. 

This one-to-one mapping from input to output is essential in many tasks, ranging from temporal action segmentation~\cite{shou2017cdc,rene2017temporal}, 
to temporal action detection~\cite{singh2016untrimmed}, to action tube detection~\cite{singh2017online}.
In all such tasks, video-level accuracy is not enough, but we need
frame-level results in terms, e.g., of detection bounding boxes and class scores.
Temporal convolution behaves in a similar way to spatial convolution: 
it results in lower resolution feature maps as compared to the input as the depth of the network increases.

Unlike the temporal deconvolution proposed in ~\cite{shou2017cdc,rene2017temporal},
our RCN inherently addresses this problem (see Table~\ref{rcn:table:netArch}).
For a fair comparison, in our tests, we adjusted our baseline I3D model for dense prediction, 
by setting the temporal stride to 1 (\S~\ref{rcn:exp:subsec:multithumos}). 
The resulting I3D model can produce dense predictions: 
given $T$ frames as input, it will generate $T$ predictions in 1-1 correspondence with the input frames.

\subsection{Causality and long-term dependencies} \label{rcn:subsec:Causality}

A size-$n$ temporal convolution operation uses a sequence $x_{t-n/2},..., x_t,... ,x_{t+n/2}$ as  input to generate an output $y_t$ at time $t$.
In our Recurrent Convolutional Network, instead, $y_t$ is a function of only the inputs $x_{0},x_{1},..., x_t$ from 
the present and the past (up to the initial time step), as shown in Figure~\ref{rcn:fig:rcn}. 
Its independence from future inputs makes the output $y_t$ at time $t$ causal.
Thus RCN, as presented here, is not only causal but poses no constraints on the modelling of 
temporal dependencies (as opposed to an upper bound of $n$ in the case of temporal convolutions).
Temporal dependencies are only limited by the input sequence length at training time. 

As in traditional RNNs, we have the option to unroll the same network to model arbitrary input sequence lengths at test time, 
thus further increasing the horizon of temporal dependencies and show a substantial gain in performance.
We will discuss it in more details while discussing the results in Section~\ref{rcn:exp:subsec:initeffect}.

\subsection{ImageNet initialisation for the 2D layers}\label{rcn:subsec:imagenetInit}

The I3D model proposed by Carreira~\etal~\cite{carreira2017quo} greatly owes its success to a good initialisation from 2D models trained on ImageNet~\cite{deng2009imagenet}.
By inflating these 2D models, we can benefit from their ImageNet pre-trained weights, as in most state-of-the-art 3D models~\cite{carreira2017quo,xie2018rethinking,nonlocal2018wang}.
We follow the same principle and initialise all 2D layers using the weights of available pre-trained 2D ResNet models~\cite{he2016deep}. 
It is noteworthy that the other state-of-the-art (2+1)D model by Tran~\etal~\cite{tran2018closer} 
cannot, instead, exploit ImageNet initialisation, because of the change in the number of kernels. 

\subsection{Identity Initialisation for the Hidden Layers}\label{rcn:subsec:idInit}

The presence of a hidden state convolution ($w_{hh}$, see Figure~\ref{rcn:fig:rcn}) layer at every depth level of the unrolled network makes initialisation a tricky issue. 
The random initialisation of the hidden state convolution component could destabilise the norm of the feature space between two 2D layers. 
In response to a similar issue, Le~\etal~\cite{le2015simple} presented a simple way to initialise RNNs when used with ReLU~\cite{glorot2011deep} activation functions. 
Most state-of-the-art 2D models~\cite{he2016deep,szegedy2015going} make indeed use of ReLU as activation function of choice for fast and optimal convergence~\cite{glorot2011deep}.

Following the example of Le~\etal~\cite{le2015simple} and others~\cite{mikolov2014learning,arjovsky2016unitary}, 
we initialise the weights of the hidden state convolution kernel ($w_{hh}$) with the identity matrix. 
Identity matrix initialisation is shown~\cite{le2015simple,mikolov2014learning} to capture longer term dependencies. 
It also helps induce forgetting capabilities in recurrent models, unlike traditional RNNs.

%% file: chapters/partIII/rcn_exp.tex
\section{Experiments} \label{rcn:sec:exp}

In this section, 
we evaluate our RCN 
on the challenging Kinetics~\cite{kay2017kinetics} and MultiThumos~\cite{yeung2015every} datasets to \textbf{answer the following questions}:
i) How does our training setup, which uses 4 GPUs, compare with the 64 GPU training setup of ~\cite{tran2014learning} (Section~\ref{subsec:trainsetupcomparision})? 
ii) How do recurrent convolutions compare against 3D convolutions in the action recognition problem (\S~\ref{rcn:sec:RCNvsI3D})?
iii) How does our RCN help solving the dense prediction task associated with action detection (\S~\ref{rcn:exp:subsec:multithumos})?
Finally, iv) we validate our claims on the temporal causality and flexibility of RCN, and check whether those features help {with longer-term temporal reasoning} (\S~\ref{rcn:exp:subsec:casuality}).
 

\paragraph{Evaluation:} For fair comparison, we computed clip-level and video-level accuracy as described in~\cite{tran2018closer,nonlocal2018wang}.
Ten regularly sampled clips were evaluated per video, and scores were averaged for video-level classification.
On videos of arbitrary length, we averaged all the predictions made by the unrolled versions of both our RCN and of I3D.
Please refer to Section~\ref{soa:datasets} to gather more details about the datasets and evaluation metrics (accuracy and mAP) used in this chapter.

\subsection{Implementation details}\label{subsec:implementationdetails}

In all our experiments we used \emph{sequences of RGB} frames as input for simplicity and computational reasons.
We used a batch size of 64 when training ResNet18-based models and 32 for models based on ResNet-34 and -50.
The initial learning rate was set to $0.01$ for batches of 64, and to $0.005$ for batches of 32. 
We reduced the learning rate by a factor of $10$ after both $250K$ and $350K$ iterations.
Moreover, training was stopped after $400K$ iterations (number of batches). 
We used standard data augmentation techniques, such as random crop, horizontal flip with 50\% probability, and temporal jitter.


\begin{table}[h]
 \centering
 \caption[Action recognition accuracy on the validation set of the Kinetics with ResNet-18 as base model.]
 {Clip-level and video-level action recognition accuracy on the validation set of the Kinetics dataset for different ResNet18 based models, trained using 8-frame-long clips as input. \vspace{1mm}}
 {\footnotesize
 {
 \begin{tabular}{lcccc}
 \toprule
 Network & \#Params & Initialisation & Clip \% & Video \%\\ 
 \midrule
 I3D~\cite{hara2018can} & 33.4M & random & -- & 54.2 \\ 
 I3D~\cite{tran2018closer} & 33.4M & random & 49.4 & 61.8 \\ 
 (2+1)D~\cite{tran2018closer} & 33.3M & random & 52.8 & 64.8 \\ 
 \midrule
 I3D${}^{\dagger}$ & 33.4M & random & 49.7 & 62.3 \\ 
 RCN [ours]${}^{\dagger}$ & \textbf{12.8}M & random & 51.0 & 63.8 \\ 
 (2+1)D${}^{\dagger}$ & 33.4M & random & 51.9 & 64.8 \\ 
 I3D${}^{\dagger}$ & 33.4M & ImageNet & 51.6 & 64.4 \\ 
 RCN [ours]${}^{\dagger}$ & \textbf{12.8}M & ImageNet & \textbf{53.4} & \textbf{65.6} \\ 
 \bottomrule
 \multicolumn{5}{l}{${}^{\dagger}$ trained with our implementation and training setup.}
 \end{tabular}
 }
 }
 \label{rcn:table:18results}
\end{table}

\subsection{Fair training setup}\label{subsec:trainsetupcomparision}
\textbf{GPU memory} consumption plays a crucial role in the design of neural network architectures.
In our training processes, a maximum of 4 GPUs was used. 
Given our GPU memory and computational constraints, we only considered training networks with 8-frame long input clips, except for ResNet34 which was trained with 16 frames long clips.

As explained above, we have GPU memory constrained, 
we will report results of previous basic 3D models~\cite{tran2018closer,carreira2017quo} re-implemented and trained by using the same amount of resources as our RCN.
The main hyperparameters involved in the training of a 3D network are learning rate, batch size, and the number of iterations. 
These parameters are interdependent, and their optimal setting depends on the computational power at disposal. 
For instance, Tran~\etal~\cite{tran2018closer} would use 64 GPUs, 
with the training process distributed across multiple machines.
In such a case, when vast computational resources are available\cite{tran2018closer,carreira2017quo,carreira2018massively}, 
training takes 10-15 hours~\cite{tran2018closer},
allowing for time to identify the optimal parameters.
The availability of such computational power, however, is scarce. 

In a bid to reproduce the training setup of~\cite{tran2018closer} on 4 GPUs, 
we re-implemented the I3D and (2+1)D models using ResNet18 and ResNet34 as a backbone. 
The ResNet18-I3D architecture is described in Table~\ref{rcn:table:netArch}. 
Based on the latter, we built a (2+1)D~\cite{tran2018closer} architecture in which 
we matched the number of parameters of separated convolutions to that of standard 3D convolutions, as explained in \cite{tran2018closer}.

The results of the I3D and (2+1)D implementations reported in Tran~\etal~\cite{tran2018closer} are shown in the top half of Table~\ref{rcn:table:18results}.
When comparing them with our implementations of the same networks in the bottom half,
it is clear that our training is as performing as that of Tran~\etal~\cite{tran2018closer}.
This allows a fair comparison of our results.

\paragraph{Why smaller clips as input:}
training a ResNet18-based model on Kinetics with 8 frame clips as input takes up to 2-3 days on 4 GPUs. 
Training a ResNet50-based model takes up to 4-5 days. 
In principle, one could train the same model for longer input clip sizes, but the amount of GPU memory and time required to train would grow linearly. 
As an estimate, it would take more than two weeks to train a ResNet50 model on 64 long frame clips, 
assuming that all the hyperparameters are known (i.e., batch size, learning rate, step size for learning rate drop, and whether batch normalisation layers should be frozen or not). 

For these reasons we stick to smaller input clips to train our models in a fair comparison setting, using the hyperparameter values from \S~\ref{subsec:implementationdetails}.

\begin{table}[t]
 \centering
 \caption[Action classification accuracy of different models on the validation set of the Kinetics dataset.]
 {Video-level action classification accuracy of different models on the validation set of the Kinetics dataset.} 
 {\footnotesize
 
 {
 
 \begin{tabular}{lccc}
 \toprule
 Model & Clip-length & Initialisation & Acc\%\\ 
 \midrule
 ResNet34-(2+1)D~\cite{tran2018closer}${}^{\dagger}$ &16 & random & 67.8 \\
 ResNet34-I3D~\cite{carreira2017quo} ${}^{\dagger}$ &16 & ImageNet & 68.2 \\
 ResNet34-RCN [ours]${}^{\star\dagger}$ &16 & ImageNet & \textbf{70.3} \\
 \midrule
 ResNet50-I3D~\cite{carreira2017quo} ${}^{\dagger}$ &8 & ImageNet & 70.0 \\
 ResNet50-RCN [ours]${}^{\star\dagger}$ &8 & ImageNet & 71.2 \\
 ResNet50-RCN-unrolled [ours]${}^{\star\dagger}$ &8 & ImageNet & \textbf{72.1} \\
 \bottomrule
 \multicolumn{4}{l}{${}^{\star}$ causal model; ${}^{\dagger}$ trained with our implementation and training setup.} 
 \end{tabular}
 }
 }
 \vspace{1mm}
 \label{rcn:table:34compare} 
\end{table}

\subsection{Results on action recognition}\label{rcn:sec:RCNvsI3D}
We compared our RCN with both I3D and (2+1)D models in the action recognition problem on the Kinetics dataset.
A fair comparison is shown in Table~\ref{rcn:table:18results} with ResNet18 as backbone architecture. 
Table~\ref{rcn:table:34compare} shows the results with ResNet34 and ResNet50 as backbone, trained on 16 frame and 8 frame clips, respectively.
It is clear from these figures that RCN significantly outperforms state-of-the-art 3D networks -- e.g. our network outperforms the equivalent
I3D network by more than 2\% across the board.

The ability to model \textbf{long-term temporal reasoning} of RCN is attested by the performance of the unrolled version (last row of Table~\ref{rcn:table:34compare}). It shows that, even though the network is trained on input clip of 8 frames, it can reason over longer temporal horizons at test time. The corresponding unrolled I3D version (the last classification layer is also convolutional, see Table~\ref{rcn:table:netArch}) showed no substantial improvement in performance -- in fact, a slight drop.

\paragraph{Comparison with I3D variants: }
the main variants of the I3D model are separated 3D convolutions with gating (S3Dg)~\cite{xie2018rethinking} and with non-local operators (NL)~\cite{nonlocal2018wang}. 
We think it appropriate to take a closer look at these variants of I3D as they provide state-of-the-art performance, albeit being all anti-causal.
In \cite{xie2018rethinking,nonlocal2018wang} the application of non-local or gating operations to I3D yields the best performances to date, mainly thanks to training on longer clips given a large amount of GPU memory at their disposal (S3Dg~\cite{xie2018rethinking} models are trained using 56 GPUs, \cite{nonlocal2018wang} uses 8 GPUs with 16GB memory each).
The best version of I3D-NL achieves an accuracy of 77.7\%, but uses 128 frames and ResNet101 as backbone network; hence we do not deem fair to compare it with our models (which only use 8 frame long clips). 
It would take almost a month to train such a network using 4 GPUs.
What needs to be stressed is that
\emph{gating and NL} operations are not at all constrained to be applied on top of I3D or S3D models: indeed, they \emph{can also be used in conjunction with (2+1)D and our own RCN model}. 
As in this work we focus on comparing our network with other 3D models, 
we chose I3D and (2+1)D as baselines (Sec. \ref{rcn:sec:RCNvsI3D}). 

\paragraph{Training on longer sequences:} 
we tried training on longer sequences (\textbf{32} frames) by reducing the batch size to 8 with ResNet50 as base network. 
Despite a sub-optimal training procedure, RCN was observed to still outperform I3D by a margin of 1.5\%. 
A closer to optimal training procedure with ResNet50 (as in \cite{tran2018closer,nonlocal2018wang}), is very likely to yield even better results.

\begin{table}[t]
 \centering
 \caption[Temporal action detection performance on MultiThumos dataset.]{Action detection/segmentation results on MultiThumos dataset, mAP computed from dense prediction at every frame (mAP@1) and every 8th frame (mAP@8).} 
 {\footnotesize
 {
 \begin{tabular}{lccc}
 \toprule
 Network & Input & mAP@1 \% & mAP@8 \%\\ 
 \midrule
 Two-stream+LSTM \cite{yeung2015every}${}^{\star}$ & \tiny{RGB+FLOW} & 28.1 & - \\
 MultiLSTM \cite{yeung2015every}${}^{\star}$ & \tiny{RGB+FLOW} & 29.7 & - \\
 Inception-I3D by~\cite{piergiovanni2018learning} & \tiny{RGB+FLOW} & - & 30.1 \\
 Inception-I3D + SE \cite{piergiovanni2018learning} & \tiny{RGB+FLOW} & - & 36.2 \\
 \midrule
 ResNet50-I3D [baseline] & \tiny{RGB} & 34.8 & 36.9 \\ 
 ResNet50-RCN [ours]${}^{\star}$ & \tiny{RGB} & 35.3 & 37.3 \\ 
 ResNet50-RCN-unrolled [ours]${}^{\star}$ & \tiny{RGB} & \textbf{36.2} & \textbf{38.3} \\ 
 \bottomrule
 \multicolumn{4}{l}{${}^{\star}$ causal model}
 \end{tabular}
 }
 }
 \label{rcn:table:multithumos} 
\end{table}


\subsection{Results on temporal action detection}\label{rcn:exp:subsec:multithumos}

We also evaluate our model on the temporal action detection problem on the MultiThumos~\cite{yeung2015every} dataset. The latter is a dense label prediction task. As a baseline, we use a temporal resolution preserving a version of I3D introduced in Section~\ref{rcn:subsec:Resolution}.
ResNet50 is employed as a backbone for both our RCN and the baseline I3D.
To capture the longer duration, we use 16 frame clips as input; the sampling period is 4 frames. 
Both networks are initialised with the respective models pretrained on Kinetics.
The initial learning rate is set to $0.001$ and dropped after $14K$ iterations, 
a batch size of $16$ is used, and trained up to $20K$ iterations. 
Similar to~\cite{piergiovanni2018learning}, we use binary cross-entropy as loss function. 

We use the evaluation setup of~\cite{yeung2015every} and ~\cite{piergiovanni2018learning}, and computed both frame-wise mean Average Precision at 1 (mAP@1)
(in which case a prediction needs to be made for each frame) 
and~\cite{piergiovanni2018learning} mAP@8 (every 8th frame).

Table~\ref{rcn:table:multithumos} shows the performance of our models along with that of other state-of-the-art methods. Two LSTM-based causal models presented by~\cite{yeung2015every} are shown in rows 1 and 2. Piergiovanni~\etal~\cite{piergiovanni2018learning} use pretrained I3D~\cite{carreira2017quo} to compute features, but do not train I3D end-to-end, hence
their performance is lower than in our version of I3D. 
Our RCN outperforms all other methods, 
including anti-causal I3D+Super-Events (SE)~\cite{piergiovanni2018learning} and the I3D baseline.
It is safe to say that RCN is well applicable to dense prediction tasks as well.

\subsection{Causality and temporal reasoning}\label{rcn:exp:subsec:casuality}

A comparison with other causal methods is a must, as we claim the causal nature of the network to be the main contributions of our work, 
making RCN best suited to online applications such as action detection and prediction. 
In Section~\ref{rcn:exp:subsec:multithumos} we have already shown that our model excels in the task of temporal action detection.

\begin{table}[t]
 \centering
 \caption[Comparison of causal and anti-causal models.]
 {Comparison between RCN and other causal models on the validation set of Mini-Kinetics~\cite{carreira2018massively} (First half of the table) and Kinetics-400 (last half of the table). 
 ResNet-50 (R50) is used as base model.}
 {\footnotesize
 {
 \begin{tabular}{lccc}
 \toprule
 Model & Clip-length & Acc\%\\ 
 \midrule
 InceptionV1-I3D~\cite{carreira2018massively} &64 & 71.8 \\
 InceptionV1-I3D-seq~\cite{carreira2018massively}${}^{\star}$ &64 & 71.1 \\
 InceptionV1-I3D-par~\cite{carreira2018massively}${}^{\star}$ &64 & 54.5 \\
 \midrule
 R50 - I3D${}^{\dagger}$ &8 & 70.0 \\
 R50 - RCN [ours]${}^{*\dagger}$ &8 & 71.2 \\
 R50 - RCN - unrolled [ours]${}^{*\dagger}$ & 8 & 72.2 \\
 \bottomrule
 \multicolumn{3}{l}{${}^{\star}$ causal model, unlike respective I3D version} \\
 \multicolumn{3}{l}{${}^{\dagger}$ trained with our implementation and training setup.} \\
 \end{tabular}
 }
 }
 \label{rcn:table:causal} 
\end{table}

Carreira~\etal~\cite{carreira2018massively} proposed two causal variants of the I3D network. 
Their sequential version of I3D, however, shows a slight drop \cite{carreira2018massively} in performance as compared to I3D, as seen in the first and second row of Table~\ref{rcn:table:causal}.
Their parallel version is much faster than the sequential one but suffers from an even more significant performance decline 71.8\% to 54.5\% \cite{carreira2018massively}, as seen in the first and third row of Table~\ref{rcn:table:causal}. 

In contrast, our causal/online model not only outperforms other causal models (see Table~\ref{rcn:table:multithumos}) but beats as well strong, 
inherently anti-causal state-of-the-art 3D networks on a large scale dataset such as Kinetics (see last half of Table~\ref{rcn:table:causal}).

\begin{figure}[t]
 \centering
 \includegraphics[scale=0.98]{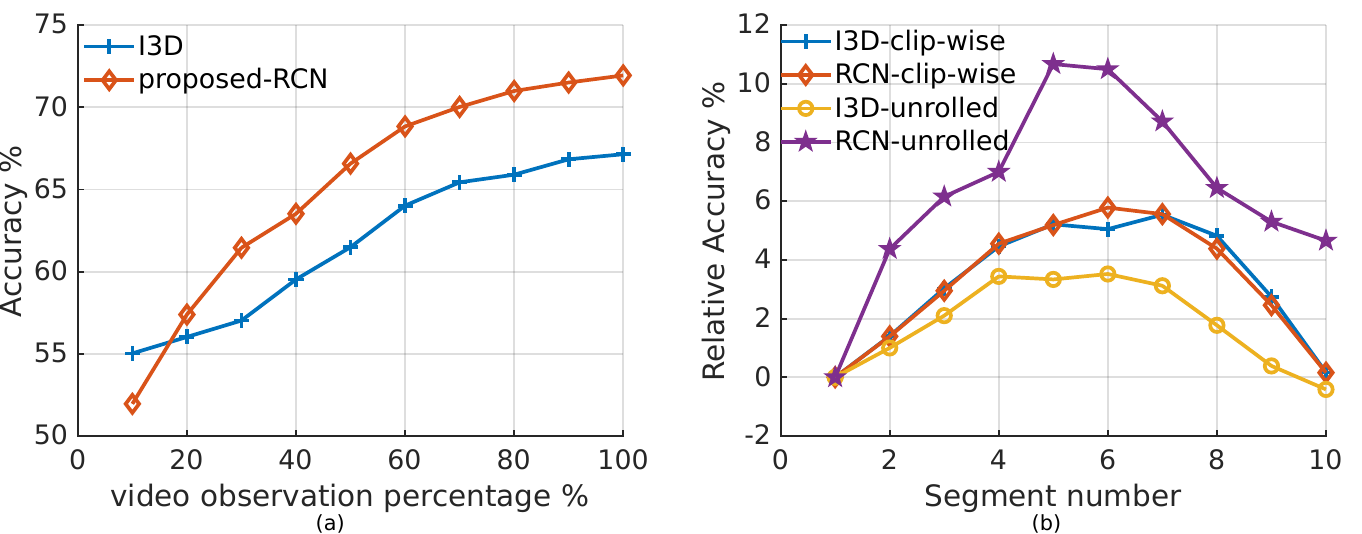}
 \vspace{1mm}
 \caption[Early action label prediction results and segment-level results on Kinetics.]
 {(a) Online/early action prediction task: accumulated scores are used to compute the accuracy against the label of the entire video as a function of the video observation percentage. 
 (b) relative (difference from first segment) accuracy of each of 10 regularly sampled segments. 
 } 

\label{rcn:fig:combinded}
\end{figure}

\vspace{2mm}
\noindent
In addition, as in~\cite{ryoo2011human,singh2017online}, we can use the
\textbf{early action prediction task} to evaluate the sequential temporal reasoning of RCN. The task consists in guessing the label of an entire action instance (or video, if containing a single action) after observing just a fraction of video frames.
Accumulated output scores up to time $t$ are used to predict the label of the entire video. 
The idea is that a system should improve its classification accuracy as it observes a larger fraction of the input video, by exploiting temporal context. 
Figure~\ref{rcn:fig:combinded}(a) shows that our RCN improves drastically as more video frames are observed when compared to I3D. It indicates that RCN has superior anticipation ability, albeit starting slowly in first 10\% of the video.

\vspace{2mm} 
\noindent
Furthermore, to provide useful cues about causality and temporal reasoning, we designed an original
\textbf{segment-level classification} evaluation setting.
Namely, the outputs of the models being tested are divided into ten regularly sampled segments and the difference between the accuracy for each segment and that for the first segment is computed,
as shown in Figure~\ref{rcn:fig:combinded}(b). 
Within this setting, we compared the I3D baseline with RCN in two different modalities, 
one considering clip-wise outputs in a sliding window fashion, 
the other obtained by unrolling both I3D and RCN over test videos of arbitrary length.

Notably, middle segments provide the best relative improvement, 
which is reasonable as it indicates that the middle part of the video is the most informative.
Secondly, the last segment (no. 10) has the lowest relative accuracy of all, except for RCN-unrolled. 
The relative accuracy of a pure causal system, though, should improve monotonically, {i.e., exploit all it has seen.}
Instead, all compared models end up at the same performance they started with, except for unrolled RCN for which the final accuracy is almost 5\% higher than the initial one. 
We can conclude that unrolled RCN has a longer-term memory than unrolled I3D or both sliding window-based I3D/RCN. 

\begin{figure}[h]
 \centering
 \includegraphics[scale=0.98]{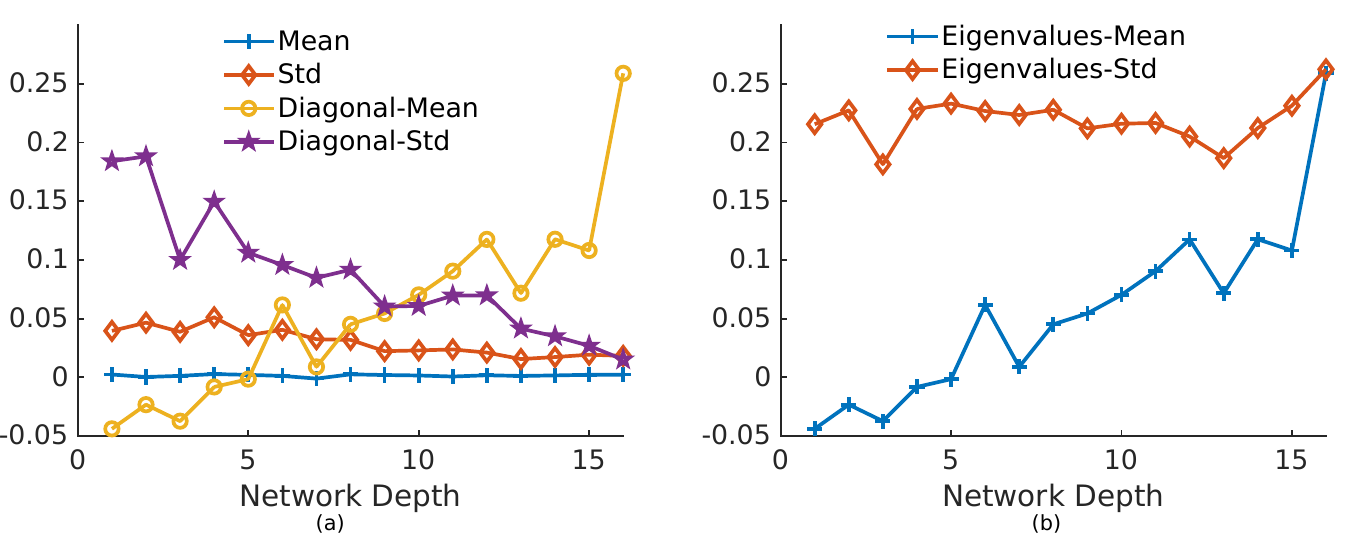}
 \vskip 1mm
 \caption[Statistics of hidden state convolution's weight matrices]{(a) Mean and standard deviation (Std) of all the entries of the weight matrices ($w_{hh}$) of the hidden state at every RCU layer of RCN, as well as those of just the diagonal elements.
 (b) Mean and Std of the eigenvalues of the hidden state {weight matrices}
 at every RCU layer of a 18-layer RCN.} 
\label{rcn:fig:combindedStats}
\end{figure}

\subsection{Evolution of recurrence with network depth}\label{rcn:exp:subsec:evolution}
It is another aspect that can provide clues about RCN's temporal flexibility.
To this purpose, we examine the statistics of the weight matrices ($w_{hh}$) associated with the hidden state at every RCU layer in the RCN network.
In Figure~\ref{rcn:fig:combindedStats}(a) we can see that the mean of the diagonal elements of the weight matrices increases and their standard deviation decreases with the depth of the network.
This means that the $w_{hh}$ matrix becomes sparser as network depth grows. 
In our view, this phenomenon is associated with RCN putting more focus on feature learning in the early part of the network, 
and emphasising temporal reasoning at later depths as the temporal reasoning horizon (`receptive field') 
increases. 
In other words, RCN learns to select the layers which should contribute towards temporal reasoning automatically.
A similar phenomenon is observed in networks based on temporal convolutions~\cite{xie2018rethinking}.

Arjovsky~\cite{arjovsky2016unitary} argue that if the eigenvalues
of the hidden state-to-hidden state weight matrix diverge from the value 1, optimisation becomes difficult due to the vanishing gradient problem.
Chang~\etal~\cite{chang2018antisymmetricrnn} explore a similar idea. 
Taking an ordinary differential equation view of RNNs, they argue that their stability for long-term memory
is related to the eigenvalues of the weight matrices. 
Figure~\ref{rcn:fig:combindedStats}(b) shows that in RCN the mean eigenvalue does rise towards 1 as network depth increases, suggesting that
later layers are more stable in terms of long-term memory whereas earlier layers 
are not concerned with long-term reasoning.


\begin{table}[h]
 \centering
 \caption[Effect of initialisation on RCN's performance.]
 {Video-level and clip-level action recognition accuracy on 
 the Kinetics validation set for different initialisation of 2D layer ($w_{xh}$) 
 and hidden state unit weights ($w_{hh}$) in RCU with ResNet-18-based RCN models 
 trained on 8 frame-long clip as an input. }
 {\footnotesize
 {
 \begin{tabular}{cccc}
 \toprule
 $w_{xh}$ init & $w_{hh}$ init & Clip-Acc\% & Video-Acc\%\\ 
 \midrule
 Random & Random & 49.2 & 61.2 \\ 
 Random & Identity & 49.7 & 62.1 \\ 
 ImageNet & Random & 50.7 & 62.9 \\ 
 ImageNet & Identity & \textbf{53.4} & \textbf{65.6} \\ 
 \bottomrule
 \end{tabular}
 }
 }
 \label{rcn:table:hiddenInit}
\end{table}

\subsection{Effect of weight initialisation.}\label{rcn:exp:subsec:initeffect}
Weights of inflated 2D layers in ResNet (base networks)~\cite{he2016deep} 
based I3D and our RCN networks 
are initialised with weights from a pre-trained ImageNet model train on RGB images.
In the case of RCN inflation in the third dimension is 1, 
practically, it is 2D weight matrix with the third dimension being 1.
As a result, we do need to replicate the weights like in the 3D layer of I3D.

Random initialisation for hidden state parameters resulted in sub-optimal training. 
Thus, in all the experiments with RCN, we used identity matrix initialisation instead.
Here we show the result of training RCN with different initialisation of 
hidden-state $w_{hh}$ convolution and spatial $w_{xh}$ convolution in Table~\ref{rcn:table:hiddenInit}.
The first row of the table where both $w_{xh}$ and $w_{hh}$ of all the RCUs in 
RCN is initialised randomly using normal initialisation process described in~\cite{he2015delving}.

Table~\ref{rcn:table:hiddenInit} show the result of different initialisation of $w_{xh}$ and $w_{hh}$ 
respectively.
It is clear from the last row of the table that RCN performs best 
if $w_{xh}$ and $w_{hh}$ are initialised with ImageNet and identity matrix respectively.
It is noteworthy that the performance difference is relatively 
small in first three rows of that table as compared to the last row jump.

\subsection{Discussion}

In the light of RCN's superior results on temporal action detection (\S~\ref{rcn:exp:subsec:multithumos}), 
early action prediction (see Figure~\ref{rcn:fig:combinded}(a)), 
long-term temporal reasoning (in its unrolled incarnation) at segment-level 
and for action recognition (\S~\ref{rcn:sec:RCNvsI3D}, see the last row of Table \ref{rcn:table:34compare}),
it is fair to say that the proposed Recurrent Convolutional Network is the best performing causal network out there. 

\paragraph{Why layer-wise design:} we tried replacing 3D CONV by RCU, (i) only in the last four layers and (ii) on four regularly sampled layers of ResNet50. This led to lower performance (69\% and 68\% respectively), compared to 71\% when RCU replaces all 18 3D CONVs. This is consistent with previous approaches~\cite{nonlocal2018wang,xie2018rethinking}, we can conclude that replacing every 3D CONV with RCU leads to the best performance.

\paragraph{The number of parameters} in our proposed RCN model is $12.8$ million (M), as opposed to 33.4M in both the I3D and (2+1)D models, see Table~\ref{rcn:table:18results}. 
It is remarkable to see that, despite a $2.6$ times reduction in the number of parameters, RCN still outperforms both I3D and (2+1)D when trained using ImageNet initialisation. 
Further, RCN surpasses I3D also under random initialisation, while using $2.6$ times fewer model parameters. 
We measured the floating-point operations (FLOPs) for I3D, R(2+1)D, and RCN, recording 41MMac, 120MMac, and 54MMac, respectively. 
Thus, RCN requires half the FLOPs as compared to (2+1)D, and is comparable to I3D because RCN preserves temporal resolution. 
We computed the average time taken to process ten-second long videos of the Kinetics dataset. This takes 0.4s, 0.8s, and 0.9s for I3D, RCN, and (2+1)D respectively.

\paragraph{ImageNet initialisation} proves to be useful for both the I3D and our RCN models.
While (2+1)D performs (Table~\ref{rcn:table:18results}, row 6) better than RCN (row 5) 
with the random initialisation, our RCN recovers to improve over (2+1)D (row 6) 
with ImageNet initialisation, whereas (2+1)D cannot make use of free ImageNet initialisation.
This seems to be a severe drawback for the (2+1)D model, and a big advantage for I3D and RCN.
One may argue that, if the purpose is to build on existing 2D models, then RCN and I3D are a better choice, 
whereas if new 3D models are preferred then (2+1)D might prove useful. 
The latter does provide better performance with the random initialisation, 
but at the price of requiring many more parameters than RCN.

\paragraph{Random initialisation} for hidden state parameters resulted in sub-optimal results. 
Thus, in all the experiments with RCN, we used identity matrix initialisation instead.
Identity matrix initialisation helps to capture forget capabilities as well, 
as suggested by~\cite{le2015simple}.
An ablation study on the effect of initialisation is provided in Section~\ref{rcn:exp:subsec:initeffect}.

\paragraph{On input clip length:} input clip length is another factor in determining the final performance of any network. 
We can see from the Table that all the Inception-based models are trained on 64 frame-long clips, one of the reasons why Inception nets work better that ResNet models while using fewer parameters. Among the latter, ResNet101-I3D-NL~\cite{nonlocal2018wang} is shown to work better with an even longer input clip length. 
Thus, the evidence supports that training on larger input sequences boosts performance while clashing with memory limitations. 

In our experiments, as mentioned, we stuck to 8 or 16 frame clips as input and compared our proposed RCN with baseline I3D models.
We think this provides enough evidence of the validity of our proposal to move away from 
temporal convolutional networks~\cite{xie2018rethinking,nonlocal2018wang,tran2018closer,carreira2017quo}, and replace them with more sophisticated \emph{and} causal structures.
As with the role of additional layers, it is fair to predict that more extensive training on more extended clips (32,64,128) has a serious potential to take RCN to outperform state of the art in absolute terms.

\paragraph{On efficient training of 3D networks: } 
two basic things are clear from our experience with training heavy 3D models (I3D, (2+1)D, RCN) on large-scale datasets such as Kinetics.
Firstly, training is very computationally expensive and memory bulky; secondly, longer input clips are crucial to achieving better optimisation which, however, renders the first issue even more severe. 
We feel that how to train these model efficiently is still a wide-open problem, whose solution is essential to speed up the process for broader adoption.
We observed that ImageNet initialisation does speed up the training procedure, and helps reach local minima much quicker.
In the case of both I3D and RCN, ImageNet initialisation improves the video classification accuracy on 
Kinetics by almost $3\%$ compared to random initialisation when using the same number of training iterations, 
as shown in the first and last row of Table~\ref{rcn:table:hiddenInit}.

The bottom line is that we should strive for more efficient implementations of 3D models for the sake of their adoption.

%% file: chapters/8_conclusion.tex
\chapter{Conclusions and Future Research Directions}
\label{chapter:conclusion}
In this thesis , 
we have introduced a number of novel deep learning based frameworks 
(Chapter~\ref{chapter:offline}, \ref{chapter:online}, \ref{chapter:tramnet}, \ref{chapter:tramnet} \& \ref{chapter:rcn})
to solve for variety video understanding problems.
In particular, we addressed the problem of action detection, online and real-time action detection, early action prediction, future action tube prediction and causal video representations.
All these problems have applications in a real-world scenario, e.g. 
online and real-time action detection system could be used in real-time video surveillance, and future action tube prediction has applications in self-driving and many others.

To conclude this thesis, we divide this chapter into two sections.
First, we summarise (\S~\ref{chapter:conclusion:sec:summary_of_contribs}) the contributions of the presented work in this thesis in chapter-wise order.
Lastly, we will conclude by introducing some the most prominent direction for the future works(\S~\ref{chapter:conclusion:sec:summary_of_contribs}).

\section{Summary of contributions of the thesis}
\label{chapter:conclusion:sec:summary_of_contribs}

\subsection{Action detection using frame-level deep features}
We presented a novel spatiotemporal action detection pipeline and temporal labelling framework, where we show that supervised proposal is important for predicting frame-level detection boxes and the associated action class scores.
We propose an original fusion strategy for merging appearance and motion cues based on the softmax probability scores and spatial overlaps of the detection of bounding boxes.
We propose to solve action tube construction by two different optimisation formulation.
Finally, we proposed an efficient dynamic programming based solution to temporal localisation of action in tubes and videos. 
As a result, our pipeline was able to outperform the previous state-of-the-art methods in action detection. 

\subsection{Online and real-time action detection}
We will describe the first online and real-time action detection system which can detect multiple co-occurring actions in untrimmed videos. 
We devise an original, greedy online algorithm capable of generating multiple action tubes incrementally, 
which also provides early action class label predictions. Further, we extend the online action detection task to untrimmed videos for the first time.
We show that such a system can work in an online and real-time fashion and also outperform previous state-of-the-art action detection methods.

\subsection{Flexible micro-tube proposals}
Moving away from the frame-level detector and tube construction methods in previous chapters, we propose two novel deep networks to incorporate information from multiple frames.
Flexible anchor proposal method is proposed with the help of a hidden Markov model (HMM) formulation, which can be approximated efficiently into a transition matrix.
We show that such a network can handle significant spatial movement in dynamic actors without penalising more static actions.
Further, we observed that it is a scalable solution for training models on both sparse and dense annotations.

\subsection{Predicting Future locations of action tubes}
We showed that a micro-tube prediction network can be extended to predict the future of each action tube individually.
Thanks to online tube construction method proposed previous chapters, we propose a future action tube prediction framework.
Here, we linked that the online nature of methods is essential to future prediction. 
We demonstrate that training a network to make predictions also helps in improving action detection performance.
We also demonstrate that feature-based fusion works better than late fusion in the context of spatiotemporal action detection, unlike late fusion in previous chapters.

\subsection{Causal 3D representations}
Motivated by the applications of online methods, 
we propose to convert temporal-convolutions based state-of-the-art anti-causal 3D CNNs to causal 3D CNNs 
by an original convolutional network with recurrent hidden states at each depth level.
We show that it can directly benefit from model initialisation via ImageNet pre-trained weights, 
as opposed to state of the art approaches, and in line with clear emerging trends in the field.
It also allows flexible temporal reasoning, as it exploits information coming from all the input sequence observed.

\section{Overall conclusion}
We presented various action detection methods in this thesis, the main thread has been online processing of videos. 
We see that online processing not only help action detector to be applicable in real-world applications but also it can help such online detection to predict action early or predict future of action tubes.
We believe that such online processing would result in much more applications.
From another main contribution point of view, we show that we can take care of the dynamic nature of action while designing the reference anchor proposals for an action detector, 
such a detector can be scaled to handle both sparse and dense annotations.
Finally, we show that we can convert anti-causal 3D CNNs into causal 3DCNNs with the help of recurrence, which leads to flexible temporal reasoning.

\section{Future research directions}
\label{chapter:conclusion:sec:future_work}
In this section, we bring forward some prominent research directions for future work based on 
the experimental results reported in this thesis and
the very latest advancements in computer vision and machine learning.

\subsection{Causal spatiotemporal action detection with 3D CNNs}
We have discussed in Chapter~\ref{chapter:related_work} in Section~\ref{related_work:subsec:3d} 
that 3D CNNs enjoy superior performance in action detection task due to their high representational power.
However, all such methods are anti-causal.
We can use causal RCN proposed in Chapter~\ref{chapter:rcn} for better online action detector, also an extension to future anticipation tasks.
RCN can also preserve the temporal resolution, which would make its application in streaming applications easier.
All the existing methods using 3D CNNs for action detection use a 2D CNNs in parallel on keyframes for bounding box detection, then they pool classification features from 3DCNNs for that box. 
It is still to be tested if RCN would need such a 2D detector in parallel or not. 
Also, it yet to be seen if we can convert latest slow-fast 3D CNN ~\cite{feichtenhofer2018slowfast} into a causal 3D CNN.

\subsection{Improving action representation}
More recent works on 3D representation~\cite{tran2018closer,xie2018rethinking} has shown that factorised convolutions not only reduce the number
of parameters in a network but also improve the performance. Subsequently, Feichtenhofer ~\etal~\cite{feichtenhofer2018slowfast},
show the two parallel streams of a network can process information 
and different rate and help gain better performance on action recognition tasks.
It would be interesting to see if these kinds of networks can be made causal.
One of the ways to make these networks causal is to pre-trained these networks the replace 3D convolutions by recurrent convolutions.
Also, it would be interesting to try to train recurrent convolutions from scratch, 
as 3D CNNs are trained from scratch in~\cite{tran2018closer,feichtenhofer2018slowfast}.

\subsection{Improving temporal action detection}
We observed in Chapter~\ref{chapter:related_work} in Section~\ref{action_cls:subsec:deep_rep} that proposal based methods 
for temporal action detection are best at finding temporal boundaries of action instance. 
It would be interesting to see if action paths (Figure.~\ref{offline:fig:action_paths_tubes}) can be trimmed with temporal boundary prediction network~\cite{lin2018bsn}.
We can see from Table~\ref{online:table:ablation-study-ucf10105} that the classes with low ratio of action instance duration over video duration have a very low performance, i.e. occupy the less temporal extent in the video. 
It means these classes needed better temporal detection, e.g. `Cricket bowling' or `Volleyball Spiking'.
We feel that the adaptation of temporal detection literature into spatiotemporal action detection method has a lot of room to improve.